%% file: main_arxiv.tex
\documentclass[11pt,a4paper]{article}

\usepackage[a4paper,margin=25mm]{geometry}

\usepackage[utf8]{inputenc}                    
\usepackage[T1]{fontenc}
\usepackage{lmodern}
\usepackage[authoryear,round]{natbib}
\setcitestyle{authoryear,round,citesep={;},aysep={,},yysep={;}}                        
\usepackage{hyperref}                   
\usepackage{url}                                    
\usepackage{booktabs}                                    
\usepackage{array}
\usepackage{amsfonts}                                
\usepackage{nicefrac}                                      
\usepackage{microtype,needspace}                       
\usepackage[dvipsnames]{xcolor}                 
\def\PhaseOneClean{1}
\input{macros}
\hypersetup{hidelinks,bookmarksdepth=section,pdftitle={Finite-Sample Approximation of Hessian-Guided Perturbed Wasserstein Gradient Flows},pdfauthor={Ryotaro Kawata, Atsushi Nitanda, Taiji Suzuki},pdfsubject={Finite-particle approximation of perturbed Wasserstein gradient flows},pdfkeywords={Wasserstein gradient flow, mean-field optimization, finite-particle approximation, formal verification}}
\newcommand{\maintextheading}[1]{%
  \par\addvspace{0.5\baselineskip}%
  \Needspace{3\baselineskip}%
  \noindent\textbf{#1}\enspace\ignorespaces
}

\title{Finite-Sample Approximation of Hessian-Guided Perturbed Wasserstein Gradient Flows}

\author{%
  Ryotaro Kawata\textsuperscript{1}\thanks{Correspondence to:
    \href{mailto:kawata-ryotaro725@g.ecc.u-tokyo.ac.jp}{\nolinkurl{kawata-ryotaro725@g.ecc.u-tokyo.ac.jp}}.}%
  \quad Atsushi Nitanda\textsuperscript{2,3}%
  \quad Taiji Suzuki\textsuperscript{1,4}\\[0.5em]
  \normalsize\textsuperscript{1} The University of Tokyo\\
  \normalsize\textsuperscript{2} Agency for Science, Technology and Research (A*STAR)\\
  \normalsize\textsuperscript{3} Nanyang Technological University\\
  \normalsize\textsuperscript{4} RIKEN Center for Advanced Intelligence Project
}
\date{}

\begin{document}

\maketitle

\begin{abstract}
\setlength{\parindent}{0pt}
\noindent
Wasserstein gradient flow extends gradient descent to probability measures.
Its Hessian-guided perturbed variant (PWGF) adds Gaussian perturbations to
escape saddle points in nonconvex problems.
We investigate when its approximation by finitely many interacting particles
remains accurate over growing time horizons. Our analysis retains the curvature accumulated along
the population-driven reference path: negative curvature can amplify approximation
errors, while subsequent positive curvature can damp their influence. This
captures favorable scenarios in which temporary instability is compatible with
accurate tracking over growing horizons. Under regularity assumptions and a
prescribed common perturbation schedule, we prove particle and objective-value
tracking bounds on a high-probability event for reference paths satisfying
explicit conditions on accumulated curvature. To handle state-dependent Gaussian
jumps, we construct a population-first coupling that preserves the reference
particles' conditional independence and reduces jump errors to covariance
comparison. We verify the conditions in a variance-plus-cosine model, where
curvature recovery yields a growing-horizon tracking guarantee. We also establish local attraction, transverse
descent, and positive second variation in two regions of a regularized
matrix-factorization model, motivating a positive--negative--positive curvature
pattern.
\end{abstract}

\input{main_body}

\subsection*{Acknowledgments}
RK was partially supported by JSPS KAKENHI (24K02905) and JST BOOST (JPMJBS2418).
TS was partially supported by JST CREST (JPMJCR2115) and JSPS KAKENHI (25H01107).
This work was partially supported by JST ERATO Grant Number JPMJER2601.

This research is supported by the National Research Foundation, Singapore and
the Ministry of Digital Development and Information under the AI Visiting
Professorship Programme (award number AIVP-2024-004). Any opinions, findings and
conclusions or recommendations expressed in this material are those of the
author(s) and do not reflect the views of National Research Foundation, Singapore
and the Ministry of Digital Development and Information.

\input{sections/ai_use_statement}

\bibliographystyle{plainnat}
\bibliography{references}

\clearpage
\appendix

\input{appendix_final_general/general_argument_reorg_v1_6}

\input{appendix_mf_dirac/examples_matrix_nonlinear_rewrite_v15}

\input{appendix_numerical/20260926-023032JST_sanity_check}

\end{document}

%% file: macros.tex
\usepackage{fancybox}
\usepackage{ascmac}

\usepackage{enumitem}
\usepackage{amsmath,amssymb,amsthm}
\usepackage{mathtools}
\usepackage{bbm}

\usepackage{physics}
\usepackage{graphicx}
\usepackage{natbib}
\usepackage{cleveref}
\usepackage{comment}

\usepackage{color}
\usepackage[dvipsnames]{xcolor}

\newif\ifphaseonecolors
\ifdefined\PhaseOneClean
  \phaseonecolorsfalse
\else
  \phaseonecolorstrue
\fi
\definecolor{PhaseOneBlue}{RGB}{0,70,180}

\theoremstyle{plain}
\newtheorem{theorem}{Theorem}[section]
\newtheorem{lemma}{Lemma}[section]
\newtheorem{proposition}{Proposition}[section]
\newtheorem{definition}{Definition}[section]
\newtheorem{assumption}{Assumption}[section]
\newtheorem{example}{Example}[section]
\newtheorem{remark}{Remark}[section]
\newtheorem{corollary}{Corollary}[section]

\newcommand{\fv}[2]{\frac{\delta #1}{\delta #2}}
\newcommand{\Law}{\mathrm{Law}}
\newcommand{\bs}[1]{\boldsymbol{#1}}

\newcommand{\Id}{\mathrm{Id}}

\newcommand{\GP}{\mathrm{GP}}

\newcommand{\E}{\mathbb{E}}
\newcommand{\R}{\mathbb{R}}

\usepackage{mathrsfs}

\providecommand{\Ptwo}{\mathcal P_2}
\providecommand{\Law}{\mathrm{Law}}
\providecommand{\Wtwo}{W_2}
\providecommand{\dW}{\nabla_{\!W}}
\providecommand{\HessW}{\nabla^{2}_{\!W}}
\providecommand{\xHessW}{\nabla_x \nabla_{\!W}}

\providecommand{\GP}{\mathrm{GP}}

\newcommand{\submanifoldforsaddlepoint}{\mathcal{M}_{-}}

\newcommand{\supXY}{\mathrm{D}_{\sup}}
\newcommand{\NofJ}{L_{\mathrm{jump}}}

\newcommand{\lambdaReg}{\lambda_{\mathrm{reg}}}
\newcommand{\Sym}{\mathrm{Sym}}
\newcommand{\sym}[1]{\mathrm{sym}\left({#1}\right)}
\newcommand{\Hertensor}{\mathcal{H}}
\newcommand{\HS}{\mathrm{HS}}
\newcommand{\F}{\mathrm{F}}

\newcommand{\Sg}{\mathfrak{S}}
\newcommand{\innerProd}[2]{\left\langle #1,#2\right\rangle}
\newcommand{\HessianMatrices}{\mathrm{H}}
\newcommand{\contr}[1]{\otimes_{#1}}

\newcommand{\Sphere}{\mathbb{S}}

\newcommand{\fD}{\mathrm{D}}

\newcommand{\Loperator}[1]{{#1}\to{#1}}

\newcommand{\LOG}{\mathrm{LOG}}
\newcommand{\EXP}{\mathrm{EXP}}

\crefname{assumption}{assumption}{assumptions}

\allowdisplaybreaks[3]

%% file: main_body.tex
\input{sections/01_introduction}
\input{sections/02_pwgf_and_coupling}
\input{sections/03_curvature_tracking_theory}
\input{sections/04_matrix_factorization_geometry}
\input{sections/05_proof_overview}
\input{sections/06_discussion}

%% file: sections/01_introduction.tex
\section{Introduction}
\label{sec:introduction}

We study finite-particle approximation for nonconvex optimization over probability measures,
\begin{equation}
    \min_{\mu\in\operatorname{dom}F\subseteq\Ptwo(\R^d)} F(\mu),
    \label{eq:measure-optimization}
\end{equation}
where \(\Ptwo(\R^d)\) denotes the set of Borel probability measures with finite second moment;
the domain \(\operatorname{dom}F\) for our moment models is specified in \Cref{def:moment-objective}.
The canonical first-order dynamics is the Wasserstein gradient flow
\begin{equation}
    \partial_t \mu_t + \nabla\cdot(v_t\mu_t)=0,
    \qquad
    v_t(x)=-\dW F(\mu_t,x),
    \label{eq:intro-wgf}
\end{equation}
which is gradient descent in the \(2\)-Wasserstein geometry.
In mean-field models, \eqref{eq:intro-wgf} is the population equation used to
describe interacting-particle dynamics in the infinite-particle limit.
This viewpoint underlies generative modeling~\citep{arbel2019,chu2019pfd},
sampling~\citep{jordan1998}, and mean-field neural-network
training~\citep{chizat2018,songmei2018,nitanda2022convex}.
Related particle variational methods include Stein variational
gradient descent~\citep{liu2016}.
In several convex, displacement-convex, and entropy-regularized settings,
WGF and its particle approximations admit strong convergence and propagation-of-chaos
\citep{sznitman1991} guarantees~\citep{chen2025uniform,
suzuki2023uniformintime,nitanda2025propagationensemble}.
Global convergence guarantees are also available for particle dual averaging
in mean-field neural-network optimization~\citep{nitanda2022pda}.

For nonconvex objectives, however, first-order stationarity does not rule out
saddle points. Such second-order landscape structure already appears in
mean-field models, including the attention model studied by
\citet{kim2024nonlinear}.
Inspired by perturbed gradient descent~\citep{GeHuangJinYuan2015,pmlr-v75-jin18a},
\citet{yamamoto2025hessianguided} introduce Hessian-guided perturbed Wasserstein
gradient flow (PWGF), which perturbs the population measure along
Hessian-guided Gaussian vector fields to escape saddles. Their guarantees
concern the population measure flow. A finite implementation replaces both the
drift and the Hessian-guided jump covariance by empirical quantities,
introducing errors along the flow and at each perturbation.
The interacting particles are generally dependent, so standard sampling bounds
for independent samples do not apply directly.
\Needspace{4\baselineskip}
We therefore ask:
\begin{center}
\emph{When can a finite system of interacting particles accurately approximate population PWGF?}
\end{center}

We answer this finite-sample question through the following contributions.

\Needspace{5\baselineskip}
\maintextheading{Contributions.}
\begin{enumerate}[leftmargin=1.35em,label=\arabic*.,itemsep=0.01em,topsep=0.05pt,parsep=5pt]
    \item \textbf{Tracking and objective values.}
    When the reference path satisfies a controlled-curvature path condition,
    we obtain particle and objective-value tracking bounds throughout \([0,T]\)
    under regularity assumptions and a prescribed deterministic common jump schedule
    (\Cref{theorem-main-r_t}).
    The bounds retain their \(T\)-dependence and weight ODE residuals and jump
    mismatches by separate curvature-response factors. By retaining the
    accumulated curvature, they capture how later positive curvature damps
    earlier errors. For fixed horizons, the ODE and jump prefactors are
    \(N^{-1/2}\) and \(N^{-1/4}\), respectively, up to arbitrarily small powers of \(N\).
    A variance-plus-cosine example gives a growing-horizon tracking guarantee
    (\Cref{ex:variance-cosine-curvature-recovery}).

    \item \textbf{A population-first coupling for Gaussian jumps.}
    At each jump, we construct a finite-dimensional standard Gaussian vector that
    reproduces the population-field evaluations and drives the empirical update.
    This preserves the correct marginal dynamics and the reference particles'
    conditional i.i.d. law given the population fields, while reducing the jump
    comparison to a covariance-factor mismatch
    (\Cref{subsec:auxiliary-particles,subsec:proof-sketch-after-assumptions}).

    \item \textbf{Local geometry of matrix factorization.}
    Matrix factorization provides a product model for the nonlinear-moment
    geometry arising in matrix and tensor decompositions. In a regularized
    product-measure model, we establish attraction along an anti-symmetric
    branch, descent transverse to that branch, and positive second variation
    in two regions under spherical product pushforward perturbations
    (\Cref{sec:matrix-factorization}). These local results motivate the
    positive--negative--positive curvature scenario.
\end{enumerate}

\maintextheading{Relation to prior work.}
\citet{yamamoto2025hessianguided} analyze population PWGF escape and convergence,
but do not bound its population-to-empirical approximation error.
\citet{glasgow2025meanfield} track continuous projected neural-network dynamics
using local particle curvature, without Hessian-induced Gaussian jumps.
Reference-particle couplings and Gaussian covariance-factor alignment have
precedents~\citep{sznitman1991,masarotto2019procrustes}. Here the coupling must
preserve the prescribed population fields, because the conditional
reference-particle law given those fields is the input to concentration.
We establish this compatibility and control the resulting jump covariance
residual through weighted features, then propagate continuous and jump errors
through the sampled curvature.

\maintextheading{Scope.}
The population PWGF proposed by \citet{yamamoto2025hessianguided}
selects perturbation times adaptively.
Prescribed schedules provide a reference point for studying how adaptive
timing affects finite-particle approximation.
Our results quantify finite-particle approximation errors under a common
prescribed schedule while retaining state-dependent drift and jump covariance.
The tracking bound applies along reference paths satisfying
the controlled-curvature path condition in \Cref{theorem-main-r_t},
which bounds error propagation along those paths.
The variance-plus-cosine example verifies this condition with high probability
and gives a growing-horizon tracking guarantee with a single prescribed
jump (\Cref{ex:variance-cosine-curvature-recovery}).
Extending this verification to broader models and synchronizing adaptive
jump times remain open directions.

%% file: sections/02_pwgf_and_coupling.tex
\section{Coupled population and particle PWGF}
\label{sec:pwgf-coupling}

We define the population and finite-particle dynamics and the coupling used to
compare them along a prescribed population PWGF path.

\subsection{Population WGF and Hessian-guided perturbations}
\label{subsec:population-pwgf}

Let \(F:\operatorname{dom}F\to\R\) be the objective functional.
Between perturbation times, the
population curve \((\mu_t)_{t\ge 0}\) follows the Wasserstein gradient flow
\begin{equation}
    \partial_t \mu_t + \nabla\cdot(v_t\mu_t)=0,
    \qquad
    v_t(x):=-\dW F(\mu_t,x).
    \label{eq:population-wgf}
\end{equation}
The original population PWGF triggers perturbations adaptively when
\(\|\dW F(\mu_t)\|_{L^2(\mu_t)}\) falls below a prescribed first-order threshold;
our finite-sample comparison uses deterministic jump times
$0<t_1<\cdots<t_{\NofJ}\le T$, fixed in advance and common to both systems.
At each such time, \citet{yamamoto2025hessianguided} perturb the measure by pushing it forward along a Hessian-guided Gaussian vector field.  The idea parallels perturbed
gradient methods in Euclidean space~\citep{GeHuangJinYuan2015,pmlr-v75-jin18a}: near a saddle,
randomness should reach negative-curvature directions.  In
\(\mathcal P_2(\mathbb R^d)\), (i) the definition of such perturbations and (ii) their
alignment with descent directions are both nontrivial, motivating the
Hessian-induced covariance.
In Wasserstein space, the second variation
along the pushforward \((\Id+hv)_\#\mu\) is represented by
\(\mathcal H_\mu=H_\mu+H_\mu'\), where
\[
    (H_\mu v)(x):=\int \HessW F(\mu;x,y)v(y)\,\mu(dy),
    \qquad
    (H_\mu'v)(x):=\xHessW F(\mu,x)v(x).
\]
For general Wasserstein gradient-flow background, see
\citet{ambrosio2005gradient}; the above second-variation decomposition is given
in \citet[Propositions~3.2 and~C.7]{yamamoto2025hessianguided}.
The perturbation studied here uses the squared integral
Hessian $H_\mu^2$ as covariance. The tracking curvature uses the full operator
$\mathcal H_\mu$; first-order stationarity alone does not imply $H_\mu'=0$.
The covariance kernel of \(H_\mu^2\) is
\[
    K_\mu(x,y)
    :=
    \int
        \HessW F(\mu;x,z)\HessW F(\mu;z,y)\,\mu(\dd z).
    \label{eq:hessian-kernel}
\]
This covariance preserves the eigenspaces of $H_\mu$ and assigns variance proportional to its squared eigenvalues. Relating this covariance to descent directions of the full Hessian requires additional structural assumptions. The population jump is
\begin{equation}
    \mu_{t_\ell}
    =
    (\Id+\eta_\ell \xi_{\mu_{t_\ell-}})_\#\mu_{t_\ell-},
    \qquad
    \xi_{\mu_{t_\ell-}}\sim \GP(0,K_{\mu_{t_\ell-}}),
    \label{eq:population-jump}
\end{equation}
with perturbation size \(\eta_\ell>0\).

\subsection{Empirical particle dynamics}
\label{subsec:particle-pwgf}

For \(\bs{X}=(X^1,\ldots,X^N)\in(\R^d)^N\), write $\rho_{\bs{X}}\coloneq \frac1N\sum_{i=1}^N \delta_{X^i}.$
The empirical PWGF evolves by the particle ODE
\begin{equation}
    \frac{\dd}{\dd t}X^i_t
    =
    -\dW F(\rho_{\mathbf X_t},X^i_t),
    \qquad i=1,\ldots,N,
    \label{eq:particle-flow}
\end{equation}
between jump times.  At \(t_\ell\), the particles are updated by
\begin{equation}
    X^i_{t_\ell}
    =
    X^i_{t_\ell-}+\eta_\ell\xi_{\rho_{\mathbf X_{t_\ell-}}}(X^i_{t_\ell-}),
    \qquad i=1,\ldots,N,
    \label{eq:particle-jump}
\end{equation}
where \(\xi_{\rho_{\mathbf X_{t_\ell-}}} \sim \GP(0,K_{\rho_{\mathbf X_{t_\ell-}}})\) is a Gaussian process induced by the empirical Hessian-guided kernel.  The covariance of \(\xi_{\rho_{\mathbf X}}(X^1),\ldots,\xi_{\rho_{\mathbf X}}(X^N)\) is
a block matrix whose \((i,j)\)-block is \(K_{\rho_{\mathbf X}}(X^i,X^j)\).
For a fixed configuration, this jump is finite dimensional: one applies an
\(Nd\times Nd\) covariance factor to a standard Gaussian vector in
\(\mathbb R^{Nd}\). It can therefore be simulated without access to the
population path.

\subsection{The auxiliary reference particles}
\label{subsec:auxiliary-particles}

The comparison is cleaner if we introduce a reference particle system
\(Y_t=(Y^1_t,\ldots,Y^N_t)\). Conditional on the population fields,
\(Y_t^1,\ldots,Y_t^N\) are i.i.d. with law \(\mu_t\). The reference particles
start from the same initial positions as the empirical particles and evolve
with the population drift and jump covariance. Their Gaussian increments are
coupled to the empirical increments through the common Gaussian vector
described below.
Specifically, we set
\begin{equation}
    X^i_0=Y^i_0,
    \qquad
    Y^i_0\stackrel{\mathrm{i.i.d.}}\sim \mu_0.
    \label{eq:same-initial-particles}
\end{equation}
Between jumps,
\begin{equation}
    \frac{\dd}{\dd t}Y^i_t
    =
    -\dW F(\mu_t,Y^i_t),
    \qquad i=1,\ldots,N.
    \label{eq:reference-flow}
\end{equation}
At \(t_\ell\), the reference jump is
\begin{equation}
    Y^i_{t_\ell}
    =
    Y^i_{t_\ell-}+\eta_\ell\xi_{\mu_{t_\ell-}}(Y^i_{t_\ell-}),
    \qquad i=1,\ldots,N,
    \label{eq:reference-jump}
\end{equation}

At each jump, the population Gaussian field is generated from a fresh
innovation independent of the initial sample and the previous innovations.
Conditional on these fields, the flow and jump maps are deterministic;
applying their common composition to the independent initial positions
preserves the reference particles' conditional i.i.d. law
(\Cref{lem:population-first-finite-coupling,eq:population-pushforward-product-law}).
For the comparison, \Cref{lem:population-first-finite-coupling} constructs a common
standard Gaussian vector \(\mathbf g_\ell^N\in\mathbb R^{Nd}\) from the given
population-field evaluations and independent auxiliary Gaussian noise,
leaving the population path unchanged. If
\(B_{\rm pop}\) denotes the aligned factor reproducing the population-field
values at the reference particles and \(M_X\) denotes the empirical Hessian
factor, then
\[
    \bigl(\xi_{\mu_{t_\ell-}}(Y^i_{t_\ell-})\bigr)_{i=1}^N
    =B_{\rm pop}\mathbf g_\ell^N,
    \qquad
    \bigl(\xi_{\rho_{\mathbf X_{t_\ell-}}}(X^i_{t_\ell-})\bigr)_{i=1}^N
    =M_X\mathbf g_\ell^N.
\]
Both jumps have the prescribed conditional laws, and the Gaussian
field-value vectors differ by \((M_X-B_{\rm pop})\mathbf g_\ell^N\).
The precise construction, including
singular covariances, and the conditional i.i.d. law of the reference particles
are given in
\Cref{lem:population-first-finite-coupling,lemma-gp-jump-implies-conditional-iid}.

\subsection{Tracking metric and decomposition}
\label{subsec:tracking-metric}

We measure the pathwise discrepancy by
$r_t
    \coloneq 
    \left(\frac1N\sum_{i=1}^N \|X^i_t-Y^i_t\|^2\right)^{1/2}
    =
    \frac1{\sqrt N}\|X_t-Y_t\|_2.
    \label{eq:tracking-metric}
$
Since the empirical coupling \(N^{-1}\sum_i\delta_{(X^i_t,Y^i_t)}\) is admissible, $\Wtwo(\rho_{\mathbf X_t},\rho_{\mathbf Y_t})\le r_t$.
To compare objective values with the population flow, we combine the
paired-particle estimate with concentration of the reference cloud's feature
moments (\Cref{subsec:functional-gap}).
The next section gives joint tracking and objective bounds under a prescribed
common schedule.

%% file: sections/03_curvature_tracking_theory.tex
\section{Curvature-driven tracking theory}
\label{sec:tracking-theory}

This section proves a finite-sample comparison theorem for perturbed WGF.
The central object is the curvature propagator \(G_\kappa^Y(t,s)\), defined below.
It transports an error created at time \(s\) to time \(t\), discounting or
amplifying it according to the curvature accumulated after its creation.

\subsection{Curvature-response factors}
\label{subsec:curvature-budgets}

\begin{definition}[Curvature-response factors]
\label{def:curvature-responses}
Let  $\HessianMatrices_t^Y$ denote the empirical discretization of $\mathcal{H}_{\mu_t}$ along the
reference particles, acting on the indexed particle space $(\mathbb R^d)^N$:
\begin{equation}
    \HessianMatrices_t^Y\coloneq
    \left\{
    \frac{1}{N}
    \nabla_W^2 F\!\left(\mu_t,Y^i_t,Y^j_t\right)
    + \delta_{ij}\nabla_x\nabla_W F\!\left(\mu_t,Y^i_t\right)\in \R^{d\times d}\right\}_{i,j=1}^N.
\end{equation}
We define the sampled curvature by
\begin{equation}
    \kappa^{Y,N}_t := \lambda_{\min}(\HessianMatrices^Y_t) \coloneq 
    \inf_{\mathbf v \in (\R^d)^N, \norm{\mathbf v}_2=1}
    \langle \mathbf v,\HessianMatrices^Y_t \mathbf v\rangle
    \label{eq:curvature-main}
\end{equation}
Positive \(\kappa_t^{Y,N}\) corresponds to instantaneous contraction of the
linearized discrepancy, while negative \(\kappa_t^{Y,N}\) corresponds to local
expansion.  This is the same role played by a local strong-convexity parameter in
ordinary gradient descent, except that here the parameter may change sign along
the trajectory. For example, $F(\mu)=\int \frac{\norm{x}^2}{2}\,\dd\mu$ implies $\HessianMatrices_t^Y = \boldsymbol{I}_{dN}$ and $\kappa_t^{Y,N} = 1>0$.

On any reference path satisfying
\(\int_0^T|\kappa_u^{Y,N}|\,\dd u<\infty\), we define the continuous
curvature-response factor
\[
    \mathfrak C_t^Y
    \coloneq 
    \int_0^t
    G_\kappa^Y(t,s)
    \dd s ,\quad G_\kappa^Y(t,s)
    \coloneq 
    \exp\left\{-\int_s^t \kappa_u^{Y,N}\,\dd u\right\},
    \qquad 0\le s\le t
\]
and the jump curvature-response factor
\[
    \mathfrak J_t^Y
    :=
    \sum_{\ell=1}^{L_t}
    \eta_\ell
    G_\kappa^Y(t,t_\ell)
    ,\quad L_t:=\#\{\ell:t_\ell\le t\}.
\]
\end{definition}
\begin{remark}[Conditioning used below.]
Let \(\mathcal G_{\mathrm{pop}}\) be the sigma-field generated by the population
Gaussian fields and the prescribed jump schedule. Conditional on
\(\mathcal G_{\mathrm{pop}}\), the reference particles are i.i.d. with law
\(\mu_t\), while the finite-dimensional coupling vectors
\(\mathbf g_\ell^N\) are not included in this conditioning; see
\Cref{subsec:auxiliary-particles}.
\end{remark}

\subsection{Main theorem}
\label{subsec:main-theorem}

We now quantify how curvature controls the accumulation of drift
fluctuations and jump mismatches over a growing horizon, under the following
assumptions:
\begin{itemize}[leftmargin=1.25em,itemsep=0pt,topsep=2pt,parsep=0pt]
    \item a structured moment representation of \(F\), with weighted smoothness
    controlling its drift and Hessian
    (\Cref{assumption-main-general-moment-smooth-compressed});
    \item uniform particle moments, branchwise Lipschitz paths, and the stated
    non-spikiness bound for the paired errors
    (\Cref{assumption-main-coupled-particle-regularity}); and
    \item a prescribed deterministic common schedule and the population-first
    Gaussian coupling (\Cref{assumption-main-jump-time}).
\end{itemize}
Under these conditions, we obtain the following comparison, valid
simultaneously for all \(t\in[0,T]\).

\begin{theorem}[Particle and objective tracking along controlled-curvature paths]
\label{theorem-main-r_t}
Assume \Cref{assumption-main-general-moment-smooth-compressed,assumption-main-coupled-particle-regularity,assumption-main-jump-time}, with fixed structural
parameters. Fix $\upsilon\ge0$, $\zeta>0$ with $\upsilon+\zeta<1/4$, and constants $C_p,C_0,L_0>0$.
Let $T\le N^{1/4-\upsilon-\zeta}$, $p\in[N^{-C_p},1)$, and use a prescribed deterministic
schedule with at most $L_0$ jumps and amplitudes $0\le\eta_\ell\le C_0$.
For brevity, write
$\Xi_{N,T,p}:=N(e+T)/p$.
For each \(N,T,p\), there is an event
$\Omega_p=\Omega_{N,T,p}$, independent of \(\varepsilon\), with probability at
least $1-p$. For every $\varepsilon>0$ and all sufficiently large $N$, the
following holds on $\Omega_p$. Suppose that the realized reference path satisfies
\[
\int_0^T|\kappa_t^{Y,N}|\,\dd t<\infty,\qquad
\sup_{t\le T}\mathfrak C_t^Y\le C_0N^\upsilon(1+T),\qquad
\sup_{t\le T}\mathfrak J_t^Y\le C_0N^\upsilon.
\]
Then, with $r_t=(N^{-1}\sum_{i=1}^N
\|X_t^i-Y_t^i\|_{\mathbb R^d}^2)^{1/2}$, simultaneously for $t\in[0,T]$,
\begin{equation}
\label{eq:objective-tracking-bound}
\begin{aligned}
&r_t\vee|F(\rho_{\mathbf X_t})-F(\mu_t)|\\
&\quad\le C_\varepsilon\Xi_{N,T,p}^{\varepsilon}
\left[(N^{-1/2}+T/N)(1+\mathfrak C_t^Y)
 +(N^{-1/4}+\sqrt{T/N})\mathfrak J_t^Y\right].
\end{aligned}
\end{equation}
Here $\vee$ denotes the maximum. The constants may depend on the fixed
parameters, including $\upsilon,\zeta,C_p,C_0,L_0$.
\end{theorem}

We also report numerical sanity checks of non-spikiness and adaptive PWGF in \Cref{sec:numerical-sanity}.

\begin{figure}[!h]
    \centering
    \includegraphics[width=0.7\linewidth]{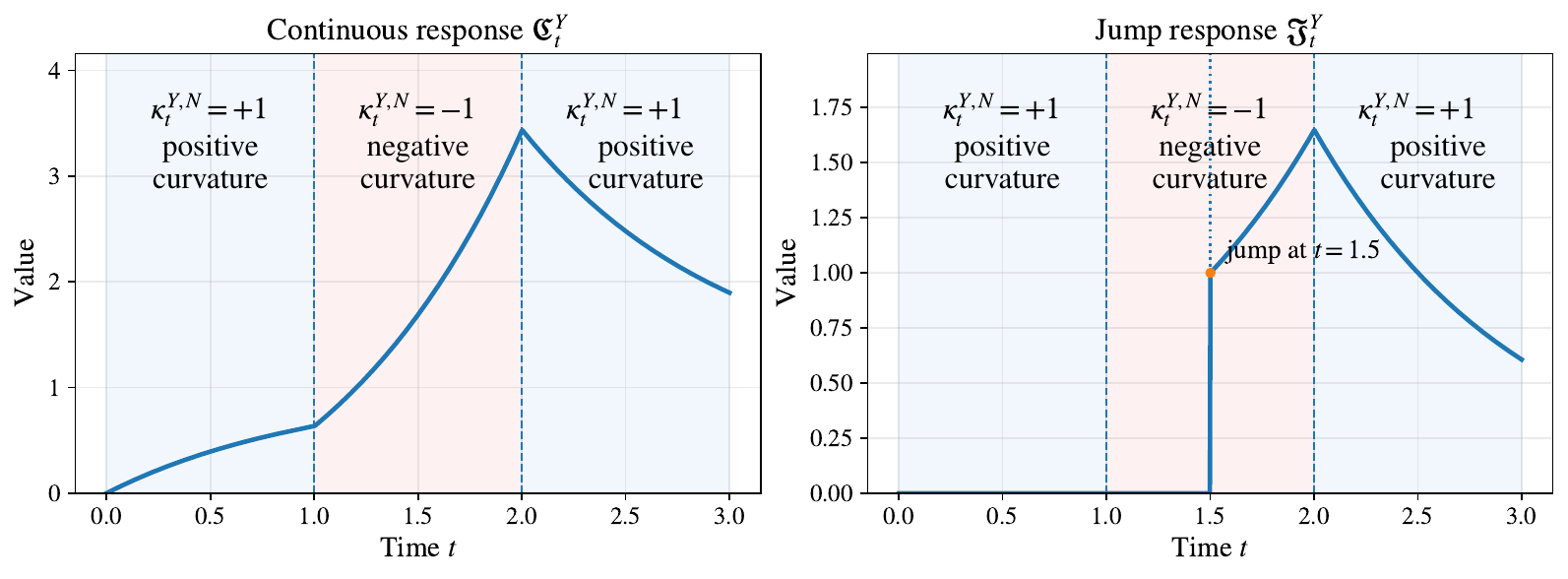}
    \caption{
    Illustration of the curvature-response factors in the tracking bound. We assume that $\kappa_t^{Y,N} = 1$ (positive curvature) if $t\in[0,1)\cup[2,3]$ and $\kappa_t^{Y,N} = -1$ (negative curvature) if $t\in[1,2)$.
    The factor \(\mathfrak C_t^Y\) measures the remaining effect of unit ODE forcing,
    while \(\mathfrak J_t^Y\) records a unit jump at \(t=1.5\), followed by
    amplification under negative curvature and decay under positive curvature.
    }
    \label{fig:curvature-response}
\end{figure}

\maintextheading{Curvature determines how long finite-sample errors persist.}
The ODE residual has scale \(N^{-1/2}+T/N\), while the jump mismatch has scale
\(N^{-1/4}+\sqrt{T/N}\), apart from the common subpolynomial factor
\(C_\varepsilon\Xi_{N,T,p}^{\varepsilon}\). The \(T/N\) and
\(\sqrt{T/N}\) terms arise from controlling the residuals over the entire
interval \([0,T]\); under
\(T\le N^{1/4-\upsilon-\zeta}\), they are dominated by \(N^{-1/2}\) and \(N^{-1/4}\),
respectively. These scales are then weighted by the actual curvature
propagator \(G_\kappa^Y\): \(\mathfrak C_t^Y\) accumulates errors generated
continuously, whereas \(\mathfrak J_t^Y\) propagates jump mismatches. Keeping this
propagator avoids replacing the accumulated curvature by a worst-case bound
that can grow exponentially in \(T\). Positive curvature makes past errors
decay, while negative curvature can preserve or amplify them
(\Cref{fig:curvature-response}). A jump mismatch may therefore dominate just
after a perturbation and later be forgotten, whereas continuous errors are
generated throughout the flow.

The following example makes this curvature-dependent tracking bound concrete.

\begin{example}[Growing-horizon tracking in a variance-plus-cosine model]
\label{ex:variance-cosine-curvature-recovery}
In dimension one, consider the model in \Cref{subsec:canonical-variance-cosine},
\[
 F(\mu)=\frac a2\operatorname{Var}(\mu)+\beta\cos(\pi m_\mu),
 \qquad m_\mu=\int x\,\mu(dx),\qquad a,\beta>0.
\]
Take a fixed bounded mean-zero initial law and one prescribed Gaussian jump
at a fixed time $\tau>0$ with fixed amplitude $\eta>0$, independently of $N$.
Writing $m_t=m_{\mu_t}$ and setting $T=N^{1/8}$,
\Cref{prop:canonical-model-assumptions} gives an event of probability at least
$1-N^{-1/16}$ on which
\Cref{lem:canonical-response-control,eq:canonical-sampled-curvature,eq:canonical-response-bounds} give
\[
 \kappa_t^{Y,N}=\min\{a,-\beta\pi^2\cos(\pi m_t)\},
 \qquad
 \sup_{0\le t\le T}(\mathfrak C_t^Y+\mathfrak J_t^Y)\le CN^{1/16}.
\]
\Cref{theorem-main-r_t} then gives, on the same event,
\[
 \sup_{0\le t\le T}
 \left[r_t+\left|F(\rho_{\mathbf X_t})-F(\mu_t)\right|\right]
 \le C_\varepsilon N^{-3/16+\varepsilon}.
\]
The bounds hold for each fixed $\varepsilon>0$ and sufficiently large $N$,
with fixed model parameters.
\end{example}

%% file: sections/04_matrix_factorization_geometry.tex
\section{Motivating geometry in a matrix factorization model}
\label{sec:matrix-factorization}

Matrix factorization is a product model for nonlinear-moment geometry
\citep{kolda2009tensor,sidiropoulos2017tensor,anandkumar2014tensor}.
Following the motivating matrix-factorization application of
\citet{yamamoto2025hessianguided}, we study a regularized product-measure
extension and prove attraction along the anti-symmetric branch, transverse
descent, and positivity of the second variation along spherical product
pushforwards on an annulus and near the teacher. These population statements motivate the
positive--negative--positive curvature-response scenario, but do not construct
a connecting PWGF trajectory, transfer the second variations to the sampled
curvature \(\kappa_t^{Y,N}\), or verify its controlled-curvature path condition.

\subsection{Product-measure model}
\label{subsec:mf-model}

Let \(\mu=\mu_1\otimes\mu_2\) have marginals on \(\Sphere^{d_w-1}\),
and let \(z\sim N(0,I_{d_w})\).
For a vector-valued activation \(\sigma:\mathbb R\to\mathbb R^{d_h}\), define
the mean-field feature map
\(h_\nu(z):=\int\sigma(w^\top z)\,\nu(\dd w)\),
which averages features over weights \(w\sim\nu\).
Given a teacher \(\mu^o=\mu_1^o\otimes\mu_2^o\), consider
\begin{equation}
    F(\mu_1\otimes\mu_2)
    :=
    \frac12\E_z\left[
        \left\|h_{\mu^o_1}(z)h_{\mu^o_2}(z)^\top
        -h_{\mu_1}(z)h_{\mu_2}(z)^\top\right\|_{\mathrm F}^2
    \right]
    +
    \frac{\lambda_{\mathrm{reg}}}{p_W}Z_\pm(\mu_1,\mu_2)^{p_W/2},
    \label{eq:mf-objective}
\end{equation}
with \(\lambda_{\mathrm{reg}}>0\), \(p_W>2\),
\(\nu_\star^\pm:=\tfrac12\delta_{a_\star}+\tfrac12\delta_{-a_\star}\),
and \(Z_\pm(\mu):=Z_\pm(\mu_1,\mu_2):=\sum_{i=1}^2
W_{2,\Sphere}^2(\mu_i,\nu_\star^\pm)\), where
\(a_\star\in\Sphere^{d_w-1}\).
The spherical distance convention is given in
\Cref{subsec:product-spherical-model}.

\subsection{Branches and standing conditions}
\label{subsec:active-moments}

For \(0<r_{\mathrm{cap}}<\pi/2\), let \(\mathcal P_\pm(r_{\mathrm{cap}})\)
denote the class of probability measures whose closed supports lie inside
the two open spherical caps of radius \(r_{\mathrm{cap}}\) centered at
\(\pm a_\star\), with mass \(1/2\) in each cap.
Throughout this section, impose the following restrictions.
\begin{assumption}[Model restrictions; see \Cref{subsubsection:spherical-model-assumptions}]
\label{assumption:mf-model-restrictions}
\textbf{(i)} \(\mu_i,\mu_i^o\in\mathcal P_\pm(r_{\mathrm{cap}})\) for \(i=1,2\).
\textbf{(ii)} The nonzero Hermite coefficients of \(\sigma\) occur at a finite
set \(\mathcal O\) of odd degrees.
\textbf{(iii)} \(\mu_1^o=\mu_2^o=:\nu^o\).
\end{assumption}

For transverse descent in \Cref{prop:mf-local-geometry}\ref{item:mf-transverse-descent},
we additionally require a consistent average bias orthogonal to \(a_\star\)
across the two-sided latitude slices \(|a_\star^\top w|\), with a positive
margin away from the equator
(\Cref{assumption:quantitative-unweighted-two-sided-transverse-imbalance}).

Using standard Gaussian-input moment coordinates
\citep{arous2021online,abbe23a,bietti2023}, set
\(T_{i,n}:=\int w^{\otimes n}\mu_i(\dd w)\),
\(S_n:=(T_{1,n}+T_{2,n})/2\), and
\(A_n:=(T_{1,n}-T_{2,n})/2\) for \(n\in\mathcal O\).
Define
\begin{equation}
    \mathcal M_-:=\{\mu_1\otimes\mu_2:\mu_2=(-\Id)_\#\mu_1\},
    \qquad
    \mathcal M_+:=\{\mu_1\otimes\mu_2:\mu_1=\mu_2\}.
    \label{eq:branches}
\end{equation}
Membership in \(\mathcal M_-\) implies \(S_n=0\), and membership in
\(\mathcal M_+\) implies \(A_n=0\); the converses are not assumed.
The central target is \(\mu^\dagger:=\nu_\star^\pm\otimes\nu_\star^\pm\).

\begin{figure}
    \centering
    \includegraphics[width=0.7\linewidth]{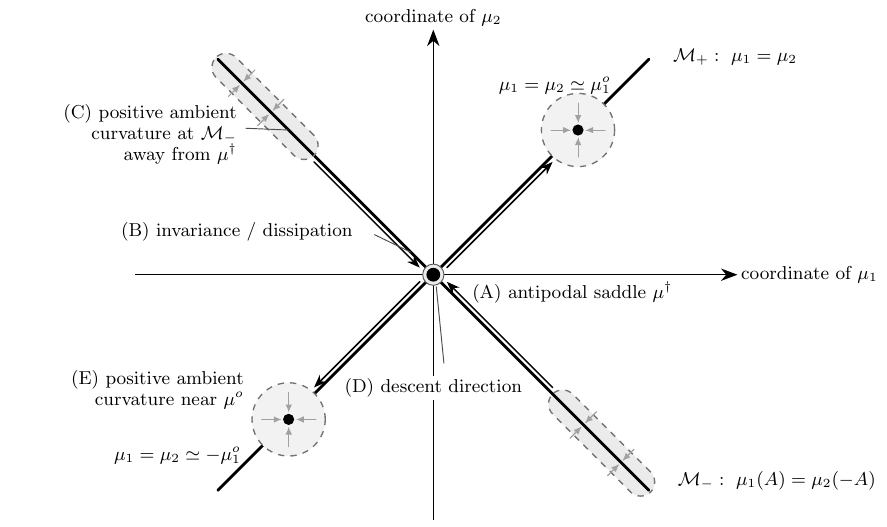}
    \caption{Schematic local geometry in \Cref{prop:mf-local-geometry}: (A)--(B) the stationary target and its local attracting branch \ref{item:mf-local-attraction};
    (C) annular positive curvature \ref{item:mf-annular-curvature};
    (D) transverse descent \ref{item:mf-transverse-descent}; and
    (E) positive curvature near the teacher \ref{item:mf-teacher-curvature}.
    The diagram's \(-\mu_1^o\) denotes the reflected measure \((-\Id)_\#\mu_1^o\).
    Pale arrows depict curvature, not gradient-flow directions.
    The arrows do not assert a connecting trajectory.}
    \label{fig:eclair}
\end{figure}

\subsection{Local geometry}
\label{subsec:mf-curvature-pieces}

We use the spherical product pushforward second variation defined in
\Cref{lem:mf-full-spherical-Hessian}.
Admissible fields are smooth product tangent fields preserving the caps for
small perturbations; they need not be tangent to either branch. When the
prescribed pushforward path is a product Wasserstein geodesic, this second variation
is its usual geodesic second derivative.

\Needspace{5\baselineskip}
\begin{proposition}[Local geometry of the matrix factorization model]
\label{prop:mf-local-geometry}
Under \Cref{assumption:mf-model-restrictions}, with \(\lambda_{\mathrm{reg}}>0\) and \(p_W>2\), the following statements hold,
each under the additional conditions stated in its item.
\begin{enumerate}[label=(\roman*),ref=(\roman*),leftmargin=*]
\item \label{item:mf-local-attraction}
\textbf{Local attraction on the anti-symmetric branch (A--B).}
For \(2<p_W<3\), every globally well-posed characteristic product WGF in the sense of \Cref{def:mf-characteristic-solution} on
\(\mathcal M_-\) that starts sufficiently close to \(\mu^\dagger\) within the
cap class remains in that class and converges to \(\mu^\dagger\) in product
Wasserstein distance.
The precise initialization conditions and rate are given in
\Cref{prop:reduced-dissipation-antisymmetric-branch}.

\item \label{item:mf-annular-curvature}
\textbf{Positive ambient curvature on an annulus (C).}
For every \(\zeta_->0\), if \(\lambda_{\mathrm{reg}}\) is sufficiently large,
depending on \(\zeta_-\) and the fixed model parameters, then at every
\(\mu\in\mathcal M_-\) with \(Z_\pm(\mu)\ge\zeta_-^2\), the spherical product
pushforward second variation is positive in every nonzero admissible direction.
The quantitative threshold and lower bound are given in
\Cref{cor:branchwise-positivity-Wasserstein-Mminus}.

\item \label{item:mf-transverse-descent}
\textbf{Transverse descent at the target (D).}
If \(\mathcal O\ne\varnothing\) and the teacher satisfies
\Cref{assumption:quantitative-unweighted-two-sided-transverse-imbalance}
at \(a_\star\), ensuring a non-cancelling transverse teacher signal,
there exists a same-density transverse perturbation \(\mu_h\) from
\(\mu^\dagger\) such that, as \(h\to0\),
\(F(\mu_h)=F(\mu^\dagger)-\gamma_{\mathrm{esc}}h^2+o(h^2)\),
with \(\gamma_{\mathrm{esc}}>0\).
Its construction is given in
\Cref{lemma:descent-direction-antipodal-dirac-from-unweighted-imbalance}.
The regularizer contributes \(O(|h|^{p_W})=o(h^2)\).

\item \label{item:mf-teacher-curvature}
\textbf{Positive ambient curvature near the teacher (E).}
If \(Z_\pm(\mu_+^o)>0\), where \(\mu_+^o:=\nu^o\otimes\nu^o\),
the spherical product pushforward second variation has a positive uniform
margin in a sufficiently
small product Wasserstein neighborhood of \(\mu_+^o\), relative to the cap class.
The existence of the margin and neighborhood is proved in
\Cref{prop:Mplus-teacher-local-positive-curvature}.
The teacher need not be stationary for the regularized objective.
\end{enumerate}
\end{proposition}

%% file: sections/05_proof_overview.tex
\section{Technical assumptions and proof overview}
\label{sec:proof-overview}

\subsection{Technical Assumptions}
\label{subsec:tracking-assumptions}

We state the conditions used in the main theorem and proof sketch;
\Cref{subsec:general-argument-bound-settings} gives their full analytic formulation,
and \Cref{tab:assumption-evidence-map} records how each model section relates to
the theorem's assumptions and conclusions.
The conditions below are verified for
\Cref{ex:variance-cosine-curvature-recovery} in
\Cref{lem:canonical-comparison-assumptions}.
\begin{assumption}[Structured moment representation]
\label{assumption-main-general-moment-smooth-compressed}
The full conditions of \Cref{assumption-structured-moment-smoothness} hold.
We use \(F(\rho)=G(M(\rho))\), with
\(M(\rho):=(\int\phi_m\,d\rho)_{m\ge1}\), as defined in
\Cref{def:moment-objective}.
The features have controlled polynomial growth through third spatial
derivatives, dominated by the decay of the moment weights
\(a_m=e^{-c_am^{\kappa_{\rm dec}}}\), with \(c_a>0\) and
\(\kappa_{\rm dec}>1\); see
\Cref{eq:weighted-moment-space,assumption-structured-moment-smoothness}.
The functional \(G\) is \(C^3\) on one open weighted-moment domain
containing all finite-scale feasible moment regions, with a uniform
weighted third-derivative bound.
All structural parameters, including the anchor derivative norms and this
uniform bound, are fixed independently of \(N,T,p\).
\end{assumption}

The localized drift and Hessian bounds derived in
\Cref{lem:conditional-local-coefficient-growth,lemma-localized-moment-smoothness-consequences}
may depend on the spatial localization radius \(D\).

\begin{assumption}[Regularity of the coupled particle systems]
\label{assumption-main-coupled-particle-regularity}
Fix \(T>0\). The coupled particle systems
\(\mathbf X_t=(X_t^1,\dots,X_t^N)\) and
\(\mathbf Y_t=(Y_t^1,\dots,Y_t^N)\) satisfy the following conditions uniformly
on \([0,T]\). The \(O(1)\) bounds below are uniform in \(N\) over the
horizons considered in \Cref{theorem-main-r_t}.

\begingroup\raggedright
\noindent
\textbf{(i) Unconditional moment bound.}
For fixed $\alpha\in(0,2]$,
\(\sup_{t\le T,q\ge1}q^{-1/\alpha}
\max\{(\mathbb E\|X_t^1\|^q)^{1/q},
       (\mathbb E\|Y_t^1\|^q)^{1/q}\}=O(1)\).\par
\endgroup

\noindent
\textbf{(ii) Temporal regularity.}
All particle paths are right-continuous with left limits at the jumps and
share a deterministic \(O(1)\)-Lipschitz bound on every ODE branch, almost surely.

\noindent
\textbf{(iii) Non-spikiness of $r_t$.}
The coupling error \(r_t\) from \Cref{subsec:tracking-metric} is assumed,
almost surely, not to concentrate on a small number of particles: for some
\(A_{\mathrm{sp}} = O(1)\),
\(\frac1N\sum_{i=1}^N\|X_t^i-Y_t^i\|^3 \le A_{\mathrm{sp}} r_t^3\;
    (0\le t\le T)\).
This condition controls the cubic Taylor remainder in the particle positions by \(O(r_t^3)\);
see \Cref{lemma-directional-point-slot-error,eq-directional-point-slot-bound}.
In \Cref{ex:variance-cosine-curvature-recovery}, all paired errors coincide, giving
\(A_{\mathrm{sp}}=1\) (\Cref{lem:canonical-comparison-assumptions}(v)).
\end{assumption}

\begin{assumption}[Prescribed schedule and population-first Gaussian jumps]
\label{assumption-main-jump-time}
The common times $(t_\ell)_{\ell=1}^{\NofJ}$ and amplitudes $\eta_\ell\ge0$
are deterministic, with $\NofJ=O(1)$ in \Cref{theorem-main-r_t}.
The initial law \(\mu_0\) is deterministic; the common initial positions
\(X_0^i=Y_0^i\), \(i=1,\ldots,N\), are sampled independently from \(\mu_0\).
Population innovations are independent of that sample and fresh at each
jump. The population Gaussian fields and the empirical transitions satisfy
Assumption~\ref{assumption-common-seed}.
\end{assumption}

\subsection{Proof sketch}
\label{subsec:proof-sketch-after-assumptions}

We explain how the ODE and jump comparisons produce the response factors
in \Cref{def:curvature-responses}. The calculations below are schematic:
\(\simeq\) retains the leading linearized drift, and \(\lesssim\) suppresses
localized coefficients and Gaussian probability factors quantified in the appendix.

The reference particles are i.i.d. with law \(\mu_t\) conditional on
\(\mathcal G_{\rm pop}\) by
\Cref{lemma-gp-jump-implies-conditional-iid,lem:population-first-finite-coupling},
enabling concentration (background: \citet{sznitman1991,fournier2015rate}).
We use \(\Gamma_t^N\) and \(\widetilde\Gamma_t^N\) from
\Cref{subsec:general-argument-overview} for drift/Hessian fluctuations
and covariance-sampling error at jumps, respectively.

\maintextheading{ODE intervals (\Cref{subsec:general-argument-bound-r_t-ode-continuous}).}
Write \(\Delta_t:=\mathbf X_t-\mathbf Y_t\), with particle blocks indexed by \(i\),
so that \(r_t=N^{-1/2}\|\Delta_t\|_{\mathbb R^{dN}}\).
For \(r_t>0\), the chain rule and the ODEs in
\cref{eq:particle-flow,eq:reference-flow} give the equalities below. After adding and subtracting the reference empirical drift,
then linearizing and comparing Hessians
(\Cref{lemma-mean-value-theorem,lemma-bound-r_t-given-eta-eps}), we informally have
\[
 r_t\dot r_t=\frac12\frac{\mathrm d}{\mathrm dt}r_t^2
 =-\frac1N\sum_{i=1}^N\bigl\langle\Delta_t^i,
 \underbrace{\nabla_WF(\rho_{\mathbf X_t},X_t^i)-\nabla_WF(\mu_t,Y_t^i)}_{
 \simeq\,(\HessianMatrices_t^Y\Delta_t)^i\;\text{(Taylor)}}\bigr\rangle
 \simeq-\frac1N\Delta_t^\top\HessianMatrices_t^Y\Delta_t.
\]
By \Cref{def:curvature-responses}, the quadratic form is bounded above by
\(-\kappa_t^{Y,N}r_t^2\); dividing by \(r_t\) gives the curvature term
\(-\kappa_t^{Y,N}r_t\).
Restoring the errors, on the localized good events and while \(r_t\le R\),
\Cref{lemma-common-empirical-moment-residual-directional-rt},
\eqref{eq-rt-directional-stopped}, gives
\(\dot r_t\le-\kappa_t^{Y,N}r_t+c_DRr_t+\Gamma_t^N(1+r_t)\).
Here \(c_D\) is defined in \eqref{eq-Csp-definition}; the same lemma
also handles \(r_t=0\).

\maintextheading{Jump times (\Cref{subsec:general-argument-r_t-jump-errors}).}
Using the empirical and aligned population factors \(M_X,B_{\rm pop}\)
and shared Gaussian vector \(\mathbf g_\ell^N\) from
\Cref{subsec:auxiliary-particles}, we substitute the jump updates in
\cref{eq:particle-jump,eq:reference-jump} and apply the triangle inequality:
\[
 r_{t_\ell}
 =\frac1{\sqrt N}\bigl\|\Delta_{t_\ell-}
 +\eta_\ell(M_X-B_{\rm pop})\mathbf g_\ell^N\bigr\|_{\mathbb R^{dN}}
 \le r_{t_\ell-}
 +\frac{\eta_\ell}{\sqrt N}
 \bigl\|(M_X-B_{\rm pop})\mathbf g_\ell^N\bigr\|_{\mathbb R^{dN}}.
\]
Let \(M_Y\) be the empirical Hessian factor at the reference configuration
just before the jump. We split
\(M_X-B_{\rm pop}=(M_X-M_Y)+(M_Y-B_{\rm pop})\).
Smoothness controls the contribution of the first difference by a multiple
of \(r_{t_\ell-}\). For the remaining mismatch, Gaussian concentration and
the aligned coupling give schematically, with all time-\(t\) quantities
evaluated at \(t_\ell-\),
\[
 \frac{\|(M_Y-B_{\rm pop})\mathbf g_\ell^N\|_{\mathbb R^{dN}}}{\sqrt N}
 \lesssim
 \underbrace{
 \sqrt{\frac1N
 \|M_Y^2-B_{\rm pop}B_{\rm pop}^{\top}\|_{*;\,\mathbb R^{dN}\to\mathbb R^{dN}}}
 }_{\text{square root of covariance mismatch}}
 \lesssim \sqrt{\Gamma_t^N}+\sqrt{\widetilde\Gamma_t^N}.
\]
Here \(*\) denotes the nuclear norm. The \((i,j)\)-block of the covariance
mismatch is \(K_{\rho_{\mathbf Y_t}}(Y_t^i,Y_t^j)-K_{\mu_t}(Y_t^i,Y_t^j)\).
Using the kernel definition in \Cref{subsec:population-pwgf}, the difference
\(K_{\rho_{\mathbf Y_t}}(x,y)-K_{\mu_t}(x,y)\) consists schematically of
\[
 \underbrace{\text{Hessian approximation}}_{\text{as in the ODE step}}
 +\int\nabla_W^2F(\mu_t;x,z)\,
 \nabla_W^2F(\mu_t;z,y)\,
 (\rho_{\mathbf Y_t}-\mu_t)(dz).
\]
Expanding the difference of the two Hessian products bounds the first term
linearly in the Hessian approximation error, with localized coefficients suppressed.
The second term is an empirical integration error.
These comparisons are established in
\Cref{lemma-aligned-finite-square-root,lemma-kernel-mismatch-bound,lem:jump-bound-common-seed}.
Combining the two contributions yields
\(r_{t_\ell}\lesssim(1+\eta_\ell)r_{t_\ell-}
+\eta_\ell(\sqrt{\Gamma_{t_\ell-}^N}+\sqrt{\widetilde\Gamma_{t_\ell-}^N})\):
a jump amplifies past error and adds a new error.
The square-root dependence explains why an \(N^{-1/2}\) covariance-sampling
bound contributes \(N^{-1/4}\) to the jump estimate.
For background on Gaussian covariance alignment, see \citet{masarotto2019procrustes}.

\maintextheading{From local errors to response factors.}
Between jumps, the homogeneous equation
\(\dot r_t=-\kappa_t^{Y,N}r_t\) has solution
\(r_t=G_\kappa^Y(t,s)r_s\), where
\(G_\kappa^Y(t,s)=\exp(-\int_s^t\kappa_u^{Y,N}\,du)\).
Hence an error created at time \(s\) receives this weight at time \(t\).
Starting from \(r_0=0\), integrating the continuous forcing and summing
the jump forcing gives an integral and a sum, respectively.
Taking uniform residual bounds outside them yields schematically
\[
 \textstyle r_t\lesssim
 (\sup_{s\le T}\Gamma_s^N)
 \underbrace{\textstyle\int_0^tG_\kappa^Y(t,s)\,ds}_{\mathfrak C_t^Y}
 +(\sqrt{\sup_{s\le T}\Gamma_s^N}
       +\sqrt{\max_{\ell\le\NofJ}\widetilde\Gamma_{t_\ell-}^N})
 \underbrace{\textstyle\sum_{\ell:t_\ell\le t}\eta_\ell
                  G_\kappa^Y(t,t_\ell)}_{\mathfrak J_t^Y}.
\]
The maximum is zero if there are no jumps.
The recursive comparison in \Cref{lem:piecewise-gronwall-forced-jumps},
\cref{eq:piecewise-recursive-bound,eq:tracking-stopped-propagator-bound},
retains the extra nonlinear amplification and the products of jump multipliers.

\maintextheading{Residual bounds and closure.}
Conditioning only on \(\mathcal G_{\rm pop}\), moment localization and
weighted-feature covariance concentration give, with probability at least
\(1-p\) and \(\Xi=\Xi_{N,T,p}\) from \Cref{theorem-main-r_t},
\[
 \sup_{t\le T}\Gamma_t^N
 \le C_\varepsilon\Xi^\varepsilon(N^{-1/2}+T/N),\qquad
 \max_{\ell\le\NofJ}\widetilde\Gamma_{t_\ell-}^N
 \le C_\varepsilon\Xi^\varepsilon N^{-1/2};
\]
see \Cref{lemma-uniform-sup-gamma-bound-convenient,cor:tilde-gamma-jump-time-union-bound}.
Under the theorem's conditions,
\Cref{thm:fixed-schedule-high-prob-r-v15,cor:fixed-schedule-high-prob-r-rate-v15}
close the localized estimate and absorb the extra amplification into
\(C_\varepsilon\Xi^\varepsilon\).
\Cref{lem:functional-gap-fixed-time-v15,eq:functional-gap-fixed-time-v15}
give the objective bound.

%% file: sections/06_discussion.tex
\section{Conclusion}
\label{sec:conclusion}

We quantified finite-particle tracking for Hessian-guided PWGF under prescribed
common jump schedules. The coupling reduces Gaussian jump error to covariance
comparison, while sampled curvature propagates residuals into tracking and
objective bounds.

\maintextheading{Limitations \& future directions.}
Beyond the variance-plus-cosine example
(\Cref{subsec:canonical-variance-cosine}), deriving the controlled-curvature
path condition and non-spikiness from model primitives remains open,
as does synchronizing adaptive jump times.

\maintextheading{Formal verification and reproducibility.}
The accompanying Lean~4 development formalizes the stated fixed-schedule
and growing-horizon tracking results, the variance-plus-cosine example,
and the four local-geometry items in \Cref{prop:mf-local-geometry},
under their explicit hypotheses. The build and axiom audit
cover 4,274 registered declarations in 661 modules, including definitions and instances.
Claim-by-claim correspondence and solution-class boundaries are documented
separately from kernel checking. The formalization does not establish
unconditional WGF well-posedness, a deterministic characteristic representation
of arbitrary weak PDE solutions, or the numerical experiments.
Pinned environments, verification commands, source hashes, and experiment
records are provided in the public reproducibility repository:
\url{https://github.com/ryotaro-kawata-wa/pwgf-finite-sample}.

%% file: sections/ai_use_statement.tex
\subsection*{AI use statement}
ChatGPT models assisted with brainstorming, writing and editing, locating
candidate references, and checking mathematical arguments. This included
formulating and refining mathematical claims, detailed proof calculations,
revising author-written drafts, supplementing incomplete derivations,
exploring proof repairs, preparing LaTeX and verification materials,
editing schematic figures, designing the experimental setup,
developing code for the synthetic experiments, and creating experimental
figures. Astra and 6 Pro also assisted with developing the Lean~4 implementation
and running its compilation, kernel proof checking, and axiom audits.
These models are not components of the
optimization method studied in this paper. The authors have checked the proofs.
Responsibility for the final
statements and proofs remains with the authors.

%% file: appendix_final_general/general_argument_reorg_v1_6.tex
\section{A General Comparison Argument via the Auxiliary Particle System}
\label{sec:general-comparison-argument}

\input{appendix_final_general/notations}
\input{appendix_final_general/01_overview}

\input{appendix_final_general/general_bound_r_t_v1_1}
\input{appendix_final_general/general_bound_r_t_bound_including_jump_bounded_by_gamma}
\input{appendix_final_general/general_bound_r_t_bound_stochastic_errors}

\input{appendix_final_general/02_fixed_schedule_statement}

\input{appendix_final_general/04_functional_gap}
\input{appendix_final_general/canonical_variance_cosine}

\input{appendix_final_general/03_trigger_stability}

\input{appendix_final_general/kappa_concentration_on_saddle_point}

%% file: appendix_final_general/notations.tex
\paragraph{Notations.}
For nonnegative quantities \(a,b\), we write \(a\lesssim b\) if
\(a\le Cb\) for a constant \(C>0\) independent of the main asymptotic
parameters, such as the particle number \(N\), horizon \(T\), and
failure probability \(p\).  Equivalently, \(a=O(b)\) means \(a\lesssim b\).
We write \(a=o(b)\) if \(a/b\to0\) in the prescribed asymptotic regime.
Similarly, \(a=\Omega(b)\) means \(b=O(a)\), and \(a=\omega(b)\) means
\(b=o(a)\).  We write \(a=\Theta(b)\) if both \(a=O(b)\) and
\(a=\Omega(b)\) hold.  Finally, \(a=\widetilde O(b)\) hides
polylogarithmic factors in the main asymptotic parameters.  The hidden
constants may depend on fixed problem parameters, the dimension, and structural smoothness constants, but not on \(N,T\), or \(p\).

For normed spaces \(E,F\), \(\mathcal L(E;F)\) denotes the space of bounded
linear maps from \(E\) to \(F\), and \(\mathcal L(E_1,\ldots,E_k;F)\) denotes
the space of bounded \(k\)-linear maps from \(E_1\times\cdots\times E_k\) to
\(F\).  When \(E=F\), we write \(\mathcal L(E)\).  For a linear map
\(A:\mathbb R^d\to\mathbb R^d\), 
$
    \norm{A}_{\Loperator{d}}
    :=
    \sup_{\norm{v}_2=1}\norm{Av}_2
$
denotes its Euclidean operator norm.  We use \(\norm{\cdot}_{\HS}\) for the
Hilbert--Schmidt norm and \(\norm{\cdot}_{\F}\) for the Frobenius norm; in
finite-dimensional matrix spaces these coincide, but we reserve
\(\norm{\cdot}_{\HS}\) for Hilbert-space operators or tensor-valued feature
maps, and \(\norm{\cdot}_{\F}\) for ordinary finite matrices.

\paragraph{Vector fields, configurations, and empirical measures.}
For a measure \(\mu\in\Ptwo(\mathbb R^d)\) and vector fields
\(v,w:\mathbb R^d\to\mathbb R^d\), define
\[
 \|v\|_{L^2(\mu)}^2:=\int\|v(x)\|^2\,\mu(dx),
 \qquad
 \langle v,w\rangle_{L^2(\mu)}:=\int\langle v(x),w(x)\rangle\,\mu(dx).
\]
For a configuration \(\mathbf x=(x^1,\ldots,x^N)\in(\mathbb R^d)^N\), set
\[
 \|\mathbf x\|_N^2:=\frac1N\sum_{i=1}^N\|x^i\|^2,
 \qquad
 \rho_{\mathbf x}:=\frac1N\sum_{i=1}^N\delta_{x^i},
 \qquad
 \xi^N(\mathbf x):=(\xi(x^1),\ldots,\xi(x^N)).
\]
We abbreviate \(\rho_{X_t}:=\rho_{\mathbf X_t}\) and
\(\rho_{Y_t}:=\rho_{\mathbf Y_t}\). The coupled discrepancy is
\[
 \delta_{t,i}:=X_t^i-Y_t^i,
 \qquad
 r_t^2:=\frac1N\sum_{i=1}^N\|\delta_{t,i}\|^2
       =\|\mathbf X_t-\mathbf Y_t\|_N^2.
\]
The normalized configuration inner product is
\[
 \langle\mathbf u,\mathbf v\rangle_N
 :=\frac1N\sum_{i=1}^N\langle u^i,v^i\rangle.
\]

\paragraph{Gradient, Hessian, and jump covariance.}
The Wasserstein gradient is \(\dW F(\mu,x)\in\mathbb R^d\).
We distinguish its measure derivative from its spatial Jacobian:
\begin{align*}
 \HessW F(\mu;x,y)&:=\nabla_W^2F(\mu;x,y)\in\mathbb R^{d\times d},\\
 \xHessW F(\mu,x)&:=\nabla_x\nabla_WF(\mu,x)\in\mathbb R^{d\times d}.
\end{align*}
Use the following operators on their natural domains in
\(L^2(\mu;\mathbb R^d)\):
\begin{align*}
 (H_\mu v)(x)&:=\int\HessW F(\mu;x,y)v(y)\,\mu(dy),\\
 (H_\mu'v)(x)&:=\xHessW F(\mu,x)v(x),
 &\mathcal H_\mu&:=H_\mu+H_\mu'.
\end{align*}
In particular, the spatial multiplication operator has domain
\[
 \mathcal D(H_\mu')
 :=\{v\in L^2(\mu;\mathbb R^d):
       \xHessW F(\mu,\cdot)v(\cdot)\in L^2(\mu;\mathbb R^d)\},
 \qquad
 \mathcal D(\mathcal H_\mu)=\mathcal D(H_\mu)\cap\mathcal D(H_\mu').
\]
Here \(\mathcal D(H_\mu)\) consists of the fields for which the displayed
kernel integral exists and defines an \(L^2(\mu)\) field.
For $v\in\mathcal D(\mathcal H_\mu)$ for which the pushforward perturbation and
the indicated derivatives are defined, the full second-variation identity is
\[
 \frac{d^2}{dh^2}\Big|_{h=0}F((\Id+hv)_\#\mu)
 =\langle v,\mathcal H_\mu v\rangle_{L^2(\mu)}.
\]
The tracking curvature retains both terms of \(\mathcal H_\mu\).
The additional spatial-Hessian bound in
\cref{prop:finite-feature-negative-curvature} makes \(H_\mu'\) bounded;
no such global bound is imposed here.
In contrast, the Hessian-guided jump covariance uses the integral term:
\[
 K_\mu(x,y):=\int\HessW F(\mu;x,z)\HessW F(\mu;z,y)\,\mu(dz),
 \qquad \xi_\mu\sim\GP(0,K_\mu).
\]
Here \(\GP(0,K_\mu)\) denotes a centered Gaussian process with covariance
kernel \(K_\mu\). The finite Gaussian coordinates used to compare two
jumps are constructed from the population field in
\cref{lem:population-first-finite-coupling}.
For a matrix or trace-class operator \(A\), we use the nuclear norm
\[
 \|A\|_*:=\operatorname{tr}((A^*A)^{1/2}),
\]
where \(A^*\) denotes the Hilbert adjoint and equals \(A^\top\) for real matrices.

\paragraph{Conventions for the fixed-schedule comparison.}
The jump count is \(\NofJ\). We take
\(0<t_1<\cdots<t_{\NofJ}\le T\), put \(t_0=0\) and
\(t_{\NofJ+1}=T\), and allow \(\NofJ=0\).
Paths are right-continuous: \(Z_{t_\ell}\) is the post-jump value and
\(Z_{t_\ell-}=\lim_{s\uparrow t_\ell}Z_s\) is the pre-jump value.
Jump covariances and jump residuals are evaluated at \(t_\ell-\).
Empty sums and products are zero and one, respectively.

\paragraph{Population conditioning and comparison history.}
The population initial law and schedule are deterministic. Population
Gaussian innovations are independent of the iid initial sample.
Let \(\mathcal G_\ell\) be the sigma-field generated by the population
fields through jump \(\ell\), and let
\(\mathcal G_{\rm pop}=\mathcal G_{\NofJ}\).
The population path satisfies
\(\mu_t=\Law(Y_t^1\mid\mathcal G_{\rm pop})\);
the conditional product law is proved in
\cref{lemma-gp-jump-implies-conditional-iid}.
The notation \(\mathcal H_{\ell-}\) denotes the full pre-jump history,
including \(X_{t_\ell-},Y_{t_\ell-},\mu_{t_\ell-}\), all previous
population innovations and auxiliary normals, and the initial sample.
It does not include the current or future innovations. Finite Gaussian comparison
coordinates are standard Gaussian conditionally on this history;
they are not the conditioning object in the product law for \(Y\).

%% file: appendix_final_general/01_overview.tex
\subsection{Overview of the finite-particle comparison}
\label{subsec:general-argument-overview}

This appendix expands the proof architecture summarized in
\cref{sec:proof-overview}, especially the
four-step sketch in \cref{subsec:proof-sketch-after-assumptions}.  The population PWGF mechanism itself is treated
as the reference dynamics, in the sense of the Hessian-guided construction of
\citet{yamamoto2025hessianguided}.  The purpose of the present appendix is the
finite-particle empirical error analysis: we quantify how the implementable
particle system tracks the prescribed mean-field PWGF path.

\begin{table}[t]
\centering
\footnotesize
\setlength{\tabcolsep}{3pt}
\renewcommand{\arraystretch}{1.12}
\begin{tabular}{>{\raggedright\arraybackslash}p{0.19\linewidth}
                >{\raggedright\arraybackslash}p{0.22\linewidth}
                >{\raggedright\arraybackslash}p{0.23\linewidth}
                >{\raggedright\arraybackslash}p{0.25\linewidth}}
\toprule
Condition or conclusion & General comparison & Variance--cosine example & Matrix-factorization case study \\
\midrule
Moment structure and weighted smoothness
& Assumed in \cref{assumption-structured-moment-smoothness}
& Verified in \cref{lem:canonical-comparison-assumptions}(i)
& Separate product-model conditions in \cref{assumption:mf-model-restrictions} \\
Gaussian jumps and schedule
& Prescribed by \cref{assumption-common-seed}
& One prescribed common jump, verified in \cref{lem:canonical-comparison-assumptions}(ii)
& --- \\
Moment growth and ODE regularity
& Assumed in \cref{assumption-moment-growth,assumption-uniform-concentration-piecewise-lipschitz}
& Verified in \cref{lem:canonical-comparison-assumptions}(iii)--(iv)
& --- \\
Non-spikiness
& Assumed in \cref{assumption-anti-spikiness-event}
& Verified with $A_{\rm sp}=1$ in \cref{lem:canonical-comparison-assumptions}(v)
& --- \\
Curvature-response mechanism
& Nonlinear stability in \cref{thm:fixed-schedule-high-prob-r-v15}; \(N^\upsilon\)-scaled curvature-response bounds in \cref{cor:fixed-schedule-high-prob-r-rate-v15}
& High-probability response bounds in \cref{lem:canonical-response-control}; \cref{cor:fixed-schedule-high-prob-r-rate-v15} applies with \(\upsilon=\zeta=1/16\)
& Population-level local geometry in \cref{prop:mf-local-geometry}: attraction, transverse descent, and positive spherical product pushforward second variation in two regions \\
Finite-particle conclusion
& Conditional tracking and objective comparison
& Growing-horizon conclusion in \cref{prop:canonical-model-assumptions}
& Open \\
\bottomrule
\end{tabular}
\caption{Relation of the model sections to the comparison theory. The
matrix-factorization column records a separate population curvature-response
case study; --- denotes items outside its scope, and ``Open'' denotes the bridge
to finite-particle tracking.}
\label{tab:assumption-evidence-map}
\end{table}

Classical globally Lipschitz propagation-of-chaos arguments close the
reference-particle stability estimate by Gronwall with a worst-case Lipschitz
constant~\citep{sznitman1991}.
Here the same architecture is driven instead by the curvature propagator
associated with the sampled curvature.

The comparison is organized through the auxiliary particle system
$\bs Y_t=(Y_t^1,\dots,Y_t^N)$:
\[
    \rho_{\bs X_t}
    \quad\longleftrightarrow\quad
    \rho_{\bs Y_t}
    \quad\longleftrightarrow\quad
    \mu_t .
\]
Here $\bs X_t=(X_t^1,\dots,X_t^N)$ is the actual empirical system, whose drift
and jump covariance are evaluated at $\rho_{\bs X_t}$; $\mu_t$ is the population
PWGF path; and $\bs Y_t$ starts from the same initial sample and uses the
population drift and the original population Gaussian field along $\mu_t$.
Procrustes/Wasserstein comparisons of covariance factors provide related
mathematical background \citep{masarotto2019procrustes}. The population-first
conditional construction required here is proved in
\cref{lem:population-first-finite-coupling}: it constructs the empirical
Gaussian coordinates jointly with the field
evaluations, preserving both marginal updates. The reference conditional
product law conditions only on the population fields. Thus the comparison
separates two effects: the pathwise stability of $\bs X_t$ relative to
$\bs Y_t$, and the concentration of the conditionally i.i.d. reference cloud
$\bs Y_t$ around $\mu_t$.

\paragraph{How this appendix matches the main proof sketch.}
The detailed proof follows the same four pieces as
\cref{subsec:proof-sketch-after-assumptions}.
First, \cref{subsec:general-argument-bound-settings} defines the common initial coupling, the common-seed Gaussian
jump construction, the moment-Lipschitz smoothness assumptions, and the
localization event.  Second,
\cref{subsec:general-argument-bound-r_t-ode-continuous} proves the continuous-time contraction-versus-residual bound for
\[
    r_t
    \coloneq
    \frac1{\sqrt N}\norm{\bs X_t-\bs Y_t}_2 .
\]
Third, \cref{subsec:general-argument-r_t-jump-errors} controls the perturbation mismatch at prescribed jump times.  Fourth,
\cref{subsec:general-argument-bound-stochastic-gamma-and-tilde-gamma}, together with
\cref{subsubsec:general-argument-bound-gamma} and
\cref{subsubsec:weighted-factorization-tilde-gamma}, supplies the concentration estimates needed to
close the deterministic inequality.

\paragraph{Core theorem: prescribed common jump schedule.}
The core result of this appendix is the fixed-schedule comparison in
\cref{subsec:fixed-schedule-package}.  In this
layer the population and empirical systems are compared along a prescribed
common jump schedule
\(
0=t_0<t_1<\cdots<t_{\NofJ}\le T
\),

with $\NofJ=O(1)$ in
\Cref{cor:fixed-schedule-high-prob-r-rate-v15}. On the population moment
and sample localization events, and under anti-spikiness, the ODE estimate is
\[
\dot r_t\le-\kappa_t^{Y,N}r_t+c_Dr_t^2+\Gamma_t^Nr_t+\Gamma_t^N.
\]
At each prescribed jump, the comparison uses only pre-jump residuals:
\[
r_{t_\ell}\le(1+a_\ell)r_{t_\ell-}
+a_\ell(\sqrt{\Gamma_{t_\ell-}^N}+\sqrt{\widetilde\Gamma_{t_\ell-}^N}),
\qquad a_\ell=c_J(D)q_p\eta_\ell.
\]
The finite Gaussian coupling is constructed after generating the population
field; the reference product law conditions only on the population fields.
The exact result in \Cref{thm:fixed-schedule-high-prob-r-v15} retains the
nonlinear condition $T(\delta+c_DR)\le1/4$. Its failure probability is bounded
within that nonlinear stability event and the regularity event. The sufficient horizon
$T\le N^{1/4-\upsilon-\zeta}$ in \cref{cor:fixed-schedule-high-prob-r-rate-v15}
also specifies the curvature-response bounds and uniform structural constants it uses.

\paragraph{The two stochastic residuals.}
The continuous residual is
\[
    \Gamma_t^N
    \coloneq
    \sum_{m\ge1}\gamma_m\Delta_{m,t}^N,
    \qquad
    \Delta_{m,t}^N
    \coloneq
    \norm{M_m(\rho_{\bs Y_t})-M_m(\mu_t)} .
\]
It controls the population-to-reference empirical error in the drift, Hessian,
and functional terms.  The jump covariance residual is
\[
    \widetilde\Gamma_t^N
    \coloneq
    \frac1N
    \left\|
    \int
    \mathbf U_t(z)\mathbf U_t(z)^\top
    (\mu_t-\rho_{\bs Y_t})(\dd z)
    \right\|_* ,
\]
where
\[
    \mathbf U_t(z)
    \coloneq
    \bigl(
    \HessW F(\mu_t;Y_t^1,z),\ldots,
    \HessW F(\mu_t;Y_t^N,z)
    \bigr)^\top .
\]
The first residual enters additively along ODE intervals, whereas the second
enters through a square-root covariance perturbation at jumps.  Consequently,
even when both residuals concentrate at nearly $N^{-1/2}$ scale, the jump
contribution generally produces an $N^{-1/4}$ bottleneck.
\Cref{subsubsec:general-argument-bound-gamma} and
\cref{subsubsec:weighted-factorization-tilde-gamma} supply the concentration
estimates for $\Gamma_t^N$ and $\widetilde\Gamma_t^N$, respectively.

\paragraph{From particle tracking to functional error.}
After the fixed-schedule $r_t$ estimate is proved,
\cref{subsec:functional-gap} converts the
tracking bound into a bound on
\[
    \bigl|F(\rho_{\bs X_t})-F(\mu_t)\bigr| .
\]
The key point is that no new stochastic object is introduced: the functional gap
is bounded by the same pair $(r_t,\Gamma_t^N)$. The common weights
\[
 \gamma_m(D)=\mathrm L_{K_{\rm pop},D}a_m
\]
from \cref{lemma-localized-moment-smoothness-consequences} control both
the dynamical comparison and the functional comparison.

\paragraph{An example on a growing horizon.}
\Cref{prop:canonical-model-assumptions} treats a variance-plus-cosine model
with bounded mean-zero initialization and one prescribed Gaussian jump.
Its proof computes the sampled curvature, bounds the probability of
escape, and checks finite-$N$ stability. After positive curvature returns,
the jump curvature-response factor decays, yielding tracking and a bound on the objective
gap to the minimum for $C\log N\le t\le N^{1/8}$.

\paragraph{Fixed-time supplements.}
\Cref{lem:fixed-time-trigger-squared-gap} compares
$\mathcal G_t=\int\|\nabla_WF(\mu_t,x)\|^2\,\mu_t(dx)$ and
$\mathcal G_t^N=N^{-1}\sum_i\|\nabla_WF(\rho_{\mathbf X_t},X_t^i)\|^2$
at one deterministic non-jump time. The margin condition
\eqref{eq:trigger-margin-agreement} gives agreement of threshold decisions
at that time. Separately,
\cref{prop:finite-feature-negative-curvature} transfers negative
population curvature to the reference sample for finite-feature
objectives under a spatial-Hessian condition.
These results are not inputs to the tracking proof and do not control
whole-path curvature-response factors or synchronize adaptive jump times.

%% file: appendix_final_general/general_bound_r_t_v1_1.tex
\subsection{Settings and Assumptions}
\label{subsec:general-argument-bound-settings}

Before defining the estimates, we fix the coupled population-reference-empirical
dynamics and the common jump schedule used throughout the comparison.

\begin{definition}
\label{definition-ode-and-jump}
Let $U_1,\ldots,U_N$ be i.i.d. with deterministic law $\mu_0$, independently of
the population innovations, and set $X_0^i=Y_0^i=U_i$. Fix deterministic jump
times $0<t_1<\cdots<t_{\NofJ}\le T$ and amplitudes $\eta_\ell\ge0$.
Write $t_0=0$ and $t_{\NofJ+1}=T$; zero jumps are allowed.
On each nonempty interval $(t_\ell,t_{\ell+1})$, $0\le\ell\le\NofJ$, let
\[
\partial_tX_t^i=-\nabla_WF(\rho_{\mathbf X_t},X_t^i),\qquad
\partial_tY_t^i=-\nabla_WF(\mu_t,Y_t^i),\qquad
\partial_t\mu_t=\operatorname{div}(\mu_t\nabla_WF(\mu_t)).
\]
The paths are right-continuous, and $\mathbf X_{t_\ell-},\mathbf Y_{t_\ell-}$
denote their actual left limits. At a jump,
\begin{equation}
\label{eq:jump-dynamics-XY}
\mathbf X_{t_\ell}=\mathbf X_{t_\ell-}+\eta_\ell\xi_{\ell,X}^N(\mathbf X_{t_\ell-}),
\qquad
\mathbf Y_{t_\ell}=\mathbf Y_{t_\ell-}+\eta_\ell\xi_{\ell,Y}^N(\mathbf Y_{t_\ell-}),
\end{equation}
where $\xi_{\ell,Y}^N$ evaluates the population field with covariance
$K_{\mu_{t_\ell-}}$, and the empirical perturbation has the transition specified
in Assumption~\ref{assumption-common-seed}. The population law is pushed forward
by the same population field. Fresh innovations are independent across jumps;
the resulting fields can have covariances depending on earlier population
randomness. Their finite coupling is constructed in
Lemma~\ref{lem:population-first-finite-coupling}.
We assume that the population path admits the following characteristic
realization: conditional on
the population innovations, the ODE and jump maps compose into a common
measurable map \(\mathcal T_t\) (and the corresponding left-trace maps) such that
\[
Y_t^i=\mathcal T_t(U_i),\qquad
\mu_t=(\mathcal T_t)_\#\mu_0.
\]
On each ODE branch, \(\mathcal T_t\) is generated by the population
characteristic ODE; at a jump, it includes the population pushforward stated
above. This is the realization used in
\cref{lem:population-first-finite-coupling,lemma-gp-jump-implies-conditional-iid}.
\end{definition}

\input{appendix_final_general/general_assumptions}

In the local comparison estimates below we work on $E_{\rm pop}$ and use the deterministic conditional moment scale and coefficient convention from the assumptions. Spatial localization and anti-spikiness are imposed where stated.

\subsection{Bound $r_t = \frac{1}{\sqrt{N}}\norm{\bs{X}_t-\bs{Y}_t}_{2}$ between jumps}
\label{subsec:general-argument-bound-r_t-ode-continuous}

Throughout this subsection, use $\mathrm L_D=\mathrm L_{K_{\rm pop},D}$
and $\gamma_m(D)=\mathrm L_Da_m$ from
\cref{lemma-localized-moment-smoothness-consequences}. We suppress the
argument $D$ of $\gamma_m$ when the localization radius is fixed.

On each ODE interval, the discrepancy between \(\bs X_t\) and \(\bs Y_t\) is read through the segment joining the two configurations.

For $s\in[0,1]$, define the interpolating particle configuration
\begin{equation}
    \bs Z_t^{(s)} \coloneq (1-s)\bs Y_t + s\bs X_t \in \R^{dN},\quad (Z_t^{(s)})^i \coloneq (1-s)Y_t^i + sX_t^i 
\end{equation}

The linearization of the empirical vector field produces the following block Hessian matrix on \((\mathbb R^d)^N\).

For $\mu \in \mathcal{P}(\R^d)$ and $\bs{v} \in (\R^{d})^{N}$, define a block matrix
\begin{equation}
    \HessianMatrices(\mu,\bs{v})\coloneq
    \left\{
    \frac{1}{N}
    \nabla_W^2 F\!\left(\mu,v_i,v_j\right)
    + \delta_{ij}\nabla_x\nabla_W F\!\left(\mu,v_i\right)\in \R^{d\times d}\right\}_{i,j=1}^N.
\end{equation}
For each $(t,s)$, define a $\R^{dN} \times \R^{dN}$ block matrix
\begin{align}
    \HessianMatrices_t^{(s)}
    \coloneq&
    \HessianMatrices(\rho_{\bs Z_t^{(s)}},\bs{Z}_t^{(s)}).
\end{align}
For each $t\ge 0$, define the population-sampled Hessian matrix along the coupled
infinite-particle trajectory by
\begin{equation}
    \HessianMatrices_t^{Y}
    \coloneq
    \HessianMatrices(\mu_t,\bs{Y}_t).
\end{equation}

The smallest eigenvalue of the population-sampled block Hessian is the local curvature quantity that drives contraction or expansion of the coupling error.

Let
\begin{equation}
    \kappa_t^{Y,N}
    \coloneq
    \lambda_{\min}\!\bigl(\HessianMatrices_t^{Y}\bigr).
\end{equation}

\paragraph{Bound the deterministic dynamics}

The first lemma identifies the exact Jacobian of the empirical vector field, so that the particlewise difference can be expressed through a block Hessian.

\begin{lemma}[Mean-value theorem for the empirical vector field]
\label{lemma-mean-value-theorem}
Let $V_N(\bs X)
:=
\bigl(V(\rho_{\bs X},X_i)\bigr)_{i=1}^N,\;
\rho_{\bs X}
:=
\frac1N\sum_{k=1}^N \delta_{X_k}$,
where \(V:\mathcal P_2(\R^d)\times \R^d\to\R^d\) is \(C^1\) in the spatial variable and Wasserstein differentiable in the measure variable, with continuous derivatives \(\nabla_x V\) and \(\nabla_W V\).
For \(\bs X_t,\bs Y_t\in(\R^d)^N\), define
$\Delta_t:=\bs X_t-\bs Y_t,\;
\bs Z_t^{(s)}:=\bs Y_t+s\Delta_t.
$
Then
\[
V_N(\bs X_t)-V_N(\bs Y_t)
=
\int_0^1 DV_N(\bs Z_t^{(s)})\,\Delta_t\,\dd s,
\]
where \(DV_N(\bs Z)\in\R^{Nd\times Nd}\) is the block matrix whose \((i,j)\)-block is given by
\[
\bigl[DV_N(\bs Z)\bigr]_{ij}
=
\delta_{ij}\,\nabla_x V(\rho_{\bs Z},Z_i)
+
\frac1N\,\nabla_W V(\rho_{\bs Z},Z_i)(Z_j).
\]
For the moment objective \(F\) in \cref{def:moment-objective}, the same
mean-value identity holds for
\(V_N(\bs X)=(-\nabla_WF(\rho_{\bs X},X_i))_{i=1}^N\), with $\bigl[DV_N(\bs Z)\bigr]_{ij}
=
-\delta_{ij}\,\nabla_x\nabla_W F(\rho_{\bs Z},Z_i)
-\frac1N\,\nabla_W^2F(\rho_{\bs Z},Z_i,Z_j).$
\end{lemma}
\begin{proof}
\noindent\textbf{Step 1: integrate the configuration derivative.}
Fix \(t\) and define $ \Phi(s):=V_N(\bs Z_t^{(s)})\in(\R^d)^N.$
Since \(V_N\) is \(C^1\) on \((\R^d)^N\), the fundamental theorem of calculus gives
\[
V_N(\bs X_t)-V_N(\bs Y_t)
=
\Phi(1)-\Phi(0)
=
\int_0^1 \Phi'(s)\,\dd s
=
\int_0^1 DV_N(\bs Z_t^{(s)})\,\Delta_t\,\dd s.
\]
\par\noindent\textbf{Step 2: compute the spatial and measure derivatives.}
It remains to identify the Jacobian \(DV_N\).
Assume that $V(\mu,x)$ is $C^1$ in $x$ and differentiable in $\mu$ in the Wasserstein sense, with derivative denoted by $\nabla_W V(\mu,x)(\cdot)$.
Then, by the chain rule for empirical measures,
\[
\partial_{X_j}\bigl[V(\rho_{\bs X},x)\bigr]
=
\frac1N \nabla_W V(\rho_{\bs X},x)(X_j).
\]
Applying the usual chain rule to $V(\rho_{\bs X},X_i)$ yields
\[
\bigl[DV_N(\bs X)\bigr]_{ij} = \partial_{X_j}V(\rho_{\bs X},X_i)
=
\delta_{ij}\,\nabla_x V(\rho_{\bs X},X_i)
+
\frac1N\,\nabla_W V(\rho_{\bs X},X_i)(X_j).
\]
Finally, if \(V=-\nabla_W F\), then $\nabla_x V=-\nabla_x\nabla_W F,
\;
\nabla_W V=-\nabla_W^2F,$
which yields the stated block formula.
For the moment objective in \cref{def:moment-objective}, every empirical
law lies in its domain. The identity
\(M(\rho_{\bs X})=N^{-1}\sum_k J_0(X_k)\),
\cref{lem:localized-weighted-feature-envelopes}, and
\eqref{eq:weighted-gradient-representation} give a \(C^1\) empirical
field \(V_N\). Its block derivative follows directly as below. Moving only
particle $j$ in direction $h$ gives
\[
 \left.\frac d{d\varepsilon}\right|_0
 M_m\left(\frac1N\sum_{k\ne j}\delta_{X_k}
              +\frac1N\delta_{X_j+\varepsilon h}\right)
 =\frac1N\fD\phi_m(X_j)h.
\]
In the weighted space, the corresponding derivative of $M(\rho_{\bs X})$
is $N^{-1}\fD J_0(X_j)h$ by
\cref{lem:localized-weighted-feature-envelopes}. Keep the evaluation point
$x$ fixed and pair with $v\in\mathbb R^d$ in
\eqref{eq:weighted-gradient-representation}:
\[
\begin{aligned}
 \partial_{X_j}\langle\nabla_WF(\rho_{\bs X},x),v\rangle_{\mathbb R^d}[h]
 &=\fD^2G(M(\rho_{\bs X}))[N^{-1}\fD J_0(X_j)h,\fD J_0(x)v]\\
 &=N^{-1}\langle v,\nabla_W^2F(\rho_{\bs X};x,X_j)h\rangle_{\mathbb R^d}.
\end{aligned}
\]
The last equality uses symmetry of $\fD^2G$ and
\eqref{eq:weighted-integral-hessian-representation}. Since $v$ is arbitrary, set $x=X_i$ to obtain
the measure contribution $N^{-1}\nabla_W^2F(\rho_{\bs X};X_i,X_j)h$.
Moving the evaluation point contributes
$\delta_{ij}\nabla_x\nabla_WF(\rho_{\bs X},X_i)h$.
Both contributions change sign when $V=-\nabla_WF$.
\end{proof}
Using this linearization, the next lemma isolates the three forces in the ODE error: sampled curvature, Hessian mismatch, and empirical drift residual.

\begin{lemma}[Upper bound of $\dot{r}_t$ under the ODE flow given error terms $\iota_t^N$ and $\varepsilon_t^N$.]
\label{lemma-bound-r_t-given-eta-eps}
Let $V(\mu,x)\coloneq -\nabla_W F(\mu,x)$.
Consider the coupled systems
\begin{equation}
    \frac{\dd}{\dd t}X_t^i = V(\rho_{\bs X_t},X_t^i),
    \qquad
    \frac{\dd}{\dd t}Y_t^i = V(\mu_t,Y_t^i).
\end{equation}
Define
\begin{align}
    &\Delta_t \coloneq \bs X_t-\bs Y_t \in (\R^{d})^N,
    \qquad
    r_t \coloneq \frac{1}{\sqrt N}\norm{\Delta_t}_2,\\
    &\label{eq:ode-hessian-discrepancy-definition}
    \iota_t^N \coloneq \frac{1}{ N} \sup_{s\in [0,1]} |\Delta_t^\top (\HessianMatrices_t^{(s)} - \HessianMatrices_t^Y) \Delta_t|,\quad 
    \varepsilon_t^N
    \coloneq
    \left(
        \frac{1}{N}\sum_{i=1}^N
        \bigl|V(\rho_{\bs Y_t},Y_t^i)-V(\mu_t,Y_t^i)\bigr|^2
    \right)^{1/2}.
\end{align}
Assume that the map
\(
\bs x\mapsto (V(\rho_{\bs x},x^i))_{i=1}^N
\)
is differentiable along the segment joining $\bs X_t$ and $\bs Y_t$.
Then, at non-jump times with $r_t>0$,
\begin{equation}
    \dot r_t
    \le
    - \kappa_t^{Y,N} r_t
    + r_t^{-1} \iota_t^N 
    + \varepsilon_t^N.
\end{equation}
\end{lemma}

\begin{proof}
\noindent\textbf{Step 1: differentiate and split the squared discrepancy.}

First,
\begin{equation}
    \frac{1}{2}\frac{\dd}{\dd t}r_t^2
    =
    \frac{1}{N}
    \innerProd{\Delta_t}{V_N(\bs X_t)-V_\infty(\bs Y_t)},
\end{equation}
where
\begin{equation}
    V_N(\bs X_t)
    \coloneq
    \bigl(V(\rho_{\bs X_t},X_t^i)\bigr)_{i=1}^N,
    \qquad
    V_\infty(\bs Y_t)
    \coloneq
    \bigl(V(\mu_t,Y_t^i)\bigr)_{i=1}^N.
\end{equation}
We decompose
\begin{align}
    \frac{1}{2}\frac{\dd}{\dd t}r_t^2
    =&
    \underbrace{
        \frac{1}{N}
        \innerProd{\Delta_t}{V_N(\bs X_t)-V_N(\bs Y_t)}
    }_{\mathrm{(A)}}
    +
    \underbrace{
        \frac{1}{N}
        \innerProd{\Delta_t}{V_N(\bs Y_t)-V_\infty(\bs Y_t)}
    }_{\mathrm{(B)}}.
\end{align}

\par\noindent\textbf{Step 2: isolate sampled curvature and the Hessian error.}
For \(\mathrm{(A)}\), by the mean-value theorem (\cref{lemma-mean-value-theorem}),
\begin{equation}
    V_N(\bs X_t)-V_N(\bs Y_t)
    =
    -\int_0^1 \HessianMatrices_t^{(s)} \Delta_t\,\dd s.
\end{equation}
Hence
\begin{align}
    \mathrm{(A)}
    &=
    -\int_0^1
    \frac{1}{N}\Delta_t^\top \HessianMatrices_t^{(s)}\Delta_t\,\dd s \\
    &=
    -\int_0^1
    \frac{1}{N}\Delta_t^\top \HessianMatrices_t^Y \Delta_t\,\dd s
    -
    \int_0^1
    \frac{1}{N}\Delta_t^\top
    \bigl(\HessianMatrices_t^{(s)}-\HessianMatrices_t^Y\bigr)
    \Delta_t\,\dd s.
\end{align}
By the definition of \(\kappa_t\),
\begin{equation}
 \begin{aligned}
    -\frac{1}{N}\Delta_t^\top \HessianMatrices_t^{Y}\Delta_t
    &=-\innerProd{\frac{1}{\sqrt{N}}\Delta_t}{\HessianMatrices_t^Y \frac{1}{\sqrt{N}}\Delta_t}_2\\
    &\le-\kappa_t^{Y,N} \norm{\frac{1}{\sqrt{N}}\Delta_t}_{\mathbb R^{dN}}^2
     =-\kappa_t^{Y,N}r_t^2\qquad(\forall s\in[0,1]),
 \end{aligned}
\end{equation}
and                                     
\begin{equation}
    \left|\frac{1}{N}\Delta_t^\top
    \bigl(\HessianMatrices_t^{(s)}- \HessianMatrices_t^{Y}\bigr)\Delta_t
    \right|
    \le
    \iota_t^N.
\end{equation}
Therefore,
\begin{equation}
    \mathrm{(A)}
    \le
    -\kappa_t^{Y,N} r_t^2 
    + \iota_t^N.
\end{equation}

\par\noindent\textbf{Step 3: bound the drift residual and divide for positive error.}
For \(\mathrm{(B)}\), by the Cauchy--Schwarz inequality,
\begin{align}
    \mathrm{(B)}
    &\le
    \left(
        \frac{1}{N}\norm{\Delta_t}_2^2
    \right)^{1/2}
    \left(
        \frac{1}{N}
        \sum_{i=1}^N
        \bigl|
            V(\rho_{\bs Y_t},Y_t^i)-V(\mu_t,Y_t^i)
        \bigr|^2
    \right)^{1/2} \\
    &=
    r_t\varepsilon_t^N.
\end{align}

Combining the estimates for \(\mathrm{(A)}\) and \(\mathrm{(B)}\), we obtain
\begin{equation}
    \frac{1}{2}\frac{\dd}{\dd t}r_t^2
    \le
    -\kappa_t^{Y,N} r_t^2 
    + \iota_t^N
    + r_t\varepsilon_t^N.
\end{equation}
Hence, whenever \(r_t>0\),
\begin{equation}
    \dot r_t
    \le
    -\kappa_t^{Y,N} r_t 
    + \iota_t^N r_t^{-1}
    + \varepsilon_t^N.
\end{equation}
At $r_t=0$ the quotient is not used; the norm derivative is treated directly in Lemma~\ref{lemma-common-empirical-moment-residual-directional-rt}.
\end{proof}

\paragraph{Bound the error terms $\iota_t^N$ and $\varepsilon_t^N$ by moment errors.}

We will repeatedly convert block-operator estimates into scalar matrix estimates by bounding each block norm separately.

\begin{lemma}[operator norm bound of blocked matrices.]
\label{lemma-operator-norm-block-matrix}
    Let 
    \begin{equation}
        A = \begin{bmatrix}
            B_{1,1} & \dots& B_{1,N}\\
            \vdots &\ddots & \vdots \\
            B_{N,1}& \dots &B_{N,N}
        \end{bmatrix},
        B_{i,j} \in \R^{d\times d}
    \end{equation}
    be a $\R^{dN}\times \R^{dN}$ block matrix and
    \begin{equation}
        C = (\norm{B_{i,j}}_{\mathbb R^d\to\mathbb R^d})_{i,j=1}^N
    \end{equation}
    be a corresponding $\R^{N}\times \R^{N}$ matrix.
    then,
    \begin{equation}
        \norm{A}_{\mathbb R^{dN}\to\mathbb R^{dN}} \leq \norm{C}_{\mathbb R^{N}\to\mathbb R^{N}}
    \end{equation}
\end{lemma}
\begin{proof}
    For any vector $x\in \R^{dN}$ such that $\norm{x}_2=1$, we write $x = (x_1,\dots,x_N)$, $x_i\in \R^d$.
    First note that
    \begin{equation}
        1 = \norm{x}_2^2 = \sum_i (\sum_j x^2_{i,j}) = \sum_i z_i^2 = \norm{z}_2^2,\quad z \coloneq (\norm{x_1}_2,\dots,\norm{x_N}_2) \in \R^N.
    \end{equation}
    Next, we define $[Ax]_i \coloneq ((Ax)_j)_{j=(i-1)d+1}^{id}$, $i=1,\dots,N$. For all $i$, $[Ax]_i = \sum_{j=1}^N B_{ij}x_j.$
    The norm is bounded as $\norm{[Ax]_i}_2\leq \sum_j \norm{B_{ij}x_j}_{\mathbb R^d} \leq \sum_j \norm{B_{i,j}}_{\mathbb R^d\to\mathbb R^d}\norm{x_j}_{\mathbb R^d}.$
    Therefore,
\begin{equation}
    \|Ax\|_2^2
    =
    \sum_{i=1}^N\|[Ax]_i\|_2^2 
    \le
    \sum_{i=1}^N (Cz)_i^2
    =
    \|Cz\|_2^2 
    \le
    \|C\|_{\mathbb R^{N}\to\mathbb R^{N}}^2\|z\|_2^2
    =
    \|C\|_{\mathbb R^{N}\to\mathbb R^{N}}^2\|x\|_2^2.
\end{equation}
Taking square roots and then the supremum over \(x\neq0\) proves the claim.
\end{proof}

The first Hessian mismatch comes from replacing the empirical measure
\(\rho_{\bs Z_t^{(s)}}\) by the population law \(\mu_t\) while keeping the particle locations fixed.

\begin{lemma}[Directional measure-slot error]
\label{lemma-directional-measure-slot-error}
Use the common weights $\gamma_m(D)=\mathrm L_Da_m$ from
\cref{lemma-localized-moment-smoothness-consequences}, and define
\begin{align}
 \widetilde\HessianMatrices_t^{(s)}
 &:=\HessianMatrices(\mu_t,\bs Z_t^{(s)}),\notag\\
 \Delta_{m,t}^N&:=\|M_m(\rho_{\bs Y_t})-M_m(\mu_t)\|_{\mathsf H_m},\notag\\
 \Gamma_t^N(D)&:=\sum_{m\ge1}\gamma_m(D)\Delta_{m,t}^N,\qquad
 \Lambda(D):=\sum_{m\ge1}\gamma_m(D)L_m(1+2D^{m-1}).
 \label{eq:common-moment-residual-and-lipschitz-sum}
\end{align}
On $E_{\rm pop}\cap E_{T,D}$, for every $s\in[0,1]$,
\begin{equation}
\label{eq-directional-measure-slot-bound}
 \left|\frac1N\Delta_t^\top
    (\HessianMatrices_t^{(s)}-\widetilde\HessianMatrices_t^{(s)})\Delta_t\right|
 \le\{\Lambda(D)r_t+\Gamma_t^N(D)\}r_t^2.
\end{equation}
\end{lemma}
\begin{proof}
\noindent\textbf{Step 1: bound the Hessian difference by moment displacement.}
Fix $s\in[0,1]$ and abbreviate $z_i:=Z_t^{(s),i}$, $y_i:=Y_t^i$.
Let $b_s:=\sum_m\gamma_m(D)\|M_m(\rho_{\bs Z_t^{(s)}})-M_m(\mu_t)\|$.
The two Hessian bounds in \cref{eq:local-two-hessians-moment-bound} give
\[
\begin{aligned}
 \|\nabla_W^2F(\rho_{\bs Z_t^{(s)}};z_i,z_j)
             -\nabla_W^2F(\mu_t;z_i,z_j)\|_{\mathbb R^d\to\mathbb R^d}&\le b_s/2,\\
 \|\nabla_x\nabla_WF(\rho_{\bs Z_t^{(s)}},z_i)
             -\nabla_x\nabla_WF(\mu_t,z_i)\|_{\mathbb R^d\to\mathbb R^d}&\le b_s/2.
\end{aligned}
\]
The first term of the block Hessian has the additional factor $1/N$;
the second is block diagonal. Applying
\cref{lemma-operator-norm-block-matrix} to each part therefore yields
\[
 \|\HessianMatrices_t^{(s)}-\widetilde\HessianMatrices_t^{(s)}\|_{\mathbb R^{dN}\to\mathbb R^{dN}}
 \le\left\|\frac{b_s}{2N}\mathbf1\mathbf1^\top\right\|_{\mathbb R^N\to\mathbb R^N}
      +\left\|\frac{b_s}{2}I_N\right\|_{\mathbb R^N\to\mathbb R^N}=b_s.
\]

\par\noindent\textbf{Step 2: compare the empirical moments along the interpolation.}
For the moment displacement, integrate $\fD\phi_m$ on the segment
joining $y_i$ and $z_i$, which stays in the ball of radius $D$.
The feature bound in
\cref{assumption-structured-moment-smoothness,eq:coordinate-feature-growth} gives
\[
\begin{aligned}
 \|\phi_m(z_i)-\phi_m(y_i)\|_{\mathsf H_m}
 &\le\int_0^1 L_m(1+D^{(m-1)_+})\|z_i-y_i\|_{\mathbb R^d}\,du\\
 &\le L_m(1+2D^{m-1})\|z_i-y_i\|_{\mathbb R^d}.
\end{aligned}
\]
Since $z_i-y_i=s(X_t^i-Y_t^i)$, Cauchy--Schwarz gives
\begin{equation}
\label{eq:paired-empirical-moment-displacement}
\begin{aligned}
 \|M_m(\rho_{\bs Z_t^{(s)}})-M_m(\rho_{\bs Y_t})\|_{\mathsf H_m}
 &\le\frac{L_m(1+2D^{m-1})}{N}\sum_i\|z_i-y_i\|_{\mathbb R^d}\\
 &\le L_m(1+2D^{m-1})
       \left(\frac1N\sum_i\|z_i-y_i\|_{\mathbb R^d}^2\right)^{1/2}\\
 &\le L_m(1+2D^{m-1})r_t.
\end{aligned}
\end{equation}
The triangle inequality, followed by summation with weights $\gamma_m(D)$,
now gives $b_s\le\Lambda(D)r_t+\Gamma_t^N(D)$.
Finally $N^{-1}\|\Delta_t\|_{\mathbb R^{dN}}^2=r_t^2$, so the bound on the block operator
norm proves \cref{eq-directional-measure-slot-bound}.
\end{proof}

The second Hessian mismatch comes from moving the particle locations from \(\bs Y_t\) to the interpolation \(\bs Z_t^{(s)}\) while keeping the population law fixed.

\begin{lemma}[Directional point-slot error under anti-spikiness]
\label{lemma-directional-point-slot-error}
Let
\[
    \delta_{t,i}:=X_t^i-Y_t^i,
    \qquad
    r_t^2:=\frac1N\sum_{i=1}^N\norm{\delta_{t,i}}_{\mathbb R^d}^2.
\]
Recall from \cref{lemma-directional-measure-slot-error} the
population-measure Hessian
\begin{equation}
    \widetilde\HessianMatrices_t^{(s)}
    \coloneq
    \HessianMatrices(\mu_t,\bs Z_t^{(s)}).
\end{equation}
On the event
\(E_{\rm pop}\cap E_{T,\mathrm{sp}}(A_{\mathrm{sp}})\cap E_{T,D}\),
for every \(s\in[0,1]\),
\begin{equation}
\label{eq-directional-point-slot-bound}
    \left|
    \frac1N
    \Delta_t^\top
    \bigl(
        \widetilde\HessianMatrices_t^{(s)}-
        \HessianMatrices_t^Y
    \bigr)
    \Delta_t
    \right|
    \le
    \bigl((2+A_{\mathrm{sp}})\mathrm L_D\bigr)r_t^3.
\end{equation}
\end{lemma}

\begin{proof}
Fix \(s\in[0,1]\), and write \(z_i:=Z_{t,i}^{(s)}\), \(y_i:=Y_t^i\), and
\(\delta_i:=X_t^i-Y_t^i\). Since \(z_i-y_i=s\delta_i\), we have
\(\norm{z_i-y_i}_{\mathbb R^d}\le \norm{\delta_i}_{\mathbb R^d}\).

\par\noindent\textbf{Step 1: bound the integral-Hessian contribution.}
Split the quadratic form into the integral-Hessian contribution
$Q_{\mathrm{off}}$, whose sum includes $i=j$, and the spatial-Jacobian
contribution $Q_{\mathrm{diag}}$:
\begin{align}
    Q_{\mathrm{off}}
    &:=
    \frac1{N^2}\sum_{i,j=1}^N
    \left\langle
        \delta_i,
        \bigl[
            \nabla_W^2F(\mu_t,z_i,z_j)-
            \nabla_W^2F(\mu_t,y_i,y_j)
        \bigr]
        \delta_j
    \right\rangle,\\
    Q_{\mathrm{diag}}
    &:=
    \frac1N\sum_{i=1}^N
    \left\langle
        \delta_i,
        \bigl[
            \nabla_x\nabla_WF(\mu_t,z_i)-
            \nabla_x\nabla_WF(\mu_t,y_i)
        \bigr]
        \delta_i
    \right\rangle.
\end{align}
The spatial Lipschitzness (\cref{lemma-localized-moment-smoothness-consequences}) gives
\begin{equation}
 \begin{aligned}
    \norm{
        \nabla_W^2F(\mu_t,z_i,z_j)-
        \nabla_W^2F(\mu_t,y_i,y_j)
    }_{\Loperator d}
    &\le \mathrm L_D\norm{z_i-y_i}_{\mathbb R^d}+\mathrm L_D\norm{z_j-y_j}_{\mathbb R^d}\\
    &\le \mathrm L_D\norm{\delta_i}_{\mathbb R^d}+\mathrm L_D\norm{\delta_j}_{\mathbb R^d}.
 \end{aligned}
\end{equation}
Hence
\begin{align}
    |Q_{\mathrm{off}}|
    &\le
    \frac1{N^2}\sum_{i,j=1}^N
    \norm{\delta_i}_{\mathbb R^d}\norm{\delta_j}_{\mathbb R^d}
    \bigl(\mathrm L_D\norm{\delta_i}_{\mathbb R^d}+\mathrm L_D\norm{\delta_j}_{\mathbb R^d}\bigr)\\
    &=
    \frac{\mathrm L_D}{N^2}
    \left(\sum_i\norm{\delta_i}_{\mathbb R^d}^2\right)
    \left(\sum_j\norm{\delta_j}_{\mathbb R^d}\right)
    +
    \frac{\mathrm L_D}{N^2}
    \left(\sum_i\norm{\delta_i}_{\mathbb R^d}\right)
    \left(\sum_j\norm{\delta_j}_{\mathbb R^d}^2\right).
\end{align}
By Cauchy--Schwarz,
\[
    \sum_i\norm{\delta_i}_{\mathbb R^d}
    \le
    \sqrt N\left(\sum_i\norm{\delta_i}_{\mathbb R^d}^2\right)^{1/2}
    =Nr_t,
    \qquad
    \sum_i\norm{\delta_i}_{\mathbb R^d}^2=Nr_t^2.
\]
Therefore
\begin{equation}
    |Q_{\mathrm{off}}|
    \le
    2\mathrm L_Dr_t^3.
\end{equation}
\par\noindent\textbf{Step 2: bound the spatial-Jacobian contribution.}
The spatial Lipschitzness (\cref{lemma-localized-moment-smoothness-consequences}) gives
\begin{equation}
    \norm{
        \nabla_x\nabla_WF(\mu_t,z_i)-
        \nabla_x\nabla_WF(\mu_t,y_i)
    }_{\Loperator d}
    \le
    \mathrm L_D\norm{z_i-y_i}_{\mathbb R^d}
    \le
    \mathrm L_D\norm{\delta_i}_{\mathbb R^d}.
\end{equation}
Consequently,
\begin{equation}
    |Q_{\mathrm{diag}}|
    \le
    \frac{\mathrm L_D}{N}\sum_{i=1}^N\norm{\delta_i}_{\mathbb R^d}^3
    \le
    \mathrm L_DA_{\mathrm{sp}}r_t^3
\end{equation}
on \(E_{T,\mathrm{sp}}(A_{\mathrm{sp}})\). Combining the two estimates proves
\cref{eq-directional-point-slot-bound}.
\end{proof}

Combining the measure-slot and point-slot estimates gives the directional control of the Hessian mismatch that appears in the differential inequality for \(r_t\).

\begin{corollary}[Combined directional bound for \(\iota_t^N\)]
\label{corollary-bound-iota-combined}
On $E_{\rm pop}\cap E_{T,D}\cap E_{T,\mathrm{sp}}(A_{\mathrm{sp}})$,
with the population moment event of \cref{lem:population-moment-event},
sample localization \eqref{eq:sample-localization-event}, and
anti-spikiness \eqref{eq-cubic-antispike-bound}, it holds that
\begin{equation}
\label{eq-iota-combined-bound}
    \iota_t^N
    \le
    \Bigl(\Lambda(D)+(2+A_{\mathrm{sp}})\mathrm L_D\Bigr)r_t^3
    +
    \Gamma_t^Nr_t^2.
\end{equation}
The same deterministic radius $D$ is used in the event and all coefficients.
Define $q_{4,t}:=(N^{-1}\sum_i\|X_t^i-Y_t^i\|_{\mathbb R^d}^4)^{1/4}$. In particular, if \(q_{4,t}\le A_4r_t\) for all \(t\), with $A_4\ge1$, then the same bound holds with
\(A_{\mathrm{sp}}=A_4^2\).
\end{corollary}

\begin{proof}
The directional Hessian discrepancy, recalled from
\cref{lemma-bound-r_t-given-eta-eps},
\eqref{eq:ode-hessian-discrepancy-definition}, is
$\iota_t^N=N^{-1}\sup_{s\in[0,1]}
 |\Delta_t^\top(\HessianMatrices_t^{(s)}-\HessianMatrices_t^Y)\Delta_t|$,
where $\Delta_t=\mathbf X_t-\mathbf Y_t$.
\noindent\textbf{Step 1: combine the two Hessian contributions.}
For each \(s\in[0,1]\), decompose
\begin{align}
    \HessianMatrices_t^{(s)}-
    \HessianMatrices_t^Y
    &=
    \bigl(
        \HessianMatrices_t^{(s)}-
        \widetilde\HessianMatrices_t^{(s)}
    \bigr)
    +
    \bigl(
        \widetilde\HessianMatrices_t^{(s)}-
        \HessianMatrices_t^Y
    \bigr).
\end{align}
Taking the associated quadratic form against \(\Delta_t\), dividing by \(N\), and applying the triangle inequality gives
\begin{align}
    \left|
    \frac1N\Delta_t^\top
    \bigl(\HessianMatrices_t^{(s)}-\HessianMatrices_t^Y\bigr)
    \Delta_t
    \right|
    &\le
    \left|
    \frac1N\Delta_t^\top
    \bigl(
        \HessianMatrices_t^{(s)}-
        \widetilde\HessianMatrices_t^{(s)}
    \bigr)
    \Delta_t
    \right|\\
    &\quad+
    \left|
    \frac1N\Delta_t^\top
    \bigl(
        \widetilde\HessianMatrices_t^{(s)}-
        \HessianMatrices_t^Y
    \bigr)
    \Delta_t
    \right|.
\end{align}
The first term is bounded by \cref{lemma-directional-measure-slot-error} as
\((\Lambda(D)r_t+\Gamma_t^N)r_t^2\). The second term is bounded by
\cref{lemma-directional-point-slot-error} as
\(((2+A_{\mathrm{sp}})\mathrm L_D)r_t^3\). Taking the supremum over
\(s\in[0,1]\) proves \cref{eq-iota-combined-bound}.

\par\noindent\textbf{Step 2: deduce anti-spikiness from the fourth moment.}
For $\delta_{t,i}=X_t^i-Y_t^i$, Cauchy--Schwarz gives
\[
 \frac1N\sum_i\|\delta_{t,i}\|_{\mathbb R^d}^3
 \le\left(\frac1N\sum_i\|\delta_{t,i}\|_{\mathbb R^d}^2\right)^{1/2}
     \left(\frac1N\sum_i\|\delta_{t,i}\|_{\mathbb R^d}^4\right)^{1/2}
 =r_tq_{4,t}^2\le A_4^2r_t^3.
\]
The calculation also holds for $r_t=0$, when every $\delta_{t,i}=0$.
Thus \eqref{eq-cubic-antispike-bound} holds with $A_{\mathrm{sp}}=A_4^2$.
\end{proof}

\paragraph{Bound $r_t$ by moment residual.}

The final ODE estimate packages all continuous-time residuals into the common moment fluctuation \(\Gamma_t^N\), yielding the stopped curvature-response inequality used in the main theorem.

\begin{lemma}[Common empirical moment residual and directional differential inequality for \(r_t\)]
\label{lemma-common-empirical-moment-residual-directional-rt}
Work on the population moment event of Lemma~\ref{lem:population-moment-event}, and on the events
$E_{T,\mathrm{sp}}(A_{\mathrm{sp}})$ and $E_{T,D}$, and fix an open ODE interval
\(I_\ell=(t_\ell,t_{\ell+1})\). Recall from
\cref{eq:common-moment-residual-and-lipschitz-sum} the coordinate residuals
and their weighted sum:
\begin{equation}
    \Delta_{m,t}^N
    \coloneq
    \norm{M_m(\rho_{\bs Y_t})-M_m(\mu_t)}_{\mathsf H_m},
    \qquad
    \Gamma_t^N(D)
    \coloneq
    \sum_{m\ge1}\gamma_m(D)\Delta_{m,t}^N.
\end{equation}
Here, as in \cref{eq:common-local-moment-weights},
\begin{equation}
    \gamma_m(D)
    \coloneq
    \mathrm L_Da_m.
\end{equation}
Let
\begin{equation}
\label{eq-Csp-definition}
    c_D
    \coloneq
    \Lambda(D)+(2+A_{\mathrm{sp}})\mathrm L_D.
\end{equation}
Then
\begin{equation}
    \varepsilon_t^N\le \Gamma_t^N,
\end{equation}
and, for a.e. \(t\in I_\ell\) with \(r_t>0\),
\begin{equation}
\label{eq-rt-directional-final}
    \dot r_t
    \le
    -\kappa_t^{Y,N}r_t
    +
    c_Dr_t^2
    +
    \Gamma_t^N(D)r_t
    +
    \Gamma_t^N(D).
\end{equation}
The right upper Dini derivative is
\begin{equation}
\label{eq:right-upper-dini-derivative}
 \fD^+r_t:=\limsup_{h\downarrow0}\frac{r_{t+h}-r_t}{h},
 \qquad t,t+h\in I_\ell.
\end{equation}
It agrees with $\dot r_t$ where the ordinary derivative exists.
The same upper bound holds with $\fD^+r_t$, including at $r_t=0$
as justified below. This definition is used within one ODE branch,
not across a jump.

Moreover, for the pathwise first-exit time
\begin{equation}
    \tau_R^\ell
    \coloneq
    \inf\{t\in I_\ell:r_t\ge R\},
\end{equation}
we have, for \(t<\tau_R^\ell\),
\begin{equation}
\label{eq-rt-directional-stopped}
    \dot r_t
    \le
    -\bigl(\kappa_t^{Y,N}-c_DR-\Gamma_t^N(D)\bigr)r_t
    +
    \Gamma_t^N(D).
\end{equation}
\end{lemma}

\begin{proof}
\noindent\textbf{Step 1: control the residual and positive discrepancy.}
For each \(i\), the gradient bound in \cref{eq:local-value-gradient-moment-bound} gives
\begin{equation}
\begin{aligned}
 \norm{V(\rho_{\bs Y_t},Y_t^i)-V(\mu_t,Y_t^i)}_{\mathbb R^d}
 &=\norm{\nabla_WF(\rho_{\bs Y_t},Y_t^i)-\nabla_WF(\mu_t,Y_t^i)}_{\mathbb R^d}\\
 &\le\sum_{m\ge1}\gamma_m(D)\Delta_{m,t}^N=\Gamma_t^N.
\end{aligned}
\end{equation}
Therefore
\begin{equation}
    \varepsilon_t^N
    =
    \left(
        \frac1N\sum_{i=1}^N
        \norm{V(\rho_{\bs Y_t},Y_t^i)-V(\mu_t,Y_t^i)}_{\mathbb R^d}^2
    \right)^{1/2}
    \le
    \Gamma_t^N.
\end{equation}
Next, by \cref{corollary-bound-iota-combined},
\begin{equation}
    \iota_t^N
    \le
    c_Dr_t^3+
    \Gamma_t^Nr_t^2.
\end{equation}
Before applying \cref{lemma-bound-r_t-given-eta-eps}, we verify its
differentiability premise. On $E_{\rm pop}\cap E_{T,D}$, convexity of the
radius-$D$ ball gives $Z_t^{(s),i}\in\overline B_D(0)$ for every
$s\in[0,1]$, and \cref{lemma-reachable-moment-tube} gives
$M(\rho_{\bs Z_t^{(s)}})\in\mathcal V_{K_{\rm pop},D}\subset\mathcal U$.
The $C^2$ feature map of \cref{lem:localized-weighted-feature-envelopes},
the condition $G\in C^3(\mathcal U)$ in
\cref{assumption-structured-moment-smoothness}, and the derivative formulas
\cref{eq:weighted-gradient-representation,eq:weighted-spatial-hessian-representation,eq:weighted-integral-hessian-representation}
therefore make the empirical configuration field $C^1$ along this segment.
Applying \cref{lemma-bound-r_t-given-eta-eps}, for \(r_t>0\),
\begin{align}
    \dot r_t
    &\le
    -\kappa_t^{Y,N}r_t
    +
    \frac{\iota_t^N}{r_t}
    +
    \varepsilon_t^N\\
    &\le
    -\kappa_t^{Y,N}r_t
    +
    c_Dr_t^2
    +
    \Gamma_t^Nr_t
    +
    \varepsilon_t^N\\
    &\le
    -\kappa_t^{Y,N}r_t
    +
    c_Dr_t^2
    +
    \Gamma_t^Nr_t
    +
    \Gamma_t^N.
\end{align}

\par\noindent\textbf{Step 2: handle zero discrepancy without division.}
It remains to justify the inequality at $r_t=0$, where division by $r_t$
was not used. First take $s,t$ on the same ODE branch. The pathwise bound
$\|Y_t^1-Y_s^1\|_{\mathbb R^d}\le B|t-s|$ and the first-derivative bound in
\cref{assumption-structured-moment-smoothness} give
\[
\begin{aligned}
 \|M_m(\mu_t)-M_m(\mu_s)\|_{\mathsf H_m}
 &\le\mathbb E_{\mathcal G}\|\phi_m(Y_t^1)-\phi_m(Y_s^1)\|_{\mathsf H_m}\\
 &\le BL_m|t-s|\,\mathbb E_{\mathcal G}
        (1+\|Y_t^1\|_{\mathbb R^d}^{m-1}+\|Y_s^1\|_{\mathbb R^d}^{m-1})\\
 &\le BL_m|t-s|
        [1+2\{K_{\rm pop}m^{1/\alpha}\}^{m-1}].
\end{aligned}
\]
Here $\mathbb E_{\mathcal G}$ means expectation conditional on the
population fields; the conditional moment estimate is
\eqref{eq:population-uniform-conditional-moments}. Summing with weights
$a_m$ bounds the $\mathsf H_a$ norm, since a nonnegative sequence has
$\ell^2$ norm at most its $\ell^1$ norm:
\[
 \|M(\mu_t)-M(\mu_s)\|_{\mathsf H_a}
 \le B|t-s|\sum_{m\ge1}a_mL_m
        [1+2\{K_{\rm pop}m^{1/\alpha}\}^{m-1}]<\infty.
\]
Apply \cref{lem:weighted-series-estimate},
\eqref{eq:weighted-series-sum-bound}, with $K=1+K_{\rm pop}$ and $D=1$:
$\{K_{\rm pop}m^{1/\alpha}\}^{m-1}
\le(1+K_{\rm pop})^m m^{m/\alpha}$, so the series is finite.
For either particle system $S=X$ or $S=Y$, write
$\mathbf S_t=(S_t^1,\ldots,S_t^N)$. The event $E_{T,D}$, the
branchwise $B$-Lipschitz bound, and convexity of the radius-$D$ ball give
\[
\begin{aligned}
 \|M_m(\rho_{\mathbf S_t})-M_m(\rho_{\mathbf S_s})\|_{\mathsf H_m}
 &\le\frac1N\sum_{i=1}^N
       \|\phi_m(S_t^i)-\phi_m(S_s^i)\|_{\mathsf H_m}\\
 &\le BL_m(1+D^{(m-1)_+})|t-s|.
\end{aligned}
\]
Consequently, using $D\ge1$ and the bounded-support summability in
\cref{eq:weighted-population-moment-summability},
\[
 \|M(\rho_{\mathbf S_t})-M(\rho_{\mathbf S_s})\|_{\mathsf H_a}
 \le B|t-s|\sum_{m\ge1}a_mL_m(1+D^m)<\infty.
\]
Thus the population and both empirical moment paths are continuous.
Together with continuity of $\fD G,\fD^2G$ on $\mathcal U$ from
\cref{assumption-structured-moment-smoothness}(ii), the particle drifts
in \eqref{eq:weighted-gradient-representation} are continuous on the branch.
Recall $\Delta_t=\mathbf X_t-\mathbf Y_t$ and
$V(\nu,x)=-\nabla_WF(\nu,x)$. If $r_t=0$, then $\Delta_t=0$ and
\[
 \dot\Delta_t=
 (V(\rho_{\mathbf Y_t},Y_t^i)-V(\mu_t,Y_t^i))_{i=1}^N,
 \qquad \Delta_{t+h}=h\dot\Delta_t+o(h).
\]
By the Dini definition \eqref{eq:right-upper-dini-derivative},
\[
 \fD^+r_t
 =\lim_{h\downarrow0}\frac{\|\Delta_{t+h}\|_2}{\sqrt N h}
 =\frac{\|\dot\Delta_t\|_2}{\sqrt N}
 =\varepsilon_t^N\le\Gamma_t^N.
\]
For positive $r_t$, the preceding calculation gives the same upper Dini
inequality (or its a.e. version under absolute continuity). This proves
\cref{eq-rt-directional-final}. If \(t<\tau_R^\ell\), then \(r_t<R\), so
\begin{equation}
    c_Dr_t^2
    \le
    c_DRr_t,
\end{equation}
which yields \cref{eq-rt-directional-stopped}.
\end{proof}

%% file: appendix_final_general/general_assumptions.tex
\subsubsection{Assumptions}

The following assumptions specify the moment representation, Gaussian
perturbations, and pathwise regularity used in the comparison. The feature
parameters and the anchor norms below are fixed independently of $N,T,p$.
The localization radius and conditional population scale will be chosen
from $N,T,p$; their dependence remains in every local comparison coefficient.

\paragraph{Weighted moments and derivative norms.}
Fix $\alpha\in(0,2]$. For Hilbert spaces $\mathsf H_m$ and features
$\phi_m:\mathbb R^d\to\mathsf H_m$, write
\begin{equation}
\label{eq:weighted-moment-space}
\begin{gathered}
 M_m(\nu):=\int\phi_m(x)\,\nu(dx)\in\mathsf H_m,\qquad M(\nu):=(M_m(\nu))_{m\ge1},\\
 a_m:=e^{-c_am^{\kappa_{\rm dec}}},\qquad
 \|h\|_{\mathsf H_a}^2:=\sum_{m\ge1}a_m^2\|h_m\|_{\mathsf H_m}^2,\qquad
 \mathsf H_a:=\{h:\|h\|_{\mathsf H_a}<\infty\},
\end{gathered}
\end{equation}
where $c_a>0$, $\kappa_{\rm dec}>1$, and each feature integral is a
Bochner integral whenever it exists. Scalar features are included by
taking $\mathsf H_m=\mathbb R$ for every $m$. For finite $K>0$ and $D\ge1$, set
\begin{equation}
\label{eq:moment-classes}
\begin{aligned}
 \mathcal P_{\alpha,K}
 &:=\left\{\nu\in\mathcal P_2(\mathbb R^d):
   \sup_{q\ge1}q^{-1/\alpha}
       \left(\int\|x\|_{\mathbb R^d}^q\,\nu(dx)\right)^{1/q}\le K\right\},\\
 \mathcal P_D&:=\{\nu\in\mathcal P_2(\mathbb R^d):
                     \nu(\{x:\|x\|_{\mathbb R^d}\le D\})=1\}.
\end{aligned}
\end{equation}
Here $K$ is a variable class scale; $K_0$ in
\cref{assumption-moment-growth} will bound unconditional particle moments.

\begin{definition}[Localized moment region]
\label{def:localized-moment-region}
For finite $K>0$ and $D\ge1$, the feasible moment region is
\begin{equation}
\label{eq:localized-feasible-moment-region}
 \mathcal V_{K,D}:=\operatorname{conv}
     \{M(\nu):\nu\in\mathcal P_{\alpha,K}\cup\mathcal P_D\}.
\end{equation}
The feature condition below makes these moment vectors elements of
$\mathsf H_a$, as shown in \cref{lem:moment-integral-formulas}(i).
The convex hull here consists of finite convex combinations.
\end{definition}

For an $r$-linear map $A:E_1\times\cdots\times E_r\to E$, use the norm
\[
 \|A\|_{E_1\times\cdots\times E_r\to E}
 :=\sup_{\|v_j\|_{E_j}\le1,\ 1\le j\le r}
                  \|A[v_1,\ldots,v_r]\|_E.
\]
All Fr\'echet derivatives of $G$ below are with respect to the
weighted Hilbert norm in \eqref{eq:weighted-moment-space}.
If $\mathbf e_m u$ inserts $u\in\mathsf H_m$ in coordinate $m$ and
zero in the other coordinates, define the coordinate derivatives by
\begin{equation}
\label{eq:moment-coordinate-derivatives}
\begin{aligned}
 \langle\partial_mG(\theta),u\rangle_{\mathsf H_m}
   &:=\fD G(\theta)[\mathbf e_m u],\\
 \langle u,\partial_{mn}^2G(\theta)v\rangle_{\mathsf H_m}
   &:=\fD^2G(\theta)[\mathbf e_m u,\mathbf e_n v].
\end{aligned}
\end{equation}
Thus $\partial_mG(\theta)\in\mathsf H_m$ and
$\partial_{mn}^2G(\theta):\mathsf H_n\to\mathsf H_m$.

\begin{assumption}[Structured moments and weighted smoothness]
\label{assumption-structured-moment-smoothness}
The following conditions hold, with fixed structural parameters
independent of $N,T,p$.
\begin{enumerate}[label=(\roman*),leftmargin=*]
\item \textbf{Coordinate feature growth.}
There are $C_\phi,c_{\rm gr}>0$ and $\kappa_{\rm gr}\ge0$ with
$\kappa_{\rm dec}>\max\{1,\kappa_{\rm gr}\}$. Set
$L_m:=C_\phi m\exp(c_{\rm gr}m^{\kappa_{\rm gr}})$.
Each $\phi_m$ is $C^2$, its third Fr\'echet derivative exists, and
\begin{equation}
\label{eq:coordinate-feature-growth}
\begin{aligned}
 \|\phi_m(x)\|_{\mathsf H_m}&\le L_m(1+\|x\|_{\mathbb R^d}^m),\\
 \|\fD^r\phi_m(x)\|_{(\mathbb R^d)^r\to\mathsf H_m}
 &\le L_m(1+\|x\|_{\mathbb R^d}^{(m-r)_+}),\qquad r=1,2,3.
\end{aligned}
\end{equation}
Here $(m-r)_+:=\max\{m-r,0\}$, and $\fD^0\phi_m=\phi_m$.
For $r\ge1$ the displayed norm is the multilinear operator norm.
Continuity of $\fD^3\phi_m$ is not required.
\item \textbf{Weighted smoothness of the moment functional.}
There is one deterministic open set $\mathcal U\subset\mathsf H_a$
and one $G\in C^3(\mathcal U;\mathbb R)$ such that
\begin{equation}
\label{eq:structured-moment-open-neighbourhood}
 \bigcup_{K>0,\ D\ge1}\mathcal V_{K,D}\subset\mathcal U,\qquad
 \sup_{\theta\in\mathcal U}
 \|\fD^3G(\theta)\|_{\mathsf H_a\times\mathsf H_a\times\mathsf H_a\to\mathbb R}
 \le C_3<\infty.
\end{equation}
The set $\mathcal U$, the functional $G$, and $C_3$ do not vary with
$K,D,N,T,p$. Convexity of $\mathcal U$ is not required:
the moment segments used below lie in $\mathcal V_{K,D}$.
\end{enumerate}
\end{assumption}

\begin{definition}[Objective induced by weighted moments]
\label{def:moment-objective}
For $\nu\in\mathcal P_2(\mathbb R^d)$ whose feature integrals exist and
whose moment vector belongs to $\mathcal U$, define
\begin{equation}
\label{eq:moment-objective}
 F(\nu):=G(M(\nu)).
\end{equation}
This domain includes every $\mathcal P_{\alpha,K}$ with finite $K$ and
every $\mathcal P_D$ by \cref{lem:moment-integral-formulas}(i) and
\eqref{eq:structured-moment-open-neighbourhood}.
\end{definition}
Local bounds on $\fD G,\fD^2G$ and their coordinate derivatives will
follow from a fixed anchor and $C_3$ in
\cref{lem:moment-integral-formulas,lem:conditional-local-coefficient-growth};
no uniform bound on these lower derivatives over all of $\mathcal U$
is imposed.

\paragraph{Jump notation.}
We use the population and empirical jump models in
\eqref{eq:population-jump} and \eqref{eq:particle-jump}.
Let $\mathcal H_{\ell-}$ be the full pre-jump history and
$\mathcal G_{\rm pop}$ the population-field sigma-field. For a law $\nu$
and a particle configuration $\mathbf Z$, set
\begin{align*}
 K_\nu(x,y)&:=\int\nabla_W^2F(\nu;x,z)\nabla_W^2F(\nu;z,y)\,\nu(dz),\\
 \mathbf M^N(\rho_{\mathbf Z},\mathbf Z)_{ij}
 &:=N^{-1/2}\nabla_W^2F(\rho_{\mathbf Z};Z^i,Z^j).
\end{align*}
\begin{assumption}[Gaussian perturbation model]
\label{assumption-common-seed}
The population and empirical jumps follow the models specified above.
At each prescribed jump, the population field is generated from a fresh
innovation independent of the initial sample and all earlier innovations.
Conditional on $\mathcal H_{\ell-}$, it is centered Gaussian with covariance
$K_{\mu_{t_\ell-}}$. For every population law $\mu_t$ and empirical law
$\rho_{\mathbf X_t},\rho_{\mathbf Y_t}$ used in the comparison,
including pre-jump laws, the Hessian kernel is symmetric under transposition
of its two spatial slots, and the operator with kernel $K_\nu$ on
$L^2(\nu;\mathbb R^d)$ is positive and trace class.
A jointly measurable version of the field is used in the population pushforward.
\end{assumption}
\paragraph{Finite representation and comparison coupling.}
Kernel symmetry gives the covariance identity
\[
 [K_{\rho_{\mathbf Z}}(Z^i,Z^j)]_{i,j}
 =\mathbf M^N(\rho_{\mathbf Z},\mathbf Z)
       \mathbf M^N(\rho_{\mathbf Z},\mathbf Z)^\top.
\]
This is the finite representation of the prescribed empirical Gaussian
transition. Lemma~\ref{lem:population-first-finite-coupling} constructs a
coordinate that is standard Gaussian conditional on $\mathcal H_{\ell-}$
and realizes this transition together with the population evaluation vector.
This is a construction, rather than an additional independence assumption
on the coordinate and the reference system. Reference conditional
independence is established under $\mathcal G_{\rm pop}$ in
Lemma~\ref{lemma-gp-jump-implies-conditional-iid}.
The covariance uses the integral Hessian term; the drift curvature uses
the full Hessian.

The following tail condition ensures that the reference and empirical particles remain in a common sub-Weibull class throughout the fixed time horizon.

\begin{assumption}[Unconditional moment growth]
\label{assumption-moment-growth}
There are deterministic $K_0>0$ and $\alpha\in(0,2]$ such that
\[
\sup_{0\le t\le T}\sup_{q\ge1}q^{-1/\alpha}
\max\{(\mathbb E\|X_t^1\|_{\mathbb R^d}^q)^{1/q},
       (\mathbb E\|Y_t^1\|_{\mathbb R^d}^q)^{1/q}\}\le K_0.
\]
\end{assumption}
\paragraph{Marginal and conditional bounds.}
The coupled construction is invariant under simultaneous particle
permutations, as proved in Step 6 of
\cref{lem:population-first-finite-coupling}; hence the same bound holds
for every particle. Bounds at left
limits follow by Fatou's lemma. The class scale $K$ in
\cref{assumption-structured-moment-smoothness} ranges over all finite
values, including $K_0$. Conditional bounds with a deterministic
enlarged scale $K_{\rm pop}$ hold on the population-measurable event of
Lemma~\ref{lem:population-moment-event}; they are not assumed almost surely
with scale $K_0$. In asymptotic statements a uniform bound on $K_0$ is
specified explicitly.

We separately assume a common deterministic Lipschitz bound on each ODE
branch. This pathwise condition is used for time-grid interpolation in
\cref{lemma-uniform-sup-gamma-bound-localized-conditional} and for the statistics in \cref{subsec:trigger-stability}; it is not a consequence
of the moment assumption. The two traces at a jump belong to different
branches, so the bound does not constrain the jump increment.

\paragraph{ODE intervals.}
For fixed $T>0$, write $I_\ell=(t_\ell,t_{\ell+1})$, with $t_0=0$
and $t_{\NofJ+1}=T$. Empty intervals, including a terminal interval
after a jump at $T$, impose no condition.
\begin{assumption}[Piecewise-Lipschitz flow]
\label{assumption-uniform-concentration-piecewise-lipschitz}
There is a deterministic constant $B\ge0$ such that
\begin{equation}
\sup_{1\le i\le N}\sup_{0\le \ell\le \NofJ}\sup_{\substack{s<t\\ s,t\in I_\ell}}
\max\Bigl\{
    \frac{\norm{Y_t^i-Y_s^i}_{\mathbb R^d}}{|t-s|},
     \frac{\norm{X_t^i-X_s^i}_{\mathbb R^d}}{|t-s|},
\Bigr\}
\le
B
\qquad\text{a.s.}
\end{equation}
\end{assumption}

The final pathwise condition rules out the degenerate case where the coupling error is carried by only a few particles.

\paragraph{Discrepancy and the anti-spikiness event.}
For
\[
    \delta_{t,i}\coloneq X_t^i-Y_t^i,
    \qquad
    r_t^2\coloneq \frac1N\sum_{i=1}^N\norm{\delta_{t,i}}_{\mathbb R^d}^2,
\]
and a constant \(A_{\mathrm{sp}}\ge 1\), define $E_{T,\mathrm{sp}}(A_{\mathrm{sp}})
    \coloneq
    \left\{
    \sup_{0\le t\le T:\,r_t>0}
    \frac{\frac1N\sum_{i=1}^N\norm{\delta_{t,i}}_{\mathbb R^d}^3}{r_t^3}
    \le
    A_{\mathrm{sp}}
    \right\}.$

\begin{assumption}[Anti-spikiness on comparison paths]
\label{assumption-anti-spikiness-event}
For a fixed $A_{\mathrm{sp}}\ge1$, the pathwise comparison is restricted
to $E_{T,\mathrm{sp}}(A_{\mathrm{sp}})$. The almost-sure anti-spikiness
condition in \cref{assumption-main-coupled-particle-regularity} is the special case where this
event has probability one; no such probability bound is asserted here.
\end{assumption}
If $r_t=0$, every discrepancy is zero. Thus, including these times,
the event is equivalently specified by
\begin{equation}
\label{eq-cubic-antispike-bound}
    \frac1N\sum_{i=1}^N\norm{X_t^i-Y_t^i}_{\mathbb R^d}^3
    \le
    A_{\mathrm{sp}}r_t^3
    \qquad(0\le t\le T).
\end{equation}

\subsubsection{Local consequences used in the comparison}

We first establish summability of the moment vectors and the integral
formulas on the regions $\mathcal V_{K,D}$ in
\cref{def:localized-moment-region}. These formulas then give one common
envelope for the localized comparison.

\paragraph{A common weighted-series estimate.}
Define
\begin{equation}
\label{eq:weighted-series-growth-function}
 \chi(u):=[\log(e+u)]^{\kappa_{\rm dec}/(\kappa_{\rm dec}-1)}\qquad(u\ge0).
\end{equation}
\begin{lemma}[Weighted-series estimate]
\label{lem:weighted-series-estimate}
With $a_m,L_m$ from \cref{assumption-structured-moment-smoothness}(i),
there is a constant $C$ depending only on the fixed feature parameters
and $\alpha$ such that, for $K>0$, $D\ge1$, and $m\ge1$,
\begin{equation}
\label{eq:weighted-series-summand-bound}
 a_mL_m\{2+D^m+K^m m^{m/\alpha}\}
 \le C\exp\{C\chi(K+D)-(c_a/4)m^{\kappa_{\rm dec}}\}.
\end{equation}
In particular,
\begin{equation}
\label{eq:weighted-series-sum-bound}
 \sum_{m\ge1}a_mL_m\{2+D^m+K^m m^{m/\alpha}\}
 \le Ce^{C\chi(K+D)}.
\end{equation}
For each fixed $c>0$ there is also a constant $C_c$ such that
\begin{equation}
\label{eq:weighted-series-scale-comparison}
 \chi(c(1+u)+1)\le C_c\chi(u)\qquad(u\ge0).
\end{equation}
\end{lemma}
\begin{proof}
Write $\kappa=\kappa_{\rm dec}$ and $s=\kappa/(\kappa-1)$ in this proof.
For $u,v\ge0$ and $\varepsilon>0$, Young's inequality with conjugate
exponents $\kappa,s$ gives
\begin{equation}
\label{eq:young-weighted-series}
 uv\le\varepsilon u^\kappa+
       \frac{\kappa-1}{\kappa}(\kappa\varepsilon)^{-1/(\kappa-1)}v^s.
\end{equation}
Indeed, substitute $a=(\kappa\varepsilon)^{1/\kappa}u$ and
$b=(\kappa\varepsilon)^{-1/\kappa}v$ into
$ab\le a^\kappa/\kappa+b^s/s$.
Taking $u=m$, $v=\log(e+K+D)$, and $\varepsilon=c_a/4$ yields
\[
 m\log(e+K+D)\le\tfrac{c_a}{4}m^\kappa+C\chi(K+D).
\]
The two remaining growth terms satisfy
\[
 c_{\rm gr}m^{\kappa_{\rm gr}}\le\tfrac{c_a}{4}m^\kappa+C,
 \qquad (m/\alpha)\log m+\log(4C_\phi m)
                  \le\tfrac{c_a}{4}m^\kappa+C.
\]
These inequalities hold for all $m\ge1$ after increasing the fixed $C$:
the ratios $m^{\kappa_{\rm gr}}/m^\kappa$ and
$m\log m/m^\kappa$ tend to zero, and the finitely many remaining values
have a finite maximum.
Since
$2+D^m+K^m m^{m/\alpha}\le4(e+K+D)^m m^{m/\alpha}$, substitution of
these three bounds gives
\[
\begin{aligned}
 &\log\!\left[a_mL_m\{2+D^m+K^m m^{m/\alpha}\}\right]\\
 &\quad\le-c_am^\kappa+c_{\rm gr}m^{\kappa_{\rm gr}}
       +m\log(e+K+D)+(m/\alpha)\log m+\log(4C_\phi m)\\
 &\quad\le(-c_a+c_a/4+c_a/4+c_a/4)m^\kappa+C\chi(K+D)+C\\
 &\quad=-(c_a/4)m^\kappa+C\chi(K+D)+C.
\end{aligned}
\]
Exponentiating and absorbing the fixed factor $e^C$ yields
\eqref{eq:weighted-series-summand-bound}.

Summing leaves $\sum_{m\ge1}e^{-(c_a/4)m^\kappa}<\infty$, proving
\eqref{eq:weighted-series-sum-bound}. Finally,
\[
 \log(e+c(1+u)+1)\le\log(c+2)+\log(e+u)
 \le(1+\log(c+2))\log(e+u).
\]
Raising to the power $s$ proves \eqref{eq:weighted-series-scale-comparison}.
The constants here do not depend on the anchor norms or on $K,D,N,T,p$.
\end{proof}

\paragraph{Fixed anchor.}
Let $\delta_0$ be the Dirac measure at the origin. Since
$\delta_0\in\mathcal P_D$, set
\begin{equation}
\label{eq:moment-anchor}
\begin{gathered}
 \theta_0:=M(\delta_0)=(\phi_m(0))_{m\ge1}\in\mathcal V_{K,D}\subset\mathcal U,\\
 g_0:=\|\fD G(\theta_0)\|_{\mathsf H_a\to\mathbb R},\qquad
 h_0:=\|\fD^2G(\theta_0)\|_{\mathsf H_a\times\mathsf H_a\to\mathbb R}.
\end{gathered}
\end{equation}
Moment summability in the next lemma justifies $\theta_0\in\mathsf H_a$.
The $C^3$ property in \cref{assumption-structured-moment-smoothness}(ii)
makes $g_0,h_0$ finite; they are fixed independently of $K,D,N,T,p$.

\begin{lemma}[Moment summability and integral formulas]
\label{lem:moment-integral-formulas}
Under \cref{assumption-structured-moment-smoothness}, for finite $K>0$
and $D\ge1$ the following statements hold.
\begin{enumerate}[label=(\roman*),leftmargin=*]
\item \textbf{Moment vectors.}
All feature integrals exist for $\nu\in\mathcal P_{\alpha,K}\cup\mathcal P_D$,
and
\begin{equation}
\label{eq:weighted-population-moment-summability}
\begin{aligned}
 \|M(\nu)\|_{\mathsf H_a}
 &\le\sum_{m\ge1}a_mL_m(1+K^m m^{m/\alpha})<\infty
       &&(\nu\in\mathcal P_{\alpha,K}),\\
 \|M(\nu)\|_{\mathsf H_a}
 &\le\sum_{m\ge1}a_mL_m(1+D^m)<\infty
       &&(\nu\in\mathcal P_D).
\end{aligned}
\end{equation}
\item \textbf{Integral formulas.}
For $\theta,\vartheta\in\mathcal V_{K,D}$, put
$\Delta:=\vartheta-\theta$ and $\theta_s:=(1-s)\theta+s\vartheta$.
Then
\begin{align}
 G(\vartheta)-G(\theta)
 &=\int_0^1\fD G(\theta_s)[\Delta]\,ds,
 \label{eq:moment-value-integral}\\
 \fD G(\vartheta)-\fD G(\theta)
 &=\int_0^1\fD^2G(\theta_s)[\Delta,\cdot]\,ds,
 \label{eq:moment-gradient-integral}\\
 \|\fD^2G(\vartheta)-\fD^2G(\theta)\|_{\mathsf H_a\times\mathsf H_a\to\mathbb R}
 &\le C_3\|\vartheta-\theta\|_{\mathsf H_a}.
 \label{eq:moment-hessian-lipschitz}
\end{align}
\item \textbf{Bounds from the anchor.}
With $\theta_0,g_0,h_0$ from \eqref{eq:moment-anchor}, for
$\theta\in\mathcal V_{K,D}$,
\begin{equation}
\label{eq:anchored-weighted-derivative-bounds}
\begin{aligned}
 \|\fD^2G(\theta)\|_{\mathsf H_a\times\mathsf H_a\to\mathbb R}
 &\le h_0+C_3\|\theta-\theta_0\|_{\mathsf H_a},\\
 \|\fD G(\theta)\|_{\mathsf H_a\to\mathbb R}
 &\le g_0+h_0\|\theta-\theta_0\|_{\mathsf H_a}
          +\tfrac12C_3\|\theta-\theta_0\|_{\mathsf H_a}^2.
\end{aligned}
\end{equation}
\end{enumerate}
\end{lemma}
\begin{proof}
\noindent\textbf{Step 1: sum the weighted coordinate moments.}
For $\nu\in\mathcal P_{\alpha,K}$, the $q=m$ case of
\eqref{eq:moment-classes} and the feature bound
\eqref{eq:coordinate-feature-growth} give
\[
 \int\|\phi_m(x)\|_{\mathsf H_m}\,\nu(dx)
 \le L_m\left(1+\int\|x\|_{\mathbb R^d}^m\,\nu(dx)\right)
 \le L_m(1+K^m m^{m/\alpha}).
\]
Thus the Bochner integrals exist and obey the same bound in norm.
For their sum, apply \cref{lem:weighted-series-estimate},
\eqref{eq:weighted-series-sum-bound}, with $D=1$ and the given $K$.
For every integer $n$,
\[
 \sum_{m\le n}a_m^2\|M_m(\nu)\|_{\mathsf H_m}^2
 \le\left(\sum_{m\le n}a_m\|M_m(\nu)\|_{\mathsf H_m}\right)^2
 \le\left(\sum_{m\ge1}a_mL_m(1+K^m m^{m/\alpha})\right)^2.
\]
Letting $n\to\infty$ proves the first line of
\eqref{eq:weighted-population-moment-summability}.
For $\nu\in\mathcal P_D$, replace the moment bound by
$\int\|x\|_{\mathbb R^d}^m\,\nu(dx)\le D^m$; apply \eqref{eq:weighted-series-sum-bound} with $K=1$ and the
given $D$ to get a finite weighted sum. The same finite-partial-sum
inequality proves the second line of \eqref{eq:weighted-population-moment-summability}.

\par\noindent\textbf{Step 2: integrate on a feasible moment segment.}
By \cref{def:localized-moment-region} and
\eqref{eq:structured-moment-open-neighbourhood},
$\theta_s\in\mathcal V_{K,D}\subset\mathcal U$ for every $s\in[0,1]$.
Since $\dot\theta_s=\Delta$, the chain rule gives
$\frac{d}{ds}G(\theta_s)=\fD G(\theta_s)[\Delta]$ and
$\frac{d}{ds}\fD G(\theta_s)=\fD^2G(\theta_s)[\Delta,\cdot]$.
Integrating gives
\cref{eq:moment-value-integral,eq:moment-gradient-integral}.
For $u,v\in\mathsf H_a$, the same calculation and assumption (ii) give
\[
\begin{aligned}
 (\fD^2G(\vartheta)-\fD^2G(\theta))[u,v]
 &=\int_0^1\fD^3G(\theta_s)[\Delta,u,v]\,ds,\\
 |(\fD^2G(\vartheta)-\fD^2G(\theta))[u,v]|
 &\le C_3\|\Delta\|_{\mathsf H_a}\|u\|_{\mathsf H_a}\|v\|_{\mathsf H_a}.
\end{aligned}
\]
The supremum over unit $u,v$ is
\eqref{eq:moment-hessian-lipschitz}.

\par\noindent\textbf{Step 3: integrate from the fixed anchor.}
Both $\theta_0$ and $\theta$ lie in $\mathcal V_{K,D}$.
Equation \eqref{eq:moment-hessian-lipschitz} first gives
$\|\fD^2G(\theta)\|_{\mathsf H_a\times\mathsf H_a\to\mathbb R}\le h_0+C_3\|\theta-\theta_0\|_{\mathsf H_a}$.
Apply this bound at $\theta_0+s(\theta-\theta_0)$ in
\eqref{eq:moment-gradient-integral}:
\[
\begin{aligned}
 \|\fD G(\theta)\|_{\mathsf H_a\to\mathbb R}
 &\le g_0+\|\theta-\theta_0\|_{\mathsf H_a}
       \int_0^1(h_0+C_3s\|\theta-\theta_0\|_{\mathsf H_a})\,ds\\
 &=g_0+h_0\|\theta-\theta_0\|_{\mathsf H_a}
          +\tfrac12 C_3\|\theta-\theta_0\|_{\mathsf H_a}^2.
\end{aligned}
\]
This proves \eqref{eq:anchored-weighted-derivative-bounds}.
\end{proof}

\paragraph{Particle localization and the conditional population scale.}
For $D\ge1$, define
\begin{equation}
\label{eq:sample-localization-event}
 E_{T,D}:=\left\{\max_{1\le i\le N}\sup_{0\le t\le T}
                    (\|X_t^i\|_{\mathbb R^d}\vee\|Y_t^i\|_{\mathbb R^d})\le D\right\}.
\end{equation}
The population quantities used next are supplied by
\cref{lem:population-moment-event}:
\begin{itemize}[leftmargin=*]
\item $K_{\rm pop}$ is the deterministic conditional moment scale in
\eqref{eq:population-exponential-scale}. It depends on the probability
budget and horizon, and need not equal the unconditional scale $K_0$.
\item $E_{\rm pop}\in\mathcal G_{\rm pop}$ is the population-measurable
event constructed there. For a budget $p_0\in(0,1)$ it has probability
at least $1-p_0$, and \eqref{eq:population-uniform-conditional-moments}
gives $\mu_t\in\mathcal P_{\alpha,K_{\rm pop}}$ on this event.
\end{itemize}
All time statements include pre-jump and post-jump values. The next
lemma records how these conclusions place the comparison paths in
the domain of $G$; the quantitative construction of $K_{\rm pop}$
and $E_{\rm pop}$ remains in \cref{lem:population-moment-event}.

\begin{lemma}[Moment points of the population, particle clouds, and interpolations]
\label{lemma-reachable-moment-tube}
With $K=K_{\rm pop}$ in \cref{def:localized-moment-region}:
\begin{enumerate}[label=(\roman*),leftmargin=*]
\item \textbf{Population.} On $E_{\rm pop}$, for every $t\le T$,
\[
 \sup_{q\ge1}q^{-1/\alpha}
       \left(\int\|x\|_{\mathbb R^d}^q\mu_t(dx)\right)^{1/q}\le K,
 \qquad \mu_t\in\mathcal P_{\alpha,K}.
\]
\item \textbf{Empirical measures and particle interpolation.}
On $E_{T,D}$, set $Z_t^{(s),i}:=(1-s)Y_t^i+sX_t^i$. Then
\[
 \rho_{\mathbf X_t},\quad \rho_{\mathbf Y_t},\quad
 \rho_{\mathbf Z_t^{(s)}}\in\mathcal P_D
 \qquad(t\le T,\ 0\le s\le1).
\]
\item \textbf{Segments in moment space.}
On $E_{\rm pop}\cap E_{T,D}$, let $\rho,\nu$ be any two measures in
\[
 \{\mu_t,\rho_{\mathbf X_t},\rho_{\mathbf Y_t}\}
       \cup\{\rho_{\mathbf Z_t^{(s)}}:0\le s\le1\}.
\]
For every $u\in[0,1]$,
\[
 (1-u)M(\rho)+uM(\nu)
 \in\mathcal V_{K,D}\subset\mathcal U.
\]
Thus the integral formulas and Hessian-difference bound in
\cref{lem:moment-integral-formulas} apply on this segment.
\end{enumerate}
\end{lemma}
\begin{proof}
For (i), use $\mu_t=\operatorname{Law}(Y_t^1\mid\mathcal G_{\rm pop})$
and the conditional moment bound
\eqref{eq:population-uniform-conditional-moments} of
\cref{lem:population-moment-event}.
For (ii), $E_{T,D}$ gives
\[
 \|Z_t^{(s),i}\|_{\mathbb R^d}\le(1-s)\|Y_t^i\|_{\mathbb R^d}+s\|X_t^i\|_{\mathbb R^d}\le D.
\]
Each stated empirical measure is therefore supported in the closed ball.
The same holds for left limits because that ball is closed.
For (iii), each endpoint belongs to \eqref{eq:localized-feasible-moment-region}
by (i)--(ii), and
\[
 (1-u)M(\rho)+uM(\nu)=M((1-u)\rho+u\nu)\in\mathcal V_{K,D}.
\]
The inclusion $\mathcal V_{K,D}\subset\mathcal U$ is exactly
\eqref{eq:structured-moment-open-neighbourhood}; no passage to a
closure is needed.
The particle interpolation in (ii) is handled by its support bound;
it is not identified with the measure mixture in the last display.
\end{proof}

\begin{definition}[One local comparison envelope]
\label{def:localized-envelope}
For $K>0$, $D\ge1$, with $a_m,L_m$ from
\cref{eq:weighted-moment-space,eq:coordinate-feature-growth} and the
anchor norms $g_0,h_0$ from \eqref{eq:moment-anchor}, define
\begin{equation}
\label{eq:conditional-comparison-envelope}
 \mathrm L_{K,D}:=4(1+g_0+h_0+C_3)
 \left[1+\sum_{m\ge1}a_mL_m
       \{2+D^m+K^m m^{m/\alpha}\}\right]^3.
\end{equation}
In the localized comparison $K=K_{\rm pop}$ is deterministic, and we write
$\mathrm L_D:=\mathrm L_{K_{\rm pop},D}$. This abbreviation retains the
dependence of $K_{\rm pop}$ on $N,T,p$; $\mathrm L_D$ is not assumed $O(1)$.
\end{definition}

\begin{lemma}[Growth and derivative bounds for the local envelope]
\label{lem:conditional-local-coefficient-growth}
Assume \cref{assumption-structured-moment-smoothness}. For $K>0$, $D\ge1$:
\begin{enumerate}[label=(\roman*),leftmargin=*]
\item \textbf{Growth in the localization scales.} There is a constant $C$
depending only on the fixed feature parameters, $\alpha$, $g_0,h_0,C_3$,
and the fixed dimension, such that
\begin{equation}
\label{eq:local-envelope-growth}
 1\le\mathrm L_{K,D}\le C\exp\{C\chi(K+D)\}.
\end{equation}
In particular $C$ is independent of $K,D,N,T,p$.
\item \textbf{Moment derivatives on the feasible region.} For
$\theta\in\mathcal V_{K,D}$,
\begin{equation}
\label{eq:local-moment-derivative-bound}
 \|\fD G(\theta)\|_{\mathsf H_a\to\mathbb R}
 \vee\|\fD^2G(\theta)\|_{\mathsf H_a\times\mathsf H_a\to\mathbb R}
 \le\mathrm L_{K,D}.
\end{equation}
\end{enumerate}
\end{lemma}
\begin{proof}
Use the following quantities only within this proof:
\[
 R:=1+\sum_{m\ge1}a_mL_m\{2+D^m+K^m m^{m/\alpha}\},
 \qquad Q:=1+g_0+h_0+C_3.
\]
Thus $\mathrm L_{K,D}=4QR^3$.

\noindent\textbf{Step 1: apply the common series estimate.}
By \cref{lem:weighted-series-estimate}, \eqref{eq:weighted-series-sum-bound},
$R\le Ce^{C\chi(K+D)}$. Substitution into
\eqref{eq:conditional-comparison-envelope} gives
$\mathrm L_{K,D}=4QR^3\le Ce^{C\chi(K+D)}$, with the fixed anchor
factor absorbed into $C$. This proves (i) without changing the defined
finite-sample coefficient $\mathrm L_{K,D}$.

\noindent\textbf{Step 2: apply the anchor estimates on the specified region.}
By \eqref{eq:coordinate-feature-growth}, $\|\phi_m(0)\|_{\mathsf H_m}\le L_m$.
For either class in \eqref{eq:moment-classes},
\[
\begin{aligned}
 \|M_m(\nu)-\phi_m(0)\|_{\mathsf H_m}
 &\le\int\|\phi_m(x)\|_{\mathsf H_m}\,\nu(dx)+\|\phi_m(0)\|_{\mathsf H_m}\\
 &\le L_m\left(2+\int\|x\|_{\mathbb R^d}^m\,\nu(dx)\right)\\
 &\le
 \begin{cases}
 L_m(2+D^m),&\nu\in\mathcal P_D,\\
 L_m(2+K^m m^{m/\alpha}),&\nu\in\mathcal P_{\alpha,K}.
 \end{cases}
\end{aligned}
\]
For $\theta\in\mathcal V_{K,D}$, the finite-convex-hull definition
\eqref{eq:localized-feasible-moment-region} gives
$\theta=\sum_{j=1}^J\lambda_jM(\nu_j)$, where
$\lambda_j\ge0$, $\sum_j\lambda_j=1$, and
$\nu_j\in\mathcal P_{\alpha,K}\cup\mathcal P_D$.
Using the weighted $\ell^2\le\ell^1$ estimate from
\cref{lem:moment-integral-formulas}(i), we get
\[
\begin{aligned}
 \|\theta-\theta_0\|_{\mathsf H_a}
 &\le\sum_{j=1}^J\lambda_j
           \sum_{m\ge1}a_m\|M_m(\nu_j)-\phi_m(0)\|_{\mathsf H_m}\\
 &\le\sum_{j=1}^J\lambda_j
           \sum_{m\ge1}a_mL_m\{2+D^m+K^m m^{m/\alpha}\}\\
 &=R-1\le R.
\end{aligned}
\]
Substitute this bound into
\eqref{eq:anchored-weighted-derivative-bounds}. Since $R\ge1$,
$h_0+C_3R\le(h_0+C_3)R\le QR$ and
$g_0+h_0R+(C_3/2)R^2\le(g_0+h_0+C_3)R^2\le QR^2$.
Thus
\begin{equation}
\label{eq:local-anchor-intermediate}
\begin{aligned}
 \sup_{\theta\in\mathcal V_{K,D}}
     \|\fD G(\theta)\|_{\mathsf H_a\to\mathbb R}&\le QR^2,\\
 \sup_{\theta\in\mathcal V_{K,D}}
     \|\fD^2G(\theta)\|_{\mathsf H_a\times\mathsf H_a\to\mathbb R}&\le QR.
\end{aligned}
\end{equation}
Both are at most $4QR^3=\mathrm L_{K,D}$, proving (ii).
To obtain coordinate decay, recall
\eqref{eq:moment-coordinate-derivatives} and
$\|\mathbf e_m u\|_{\mathsf H_a}=a_m\|u\|_{\mathsf H_m}$.
For unit $u\in\mathsf H_m$, $v\in\mathsf H_n$,
\[
\begin{aligned}
 |\langle\partial_mG(\theta),u\rangle|
 &=|\fD G(\theta)[\mathbf e_m u]|
   \le\mathrm L_{K,D}a_m,\\
 |\langle u,\partial_{mn}^2G(\theta)v\rangle|
 &=|\fD^2G(\theta)[\mathbf e_m u,\mathbf e_n v]|
   \le\mathrm L_{K,D}a_ma_n.
\end{aligned}
\]
Taking the unit suprema gives
$\|\partial_mG(\theta)\|_{\mathsf H_m}\le\mathrm L_{K,D}a_m$ and
$\|\partial_{mn}^2G(\theta)\|_{\mathsf H_n\to\mathsf H_m}
\le\mathrm L_{K,D}a_ma_n$.
\end{proof}

\begin{lemma}[Weighted feature regularity from coordinate bounds]
\label{lem:localized-weighted-feature-envelopes}
Under item (i) of \cref{assumption-structured-moment-smoothness}, the map
$J_0(x):=(\phi_m(x))_{m\ge1}$ is a $C^2$ map from $\mathbb R^d$ to
$\mathsf H_a$, with
\[
 \fD J_0(x)[h]=(\fD\phi_m(x)[h])_{m\ge1},\qquad
 \fD^2J_0(x)[h,v]=(\fD^2\phi_m(x)[h,v])_{m\ge1}.
\]
For $D\ge1$,
\begin{align}
 \sup_{\|x\|_{\mathbb R^d}\le D}\|\fD^rJ_0(x)\|_{(\mathbb R^d)^r\to\mathsf H_a}
 &\le\sum_{m\ge1}a_mL_m(1+D^m)<\infty
       \quad(r=0,1,2),\label{eq:weighted-feature-derivative-bound}\\
 \|\fD^rJ_0(x)-\fD^rJ_0(y)\|_{(\mathbb R^d)^r\to\mathsf H_a}
 &\le\left(\sum_{m\ge1}a_mL_m(1+D^m)\right)\|x-y\|_{\mathbb R^d}
       \quad(r=1,2).
 \label{eq:weighted-feature-spatial-difference}
\end{align}
The difference bound holds for $\|x\|_{\mathbb R^d},\|y\|_{\mathbb R^d}\le D$.
The $r=0$ subscript denotes the $\mathsf H_a$ norm; for $r=1,2$ it is the
multilinear operator norm from $(\mathbb R^d)^r$ to $\mathsf H_a$.
\end{lemma}
\begin{proof}
\noindent\textbf{Step 1: bound the coordinate maps and their differences.}
For fixed $D$, summability of
$\sum_m a_mL_m(1+D^m)$ is
\eqref{eq:weighted-population-moment-summability} for bounded support.
For $r=0,1,2$, \eqref{eq:coordinate-feature-growth} and $D\ge1$ give, for unit directions,
\[
 \|(\fD^r\phi_m(x)[v_1,\ldots,v_r])_m\|_{\mathsf H_a}
 \le\left(\sum_m a_m^2L_m^2(1+D^m)^2\right)^{1/2}
 \le\sum_m a_mL_m(1+D^m).
\]
For $r=0$ the left-hand side means $\|J_0(x)\|_{\mathsf H_a}$.
Thus the first- and second-derivative candidate sequences define bounded
linear and bilinear maps, respectively.
If $\|x\|_{\mathbb R^d},\|y\|_{\mathbb R^d}\le D$, then
$\|(1-s)y+sx\|_{\mathbb R^d}\le(1-s)\|y\|_{\mathbb R^d}+s\|x\|_{\mathbb R^d}\le D$.
The mean-value inequality, applied to $\fD^r\phi_m$ for $r=1,2$, gives
\begin{equation}
\label{eq:coordinate-feature-spatial-difference}
\begin{aligned}
 \|\fD^r\phi_m(x)-\fD^r\phi_m(y)\|_{(\mathbb R^d)^r\to\mathsf H_m}
 &\le\sup_{\|z\|_{\mathbb R^d}\le D}\|\fD^{r+1}\phi_m(z)\|_{(\mathbb R^d)^{r+1}\to\mathsf H_m}\,\|x-y\|_{\mathbb R^d}
 \\ &\le L_m(1+D^m)\|x-y\|_{\mathbb R^d}.
\end{aligned}
\end{equation}
The norms here are from $(\mathbb R^d)^r$ to $\mathsf H_m$.
To apply the scalar mean-value theorem, evaluate on unit directions
and pair the output with a unit vector of $\mathsf H_m$; then take the
supremum. This uses the bound in \eqref{eq:coordinate-feature-growth},
including $r=3$, but no continuity of $\fD^3\phi_m$.

\par\noindent\textbf{Step 2: sum the coordinate Taylor remainders.}
Fix $D>\max\{1,\|x\|_{\mathbb R^d}\}$ and $\|h\|_{\mathbb R^d}<D-\|x\|_{\mathbb R^d}$; the segment $x+sh$
is in the ball by the convexity calculation in Step 1.
Coordinate-wise integration gives
\[
\begin{aligned}
 \phi_m(x+h)-\phi_m(x)-\fD\phi_m(x)[h]
 &=\int_0^1\{\fD\phi_m(x+sh)-\fD\phi_m(x)\}[h]\,ds,\\
 \{\fD\phi_m(x+h)-\fD\phi_m(x)\}[v]-\fD^2\phi_m(x)[h,v]
 &=\int_0^1\{\fD^2\phi_m(x+sh)-\fD^2\phi_m(x)\}[h,v]\,ds.
\end{aligned}
\]
Apply \eqref{eq:coordinate-feature-spatial-difference} with points $x+sh,x$,
whose distance is $s\|h\|_{\mathbb R^d}$. For unit $v$, each right-hand side has norm
at most $L_m(1+D^m)\|h\|_{\mathbb R^d}^2\int_0^1s\,ds
=\tfrac12 L_m(1+D^m)\|h\|_{\mathbb R^d}^2$.
The weighted norm of a sequence is at most the weighted sum of its
coordinate norms, by the finite-partial-sum inequality
$(\sum_{m\le n}b_m^2)^{1/2}\le\sum_{m\le n}b_m$ for $b_m\ge0$
and then $n\to\infty$. Consequently,
\[
\begin{aligned}
 &\|J_0(x+h)-J_0(x)-(\fD\phi_m(x)[h])_m\|_{\mathsf H_a}\\
 &\qquad\le\tfrac12\left(\sum_m a_mL_m(1+D^m)\right)\|h\|_{\mathbb R^d}^2,\\
 &\sup_{\|v\|_{\mathbb R^d}\le1}
 \|(\{\fD\phi_m(x+h)-\fD\phi_m(x)\}[v]-\fD^2\phi_m(x)[h,v])_m\|_{\mathsf H_a}\\
 &\qquad\le\tfrac12\left(\sum_m a_mL_m(1+D^m)\right)\|h\|_{\mathbb R^d}^2.
\end{aligned}
\]
Dividing by $\|h\|_{\mathbb R^d}$ proves the two derivative identities. Summing
\eqref{eq:coordinate-feature-spatial-difference} in the same weighted
norm gives \eqref{eq:weighted-feature-spatial-difference}, so these
derivatives are continuous and $J_0$ is $C^2$.
\end{proof}

\begin{lemma}[Local comparison bounds with common moment weights]
\label{lemma-localized-moment-smoothness-consequences}
Assume \cref{assumption-structured-moment-smoothness}. Fix $K>0$, $D\ge1$,
and define
\begin{equation}
\label{eq:common-local-moment-weights}
 \gamma_m(K,D):=\mathrm L_{K,D}a_m
              =\mathrm L_{K,D}e^{-c_am^{\kappa_{\rm dec}}}.
\end{equation}
Let $\rho,\nu$ be in the domain of $F$ in
\cref{eq:moment-objective}, with
$M(\rho),M(\nu)\in\mathcal V_{K,D}$, and let $x,x',y,y'$ be in
the ball of radius $D$. Write
$\Delta_m:=\|M_m(\rho)-M_m(\nu)\|_{\mathsf H_m}$.
\begin{enumerate}[label=(\roman*),leftmargin=*]
\item \textbf{Dependence on the measure.}
\begin{align}
 |F(\rho)-F(\nu)|\vee
 \|\nabla_WF(\rho,x)-\nabla_WF(\nu,x)\|_{\mathbb R^d}
 &\le\sum_m\gamma_m(K,D)\Delta_m,
 \label{eq:local-value-gradient-moment-bound}\\
 \max\bigl\{\|\nabla_x\nabla_WF(\rho,x)-\nabla_x\nabla_WF(\nu,x)\|_{\mathbb R^d\to\mathbb R^d},
 &\notag\\[-0.5ex]
 \|\nabla_W^2F(\rho;x,y)-\nabla_W^2F(\nu;x,y)\|_{\mathbb R^d\to\mathbb R^d}\bigr\}
 &\le\tfrac12\sum_m\gamma_m(K,D)\Delta_m.
 \label{eq:local-two-hessians-moment-bound}
\end{align}
In particular the \emph{sum} of the two Hessian differences in
\eqref{eq:local-two-hessians-moment-bound} is at most
$\sum_m\gamma_m(K,D)\Delta_m$.
\item \textbf{Magnitude and spatial dependence.}
\begin{equation}
\label{eq:local-field-magnitude-bound}
 \|\nabla_WF(\nu,x)\|_{\mathbb R^d}\vee
 \|\nabla_x\nabla_WF(\nu,x)\|_{\mathbb R^d\to\mathbb R^d}\vee
 \|\nabla_W^2F(\nu;x,y)\|_{\mathbb R^d\to\mathbb R^d}\le\mathrm L_{K,D}.
\end{equation}
The spatial differences satisfy
\begin{align}
 \|\nabla_WF(\nu,x)-\nabla_WF(\nu,x')\|_{\mathbb R^d}
 &\le\mathrm L_{K,D}\|x-x'\|_{\mathbb R^d},
 \label{eq:local-gradient-spatial-bound}\\
 \|\nabla_x\nabla_WF(\nu,x)-\nabla_x\nabla_WF(\nu,x')\|_{\mathbb R^d\to\mathbb R^d}
 &\le\mathrm L_{K,D}\|x-x'\|_{\mathbb R^d},
 \label{eq:local-jacobian-spatial-bound}\\
 \|\nabla_W^2F(\nu;x,y)-\nabla_W^2F(\nu;x',y')\|_{\mathbb R^d\to\mathbb R^d}
 &\le\mathrm L_{K,D}(\|x-x'\|_{\mathbb R^d}+\|y-y'\|_{\mathbb R^d}).
 \label{eq:local-kernel-spatial-bound}
\end{align}
\end{enumerate}
For the localized comparison we abbreviate
$\gamma_m(D):=\gamma_m(K_{\rm pop},D)=\mathrm L_Da_m$.
\end{lemma}

\begin{proof}
\noindent\textbf{Step 1: recall the derivative representations and their bounds.}
Within this proof set
\[
 Q:=1+g_0+h_0+C_3,\qquad
 R:=1+\sum_m a_mL_m\{2+D^m+K^m m^{m/\alpha}\},
 \qquad \mathrm L_{K,D}=4QR^3.
\]
On the stated moment and spatial regions,
\cref{eq:local-anchor-intermediate,eq:weighted-feature-derivative-bound} give
\begin{equation}
\label{eq:local-proof-bounds}
\begin{gathered}
 \|\fD G\|_{\mathsf H_a\to\mathbb R}\le QR^2,\qquad
 \|\fD^2G\|_{\mathsf H_a\times\mathsf H_a\to\mathbb R}\le QR,\\
 \|\fD^rJ_0(x)\|_{(\mathbb R^d)^r\to\mathsf H_a}\le R
       \qquad(r=1,2).
\end{gathered}
\end{equation}
For $\theta_\nu:=M(\nu)$ and $u,v\in\mathbb R^d$, the moment chain rule for
\cref{eq:moment-objective} is
\begin{align}
 \langle\nabla_WF(\nu,x),v\rangle
 &=\fD G(\theta_\nu)[\fD J_0(x)v],
 \label{eq:weighted-gradient-representation}\\
 \langle u,\nabla_x\nabla_WF(\nu,x)v\rangle
 &=\fD G(\theta_\nu)[\fD^2J_0(x)[u,v]],
 \label{eq:weighted-spatial-hessian-representation}\\
 \langle u,\nabla_W^2F(\nu;x,y)v\rangle
 &=\fD^2G(\theta_\nu)[\fD J_0(x)u,\fD J_0(y)v].
 \label{eq:weighted-integral-hessian-representation}
\end{align}
These are the coordinate chain-rule sums, with the directions collected
in $\mathsf H_a$. For example the first sum is
$\sum_m\langle\partial_mG(\theta_\nu),\fD\phi_m(x)v\rangle$.
Finite coordinate sums converge to the displayed forms because the
directions converge in $\mathsf H_a$ by
\cref{lem:localized-weighted-feature-envelopes}, and $\fD G,\fD^2G$ are
bounded forms on $\mathsf H_a$ by
\cref{assumption-structured-moment-smoothness}(ii).

\noindent\textbf{Step 2: change the measure along its moment segment.}
Set $\theta_\rho:=M(\rho)$, $\Delta:=\theta_\rho-\theta_\nu$, and
$\theta_s:=\theta_\nu+s\Delta\in\mathcal V_{K,D}$. By
\cref{eq:moment-value-integral,eq:moment-gradient-integral,eq:local-proof-bounds},
\begin{align*}
 |F(\rho)-F(\nu)|
 &\le\int_0^1QR^2\|\Delta\|_{\mathsf H_a}\,ds
  =QR^2\|\Delta\|_{\mathsf H_a},\\
 \|\fD G(\theta_\rho)-\fD G(\theta_\nu)\|_{\mathsf H_a\to\mathbb R}
 &\le\int_0^1QR\|\Delta\|_{\mathsf H_a}\,ds
  =QR\|\Delta\|_{\mathsf H_a}.
\end{align*}
Substituting the second inequality in
\cref{eq:weighted-gradient-representation,eq:weighted-spatial-hessian-representation}
and taking suprema over unit directions gives
\[
\begin{aligned}
 \|\nabla_WF(\rho,x)-\nabla_WF(\nu,x)\|_{\mathbb R^d}
 &\le QR\|\Delta\|_{\mathsf H_a}\|\fD J_0(x)\|_{\mathbb R^d\to\mathsf H_a}
  \le QR^2\|\Delta\|_{\mathsf H_a},\\
 \|\nabla_x\nabla_WF(\rho,x)-\nabla_x\nabla_WF(\nu,x)\|_{\mathbb R^d\to\mathbb R^d}
 &\le QR\|\Delta\|_{\mathsf H_a}\|\fD^2J_0(x)\|_{\mathbb R^d\times\mathbb R^d\to\mathsf H_a}
  \le QR^2\|\Delta\|_{\mathsf H_a}.
\end{aligned}
\]
For the integral Hessian, use
\cref{eq:moment-hessian-lipschitz,eq:weighted-integral-hessian-representation}:
\[
\begin{aligned}
 &\|\nabla_W^2F(\rho;x,y)-\nabla_W^2F(\nu;x,y)\|_{\mathbb R^d\to\mathbb R^d}\\
 &\quad\le C_3\|\Delta\|_{\mathsf H_a}
       \|\fD J_0(x)\|_{\mathbb R^d\to\mathsf H_a}
       \|\fD J_0(y)\|_{\mathbb R^d\to\mathsf H_a}
 \le QR^2\|\Delta\|_{\mathsf H_a}.
\end{aligned}
\]
Finally,
\[
 \|\Delta\|_{\mathsf H_a}\le\sum_m a_m\Delta_m,
 \qquad QR^2\le\tfrac12\mathrm L_{K,D}
 \quad\text{because }\mathrm L_{K,D}=4QR^3,\ R\ge1.
\]
This proves both inequalities in (i), with room to sum the two Hessian errors.

\noindent\textbf{Step 3: change the spatial variables.}
The three representations and \eqref{eq:local-proof-bounds} give
\[
\begin{aligned}
 \|\nabla_WF(\nu,x)\|_{\mathbb R^d}&\le QR^2R,\\
 \|\nabla_x\nabla_WF(\nu,x)\|_{\mathbb R^d\to\mathbb R^d}&\le QR^2R,\\
 \|\nabla_W^2F(\nu;x,y)\|_{\mathbb R^d\to\mathbb R^d}&\le QRR^2.
\end{aligned}
\]
All three are at most $QR^3\le\mathrm L_{K,D}$.
Equation \eqref{eq:weighted-feature-spatial-difference} gives
$\|\fD^rJ_0(x)-\fD^rJ_0(x')\|_{(\mathbb R^d)^r\to\mathsf H_a}\le R\|x-x'\|_{\mathbb R^d}$ for $r=1,2$.
Inserting these bounds in the first two representations gives
\cref{eq:local-gradient-spatial-bound,eq:local-jacobian-spatial-bound},
since $QR^2R\le\mathrm L_{K,D}$.
For the kernel, change one slot at a time:
\[
\begin{aligned}
 &\fD^2G(\theta_\nu)[\fD J_0(x)u,\fD J_0(y)v]
      -\fD^2G(\theta_\nu)[\fD J_0(x')u,\fD J_0(y')v]\\
 &=\fD^2G(\theta_\nu)[(\fD J_0(x)-\fD J_0(x'))u,\fD J_0(y)v]\\
 &\quad+\fD^2G(\theta_\nu)[\fD J_0(x')u,(\fD J_0(y)-\fD J_0(y'))v].
\end{aligned}
\]
For unit $u,v$, its absolute value is at most
$QR\{R^2\|x-x'\|_{\mathbb R^d}+R^2\|y-y'\|_{\mathbb R^d}\}$.
Taking the supremum proves \eqref{eq:local-kernel-spatial-bound}.
\end{proof}

%% file: appendix_final_general/general_bound_r_t_bound_including_jump_bounded_by_gamma.tex
\subsection{Bound of $r_t$ including jump errors}
\label{subsec:general-argument-r_t-jump-errors}

The jump analysis has one purpose: compare two Gaussian perturbation vectors
using the same seed, so that the error is measured by covariance mismatch.

At a jump time \(t_\ell\), the reference particles are defined by evaluating the
population Gaussian field:
\[
Y_{t_\ell}^i
=
Y_{t_\ell-}^i
+
\eta_\ell
\xi_{\mu_{t_\ell-}}(Y_{t_\ell-}^i),
\qquad
\xi_{\mu_{t_\ell-}}\sim GP(0,K_{\mu_{t_\ell-}}).
\]
Thus, conditional on the population Gaussian fields and the jump schedule,
the reference particles are deterministic pushforwards of their i.i.d. initial
particles and remain conditionally i.i.d. with common law \(\mu_t\).

We now pass from the Gaussian field to its finite evaluation vector at the
reference particles.

For the jump comparison only, we use the following finite-dimensional
representation. Conditional on the full pre-jump history \(\mathcal H_{\ell-}\), the vector
\[
\big(\xi_{\mu_{t_\ell-}}(Y_{t_\ell-}^1),\dots,
\xi_{\mu_{t_\ell-}}(Y_{t_\ell-}^N)\big)
\]
is a centered Gaussian vector with covariance
\(C_{\mu_{t_\ell-}}(Y_{t_\ell-})\). Hence it can be represented as
\(B_{\mu_{t_\ell-}}(Y_{t_\ell-})g_\ell^N\), where
\(B_{\mu_{t_\ell-}}(Y_{t_\ell-})B_{\mu_{t_\ell-}}(Y_{t_\ell-})^\top
= C_{\mu_{t_\ell-}}(Y_{t_\ell-})\).
This representation is used only inside the covariance-mismatch estimate.

Covariance-factor couplings have a Procrustes/Wasserstein precedent in
\citet{masarotto2019procrustes}. Here we choose a factor that preserves the
population covariance and admits a direct square-root bound. We do not
need to solve the Procrustes minimization problem.
\begin{lemma}[Aligned construction of the finite square root]
\label{lemma-aligned-finite-square-root}
Let $\mu$ be a law in the Gaussian perturbation model
\cref{assumption-common-seed}, and let
$\mathbf Y=(Y^1,\dots,Y^N)\in (\R^d)^N$ have finite covariance
evaluations as in that model. Write
$M:=\mathbf M^N(\rho_{\mathbf Y}^N,\mathbf Y)\in \R^{Nd\times Nd},\;
M_{ij}
:=
\frac1{\sqrt N}\nabla_W^2F(\rho_{\mathbf Y}^N;Y^i,Y^j),$
as well as
$
C:=\bigl[K_\mu(Y^i,Y^j)\bigr]_{i,j=1}^N\in \R^{Nd\times Nd}.
$
Then $M$ and $C$ are symmetric, $C$ is positive semidefinite, and there exists
an orthogonal matrix $Q\in\R^{Nd\times Nd}$ such that
\[
M=|M|Q,
\qquad
|M|:=(M^2)^{1/2}.
\]
If we define
\[
B:=C^{1/2}Q,
\]
then
\[
BB^\top=C,
\]
and
\[
\|B-M\|_{F;\,\mathbb R^{dN}\to\mathbb R^{dN}}=\|C^{1/2}-|M|\|_{F;\,\mathbb R^{dN}\to\mathbb R^{dN}}.
\]
In particular,
\[
\|B-M\|_{F;\,\mathbb R^{dN}\to\mathbb R^{dN}}^2
\le
\|C-M^2\|_{*;\,\mathbb R^{dN}\to\mathbb R^{dN}},
\]
where $\norm{\cdot}_*$ is the nuclear norm (sum of singular values).
\end{lemma}

\begin{proof}
\noindent\textbf{Step 1: check symmetry and positivity.}
Since each block of $M$ is given by
\[
M_{ij}=\frac1{\sqrt N}\nabla_W^2F(\rho_{\mathbf Y}^N;Y^i,Y^j),
\]
the assumed symmetry of $\nabla_W^2F$ yields symmetry of $M$.
Similarly, the kernel matrix $C$ is also symmetric.

Moreover, for any $v=(v_1,\dots,v_N)\in \R^{Nd}$,
\[
v^\top C v
=
\sum_{i,j=1}^N
v_i^\top K_\mu(Y^i,Y^j)v_j.
\]
By the definition of $K_\mu$,
\[
v^\top C v
=
\int
\sum_{i,j=1}^N
v_i^\top
\nabla_W^2F(\mu;Y^i,z)\nabla_W^2F(\mu;z,Y^j)
v_j\,
\mu(dz).
\]
Since $\nabla_W^2F(\mu;z,Y^j)=\nabla_W^2F(\mu;Y^j,z)^\top$, this becomes
\[
v^\top C v
=
\int
\Bigl\|
\sum_{i=1}^N
\nabla_W^2F(\mu;Y^i,z)^\top v_i
\Bigr\|_{\mathbb R^d}^2
\mu(dz)
\ge 0.
\]
Hence $C$ is positive semidefinite.

\par\noindent\textbf{Step 2: construct the aligned factor.}
For the spectral decomposition
$M=U\operatorname{diag}(\lambda_j)U^\top$, define
\[
 Q:=U\operatorname{diag}(\operatorname{sign}_+(\lambda_j))U^\top,
 \qquad
 \operatorname{sign}_+(a):=
 \begin{cases}-1,&a<0,\\1,&a\ge0.\end{cases}
\]
Then $QQ^\top=U\operatorname{diag}(1)U^\top=I$, including on the
zero eigenspace, and
\[
 |M|Q=U\operatorname{diag}(|\lambda_j|\operatorname{sign}_+(\lambda_j))U^\top=M.
\]
Define
\[
B:=C^{1/2}Q.
\]
Then
\[
BB^\top
=
C^{1/2}QQ^\top C^{1/2}
=
C.
\]
On the other hand,
\[
B-M
=
C^{1/2}Q-|M|Q
=
(C^{1/2}-|M|)Q.
\]
Right orthogonal invariance follows from
$\|(C^{1/2}-|M|)Q\|_{F;\,\mathbb R^{dN}\to\mathbb R^{dN}}^2
=\operatorname{tr}((C^{1/2}-|M|)QQ^\top(C^{1/2}-|M|)^\top)
=\|C^{1/2}-|M|\|_{F;\,\mathbb R^{dN}\to\mathbb R^{dN}}^2$. Therefore
\[
\|B-M\|_{F;\,\mathbb R^{dN}\to\mathbb R^{dN}}
=
\|(C^{1/2}-|M|)Q\|_{F;\,\mathbb R^{dN}\to\mathbb R^{dN}}
=
\|C^{1/2}-|M|\|_{F;\,\mathbb R^{dN}\to\mathbb R^{dN}}.
\]

\par\noindent\textbf{Step 3: bound the factor difference by the covariance difference.}
The estimate proved in this step is the finite-dimensional
Powers--St{\o}rmer inequality~\citep{powers1970free}; we include the calculation
to keep the norms and constants explicit.
For completeness, set $S=C^{1/2}-|M|$ in the following calculation, and
write $S=\sum_js_je_je_j^\top$ in an orthonormal eigenbasis and define
$\operatorname{sign}(S)=\sum_j\operatorname{sign}(s_j)e_je_j^\top$,
where $\operatorname{sign}(a)$ is $1$, $0$, or $-1$ according as
$a>0$, $a=0$, or $a<0$. This differs from
$\operatorname{sign}_+(0)=1$ used to make $Q$ orthogonal.
The product identity follows by cancellation:
\[
 C^{1/2}S+S|M|
 =C-C^{1/2}|M|+C^{1/2}|M|-|M|^2=C-M^2.
\]
Thus
\[
 C-M^2=C^{1/2}S+S|M|,\qquad
 \|\operatorname{sign}(S)\|_{\mathbb R^{dN}\to\mathbb R^{dN}}\le1,\qquad
 S\operatorname{sign}(S)=\operatorname{sign}(S)S=|S|.
\]
For any symmetric matrix $A=\sum_j\lambda_je_je_j^\top$ and any
matrix $J$ with $\|J\|_{\mathbb R^{dN}\to\mathbb R^{dN}}\le1$,
$\operatorname{tr}(JA)=\sum_j\lambda_j\langle e_j,Je_j\rangle
\le\sum_j|\lambda_j|=\|A\|_{*;\,\mathbb R^{dN}\to\mathbb R^{dN}}$. Apply this fact to $A=C-M^2$
and $J=\operatorname{sign}(S)$, then use cyclicity of the trace:
\[
\begin{aligned}
 \|C-M^2\|_{*;\,\mathbb R^{dN}\to\mathbb R^{dN}}
 &\ge\operatorname{tr}(\operatorname{sign}(S)(C^{1/2}S+S|M|))\\
 &=\operatorname{tr}(S\operatorname{sign}(S)C^{1/2})
        +\operatorname{tr}(\operatorname{sign}(S)S|M|)\\
 &=\operatorname{tr}(|S|(C^{1/2}+|M|)).
\end{aligned}
\]
For each $e_j$, both $\langle e_j,C^{1/2}e_j\rangle$ and
$\langle e_j,|M|e_j\rangle$ are nonnegative. The scalar inequality
$a+b\ge|a-b|$ for $a,b\ge0$ therefore gives
\[
 \langle e_j,(C^{1/2}+|M|)e_j\rangle
 \ge|\langle e_j,(C^{1/2}-|M|)e_j\rangle|=|s_j|.
\]
Consequently,
\[
\begin{aligned}
 \|C-M^2\|_{*;\,\mathbb R^{dN}\to\mathbb R^{dN}}
 &\ge\sum_j|s_j|\langle e_j,(C^{1/2}+|M|)e_j\rangle
 \ge\sum_j|s_j|^2\\
 &=\|C^{1/2}-|M|\|_{F;\,\mathbb R^{dN}\to\mathbb R^{dN}}^2
 =\|B-M\|_{F;\,\mathbb R^{dN}\to\mathbb R^{dN}}^2.
\end{aligned}
\]
No nonsingularity or optimality of the chosen factor is used.
\end{proof}

\begin{lemma}[Population-first finite coupling]
\label{lem:population-first-finite-coupling}
Fix a prescribed jump $t_\ell$ in
\cref{definition-ode-and-jump,assumption-common-seed}. Recall that
$\mathcal H_{\ell-}$ is the pre-jump history: it includes the initial
sample, previous population innovations, previous auxiliary normals, and
the pre-jump states, but not the current or future innovations.
Generate the prescribed population field and define
\[
 v:=\bigl(\xi_{\mu_{t_\ell-}}(Y_{t_\ell-}^i)\bigr)_{i=1}^N,
 \qquad
 C:=\bigl[K_{\mu_{t_\ell-}}(Y_{t_\ell-}^i,Y_{t_\ell-}^j)\bigr]_{i,j=1}^N.
\]
Thus $v\mid\mathcal H_{\ell-}\sim N(0,C)$. Recall the Hessian factors:
\[
\begin{gathered}
 (M_Z)_{ij}:=\frac1{\sqrt N}
 \nabla_W^2F(\rho_{\mathbf Z_{t_\ell-}};Z_{t_\ell-}^i,Z_{t_\ell-}^j),
 \quad Z\in\{X,Y\},\\
 Q:=\operatorname{sign}_+(M_Y),\qquad B:=C^{1/2}Q.
\end{gathered}
\]
This local factor \(B\) is denoted by \(B_{\rm pop}\) in the main text.
On an extension of the probability space, take $h_\ell\sim N(0,I_{Nd})$
independent of the original population system, the initial sample, and all
other auxiliary normals. Define
\[
\mathbf g_\ell^N=B^\dagger v+(I-B^\dagger B)h_\ell,
\]
where $B^\dagger$ is the Moore--Penrose inverse, defined below. Then:
\begin{enumerate}
\item The finite coordinate is conditionally standard Gaussian and
reproduces the original population evaluation:
\[
\mathbf g_\ell^N\mid\mathcal H_{\ell-}\sim N(0,I_{Nd}),\qquad
B\mathbf g_\ell^N=v\quad\text{a.s.}
\]
\item The factor error satisfies
\[
\|B-M_Y\|_{F;\,\mathbb R^{dN}\to\mathbb R^{dN}}^2\le\|C-M_Y^2\|_{*;\,\mathbb R^{dN}\to\mathbb R^{dN}}.
\]
\item The perturbation $M_X\mathbf g_\ell^N$ has the prescribed empirical
conditional covariance
\[
[K_{\rho_{\mathbf X_{t_\ell-}}}
(X_{t_\ell-}^i,X_{t_\ell-}^j)]_{i,j}.
\]
Repeating the construction with
fresh $h_\ell$ couples all prescribed jumps while retaining the original
population path and the conditional product law for the reference particles.
\end{enumerate}
\end{lemma}
\begin{proof}
\noindent\textbf{Step 1: the pre-jump matrices and the inverse.}
The evaluation points and $\mu_{t_\ell-}$ are fixed after conditioning on
$\mathcal H_{\ell-}$. The fresh field therefore has the asserted finite
Gaussian law by \cref{assumption-common-seed}; joint measurability is part
of that model. All matrices $C,M_X,M_Y,Q,B$ are measurable before the
jump. For $Q$ no measurable choice of eigenvectors is needed: the sign in
\cref{lemma-aligned-finite-square-root} is the Borel map
\[
 Q=I-2\lim_{k\to\infty}
 (|M_Y|-M_Y)(|M_Y|-M_Y+k^{-1}I)^{-1}.
\]
On an eigenvalue $\lambda$, the ratio tends to $1$ if $\lambda<0$ and
to $0$ if $\lambda\ge0$. The positive square root is continuous.

For a singular-value decomposition $B=U\operatorname{diag}(s_j)V^\top$,
with $U,V$ orthogonal and $s_j\ge0$,
define its Moore--Penrose inverse by replacing each positive $s_j$ with
$s_j^{-1}$ and each zero $s_j$ with zero:
\[
 B^\dagger=V\operatorname{diag}(s_j^{-1}\mathbf1_{\{s_j>0\}})U^\top,
 \qquad
 P_B:=B^\dagger B
 =V\operatorname{diag}(\mathbf1_{\{s_j>0\}})V^\top.
\]
The displayed diagonal convention avoids evaluating $0^{-1}$.
It gives $P_B^\top=P_B$ and
\[
\begin{gathered}
 P_B^2=V\operatorname{diag}(\mathbf1_{\{s_j>0\}}^2)V^\top=P_B,
 \qquad BB^\dagger=U\operatorname{diag}(\mathbf1_{\{s_j>0\}})U^\top,\\
 B(I-P_B)=U\operatorname{diag}(s_j\mathbf1_{\{s_j=0\}})V^\top=0,
 \qquad (I-BB^\dagger)B=0.
\end{gathered}
\]
The last equality has the same diagonal entries $s_j\mathbf1_{\{s_j=0\}}$. The inverse is Borel measurable even when the rank
changes, because
\[
 B^\dagger=\lim_{k\to\infty}(B^\top B+k^{-1}I)^{-1}B^\top;
\]
the singular-value coefficient on the right is $s_j/(s_j^2+k^{-1})$.

\par\noindent\textbf{Step 2: the conditional standard Gaussian law.}
The preceding lemma gives $BB^\top=C$. Conditional on $\mathcal H_{\ell-}$,
the terms $B^\dagger v$ and $(I-P_B)h_\ell$ are independent centered
Gaussians, since $h_\ell$ is independent of $\sigma(\mathcal H_{\ell-},v)$.
Since both conditional means are zero, their covariance matrices are
\begin{equation}
\label{eq:population-whitening-covariances}
\begin{aligned}
 \mathbb E[B^\dagger vv^\top(B^\dagger)^\top\mid\mathcal H_{\ell-}]
 &=B^\dagger C(B^\dagger)^\top
 =(B^\dagger B)(B^\dagger B)^\top=P_B,\\
 \mathbb E[(I-P_B)h_\ell h_\ell^\top(I-P_B)^\top\mid\mathcal H_{\ell-}]
 &=(I-P_B)I(I-P_B)^\top=I-2P_B+P_B^2=I-P_B.
\end{aligned}
\end{equation}
Conditional independence makes the cross covariance zero, so the sum
has covariance $P_B+(I-P_B)=I$.
For every $\theta\in\mathbb R^{Nd}$, their characteristic functions give
\[
 \mathbb E[e^{i\theta^\top\mathbf g_\ell^N}\mid\mathcal H_{\ell-}]
 =e^{-\theta^\top P_B\theta/2}
    e^{-\theta^\top(I-P_B)\theta/2}
 =e^{-\|\theta\|_{\mathbb R^{dN}}^2/2}.
\]
This proves the conditional $N(0,I_{Nd})$ law, including singular $C$.

\par\noindent\textbf{Step 3: retain the original population update.}
Since $(I-BB^\dagger)B=0$ and $C=BB^\top$,
\[
 \begin{aligned}
 \mathbb E[\|(I-BB^\dagger)v\|_{\mathbb R^{dN}}^2\mid\mathcal H_{\ell-}]
 &=\operatorname{tr}((I-BB^\dagger)BB^\top(I-BB^\dagger)^\top)\\
 &=\operatorname{tr}([(I-BB^\dagger)B][(I-BB^\dagger)B]^\top)=0.
\end{aligned}
\]
The squared norm is nonnegative, so $BB^\dagger v=v$ almost surely and
\[
 B\mathbf g_\ell^N=BB^\dagger v+B(I-P_B)h_\ell=v.
\]
If $C=0$, then $v=B=B^\dagger=0$ and $\mathbf g_\ell^N=h_\ell$,
which verifies this case directly. The factor error is the bound in
\cref{lemma-aligned-finite-square-root} with $M=M_Y$.

\par\noindent\textbf{Step 4: the empirical transition.}
Since $M_X$ is pre-jump measurable,
$M_X\mathbf g_\ell^N\mid\mathcal H_{\ell-}\sim N(0,M_XM_X^\top)$.
For this calculation write $X^i=X_{t_\ell-}^i$ and
$\rho_X=N^{-1}\sum_k\delta_{X^k}$. Kernel symmetry and the definition of
$K_\nu$ in \cref{assumption-common-seed} give
\[
\begin{aligned}
 (M_XM_X^\top)_{ij}
 &=\frac1N\sum_{k=1}^N
 \nabla_W^2F(\rho_X;X^i,X^k)
 \nabla_W^2F(\rho_X;X^j,X^k)^\top\\
 &=\frac1N\sum_{k=1}^N
 \nabla_W^2F(\rho_X;X^i,X^k)
 \nabla_W^2F(\rho_X;X^k,X^j)\\
 &=\int\nabla_W^2F(\rho_X;X^i,z)\nabla_W^2F(\rho_X;z,X^j)\rho_X(dz)
 =K_{\rho_X}(X^i,X^j).
\end{aligned}
\]
This is exactly the required empirical conditional law.

\par\noindent\textbf{Step 5: iterate without changing the reference law.}
Keep the original population innovations and iid initial sample
$U_1,\ldots,U_N\sim\mu_0$, and add independent
$h_1,\ldots,h_{\NofJ}$. Suppose the two pre-jump configurations at
$t_\ell$ have been constructed from the initial sample and the first
$\ell-1$ innovations and auxiliary normals. Steps 1--4 give the next
coordinate, update $X$ by $\eta_\ell M_X\mathbf g_\ell^N$, and update
$Y$ by its original field value $\eta_\ell v$. The ODEs then give
configurations measurable with respect to the next pre-jump history.
This is a finite induction over the prescribed schedule.

The conditional standard Gaussian law is constant in the history. For
$A\in\mathcal H_{\ell-}$ and a Borel set $D\subset\mathbb R^{Nd}$,
\[
 \mathbb P(A\cap\{\mathbf g_\ell^N\in D\})
 =\mathbb E[\mathbf1_A\mathbb P(\mathbf g_\ell^N\in D\mid\mathcal H_{\ell-})]
 =\mathbb P(A)\,N(0,I_{Nd})(D).
\]
Thus each coordinate is independent of its pre-jump history, which includes
previous coordinates. Independence of the current field or post-jump
reference particles is not asserted. The population system is unaltered:
conditional on its fields, the ODE and jump maps compose into a common map
$\mathcal T_t$ with
\[
 Y_t^i=\mathcal T_t(U_i),\qquad\mu_t=(\mathcal T_t)_\#\mu_0.
\]
The initial sample is independent of all population innovations, so
$\operatorname{Law}((U_1,\ldots,U_N)\mid\mathcal G_{\rm pop})=\mu_0^{\otimes N}$.
For bounded measurable $f_1,\ldots,f_N$, integration against this product
law and the pushforward identity give
\begin{equation}
\label{eq:population-pushforward-product-law}
\begin{aligned}
 \mathbb E\left[\prod_i f_i(Y_t^i)\,\middle|\,\mathcal G_{\rm pop}\right]
 &=\int\prod_i f_i(\mathcal T_t(u_i))\prod_i\mu_0(du_i)\\
 &=\prod_i\int f_i(\mathcal T_t(u_i))\mu_0(du_i)
 =\prod_i\int f_i(y)\mu_t(dy).
\end{aligned}
\end{equation}
This is the product law used in \cref{lemma-gp-jump-implies-conditional-iid};
$\mathbf g_\ell^N$ is not included in its conditioning sigma-field.

\par\noindent\textbf{Step 6: preserve exchangeability of particle pairs.}
Let $P$ be the orthogonal matrix of any permutation of the $N$ particle
blocks. Permute the initial sample and every auxiliary normal by $P$,
so $h_\ell'=Ph_\ell\sim N(0,PIP^\top)=N(0,I)$. These transformed
inputs have the same joint law, with the population innovations held fixed. Under simultaneous permutation of $\mathbf X,\mathbf Y$, the
empirical measures are unchanged and the evaluation vector becomes $Pv$.
The covariance and Hessian factors therefore transform as
\[
\begin{gathered}
 C'=PCP^\top,\qquad M_Z'=PM_ZP^\top,\qquad Q'=PQP^\top,\\
 B'=PBP^\top,\qquad (B')^\dagger=PB^\dagger P^\top.
\end{gathered}
\]
For the square root, $(PC^{1/2}P^\top)^2=PCP^\top$ and the first matrix
is positive semidefinite; uniqueness of the positive square root gives the identity.
The formulas for $Q$ and $B^\dagger$ in Step 1 give the other identities
by conjugating each matrix and its inverse. Hence
\[
 (B')^\dagger Pv+[I-(B')^\dagger B']Ph_\ell
 =P\{B^\dagger v+(I-B^\dagger B)h_\ell\}
 =P\mathbf g_\ell^N.
\]
The empirical increment transforms to $PM_X\mathbf g_\ell^N$, and the
reference increment to $Pv$. The drift formulas also commute with $P$.
The iid initial sample and independent standard normals have
permutation-invariant joint law, with the population innovations unchanged.
Induction over ODE intervals and jumps therefore shows that
$(X_t^i,Y_t^i)_{i=1}^N$ is exchangeable. This gives identical one-particle
moment bounds; it does not assert independence of the empirical particles.
\end{proof}

With the square roots aligned, the Gaussian perturbation error reduces to a
kernel covariance mismatch.

\begin{lemma}[Reduction to kernel mismatch]
\label{lemma-gaussian-error-to-kernel-mismatch}
Recall $M=M_Y=\mathbf M^N(\rho_{\mathbf Y},\mathbf Y)$ and
$B=C^{1/2}\operatorname{sign}_+(M_Y)$ from
\cref{lemma-aligned-finite-square-root,lem:population-first-finite-coupling},
where $\mathbf Y=\mathbf Y_{t_\ell-}$. Let $\mathbf g^N\mid\mathcal H_{\ell-}\sim N(0,I_{Nd})$ be the coordinate constructed above, with $B,M,C$ measurable before the jump.
Then
\[
\mathbb E\bigl[\|(B-M)\mathbf g^N\|_N^2\mid\mathcal H_{\ell-}\bigr]
=
\frac1N\|B-M\|_{F;\,\mathbb R^{dN}\to\mathbb R^{dN}}^2
\le
\frac1N\|C-M^2\|_{*;\,\mathbb R^{dN}\to\mathbb R^{dN}},
\]
where $\norm{\cdot}_N \coloneq \frac{1}{\sqrt{N}}\norm{\cdot}_{\R^{dN}}.$
Equivalently,
\[
\mathbb E\bigl[\|(B-M)\mathbf g^N\|_N^2\mid\mathcal H_{\ell-}\bigr]
\le
\frac1N
\left\|
\bigl[
K_\mu(Y^i,Y^j)-K_{\rho_{\mathbf Y}^N}(Y^i,Y^j)
\bigr]_{i,j=1}^N
\right\|_{*;\,\mathbb R^{dN}\to\mathbb R^{dN}}.
\]
\end{lemma}

\begin{proof}
Writing $A=B-M$ in this calculation, conditional standard Gaussianity gives
\[
\begin{aligned}
 \mathbb E[\|A\mathbf g^N\|_N^2\mid\mathcal H_{\ell-}]
 &=\frac1N\sum_{a,b,c}A_{ab}A_{ac}
       \mathbb E[g_b^Ng_c^N\mid\mathcal H_{\ell-}]\\
 &=\frac1N\sum_{a,b,c}A_{ab}A_{ac}\delta_{bc}
 =\frac1N\sum_{a,b}A_{ab}^2=\frac1N\|A\|_{F;\,\mathbb R^{dN}\to\mathbb R^{dN}}^2.
\end{aligned}
\]
Apply \cref{lemma-aligned-finite-square-root}. The block identity
\[
 (M^2)_{ij}=\frac1N\sum_k\nabla_W^2F(\rho_{\mathbf Y};Y^i,Y^k)
 \nabla_W^2F(\rho_{\mathbf Y};Y^k,Y^j)
 =K_{\rho_{\mathbf Y}}(Y^i,Y^j)
\]
then gives the stated kernel difference.
\end{proof}

The kernel mismatch is split into a smooth measure-dependence term and the
empirical covariance fluctuation of the reference cloud.

\begin{lemma}[Kernel mismatch bound]
\label{lemma-kernel-mismatch-bound}
Write $H_\nu(x,y)=\nabla_W^2F(\nu;x,y)$ for the kernel in this lemma.
Work on $E_{\rm pop}\cap E_{T,D}$ from
\cref{lem:population-moment-event,eq:sample-localization-event}, with
$\mathrm L_D=\mathrm L_{K_{\rm pop},D}$ from \cref{def:localized-envelope}.
By \cref{lemma-reachable-moment-tube}, the moments of $\mu_t,\rho_{\mathbf Y_t}$ and the segment between them lie in the
reachable moment region. For $\|x\|_{\mathbb R^d},\|y\|_{\mathbb R^d}\le D$ and these measures,
Lemma~\ref{lemma-localized-moment-smoothness-consequences} gives
$\|H_\nu(x,y)\|_{\mathbb R^d\to\mathbb R^d}\le \mathrm L_D$. Then the block matrix
\[
\Delta K^N(\mu,\mathbf Y)
:=
\bigl[
K_\mu(Y^i,Y^j)-K_{\rho_{\mathbf Y}^N}(Y^i,Y^j)
\bigr]_{i,j=1}^N
\in\R^{Nd\times Nd}
\]
satisfies
\begin{align}
\frac1N\|\Delta K^N(\mu_t,\mathbf Y)\|_{*;\,\mathbb R^{dN}\to\mathbb R^{dN}}
\le{}&
2d\,\mathrm L_D\Gamma_t^N(D)+\tilde{\Gamma}_t^N
\label{eq:kernel-matrix-bound-good}
\notag
\end{align}
where
\[
\mathbf U_\nu(z)
:=
\begin{pmatrix}
H_\nu(Y^1,z)\\
\vdots\\
H_\nu(Y^N,z)
\end{pmatrix}
\in\R^{Nd\times d},
\]
\[
\tilde{\Gamma}_t^N\coloneq 
\frac1N\left\|
\int \mathbf U_{\mu_t}(z)\mathbf U_{\mu_t}(z)^\top\,
(\mu_t-\rho_{\mathbf Y_t})(dz)
\right\|_{*;\,\mathbb R^{dN}\to\mathbb R^{dN}}
\]
The explicit factor $N$ cancels from the coefficient $2d\mathrm L_D$.
Dependence through $\mathrm L_D=\mathrm L_{K_{\rm pop},D}$ is retained,
as specified in \cref{def:localized-envelope}.
\end{lemma}

\begin{proof}
\noindent\textbf{Step 1: separate the covariance and measure changes.}
Fix $t$ and write $\mathbf Y=\mathbf Y_t$, $\mu=\mu_t$, and
$\rho=\rho_{\mathbf Y_t}$. Since
\[
H_\mu(Y^i,z)^\top=H_\mu(z,Y^i),
\]
we may write
\[
K_\mu(Y^i,Y^j)
=
\int H_\mu(Y^i,z)H_\mu(z,Y^j)\,\mu(dz)
=
\int H_\mu(Y^i,z)H_\mu(Y^j,z)^\top\,\mu(dz).
\]
Therefore, with
\[
\mathbf U_\mu(z)
=
\begin{pmatrix}
H_\mu(Y^1,z)\\
\vdots\\
H_\mu(Y^N,z)
\end{pmatrix}
\in\R^{Nd\times d},
\]
we have the matrix identity
\[
\bigl[K_\mu(Y^i,Y^j)\bigr]_{i,j=1}^N
=
\int \mathbf U_\mu(z)\mathbf U_\mu(z)^\top\,\mu(dz).
\]
Hence
\begin{align*}
\Delta K^N(\mu,\mathbf Y)
&=
\int \mathbf U_\mu(z)\mathbf U_\mu(z)^\top\,(\mu-\rho)(dz)\\
&+
\int \Bigl(\mathbf U_\mu(z)\mathbf U_\mu(z)^\top
-
\mathbf U_\rho(z)\mathbf U_\rho(z)^\top\Bigr)\,\rho(dz).
\\
\end{align*}
Thus, by the triangle inequality,
\[
\|\Delta K^N(\mu,\mathbf Y)\|_{*;\,\mathbb R^{dN}\to\mathbb R^{dN}}
\le I_1+I_2,
\]
where
\[
I_1:=
\int
\bigl\|
\mathbf U_\mu(z)\mathbf U_\mu(z)^\top
-
\mathbf U_\rho(z)\mathbf U_\rho(z)^\top
\bigr\|_{*;\,\mathbb R^{dN}\to\mathbb R^{dN}}
\,\rho(dz),
\]
and
\[
I_2:=
\left\|
\int \mathbf U_\mu(z)\mathbf U_\mu(z)^\top\,(\mu-\rho)(dz)
\right\|_{*;\,\mathbb R^{dN}\to\mathbb R^{dN}}.
\]
The two terms require different estimates. The measure in $I_1$ is the
radius-$D$ empirical law $\rho$, so the localized kernel bounds apply to its
integrand. The signed measure in $I_2$ contains the unbounded population law
$\mu$; it is retained as the covariance residual
$\widetilde\Gamma_t^N$ of
\cref{eq:tilde-gamma-full-definition-weighted} and is controlled by the
weighted-feature concentration argument.

\par\noindent\textbf{Step 2: bound the sample-supported measure change.}
For $X,Y\in\mathbb R^{dN\times d}$ with columns
$X=[x_1\ \cdots\ x_d]$, $Y=[y_1\ \cdots\ y_d]$, the sole possible nonzero singular value of
$x_ky_k^\top$ is $\|x_k\|_{\mathbb R^{dN}}\|y_k\|_{\mathbb R^{dN}}$. The triangle and Cauchy--Schwarz inequalities give
\begin{equation}
\label{eq:nuclear-hilbert-schmidt-product}
\begin{aligned}
 \|XY^\top\|_{*;\,\mathbb R^{dN}\to\mathbb R^{dN}}
 &=\left\|\sum_{k=1}^d x_ky_k^\top\right\|_{*;\,\mathbb R^{dN}\to\mathbb R^{dN}}
 \le\sum_{k=1}^d\|x_k\|_{\mathbb R^{dN}}\|y_k\|_{\mathbb R^{dN}}\\
 &\le\left(\sum_k\|x_k\|_{\mathbb R^{dN}}^2\right)^{1/2}
       \left(\sum_k\|y_k\|_{\mathbb R^{dN}}^2\right)^{1/2}
 =\|X\|_{F;\,\mathbb R^d\to\mathbb R^{dN}}\|Y\|_{F;\,\mathbb R^d\to\mathbb R^{dN}}.
\end{aligned}
\end{equation}
The same argument works for any number of columns. For Hilbert--Schmidt
maps $X,Y:E\to F$ between Hilbert spaces, truncate the common input
orthonormal basis. The nuclear norm
of the omitted sum is at most the product of the two square-summable
column tails, which tends to zero. Thus
$\|XY^*\|_{*;\,F\to F}\le\|X\|_{\mathrm{HS};\,E\to F}\|Y\|_{\mathrm{HS};\,E\to F}$ also holds in that setting.
For $I_1$, use this bound and
$AA^\top-BB^\top=(A-B)A^\top+B(A-B)^\top$.
Hence
\begin{align*}
 &\|\mathbf U_\mu(z)\mathbf U_\mu(z)^\top
          -\mathbf U_\rho(z)\mathbf U_\rho(z)^\top
    \|_{*;\,\mathbb R^{dN}\to\mathbb R^{dN}}\\
 &\quad\le\|\mathbf U_\mu(z)-\mathbf U_\rho(z)\|_{F;\,\mathbb R^d\to\mathbb R^{dN}}
       \|\mathbf U_\mu(z)\|_{F;\,\mathbb R^d\to\mathbb R^{dN}}\\
 &\qquad+\|\mathbf U_\rho(z)\|_{F;\,\mathbb R^d\to\mathbb R^{dN}}
       \|\mathbf U_\mu(z)-\mathbf U_\rho(z)\|_{F;\,\mathbb R^d\to\mathbb R^{dN}}.
\end{align*}
For $\rho$-almost every $z$, \eqref{eq:sample-localization-event} gives
$\|z\|_{\mathbb R^d}\le D$ and $\|Y^i\|_{\mathbb R^d}\le D$. The following local bounds therefore
apply to the integrand of $I_1$:
\[
\|\mathbf U_\eta(z)\|_{F;\,\mathbb R^d\to\mathbb R^{dN}}^2
=
\sum_{i=1}^N \|H_\eta(Y^i,z)\|_{F;\,\mathbb R^d\to\mathbb R^d}^2
\le
d\sum_{i=1}^N \|H_\eta(Y^i,z)\|_{\mathbb R^d\to\mathbb R^d}^2
\le
Nd\,\mathrm L_D^2,
\]
so
\[
\|\mathbf U_\eta(z)\|_{F;\,\mathbb R^d\to\mathbb R^{dN}} \le \sqrt{Nd}\,\mathrm L_D
\qquad
(\eta\in\{\mu,\rho\}).
\]
Moreover, \cref{eq:local-two-hessians-moment-bound} gives the (weaker) bound
\[
\|H_\mu(Y^i,z)-H_\rho(Y^i,z)\|_{\mathbb R^d\to\mathbb R^d}
\le
\Gamma_t^N(D),
\qquad
\Gamma_t^N(D)=
\sum_{m\ge1}\gamma_m(D)\|M_m(\mu)-M_m(\rho)\|_{\mathsf H_m},
\]
where the residual is defined in \eqref{eq:common-moment-residual-and-lipschitz-sum}; hence
\[
\|\mathbf U_\mu(z)-\mathbf U_\rho(z)\|_{F;\,\mathbb R^d\to\mathbb R^{dN}}^2
\le
d\sum_{i=1}^N
\|H_\mu(Y^i,z)-H_\rho(Y^i,z)\|_{\mathbb R^d\to\mathbb R^d}^2
\le
Nd\,\Gamma_t^N(D)^2.
\]
Thus
\[
\|\mathbf U_\mu(z)-\mathbf U_\rho(z)\|_{F;\,\mathbb R^d\to\mathbb R^{dN}} \le \sqrt{Nd}\,\Gamma_t^N(D).
\]
Combining the above bounds,
\[
\bigl\|
\mathbf U_\mu(z)\mathbf U_\mu(z)^\top
-
\mathbf U_\rho(z)\mathbf U_\rho(z)^\top
\bigr\|_{*;\,\mathbb R^{dN}\to\mathbb R^{dN}}
\le
2Nd\,\mathrm L_D\,\Gamma_t^N(D).
\]
Integrating in \(z\) yields
\[
I_1 \le 2Nd\,\mathrm L_D\,\Gamma_t^N(D).
\]
Dividing by $N$ gives
\[
 N^{-1}I_1\le2d\mathrm L_D\Gamma_t^N(D).
\]
The remaining term is exactly $N^{-1}I_2=\widetilde\Gamma_t^N$.
Its integral includes the population law $\mu_t$; no bound $\|z\|_{\mathbb R^d}\le D$
is applied to that integral. This proves the claim.
\end{proof}

We now combine the Lipschitz comparison of empirical perturbations with the
covariance mismatch bound to obtain the jump recursion for \(r_t\).

\begin{lemma}[One-jump comparison]
\label{lem:jump-bound-common-seed}
Use the coupling of Lemma~\ref{lem:population-first-finite-coupling} at
$t_\ell$, and let $u\in(0,1)$. Define
\[
\Lambda(D):=\sum_{m\ge1}\gamma_m(D)L_m(1+2D^{m-1}),\qquad
q(u):=1+\sqrt{2\log(2/u)},
\]
\[
c_J(D):=1+\sqrt d\{\Lambda(D)+2\mathrm L_D\}+\sqrt{2d\mathrm L_D}.
\]
There is an event $\Omega_{\ell,u}^{\rm jump}$ with
$\mathbb P((\Omega_{\ell,u}^{\rm jump})^c\mid\mathcal H_{\ell-})\le u$
such that, on this event intersected with the population moment and sample
localization events,
\[
r_{t_\ell}\le (1+c_J(D)q(u)\eta_\ell)r_{t_\ell-}
+c_J(D)q(u)\eta_\ell
  \bigl(\sqrt{\Gamma_{t_\ell-}^N(D)}+\sqrt{\widetilde\Gamma_{t_\ell-}^N}\bigr).
\]
All residuals and kernel covariances in this statement are evaluated before
the jump. Constants may depend on the deterministic population moment scale
as specified in the localized-envelope convention.
\end{lemma}
\begin{proof}
\noindent\textbf{Step 1: control the two conditional Gaussian errors.}
By \cref{eq:jump-dynamics-XY,lem:population-first-finite-coupling},
\[
 \mathbf X_{t_\ell}-\mathbf Y_{t_\ell}
 =\mathbf X_{t_\ell-}-\mathbf Y_{t_\ell-}
   +\eta_\ell\{(M_X-M_Y)+(M_Y-B)\}\mathbf g_\ell^N.
\]
Divide its triangle inequality by $\sqrt N$ to get
\begin{equation}
\label{eq:jump-estimate-proof}
r_{t_\ell}\le r_{t_\ell-}
+\frac{\eta_\ell}{\sqrt N}\|(M_X-M_Y)\mathbf g_\ell^N\|_{\mathbb R^{dN}}
+\frac{\eta_\ell}{\sqrt N}\|(M_Y-B)\mathbf g_\ell^N\|_{\mathbb R^{dN}}.
\end{equation}
For a pre-jump measurable matrix $A$, the conditional standard Gaussian
law gives $\mathbb E[\|A\mathbf g_\ell^N\|_{\mathbb R^{dN}}^2\mid\mathcal H_{\ell-}]
=\|A\|_{F;\,\mathbb R^{dN}\to\mathbb R^{dN}}^2$ as in \cref{lemma-gaussian-error-to-kernel-mismatch}.
Hence its mean norm is at most $\|A\|_{F;\,\mathbb R^{dN}\to\mathbb R^{dN}}$ by Cauchy--Schwarz. Also
\[
 |\|Ag\|_{\mathbb R^{dN}}-\|Ag'\|_{\mathbb R^{dN}}|\le\|A(g-g')\|_{\mathbb R^{dN}}
 \le\|A\|_{\mathbb R^{dN}\to\mathbb R^{dN}}\|g-g'\|_{\mathbb R^{dN}}.
\]
The Gaussian concentration inequality for a Lipschitz function
\citep{BLM2013}, applied conditionally with deviation
$\|A\|_{\mathbb R^{dN}\to\mathbb R^{dN}}\sqrt{2\log(2/u)}$, has failure probability
$e^{-\log(2/u)}=u/2$. Using $\|A\|_{\mathbb R^{dN}\to\mathbb R^{dN}}\le\|A\|_{F;\,\mathbb R^{dN}\to\mathbb R^{dN}}$ gives
\[
\mathbb P\!\left(\|A\mathbf g_\ell^N\|_{\mathbb R^{dN}}>
(1+\sqrt{2\log(2/u)})\|A\|_{F;\,\mathbb R^{dN}\to\mathbb R^{dN}}\,\middle|\,\mathcal H_{\ell-}\right)\le u/2.
\]
Apply this to the two matrices in \eqref{eq:jump-estimate-proof} and take
a union bound. These Gaussian events are defined before imposing any sample
localization event, so no conditional independence after localization is used.

\par\noindent\textbf{Step 2: compare the two empirical Hessian factors.}
Write $X^i=X_{t_\ell-}^i$, $Y^i=Y_{t_\ell-}^i$,
$\rho_X=\rho_{\mathbf X_{t_\ell-}}$, $\rho_Y=\rho_{\mathbf Y_{t_\ell-}}$,
and $\delta_i=X^i-Y^i$. Use the paired empirical moment calculation
\eqref{eq:paired-empirical-moment-displacement} in
\cref{lemma-directional-measure-slot-error}, with $s=1$ and the actual
pre-jump configurations. On $E_{\rm pop}\cap E_{T,D}$ it gives
\[
 \sum_m\gamma_m(D)\|M_m(\rho_X)-M_m(\rho_Y)\|_{\mathsf H_m}
 \le\sum_m\gamma_m(D)L_m(1+2D^{m-1})r_{t_\ell-}
 =\Lambda(D)r_{t_\ell-},
\]
using \eqref{eq:common-moment-residual-and-lipschitz-sum}.
Insert $\nabla_W^2F(\rho_Y;X^i,X^j)$ between
$\nabla_W^2F(\rho_X;X^i,X^j)$ and
$\nabla_W^2F(\rho_Y;Y^i,Y^j)$.
Apply \eqref{eq:local-two-hessians-moment-bound} to the measure difference
and \eqref{eq:local-kernel-spatial-bound} to the spatial difference.
The two differences are bounded by
$\tfrac12\Lambda(D)r_{t_\ell-}$ and
$\mathrm L_D(\|\delta_i\|_{\mathbb R^d}+\|\delta_j\|_{\mathbb R^d})$, respectively.
Loosening $1/2$ to $1$, for every $i,j$ this yields
\[
\|\nabla_W^2F(\rho_X;X^i,X^j)
-\nabla_W^2F(\rho_Y;Y^i,Y^j)\|_{\mathbb R^d\to\mathbb R^d}
\le\Lambda(D)r_{t_\ell-}+\mathrm L_D(\|\delta_i\|_{\mathbb R^d}+\|\delta_j\|_{\mathbb R^d}).
\]
Here the same $\mathrm L_D$ bounds each spatial slot, by \cref{eq:local-kernel-spatial-bound}.
The block Frobenius norm is at most $\sqrt d$ times its operator norm.
The $N^{-1/2}$ in each block gives exactly
\[
 \frac1N\|M_X-M_Y\|_{F;\,\mathbb R^{dN}\to\mathbb R^{dN}}^2
 =\frac1{N^2}\sum_{i,j}
 \|\nabla_W^2F(\rho_X;X^i,X^j)-\nabla_W^2F(\rho_Y;Y^i,Y^j)\|_{F;\,\mathbb R^d\to\mathbb R^d}^2.
\]
Minkowski's inequality for this normalized double sum gives
\[
\begin{aligned}
 N^{-1/2}\|M_X-M_Y\|_{F;\,\mathbb R^{dN}\to\mathbb R^{dN}}
 &\le\sqrt d\left[\frac1{N^2}\sum_{i,j}
  \{\Lambda(D)r_{t_\ell-}+\mathrm L_D\|\delta_i\|_{\mathbb R^d}
                              +\mathrm L_D\|\delta_j\|_{\mathbb R^d}\}^2\right]^{1/2}\\
 &\le\sqrt d\,\Lambda(D)r_{t_\ell-}
       +\sqrt d\,\mathrm L_D\left(\frac1{N^2}\sum_{i,j}\|\delta_i\|_{\mathbb R^d}^2\right)^{1/2}\\
 &\quad+\sqrt d\,\mathrm L_D\left(\frac1{N^2}\sum_{i,j}\|\delta_j\|_{\mathbb R^d}^2\right)^{1/2}\\
 &=\sqrt d\{\Lambda(D)+2\mathrm L_D\}r_{t_\ell-},
\end{aligned}
\]
since each squared double average equals
$N^{-1}\sum_i\|\delta_i\|_{\mathbb R^d}^2=r_{t_\ell-}^2$.

\par\noindent\textbf{Step 3: use the population covariance residual.}
For the second matrix, the aligned square-root bound and the kernel mismatch
lemma give
\[
\begin{aligned}
 N^{-1/2}\|M_Y-B\|_{F;\,\mathbb R^{dN}\to\mathbb R^{dN}}
 &\le\sqrt{N^{-1}\|C-M_Y^2\|_{*;\,\mathbb R^{dN}\to\mathbb R^{dN}}}\\
 &\le\sqrt{2d\mathrm L_D\Gamma_{t_\ell-}^N(D)+\widetilde\Gamma_{t_\ell-}^N}\\
 &\le\sqrt{2d\mathrm L_D}\sqrt{\Gamma_{t_\ell-}^N(D)}
       +\sqrt{\widetilde\Gamma_{t_\ell-}^N}.
\end{aligned}
\]
The middle inequality is \cref{lemma-kernel-mismatch-bound}; the last
uses $\sqrt{a+b}\le\sqrt a+\sqrt b$. Substitute into
\eqref{eq:jump-estimate-proof} on the Gaussian event from Step 1:
\[
\begin{aligned}
 r_{t_\ell}\le r_{t_\ell-}+\eta_\ell q(u)\bigl[
 &\sqrt d\{\Lambda(D)+2\mathrm L_D\}r_{t_\ell-}\\
 &+\sqrt{2d\mathrm L_D}\sqrt{\Gamma_{t_\ell-}^N(D)}
       +\sqrt{\widetilde\Gamma_{t_\ell-}^N}\bigr].
\end{aligned}
\]
The defined $c_J(D)$ bounds each of the three coefficients in brackets,
which gives the asserted recursion. More explicitly, its leading $1$ bounds
the coefficient of $\sqrt{\widetilde\Gamma_{t_\ell-}^N}$, while the other
two summands bound the coefficients of $r_{t_\ell-}$ and
$\sqrt{\Gamma_{t_\ell-}^N(D)}$, respectively.
\end{proof}
\begin{remark}
The aligned matrix and the whitening coordinate are used to compare the
empirical transition with a finite evaluation of the population field.
Conditional i.i.d. of the reference particles is a separate population
pushforward statement, proved in
Lemma~\ref{lemma-gp-jump-implies-conditional-iid}; it does not condition on
$\mathbf g_\ell^N$.
\end{remark}

%% file: appendix_final_general/general_bound_r_t_bound_stochastic_errors.tex
\subsection{Concentration of the stochastic residuals \(\Gamma_t^N\) and \(\widetilde\Gamma_t^N\)}
\label{subsec:general-argument-bound-stochastic-gamma-and-tilde-gamma}

We first record a bounded-difference inequality used to turn sensitivity bounds
for empirical functionals into high-probability estimates.

\subsubsection{Preliminary}

McDiarmid's inequality will be applied conditionally on the population-level
randomness, with the reference particles treated as independent inputs.

\begin{lemma}[McDiarmid's inequality (textbook: \citet{BLM2013})]
\label{lemma:mcdiarmid}
Let \(X_1,\dots,X_N\) be independent random variables taking values in measurable
spaces \(\mathcal X_1,\dots,\mathcal X_N\), and let
\(
F:\mathcal X_1\times\cdots\times\mathcal X_N\to\R
\)
be measurable.
Assume that there exist constants \(c_1,\dots,c_N\ge0\) such that for every
\(i\in\{1,\dots,N\}\),
\[
|F(x_1,\dots,x_N)-F(x_1,\dots,x_i',\dots,x_N)|
\le c_i
\]
for all \(x_j,x_i'\) for which the two arguments differ only in the \(i\)-th
coordinate.
If $\sum_i c_i^2>0$, then, for every \(r>0\),
\[
\mathbb P\!\left(
F(X_1,\dots,X_N)-\mathbb E[F(X_1,\dots,X_N)] \ge r
\right)
\le
\exp\!\left(
-\frac{2r^2}{\sum_{i=1}^N c_i^2}
\right).
\]
In particular,
\[
\mathbb P\!\left(
|F(X_1,\dots,X_N)-\mathbb E[F(X_1,\dots,X_N)]| \ge r
\right)
\le
2\exp\!\left(
-\frac{2r^2}{\sum_{i=1}^N c_i^2}
\right).
\]
If every $c_i=0$, changing coordinates one at a time gives
$|F(x)-F(y)|\le\sum_i c_i=0$ for any two inputs. Thus $F$ is constant
and both displayed tail probabilities are zero for $r>0$.
The parallel statement holds for the conditionally independent variables.
\end{lemma}

The jump covariance residual is controlled by a Hilbert--Schmidt second-moment estimate and McDiarmid's inequality; see \cref{lem:tilde-gamma-weighted-covariance-input}.

\input{appendix_final_general/general_bound_r_t_bound_stochastic_errors_gamma}

\input{appendix_final_general/general_bound_r_t_tilde-gamma_rebuilt}

%% file: appendix_final_general/general_bound_r_t_bound_stochastic_errors_gamma.tex
\subsubsection{Concentration of $\Gamma_t^N$}
\label{subsubsec:general-argument-bound-gamma}

We first control \(\Gamma_t^N\), the weighted moment fluctuation of the
reference empirical cloud.  The key input is that, after conditioning on the
common population-level jump randomness, the reference particles remain i.i.d.
with law \(\mu_t\).

\noindent
\textbf{A formal proof of conditionally i.i.d. property.}

We make the conditional i.i.d. structure explicit, since all concentration
bounds below are applied after conditioning on the common jump randomness.

\begin{definition}[Conditional i.i.d. reference particles]
\label{definition-conditional-iid-common-jump}
For the population field histories $\mathcal G_\ell$ fixed in the notation
section, conditional i.i.d. means
\[
\operatorname{Law}((Y_t^1,\ldots,Y_t^N)\mid\mathcal G_\ell)
=\mu_t^{\otimes N},\qquad t\in[t_\ell,t_{\ell+1}),
\]
with the pre-jump trace at $t_{\ell+1}-$. A terminal post-jump value at $T$
uses $\mathcal G_{\NofJ}$. Conditioning instead on
$\mathcal G_{\rm pop}=\mathcal G_{\NofJ}$ gives the same identity for every
time and each of its relevant traces.
\end{definition}
\begin{lemma}[The population GP jump scheme preserves conditional i.i.d.]
\label{lemma-gp-jump-implies-conditional-iid}
The population-first construction in Definition~\ref{definition-ode-and-jump}
and Assumption~\ref{assumption-common-seed} has the conditional i.i.d.
property in Definition~\ref{definition-conditional-iid-common-jump}.
\end{lemma}
\begin{proof}
Use the conditional product calculation
\eqref{eq:population-pushforward-product-law} in
\cref{lem:population-first-finite-coupling}: the common map
$\mathcal T_t$ satisfies $Y_t^i=\mathcal T_t(U_i)$ and
$\mu_t=(\mathcal T_t)_\#\mu_0$, also at each left trace.
On $[t_\ell,t_{\ell+1})$ and its final left trace, these maps and laws
are already $\mathcal G_\ell$-measurable. For bounded measurable $f_i$,
the tower property gives
\[
\begin{aligned}
 \mathbb E\left[\prod_i f_i(Y_t^i)\,\middle|\,\mathcal G_\ell\right]
 &=\mathbb E\left[\mathbb E\left[\prod_i f_i(Y_t^i)
       \,\middle|\,\mathcal G_{\rm pop}\right]\middle|\mathcal G_\ell\right]\\
 &=\mathbb E\left[\prod_i\int f_i(y)\mu_t(dy)
       \,\middle|\,\mathcal G_\ell\right]
 =\prod_i\int f_i(y)\mu_t(dy).
\end{aligned}
\]
A terminal post-jump trace uses $\mathcal G_{\NofJ}$.
The finite Gaussian coordinate is included in neither conditioning field.
\end{proof}
\paragraph{Unconditional regularity and a conditional moment event.}
In the unconditional moment assumption, write $K_0$ for the deterministic
one-particle moment bound. Conditional moment bounds will be consequences
on the event below, rather than almost-sure assumptions with the same $K_0$.
The population sigma-field $\mathcal G_{\rm pop}$ is generated by the
population fields in the population-first coupling. In particular,
$\mu_t=\operatorname{Law}(Y_t^1\mid\mathcal G_{\rm pop})$.

\begin{lemma}[Conditional moments on a population-measurable event]
\label{lem:population-conditional-moment-localization}\label{lem:population-moment-event}
Let $0<t_1<\cdots<t_{\NofJ}\le T$ be deterministic jump times. Suppose that
the paths $X^1,Y^1$ have right values and left limits at these times and
are $B$-Lipschitz on every intervening open interval, almost surely,
with the continuous branch extensions and endpoint traces of
\cref{definition-ode-and-jump},
where $B\ge0$ is deterministic. Assume that, for $S\in\{X,Y\}$,
\begin{equation}
\label{eq:population-unconditional-moment-input}
 \sup_{0\le t\le T}\sup_{q\ge1}
 q^{-1/\alpha}\bigl(\mathbb E\|S_t^1\|_{\mathbb R^d}^q\bigr)^{1/q}
 \le K_0,
 \qquad 0<\alpha\le2.
\end{equation}
The same bounds at left limits are understood by Fatou's lemma.
For $p_0\in(0,1)$, define
\[
 Q=\left\lceil(1+B)T\right\rceil+2(\NofJ+1),\qquad
 b=\max\{1,K_0\{2e(1+\alpha)\}^{1/\alpha}\},\qquad
 A_0=\frac{4Q}{p_0}.
\]
Here $b$ is a deterministic exponential-moment scale, $Q$ bounds the
number of grid time-side pairs, and $A_0$ is the threshold allocating
failure probability $p_0$ to their conditional moments.
Set $c_\alpha=2^{(\alpha-1)_+}$ and
\begin{equation}
\label{eq:population-exponential-amplitude}
 b_{\rm flow}=c_\alpha^{1/\alpha}b,\qquad
 M_{\rm pop}=A_0\exp(b^{-\alpha})\le\frac{4eQ}{p_0},
\end{equation}
and define
\begin{equation}
\label{eq:population-exponential-scale}
 k_{\rm pop}^{\alpha}
 =\frac{c_\alpha}{\log2}\{b^\alpha\log A_0+1\},
 \qquad
 K_{\rm pop}=C_\alpha k_{\rm pop},
\end{equation}
where $C_\alpha$ depends only on $\alpha$ and can be chosen so that
an exponential moment bound with scale $k$ implies the moment bound
$C_\alpha kq^{1/\alpha}$ for all $q\ge1$.
There is an event $E_{\rm pop}\in\mathcal G_{\rm pop}$ with
$\mathbb P(E_{\rm pop})\ge1-p_0$ such that, simultaneously for
$S\in\{X,Y\}$, all $t\in[0,T]$, and both sides of every jump,
\begin{equation}
\label{eq:population-fixed-scale-exponential}
 \mathbb E\!\left[
  \left.\exp\left\{\frac{\|S_t^1\|_{\mathbb R^d}^\alpha}{b_{\rm flow}^\alpha}\right\}
  \right|\mathcal G_{\rm pop}\right]\le M_{\rm pop},
\end{equation}
as well as
\begin{equation}
\label{eq:population-uniform-conditional-exponential}
 \mathbb E\!\left[
   \left.\exp\left\{\frac{\|S_t^1\|_{\mathbb R^d}^\alpha}
                              {k_{\rm pop}^\alpha}\right\}
   \right|\mathcal G_{\rm pop}\right]\le2,
\end{equation}
and hence
\begin{equation}
\label{eq:population-uniform-conditional-moments}
 \sup_{q\ge1}q^{-1/\alpha}
 \bigl(\mathbb E[\|S_t^1\|_{\mathbb R^d}^q\mid\mathcal G_{\rm pop}]\bigr)^{1/q}
 \le K_{\rm pop}.
\end{equation}
In particular $\mu_t\in\mathcal P_{\alpha,K_{\rm pop}}$ on this event.
No conditional independence of the $X$ particles is asserted or needed.
\end{lemma}

\begin{proof}
\noindent\textbf{Step 1: an unconditional exponential moment.}
For $q<1$, Lyapunov's inequality and the $q=1$ case of
\eqref{eq:population-unconditional-moment-input} give
$\mathbb E\|S_t^1\|_{\mathbb R^d}^q\le K_0^q$.
Consequently, for each integer $j\ge1$,
\[
 \mathbb E\|S_t^1\|_{\mathbb R^d}^{\alpha j}
 \le K_0^{\alpha j}\max\{1,\alpha j\}^{j}.
\]
Since $j!\ge(j/e)^j$, $\max\{1,\alpha j\}\le(1+\alpha)j$, and
$b^\alpha\ge2e(1+\alpha)K_0^\alpha$, we have
\[
 \frac{\mathbb E\|S_t^1\|_{\mathbb R^d}^{\alpha j}}{j!b^{\alpha j}}
 \le\left(\frac{eK_0^\alpha\max\{1,\alpha j\}}{jb^\alpha}\right)^j
 \le\left(\frac{eK_0^\alpha(1+\alpha)}{b^\alpha}\right)^j
 \le2^{-j}.
\]
Monotone convergence for the nonnegative exponential series therefore gives
\[
 \mathbb E\exp\{\|S_t^1\|_{\mathbb R^d}^\alpha/b^\alpha\}
 \le1+\sum_{j\ge1}2^{-j}=2.
\]
The same inequality holds at a left trace by Fatou's lemma.

\par\noindent\textbf{Step 2: one grid event and a common conditional path law.}
The deterministic branch speed $B$ is used here to transfer moment bounds
from a finite time grid to every time on each branch. It is fixed before the
random sample-localization event $E_{T,D}$ is introduced.
Set $t_0=0$ and $t_{\NofJ+1}=T$.
On each nonempty ODE branch choose a deterministic grid of mesh at most
$(1+B)^{-1}$, including the initial right value and final left trace.
The number of entries on this branch is at most
$\lceil(1+B)(t_{\ell+1}-t_\ell)\rceil+1$.
At a jump the two traces are separate entries. If $t_{\NofJ}=T$, the
zero-length final branch contributes the single post-jump value at $T$.
Thus the set $\mathcal T$ of time-side pairs satisfies
\[
 \#\mathcal T
 \le\sum_{\ell=0}^{\NofJ}
     \left(\left\lceil(1+B)(t_{\ell+1}-t_\ell)\right\rceil+1\right)
 \le(1+B)T+2(\NofJ+1)\le Q.
\]

To make the branchwise Lipschitz inequalities hold simultaneously under the
same conditional law at every grid point and time trace, use a regular
conditional law of the pair of paths $(X^1,Y^1)$ given
$\mathcal G_{\rm pop}$. Equivalently, regard each path as its finitely
many continuous branch extensions with the prescribed endpoint traces,
together with any isolated terminal post-jump value. This is a
standard Borel path space. All conditional expectations in this proof
are evaluated using this one conditional path law.
Define the countable intersection
\[
 \mathcal L:=\bigcap_{S\in\{X,Y\}}
  \bigcap_{\ell=0}^{\NofJ}
  \bigcap_{s,t\in\mathbb Q\cap(t_\ell,t_{\ell+1})}
   \{\|S_t^1-S_s^1\|_{\mathbb R^d}\le B|t-s|\}.
\]
The assumed branchwise Lipschitz property gives $\mathbb P(\mathcal L)=1$.
The tower property implies
\[
 \mathbb E[\mathbb P(\mathcal L^c\mid\mathcal G_{\rm pop})]
 =\mathbb P(\mathcal L^c)=0.
\]
Since the conditional probability is nonnegative, it is zero almost surely.
Continuity on each branch extends the rational-time inequalities to all
time pairs and to that branch's endpoint traces. Thus these inequalities
hold simultaneously under the conditional path law outside one
population-null set.

For $s\in\mathcal T$, put
\[
 Z_s^S:=\mathbb E[\exp\{\|S_s^1\|_{\mathbb R^d}^\alpha/b^\alpha\}
                         \mid\mathcal G_{\rm pop}],
 \qquad
 E_{\rm pop}:=
 \{\mathbb P(\mathcal L\mid\mathcal G_{\rm pop})=1\}
 \cap\bigcap_{S\in\{X,Y\}}\bigcap_{s\in\mathcal T}
          \{Z_s^S\le A_0\}.
\]
The event is $\mathcal G_{\rm pop}$-measurable. Step 1, the tower property,
Markov's inequality, and the finite union bound give
\[
 \mathbb P(E_{\rm pop}^c)
 \le\sum_{S\in\{X,Y\}}\sum_{s\in\mathcal T}
       \frac{\mathbb E Z_s^S}{A_0}
 \le(2Q)\frac2{A_0}=p_0.
\]

\par\noindent\textbf{Step 3: extend both exponential bounds to all times.}
Fix a population realization in $E_{\rm pop}$.
For any time-side pair $t$ on a nonempty branch, choose $s\in\mathcal T$
on that branch with $|t-s|\le(1+B)^{-1}$. Under the conditional path law,
\[
 \|S_t^1\|_{\mathbb R^d}\le\|S_s^1\|_{\mathbb R^d}+B|t-s|\le\|S_s^1\|_{\mathbb R^d}+1.
\]
For an isolated terminal value, use $s=t$.
The inequality $(u+v)^\alpha\le c_\alpha(u^\alpha+v^\alpha)$ and
$b_{\rm flow}^\alpha=c_\alpha b^\alpha$ from
\eqref{eq:population-exponential-amplitude} give
\[
 \mathbb E[\exp\{\|S_t^1\|_{\mathbb R^d}^\alpha/b_{\rm flow}^\alpha\}
                         \mid\mathcal G_{\rm pop}]
 \le e^{b^{-\alpha}}Z_s^S
 \le e^{b^{-\alpha}}A_0=M_{\rm pop}.
\]
This proves \eqref{eq:population-fixed-scale-exponential}; its fixed scale
will be retained when estimating the clipping bias.

For the normalized exponential bound, put
$\eta=\log2/\log M_{\rm pop}\in(0,1)$. By
\cref{eq:population-exponential-amplitude,eq:population-exponential-scale},
$k_{\rm pop}^\alpha=b_{\rm flow}^\alpha/\eta$.
Apply conditional Jensen directly to the all-time estimate
\eqref{eq:population-fixed-scale-exponential}:
\[
\begin{aligned}
 \mathbb E\left[e^{\|S_t^1\|_{\mathbb R^d}^\alpha/k_{\rm pop}^\alpha}
                       \,\middle|\,\mathcal G_{\rm pop}\right]
 &=\mathbb E\left[\left(e^{\|S_t^1\|_{\mathbb R^d}^\alpha/b_{\rm flow}^\alpha}\right)^\eta
                       \,\middle|\,\mathcal G_{\rm pop}\right]\\
 &\le\left(\mathbb E\left[e^{\|S_t^1\|_{\mathbb R^d}^\alpha/b_{\rm flow}^\alpha}
                       \,\middle|\,\mathcal G_{\rm pop}\right]\right)^\eta
 \le M_{\rm pop}^\eta=2.
\end{aligned}
\]
The same conditional path law is used at all time traces, proving
\eqref{eq:population-uniform-conditional-exponential} simultaneously.

\par\noindent\textbf{Step 4: obtain every ordinary conditional moment.}
Fix $q\ge1$. For $u>0$,
\[
 \frac d{du}\log\{u^qe^{-(u/k_{\rm pop})^\alpha}\}
 =\frac q u-\frac{\alpha u^{\alpha-1}}{k_{\rm pop}^\alpha}.
\]
This derivative is positive before
$u=k_{\rm pop}(q/\alpha)^{1/\alpha}$ and negative after it.
Evaluation at that maximum, also including $u=0$ by continuity, gives
\[
 u^q\le
 k_{\rm pop}^q\left(\frac q{\alpha e}\right)^{q/\alpha}
 e^{(u/k_{\rm pop})^\alpha}\qquad(u\ge0).
\]
Integrate under the same conditional path law and use
\eqref{eq:population-uniform-conditional-exponential} to obtain
\[
 \bigl(\mathbb E[\|S_t^1\|_{\mathbb R^d}^q\mid\mathcal G_{\rm pop}]\bigr)^{1/q}
 \le2^{1/q}k_{\rm pop}\left(\frac q{\alpha e}\right)^{1/\alpha}
 \le C_\alpha k_{\rm pop}q^{1/\alpha},
 \qquad
 C_\alpha=2\max\{1,(\alpha e)^{-1/\alpha}\}.
\]
This proves \eqref{eq:population-uniform-conditional-moments} for every
$q\ge1$ and every time trace on $E_{\rm pop}$.
Finally, $\mu_t=\operatorname{Law}(Y_t^1\mid\mathcal G_{\rm pop})$
implies $\mu_t\in\mathcal P_{\alpha,K_{\rm pop}}$ there.
\end{proof}

\paragraph{Probability bookkeeping.}
The event $E_{\rm pop}$ is measurable with respect to the sigma-field
under which the reference samples are conditionally iid. Thus all
conditional concentration arguments can be applied for each population
realization in this event, with the deterministic scale $K_{\rm pop}$.
If their conditional failure probability is at most $p_1$, the
unconditional failure probability, including $E_{\rm pop}^{\rm c}$, is
at most $p_0+p_1$. Conditioning on a sample-localization event is not
used to claim independence. If only the reference moment bound is
needed, the $X$ constraints can be omitted and $A_0$ replaced by $2Q/p_0$.

\noindent
\textbf{Conditional concentration of \(\Gamma_t^N\)}

We first prove a localized uniform-in-time bound: truncate the features, apply
conditional McDiarmid at grid times, and then remove the truncation.

Throughout this subsection, the jump times are deterministic, with
$0=t_0<t_1<\cdots<t_{\NofJ}\le T$ and $t_{\NofJ+1}=T$.
Every supremum and time grid includes both traces at each jump.
More precisely, the closure of an ODE interval has its post-jump value at
the left endpoint and its pre-jump value at the right endpoint.  The last
interval ends at $T$, and a jump at $T$ has an additional terminal post-jump
value.  An interval of zero length contributes that terminal value.
We condition only on the full population randomness $\mathcal G_{\rm pop}$.
Conditional on it, the reference trajectories are independent copies of
the same population trajectory and have marginal law $\mu_t$.

Let $E_{\rm pop}\in\mathcal G_{\rm pop}$ be the event supplied by
the population-moment localization lemma.  Retain both of its conclusions:
on $E_{\rm pop}$, for deterministic $K_{\rm pop}>0$, $K_0>0$,
$c_0=(K_0/b_{\rm flow})^\alpha>0$, and $M_{\rm pop}\ge1$,
\begin{equation}
\label{eq:gamma-population-localization-input}
\begin{gathered}
 \sup_{t\le T}\sup_{r\ge1}r^{-1/\alpha}
 \left(\int\|y\|_{\mathbb R^d}^r\,\mu_t(dy)\right)^{1/r}\le K_{\rm pop},
 \\
 \sup_{t\le T}\int
 \exp\!\left(c_0\left(\frac{\|y\|_{\mathbb R^d}}{K_0}\right)^\alpha\right)
 \,\mu_t(dy)\le M_{\rm pop}.
\end{gathered}
\end{equation}
Here $K_0$ is the unconditional moment scale; $K_{\rm pop}$ and
$M_{\rm pop}$ are deterministic functions of the probability budget and
the time grid.  In particular, they are not asserted to be bounded
uniformly in the realized population Gaussian fields.

Fix $D\ge1$. The deterministic weights from
\cref{eq:common-local-moment-weights} are
\[
 \gamma_m=\gamma_m(K_{\rm pop},D)=\mathrm L_Da_m.
\]
Define the radial projection onto the radius-$D$ ball by
\[
\Pi_D(y)=y\min\{1,D/\|y\|_{\mathbb R^d}\}\quad(y\ne0),\qquad \Pi_D(0)=0.
\]
As the metric projection onto a closed convex ball,
$\|\Pi_Dx-\Pi_Dy\|_{\mathbb R^d}\le\|x-y\|_{\mathbb R^d}$.
Let $\phi_m^{(D)}=\phi_m\circ\Pi_D$. The feature assumption
\cref{assumption-structured-moment-smoothness}(i),
\eqref{eq:coordinate-feature-growth}, gives
$\|\phi_m^{(D)}(y)\|_{\mathsf H_m}\le L_m(1+D^m)$.
This amplitude bounds the effect of replacing one sample in McDiarmid's
inequality. The spatial Lipschitz bound below is used separately for
time interpolation. Define the respective envelopes
\begin{equation}
\label{eq:gamma-amplitude-lipschitz-envelopes}
 A(D,K_{\rm pop})=2\sum_{m\ge1}\gamma_mL_m(1+D^m),
 \qquad
 \Lambda(D,K_{\rm pop})=
 \sum_{m\ge1}\gamma_mL_m(1+2D^{m-1}).
\end{equation}
For $\|u\|_{\mathbb R^d},\|v\|_{\mathbb R^d}\le D$, the segment $v+s(u-v)$ remains in the ball.
Integrating the first derivative in \eqref{eq:coordinate-feature-growth}
gives
\[
 \|\phi_m(u)-\phi_m(v)\|_{\mathsf H_m}
 \le\int_0^1 L_m(1+D^{m-1})\|u-v\|_{\mathbb R^d}\,ds
 =L_m(1+D^{m-1})\|u-v\|_{\mathbb R^d}.
\]
Apply this with $u=\Pi_Dx$, $v=\Pi_Dy$ and use the $1$-Lipschitz
bound above. Thus $\phi_m^{(D)}$ is globally Lipschitz with constant
at most $L_m(1+D^{m-1})\le L_m(1+2D^{m-1})$, as used in
\eqref{eq:gamma-amplitude-lipschitz-envelopes}.
Differentiability of $\Pi_D$ on the sphere is not needed.
Define also
\begin{equation}
\label{eq:gamma-conditional-tail-sum}
\begin{aligned}
 S(D,K_{\rm pop})&=
 \sum_{m\ge1}\gamma_mL_m
 \left[1+\{K_{\rm pop}(2m)^{1/\alpha}\}^{m}\right],
 \\
 \overline\beta_T(D)&=
 2S(D,K_{\rm pop})\sqrt{M_{\rm pop}}
 \exp\!\left[-\frac{c_0}{2}
               \left(\frac{D}{K_0}\right)^\alpha\right].
\end{aligned}
\end{equation}

\begin{lemma}[Localized uniform high-probability bound under the GP jump scheme]
\label{lemma-uniform-sup-gamma-bound-localized-conditional}
Assume the conditional i.i.d. conclusion of
\cref{lemma-gp-jump-implies-conditional-iid}, the feature bounds in
\cref{assumption-structured-moment-smoothness}(i), and
\cref{assumption-uniform-concentration-piecewise-lipschitz}
with deterministic constant $B$.  Assume
\eqref{eq:gamma-population-localization-input} on $E_{\rm pop}$,
and assume the sums in
\eqref{eq:gamma-amplitude-lipschitz-envelopes} and
\eqref{eq:gamma-conditional-tail-sum} are finite.
For $N\ge2$ and $p_\Gamma\in(0,1)$, set $Q_N=(\NofJ+1)(N+1)$ and
\begin{equation}
\label{eq:gamma-explicit-envelope}
 \delta_\Gamma(D)=\overline\beta_T(D)
 +\frac{A(D,K_{\rm pop})}{\sqrt N}
 +A(D,K_{\rm pop})\sqrt{\frac{\log(Q_N/p_\Gamma)}{2N}}
 +\frac{2BT\Lambda(D,K_{\rm pop})}{N}.
\end{equation}
Then
\begin{equation}
\label{eq:gamma-localized-probability}
 \mathbb P\left(
  E_{\rm pop}\cap E_{T,D}\cap
  \left\{\sup_{t\le T}\Gamma_t^N>\delta_\Gamma(D)\right\}
 \right)\le p_\Gamma.
\end{equation}
Consequently, if
$\mathbb P(E_{\rm pop}^c)\le p_{\rm pop}$ and
$\mathbb P(E_{T,D}^c)\le p_{\rm rad}$, then
\[
 \mathbb P\left(
  E_{\rm pop}\cap E_{T,D}\cap
  \left\{\sup_{t\le T}\Gamma_t^N\le\delta_\Gamma(D)\right\}
 \right)\ge1-p_{\rm pop}-p_{\rm rad}-p_\Gamma.
\]
\end{lemma}

\begin{proof}
\noindent\textbf{Step 1: concentrate the clipped empirical moments.}
Condition on $\mathcal G_{\rm pop}$, without conditioning on $E_{T,D}$,
and write $\mathbb E_{\mathcal G}[Z]=\mathbb E[Z\mid\mathcal G_{\rm pop}]$.
At a fixed time trace, let
\[
 F_t^D=\sum_{m\ge1}\gamma_m
 \left\|\frac1N\sum_{i=1}^N\phi_m^{(D)}(Y_t^i)
       -\int\phi_m^{(D)}(y)\,\mu_t(dy)\right\|_{\mathsf H_m}.
\]
If a single sample $y$ is replaced by $y'$, the reverse triangle inequality
and $\|\phi_m^{(D)}(y)\|_{\mathsf H_m}\le L_m(1+D^m)$ give
\[
 |F_t^D(y,Y^2,\ldots)-F_t^D(y',Y^2,\ldots)|
 \le\frac1N\sum_m\gamma_m
       \|\phi_m^{(D)}(y)-\phi_m^{(D)}(y')\|_{\mathsf H_m}
 \le\frac{A(D,K_{\rm pop})}{N}.
\]
For each fixed $m$, conditional independence makes every off-diagonal term
in the squared centered sample mean zero: for $i\ne j$,
\[
\begin{aligned}
 &\mathbb E_{\mathcal G}\left\langle
 \phi_m^{(D)}(Y_t^i)-\mathbb E_{\mathcal G}\phi_m^{(D)}(Y_t^i),
 \phi_m^{(D)}(Y_t^j)-\mathbb E_{\mathcal G}\phi_m^{(D)}(Y_t^j)
 \right\rangle\\
 &\qquad=\left\langle
 \mathbb E_{\mathcal G}(\phi_m^{(D)}(Y_t^i)-\mathbb E_{\mathcal G}\phi_m^{(D)}(Y_t^i)),
 \mathbb E_{\mathcal G}(\phi_m^{(D)}(Y_t^j)-\mathbb E_{\mathcal G}\phi_m^{(D)}(Y_t^j))
 \right\rangle=0.
\end{aligned}
\]
The remaining diagonal terms give
\[
\begin{aligned}
 &\mathbb E_{\mathcal G}\left\|
    \frac1N\sum_i\phi_m^{(D)}(Y_t^i)
           -\mathbb E_{\mathcal G}\phi_m^{(D)}(Y_t^1)\right\|_{\mathsf H_m}^2\\
 &\quad=\frac1N\left(
      \mathbb E_{\mathcal G}\|\phi_m^{(D)}(Y_t^1)\|_{\mathsf H_m}^2
      -\|\mathbb E_{\mathcal G}\phi_m^{(D)}(Y_t^1)\|_{\mathsf H_m}^2\right)
 \le\frac{L_m^2(1+D^m)^2}{N}.
\end{aligned}
\]
Cauchy--Schwarz bounds the expected norm by the square root of this
quantity. Summing with the nonnegative weights $\gamma_m$ gives
\[
 \mathbb E[F_t^D\mid\mathcal G_{\rm pop}]
 \le\frac1{\sqrt N}\sum_m\gamma_mL_m(1+D^m)
 \le\frac{A(D,K_{\rm pop})}{\sqrt N}.
\]
Apply \cref{lemma:mcdiarmid} with squared-difference sum
$N\{A(D,K_{\rm pop})/N\}^2=A(D,K_{\rm pop})^2/N$. It gives, for every $u>0$,
\[
 \mathbb P\left(
 F_t^D>\frac{A(D,K_{\rm pop})}{\sqrt N}+u
 \,\middle|\,\mathcal G_{\rm pop}\right)
 \le \exp\left(-\frac{2Nu^2}{A(D,K_{\rm pop})^2}\right).
\]
When $A=0$, this conclusion is read as $F_t^D=0$.
These arguments apply first to finite partial sums and then to the
infinite weighted sum by the finite envelope and monotone convergence.

\par\noindent\textbf{Step 2: extend the estimate between grid points.}
Subdivide each closed ODE interval, with its correctly labelled endpoint
traces, into $N$ equal subintervals.  The union has at most $Q_N$ nodes.
For $s,t$ on the same ODE interval, use the clipped-feature Lipschitz bound
from \eqref{eq:gamma-amplitude-lipschitz-envelopes}. The reverse triangle
inequality separates the empirical and population changes as
\[
\begin{aligned}
 |F_t^D-F_s^D|
 &\le\sum_m\gamma_mL_m(1+2D^{m-1})
 \left(\frac1N\sum_i\|Y_t^i-Y_s^i\|_{\mathbb R^d}
             +\mathbb E_{\mathcal G}\|Y_t^1-Y_s^1\|_{\mathbb R^d}\right)\\
 &\le2B\Lambda(D,K_{\rm pop})|t-s|.
\end{aligned}
\]
At each grid node substitute
$u=A(D,K_{\rm pop})\sqrt{\log(Q_N/p_\Gamma)/(2N)}$ in the preceding
concentration inequality. When $A>0$, the union over at most $Q_N$ nodes
costs
\[
 Q_N\exp\left(-\frac{2N}{A(D,K_{\rm pop})^2}
              \frac{A(D,K_{\rm pop})^2\log(Q_N/p_\Gamma)}{2N}\right)
 =p_\Gamma.
\]
When $A=0$, $F_t^D=0$ directly. Every time trace lies within $T/N$ of a
grid node on its own branch. The time Lipschitz estimate adds at most
$2BT\Lambda(D,K_{\rm pop})/N$. Consequently, for each population realization in $E_{\rm pop}$,
\begin{equation}
\label{eq:gamma-clipped-uniform-probability}
 \mathbb P\left(\sup_{t\le T}F_t^D>
 \delta_\Gamma(D)-\overline\beta_T(D)
 \,\middle|\,\mathcal G_{\rm pop}\right)\le p_\Gamma.
\end{equation}
Averaging this conditional bound preserves its failure budget.

\par\noindent\textbf{Step 3: remove clipping on the sample event.}
On $E_{T,D}$, the clipped and original empirical feature averages agree.
For $\|y\|_{\mathbb R^d}\le D$, the population integrand difference is zero; otherwise
the feature growth bound and $D^m\le\|y\|_{\mathbb R^d}^m$ give
\[
 \|\phi_m(y)-\phi_m(\Pi_Dy)\|_{\mathsf H_m}
 \le L_m(2+\|y\|_{\mathbb R^d}^m+D^m)\le2L_m(1+\|y\|_{\mathbb R^d}^m).
\]
Thus the triangle inequality gives
\begin{equation}
\label{eq:gamma-population-clipping-bias}
 \Gamma_t^N\le F_t^D+\beta_T^{\mathcal G}(D),\qquad
 \beta_T^{\mathcal G}(D)=
 2\sum_m\gamma_mL_m\sup_{s\le T}
 \int(1+\|y\|_{\mathbb R^d}^m)\mathbf1_{\{\|y\|_{\mathbb R^d}>D\}}\,\mu_s(dy).
\end{equation}
This is a conditional population integral, not the unconditional moment
of a randomly selected population field.  \Cref{lemma-beta-tail-bound}, using only the population input
\eqref{eq:gamma-population-localization-input}, gives
$\beta_T^{\mathcal G}(D)\le\overline\beta_T(D)$ on
$E_{\rm pop}$.  Intersecting the event in
\eqref{eq:gamma-clipped-uniform-probability} with $E_{\rm pop}$ and $E_{T,D}$ proves
\eqref{eq:gamma-localized-probability} and the final probability bound.
\end{proof}

\begin{lemma}[Conditional population truncation bias]
\label{lemma-beta-tail-bound}
On $E_{\rm pop}$, the bias $\beta_T^{\mathcal G}(D)$ in
\eqref{eq:gamma-population-clipping-bias} is
bounded by $\overline\beta_T(D)$ in
\eqref{eq:gamma-conditional-tail-sum}.
If only the first, moment-bound part of
\eqref{eq:gamma-population-localization-input} is retained, the valid
alternative, for $D\ge eK_{\rm pop}$, is
\begin{equation}
\label{eq:gamma-moment-only-bias}
 \beta_T^{\mathcal G}(D)\le
 2S(D,K_{\rm pop})
 \exp\!\left[-\frac12
               \left(\frac{D}{eK_{\rm pop}}\right)^\alpha\right].
\end{equation}
\end{lemma}
\begin{proof}
Write $u=c_0(D/K_0)^\alpha$. Conditional Markov, using the exponential-moment
bound \eqref{eq:population-fixed-scale-exponential} of
\cref{lem:population-moment-event}, and Cauchy--Schwarz give,
uniformly over time traces,
\[
 \mu_t(\|y\|_{\mathbb R^d}>D)\le\min\{1,M_{\rm pop}e^{-u}\},
\]
\[
\int(1+\|y\|_{\mathbb R^d}^m)\mathbf1_{\{\|y\|_{\mathbb R^d}>D\}}\,\mu_t(dy)
 \le\left[1+\{K_{\rm pop}(2m)^{1/\alpha}\}^{m}\right]
       \sqrt{M_{\rm pop}}e^{-u/2}.
\]
Indeed, Cauchy--Schwarz followed by the $L^2$ triangle inequality gives
\begin{equation}
\label{eq:gamma-tail-cauchy-schwarz}
\begin{aligned}
 \int(1+\|y\|_{\mathbb R^d}^m)\mathbf1_{\{\|y\|_{\mathbb R^d}>D\}}\,\mu_t(dy)
 &\le\left(\int(1+\|y\|_{\mathbb R^d}^m)^2\,\mu_t(dy)\right)^{1/2}
                   \mu_t(\|y\|_{\mathbb R^d}>D)^{1/2}\\
 &\le\left[1+\left(\int\|y\|_{\mathbb R^d}^{2m}\,\mu_t(dy)\right)^{1/2}\right]
                   \sqrt{\min\{1,M_{\rm pop}e^{-u}\}}.
\end{aligned}
\end{equation}
Now use \eqref{eq:gamma-population-localization-input} with $r=2m$
and $\sqrt{\min\{1,z\}}\le\sqrt z$, also when $z>1$.
Summing with weights $2\gamma_mL_m$ and recalling $S$ in
\eqref{eq:gamma-conditional-tail-sum} gives
$\beta_T^{\mathcal G}(D)\le2S(D,K_{\rm pop})\sqrt{M_{\rm pop}}e^{-u/2}
=\overline\beta_T(D)$.
For the moment-only version, set $r=(D/(eK_{\rm pop}))^\alpha\ge1$.
Markov's inequality and the moment input give
\begin{equation}
\label{eq:moment-tail-markov-substitution}
 \mu_t(\|y\|_{\mathbb R^d}>D)\le D^{-r}\int\|y\|_{\mathbb R^d}^r\mu_t(dy)
 \le\left(\frac{K_{\rm pop}r^{1/\alpha}}D\right)^r=e^{-r}.
\end{equation}
Substitute this probability into \eqref{eq:gamma-tail-cauchy-schwarz}
and sum as in \eqref{eq:gamma-population-clipping-bias}:
\[
 \beta_T^{\mathcal G}(D)
 \le2\sum_m\gamma_mL_m[1+\{K_{\rm pop}(2m)^{1/\alpha}\}^m]e^{-r/2}
 =2S(D,K_{\rm pop})e^{-r/2}.
\]
This is \eqref{eq:gamma-moment-only-bias}.
\end{proof}

\begin{lemma}[Concrete bounds for the amplitude, Lipschitz, and tail sums]
\label{lemma-concrete-bound-A-Lambda}
Let $\kappa=\kappa_{\rm dec}>\max\{1,\kappa_{\rm gr}\}$ and
$q=\kappa/(\kappa-1)$.  \Cref{lem:conditional-local-coefficient-growth,lemma-localized-moment-smoothness-consequences} give
\begin{equation}
\label{eq:gamma-coefficient-growth-explicit}
 \gamma_m(K,D)L_m\le
 C\exp\left(C\{(\log(e+D))^q+(\log(e+K))^q\}-c m^\kappa\right),
\end{equation}
with fixed $C,c>0$.  In particular,
\begin{equation}
\label{eq:gamma-three-envelope-growth}
 A(D,K)\vee\Lambda(D,K)\vee S(D,K)
 \le C\exp\left(C\{(\log(e+D))^q+(\log(e+K))^q\}\right).
\end{equation}
Here the constants do not depend on $D,K,N,T,p$.
\end{lemma}
\begin{proof}
By \cref{eq:common-local-moment-weights,eq:local-envelope-growth},
$\gamma_m(K,D)=\mathrm L_{K,D}a_m$ and
$\mathrm L_{K,D}\le Ce^{C\chi(K+D)}$.
Use \cref{lem:weighted-series-estimate},
\eqref{eq:weighted-series-summand-bound}, first without the braces
(which are at least one). Since
\[
 \log(e+K+D)\le\log(e+K)+\log(e+D),\qquad
 \chi(K+D)\le2^{q-1}\{\chi(K)+\chi(D)\},
\]
this proves \eqref{eq:gamma-coefficient-growth-explicit}, with
$c=c_a/4$ allowed. For the three sums, substitute in
\eqref{eq:weighted-series-sum-bound} as follows:
\[
\begin{aligned}
 A(D,K)&=2\mathrm L_{K,D}\sum_m a_mL_m(1+D^m)
              \le C\mathrm L_{K,D}e^{C\chi(K+D)},\\
 \Lambda(D,K)&\le2\mathrm L_{K,D}\sum_m a_mL_m(2+D^m)
              \le C\mathrm L_{K,D}e^{C\chi(K+D)},\\
 S(D,K)&=\mathrm L_{K,D}\sum_m a_mL_m
                 [1+\{2^{1/\alpha}K\}^m m^{m/\alpha}]\\
 &\le C\mathrm L_{K,D}e^{C\chi(2^{1/\alpha}K+D)}.
\end{aligned}
\]
Here $D\ge1$ gives $1+2D^{m-1}\le2(2+D^m)$.
Use \eqref{eq:weighted-series-scale-comparison} to absorb the fixed
rescaling of $K+D$, then apply the displayed bound for $\chi(K+D)$.
This proves \eqref{eq:gamma-three-envelope-growth}.
\end{proof}

\begin{lemma}[Uniform radius bound for both particle clouds]
\label{lemma:uniform-radius-bound-XY}
Suppose each one-particle marginal in both clouds obeys the unconditional
bound
$\sup_{t\le T,r\ge1}r^{-1/\alpha}
(\mathbb E\|Z_t^i\|_{\mathbb R^d}^r)^{1/r}\le K_0$, where $Z=X,Y$,
and both clouds have deterministic ODE-interval Lipschitz bound $B$.
For $D\ge2eK_0$,
\begin{equation}
\label{eq:ETD-tail}
 \mathbb P(E_{T,D}^c)\le
 2N\left(2(\NofJ+1)+\frac{2BT}{D}\right)
 \exp\left[-\left(\frac{D}{2eK_0}\right)^\alpha\right].
\end{equation}
In particular, $\mathbb P(E_{T,D}^c)\le p_{\rm rad}$ when
\begin{equation}
\label{eq:gamma-radius-choice}
 D\ge2eK_0\left[
 1+\log\frac{2N\{2(\NofJ+1)+BT/(eK_0)\}}{p_{\rm rad}}
 \right]^{1/\alpha}.
\end{equation}
\end{lemma}
\begin{proof}
Apply the Markov calculation \eqref{eq:moment-tail-markov-substitution}
of \cref{lemma-beta-tail-bound} to the unconditional law of $Z_t^i$,
with $K_{\rm pop}$ replaced by $K_0$ and $D$ by $v\ge eK_0$:
$\mathbb P(\|Z_t^i\|_{\mathbb R^d}>v)\le\exp[-(v/(eK_0))^\alpha]$.
For $B>0$, choose a grid $\mathcal T_D$ of mesh at most $D/(2B)$ on each
closed ODE branch, including both endpoint traces and any isolated
terminal post-jump value. As in \cref{lem:population-moment-event},
$\#\mathcal T_D\le2(\NofJ+1)+2BT/D$.
If $\|Z_t^i\|_{\mathbb R^d}>D$, choose a grid node $s$ on the same branch with
$|t-s|\le D/(2B)$. Then
\[
\begin{gathered}
 \|Z_s^i\|_{\mathbb R^d}\ge\|Z_t^i\|_{\mathbb R^d}-B|t-s|>D-D/2=D/2,
 \\
 E_{T,D}^c\subseteq
 \bigcup_{Z\in\{X,Y\}}\bigcup_{i=1}^N\bigcup_{s\in\mathcal T_D}
       \{\|Z_s^i\|_{\mathbb R^d}>D/2\}.
\end{gathered}
\]
For an isolated terminal value use $s=t$. Substituting $v=D/2\ge eK_0$
in the preceding tail bound and summing over this union gives
\eqref{eq:ETD-tail}; independence is not needed.
For $B=0$, one node per constant segment suffices, and the displayed
bound remains valid.  Finally, $D\ge2eK_0$ implies
$2BT/D\le BT/(eK_0)$, and
\eqref{eq:gamma-radius-choice} implies that
$(D/(2eK_0))^\alpha$ exceeds the logarithm of the union factor divided
by $p_{\rm rad}$.  This proves the final assertion.
\end{proof}

\begin{lemma}[Convenient uniform high-probability bound]
\label{lemma-uniform-sup-gamma-bound-convenient}
Assume \cref{assumption-structured-moment-smoothness,assumption-moment-growth,assumption-uniform-concentration-piecewise-lipschitz}
and the conditional product law of
\cref{lemma-gp-jump-implies-conditional-iid}. Choose the population,
radius, and concentration failure budgets each as $p/3$.
Keep the deterministic schedule size explicit and set
\[
 \Xi=\frac{3N(e+T)(\NofJ+1)(1+B)}p.
\]
Suppose the population localization lemma gives
\[
 \mathbb P(E_{\rm pop}^c)\le p/3,\qquad
 M_{\rm pop}\le C\Xi^{c_M},\qquad
 K_{\rm pop}\le C(1+\log\Xi)^{b_K},
\]
where $C,c_M,b_K$ and $K_0,c_0,\alpha$ are fixed.
For every $\varepsilon>0$, one can choose $C_D,C_\varepsilon$ depending
only on the displayed structural constants and on $\varepsilon$ such
that, with
\begin{equation}
\label{eq:def-DNTP}
 D=C_D(1+\log\Xi)^{1/\alpha},
\end{equation}
the following holds for all sufficiently large $N$:
\[
 \mathbb P\left(
  E_{\rm pop}\cap E_{T,D}\cap
  \left\{\sup_{t\le T}\Gamma_t^N(D,K_{\rm pop})
    \le C_\varepsilon\Xi^\varepsilon
          \left(N^{-1/2}+\frac{BT}{N}\right)\right\}
 \right)\ge1-p.
\]
For a uniformly bounded $B$, $BT/N$ may be replaced by $T/N$ after
changing the constant.  If $\NofJ$ is also uniformly bounded, the factor
$\Xi^\varepsilon$ is equivalent, up to constants, to
$\{N(e+T)/p\}^\varepsilon$.
\end{lemma}
\begin{proof}
\noindent\textbf{Step 1: choose the radius and control the bias.}
Increasing $C_D$ first makes \eqref{eq:gamma-radius-choice} hold with
$p_{\rm rad}=p/3$.  By \eqref{eq:gamma-three-envelope-growth},
\[
 A(D,K_{\rm pop})\vee\Lambda(D,K_{\rm pop})\vee S(D,K_{\rm pop})
 \le C\exp\{C(1+\log\log(e+\Xi))^q\}.
\]
Consequently each of these quantities is bounded by
$C_\eta\Xi^\eta$ for every fixed $\eta>0$.
The bias formula \eqref{eq:gamma-conditional-tail-sum} gives
\[
 \overline\beta_T(D)
 \le C S(D,K_{\rm pop})
 \Xi^{c_M/2}
 \exp\left[-\frac{c_0}{2}
  \left(\frac{C_D}{K_0}\right)^\alpha(1+\log\Xi)\right].
\]
Choose $C_D$ large enough that
$(c_0/2)(C_D/K_0)^\alpha\ge c_M/2+3$.
Use the subpolynomial bound with exponent $1$, so
$S(D,K_{\rm pop})\le C\Xi$. It follows that
\[
 \overline\beta_T(D)
 \le C\Xi^{1+c_M/2-(c_M/2+3)}
 =C\Xi^{-2}\le N^{-1}
\]
for sufficiently large $N$, since $\Xi\ge3eN$.

\par\noindent\textbf{Step 2: absorb the concentration logarithm and combine budgets.}
Finally, $Q_N/(p/3)\le C\Xi$ and
$\sqrt{\log(C\Xi)}\le C_\varepsilon\Xi^{\varepsilon/2}$.
Use \eqref{eq:gamma-three-envelope-growth} with exponent
$\varepsilon/2$ for $A$ and $\Lambda$. Substitution in
\cref{lemma-uniform-sup-gamma-bound-localized-conditional},
\eqref{eq:gamma-explicit-envelope}, gives
\[
\begin{aligned}
 \delta_\Gamma(D)
 &\le N^{-1}+\frac{C_\varepsilon\Xi^{\varepsilon/2}}{\sqrt N}
       (1+\Xi^{\varepsilon/2})
       +\frac{C_\varepsilon BT\Xi^{\varepsilon/2}}N\\
 &\le C_\varepsilon\Xi^\varepsilon(N^{-1/2}+BT/N).
\end{aligned}
\]
The population event, the radius bound \eqref{eq:gamma-radius-choice},
and concentration \eqref{eq:gamma-localized-probability} each cost $p/3$.
Their sum is $p$.
\end{proof}

%% file: appendix_final_general/general_bound_r_t_tilde-gamma_rebuilt.tex
\subsubsection{Weighted-factorization control of $\widetilde\Gamma_t^N$}
\label{subsubsec:weighted-factorization-tilde-gamma}

We bound the full jump covariance residual directly in a Hilbert space of
weighted features. The proof has three steps: factor the Hessian kernel,
concentrate the feature covariance in Hilbert--Schmidt norm, and convert
that estimate to the nuclear norm of the finite jump covariance matrix.

Throughout this subsection, fix a population realization in $E_{\rm pop}$
from \cref{lem:population-moment-event}, and write
\[
 \mathbb E_{\mathcal G}[Z]:=\mathbb E[Z\mid\mathcal G_{\rm pop}],
 \qquad
 \mathbb P_{\mathcal G}(A):=\mathbb P(A\mid\mathcal G_{\rm pop}).
\]
The conditional product law in \cref{lemma-gp-jump-implies-conditional-iid}
and the conditional moment bound \eqref{eq:population-uniform-conditional-moments}
give, at every fixed time and its relevant trace,
\begin{equation}
\label{eq:weighted-feature-conditional-input}
 \operatorname{Law}((Y_t^1,\ldots,Y_t^N)\mid\mathcal G_{\rm pop})
 =\mu_t^{\otimes N},\qquad
 (\mathbb E_{\mathcal G}\|Y_t^1\|_{\mathbb R^d}^r)^{1/r}
 \le K_{\rm pop}r^{1/\alpha}\quad(r\ge1).
\end{equation}
Here $K_{\rm pop}$ is the deterministic envelope chosen for this population
event; it may depend on the horizon and its probability budget. We never
condition further on the sample localization event.

For the fixed deterministic radius $D\ge1$, recall
$\mathrm L_D=\mathrm L_{K_{\rm pop},D}$ from
\cref{def:localized-envelope}. The population moment belongs to
$\mathcal V_{K_{\rm pop},D}$ on $E_{\rm pop}$ by
\cref{lemma-reachable-moment-tube}. Thus the moment-Hessian bound
\eqref{eq:local-moment-derivative-bound} is available here without a bound
on the realized sample positions. We retain $\mathrm L_D^2$ explicitly.

\paragraph{Weighted features and the integral Hessian kernel.}
Recall $a_m=e^{-c_am^{\kappa_{\rm dec}}}$ and
$\|h\|_{\mathsf H_a}^2=\sum_ma_m^2\|h_m\|_{\mathsf H_m}^2$ from
\cref{assumption-structured-moment-smoothness}. Also recall the feature map
$J_0:\mathbb R^d\to\mathsf H_a$, $J_0(x)=(\phi_m(x))_{m\ge1}$, from
\cref{lem:localized-weighted-feature-envelopes}, with
$\mathrm D J_0(x)b=(\mathrm D\phi_m(x)b)_{m\ge1}$. Define
\[
 \mathsf E:=\bigoplus_{m\ge1}\mathsf H_m,\qquad
 \mathsf U_a:\mathsf H_a\longrightarrow\mathsf E,
 \quad(\mathsf U_a h)_m=a_mh_m.
\]
The direct sum has its unweighted Hilbert norm, so
$\|\mathsf U_a h\|_{\mathsf E}=\|h\|_{\mathsf H_a}$, and
$(\mathsf U_a^{-1}u)_m=a_m^{-1}u_m$ satisfies
$\|\mathsf U_a^{-1}u\|_{\mathsf H_a}^2=\sum_m\|u_m\|_{\mathsf H_m}^2$.
Thus $\mathsf U_a$ is an isometric isomorphism. Express the derivative
of $J_0$ in these unweighted coordinates by defining
\begin{equation}
\label{eq:weighted-feature-j0-identification}
\begin{aligned}
 \Psi_\infty(x)&:=\mathsf U_a\circ\mathrm D J_0(x)
       :\mathbb R^d\to\mathsf E,\\
 \Psi_\infty(x)b&=\mathsf U_a((\mathrm D\phi_m(x)b)_m)
       =(a_m\mathrm D\phi_m(x)b)_m.
\end{aligned}
\end{equation}
This map is Hilbert--Schmidt:
\begin{equation}
\label{eq:full-weighted-feature-norm}
 \|\Psi_\infty(x)\|_{\mathrm{HS};\,\mathbb R^d\to\mathsf E}^2
 =\sum_{m\ge1}a_m^2\|\mathrm D\phi_m(x)\|_{\mathrm{HS};\,\mathbb R^d\to\mathsf H_m}^2
 \le d\sum_{m\ge1}a_m^2L_m^2(1+\|x\|_{\mathbb R^d}^{m-1})^2<\infty.
\end{equation}
The factor $d$ follows by summing the squared operator bound over an
orthonormal basis of $\mathbb R^d$. Finiteness follows by squaring \cref{lem:weighted-series-estimate},
\eqref{eq:weighted-series-summand-bound}, with $K=1$ and
$D=1+\|x\|_{\mathbb R^d}$, and summing.

The following factorization uses the full feature space $\mathsf E$.

\begin{lemma}[Weighted Hessian factorization]
\label{lemma-structured-weighted-hessian-factorization}
For $M(\mu)\in\mathcal V_{K_{\rm pop},D}$, there is a bounded self-adjoint
operator $\overline A_\infty(\mu)$ on $\mathsf E$ such that
\begin{align}
 \langle u,\overline A_\infty(\mu)v\rangle
 &=\mathrm D^2G(M(\mu))[\mathsf U_a^{-1}u,\mathsf U_a^{-1}v],
 \label{eq:full-weighted-moment-hessian}\\
 \nabla_W^2F(\mu;x,z)
 &=\Psi_\infty(x)^*\overline A_\infty(\mu)\Psi_\infty(z),
 \qquad \|\overline A_\infty(\mu)\|_{\mathsf E\to\mathsf E}\le\mathrm L_D.
 \label{eq:full-weighted-hessian-factorization}
\end{align}
\end{lemma}
\begin{proof}
\noindent\textbf{Step 1: represent the bounded moment bilinear form.}
By the isometry $\mathsf U_a$ and \eqref{eq:local-moment-derivative-bound},
\[
 |\mathrm D^2G(M(\mu))[\mathsf U_a^{-1}u,\mathsf U_a^{-1}v]|
 \le\mathrm L_D\|u\|_{\mathsf E}\|v\|_{\mathsf E}.
\]
For fixed $v\in\mathsf E$, the map
$u\mapsto\mathrm D^2G(M(\mu))[\mathsf U_a^{-1}u,\mathsf U_a^{-1}v]$
is a continuous linear functional of norm at most
$\mathrm L_D\|v\|_{\mathsf E}$. Riesz representation gives its unique representing
vector $\overline A_\infty(\mu)v$. Uniqueness and linearity in $v$
make this a linear operator, with
\[
 \|\overline A_\infty(\mu)v\|_{\mathsf E}
 =\sup_{\|u\|_{\mathsf E}=1}
   |\mathrm D^2G(M(\mu))[\mathsf U_a^{-1}u,\mathsf U_a^{-1}v]|
 \le\mathrm L_D\|v\|_{\mathsf E}.
\]
Symmetry of the second derivative gives
$\langle u,\overline A_\infty(\mu)v\rangle
=\langle\overline A_\infty(\mu)u,v\rangle$,
so this bounded operator is self-adjoint.
\par\noindent\textbf{Step 2: identify the spatial Hessian kernel.}
For $b,c\in\mathbb R^d$, \cref{eq:weighted-feature-j0-identification}
gives $\mathsf U_a^{-1}\Psi_\infty(x)b=\mathrm D J_0(x)b$ and the
corresponding identity at $z$. Substituting these identities into
\cref{eq:weighted-integral-hessian-representation} and using
\cref{eq:full-weighted-moment-hessian} gives
\[
\begin{aligned}
 \langle b,\nabla_W^2F(\mu;x,z)c\rangle_{\mathbb R^d}
 &=\mathrm D^2G(M(\mu))[\mathrm D J_0(x)b,\mathrm D J_0(z)c]\\
 &=\mathrm D^2G(M(\mu))[
       \mathsf U_a^{-1}\Psi_\infty(x)b,\mathsf U_a^{-1}\Psi_\infty(z)c]\\
 &=\langle\Psi_\infty(x)b,
       \overline A_\infty(\mu)\Psi_\infty(z)c\rangle_{\mathsf E}.
\end{aligned}
\]
Both argument sequences belong to $\mathsf H_a$ by
\eqref{eq:full-weighted-feature-norm}; finite partial sums converge there,
so the displayed equality also identifies the infinite coordinate series.
\end{proof}

\begin{lemma}[Weighted feature bounds with explicit conditional moment dependence]
\label{lemma-weighted-feature-envelope-derived}
Assume \eqref{eq:weighted-feature-conditional-input}. Recall
$\chi(u)=[\log(e+u)]^{\kappa_{\rm dec}/(\kappa_{\rm dec}-1)}$ from
\eqref{eq:weighted-series-growth-function}, and define the single fourth-moment envelope
\begin{equation}
\label{eq:conditional-weighted-fourth-envelope}
 C_{\Phi,4}(K):=
 \left[d\sum_{m\ge1}a_m^2L_m^2
 \left\{1+K^{m-1}(4(m-1)\vee1)^{(m-1)/\alpha}\right\}^2\right]^2.
\end{equation}
All constants $C$ below depend only on the fixed feature parameters,
$d$, and $\alpha$, not on $K_{\rm pop},N,t,D$ or any probability budget.
The following assertions hold.
\begin{enumerate}
 \item The full feature map satisfies
 \begin{equation}
 \label{eq:derived-weighted-feature-growth}
 \|\Psi_\infty(x)\|_{\mathrm{HS};\,\mathbb R^d\to\mathsf E}^2
 \le C\exp\{C\chi(\|x\|_{\mathbb R^d})\}.
 \end{equation}
 The feature map is continuous in Hilbert--Schmidt norm and strongly
 measurable.
 \item Uniformly in $t$,
 \begin{equation}
 \label{eq:derived-weighted-fourth-moment}
 \mathbb E_{\mathcal G}\|\Psi_\infty(Y_t^1)\|_{\mathrm{HS};\,\mathbb R^d\to\mathsf E}^4
 \le C_{\Phi,4}(K_{\rm pop})\le C e^{C\chi(K_{\rm pop})}.
 \end{equation}
 \item The Bochner integral
 $\Sigma_t:=\mathbb E_{\mathcal G}[\Psi_\infty(Y_t^1)\Psi_\infty(Y_t^1)^*]$
 is a positive trace-class operator, and
 \begin{equation}
 \label{eq:derived-weighted-covariance-bound}
 \Sigma_t\preceq C_{\Phi,4}(K_{\rm pop})^{1/2}I_{\mathsf E}.
 \end{equation}
\end{enumerate}
\end{lemma}
\begin{proof}
\noindent\textbf{Step 1: continuity and the pointwise bound.}
For $R\ge1$ and $\|x\|_{\mathbb R^d},\|y\|_{\mathbb R^d}\le R$, integrate $\mathrm D^2\phi_m$ on the line
segment from $y$ to $x$. The derivative-growth assumption gives
\[
 \|\mathrm D\phi_m(x)-\mathrm D\phi_m(y)\|_{\mathrm{HS};\,\mathbb R^d\to\mathsf H_m}
 \le\sqrt d\,L_m(1+R^{(m-2)_+})\|x-y\|_{\mathbb R^d}.
\]
After summing squares with weights $a_m^2$, this yields
\[
 \|\Psi_\infty(x)-\Psi_\infty(y)\|_{\mathrm{HS};\,\mathbb R^d\to\mathsf E}^2
 \le d\|x-y\|_{\mathbb R^d}^2\sum_{m\ge1}a_m^2L_m^2(1+R^{(m-2)_+})^2.
\]
The coefficient is finite by squaring \eqref{eq:weighted-series-summand-bound}
with $K=1$, $D=R$ and summing. This proves continuity on every ball. Its image
lies in a separable closed subspace, since $\mathbb R^d$ is separable and
the map is continuous. Strong measurability means being an almost-sure norm limit of measurable
simple functions. A continuous map from the separable space $\mathbb R^d$
has separable image, so this property holds for $\Psi_\infty(Y_t^1)$
without assuming that the whole feature space is separable.
For the pointwise bound, $1+\|x\|_{\mathbb R^d}^{m-1}\le2+(1+\|x\|_{\mathbb R^d})^m$.
Apply \cref{lem:weighted-series-estimate},
\eqref{eq:weighted-series-summand-bound}, with $K=1$ and $D=1+\|x\|_{\mathbb R^d}$:
\begin{equation}
\label{eq:weighted-feature-growth-series}
 \|\Psi_\infty(x)\|_{\mathrm{HS};\,\mathbb R^d\to\mathsf E}^2
 \le d\sum_m[a_mL_m\{2+(1+\|x\|_{\mathbb R^d})^m\}]^2
 \le Ce^{C\chi(\|x\|_{\mathbb R^d})}\sum_m e^{-(c_a/2)m^{\kappa_{\rm dec}}}.
\end{equation}
Here \eqref{eq:weighted-series-scale-comparison} with $c=1$ absorbs
$\chi(2+\|x\|_{\mathbb R^d})$. The finite sum proves
\eqref{eq:derived-weighted-feature-growth}.

\par\noindent\textbf{Step 2: the fourth moment.}
For $m\ge2$, \eqref{eq:weighted-feature-conditional-input} with $r=4(m-1)$
gives
\[
 \|1+\|Y_t^1\|_{\mathbb R^d}^{m-1}\|_{L^4(\mathbb P_{\mathcal G})}
 \le1+(\mathbb E_{\mathcal G}\|Y_t^1\|_{\mathbb R^d}^{4(m-1)})^{1/4}
 \le1+K_{\rm pop}^{m-1}(4(m-1))^{(m-1)/\alpha}.
\]
For $m=1$ the left side is $2$, matching the convention in
\eqref{eq:conditional-weighted-fourth-envelope}. Minkowski's inequality
for a finite sum in $L^2(\mathbb P_{\mathcal G})$ now gives
\[
\begin{aligned}
 \left\|\sum_{m=1}^M a_m^2
       \|\mathrm D\phi_m(Y_t^1)\|_{\mathrm{HS};\,\mathbb R^d\to\mathsf H_m}^2\right\|_{L^2(\mathbb P_{\mathcal G})}
 &\le d\sum_{m\le M}a_m^2L_m^2
       \|1+\|Y_t^1\|_{\mathbb R^d}^{m-1}\|_{L^4(\mathbb P_{\mathcal G})}^2\\
 &\le d\sum_{m\le M}a_m^2L_m^2
 \{1+K_{\rm pop}^{m-1}(4(m-1)\vee1)^{(m-1)/\alpha}\}^2.
\end{aligned}
\]
Squaring and applying monotone convergence gives the explicit connection
between the finite sums and the full fourth moment:
\begin{equation}
\label{eq:weighted-fourth-moment-limit}
\begin{aligned}
 \mathbb E_{\mathcal G}\|\Psi_\infty(Y_t^1)\|_{\mathrm{HS};\,\mathbb R^d\to\mathsf E}^4
 &=\lim_{M\to\infty}\mathbb E_{\mathcal G}
 \left[\sum_{m=1}^M a_m^2
          \|\mathrm D\phi_m(Y_t^1)\|_{\mathrm{HS};\,\mathbb R^d\to\mathsf H_m}^2\right]^2\\
 &\le C_{\Phi,4}(K_{\rm pop}).
\end{aligned}
\end{equation}
For every $m\ge1$, including $m=1$, the braces in
\eqref{eq:conditional-weighted-fourth-envelope} satisfy
\[
 1+K^{m-1}(4(m-1)\vee1)^{(m-1)/\alpha}
 \le 1+\{4^{1/\alpha}(1+K)\}^m m^{m/\alpha}.
\]
Use \eqref{eq:weighted-series-summand-bound} with
$K$ replaced by $4^{1/\alpha}(1+K_{\rm pop})$ and $D=1$.
Squaring each summand, summing, and then squaring the sum in
\eqref{eq:conditional-weighted-fourth-envelope} yields
\[
 C_{\Phi,4}(K_{\rm pop})
 \le\left[Ce^{C\chi(K_{\rm pop})}
           \sum_m e^{-(c_a/2)m^{\kappa_{\rm dec}}}\right]^2
 \le Ce^{C\chi(K_{\rm pop})},
\]
where \eqref{eq:weighted-series-scale-comparison} absorbs the fixed
rescaling. This proves \eqref{eq:derived-weighted-fourth-moment}.
\par\noindent\textbf{Step 3: the covariance.}
For Hilbert--Schmidt maps $A,B:\mathbb R^d\to\mathsf E$, its singular
values $s_j$ give
\[
 \operatorname{tr}(AA^*)=\sum_js_j^2=\|A\|_{\mathrm{HS};\,\mathbb R^d\to\mathsf E}^2,
 \qquad
 \|AA^*\|_{\mathrm{HS};\,\mathsf E\to\mathsf E}=(\sum_js_j^4)^{1/2}\le\sum_js_j^2.
\]
Moreover, $AA^*-BB^*=(A-B)A^*+B(A-B)^*$ gives
\[
 \|AA^*-BB^*\|_{*;\,\mathsf E\to\mathsf E}
 \le(\|A\|_{\mathrm{HS};\,\mathbb R^d\to\mathsf E}+\|B\|_{\mathrm{HS};\,\mathbb R^d\to\mathsf E})\|A-B\|_{\mathrm{HS};\,\mathbb R^d\to\mathsf E}.
\]
Thus $A\mapsto AA^*$ is continuous from Hilbert--Schmidt norm to trace
norm, using \eqref{eq:nuclear-hilbert-schmidt-product}.
Composition with the strongly measurable feature map from Step 1
therefore gives strong measurability in trace norm. For integrability,
positivity of $\Psi_\infty(Y_t^1)\Psi_\infty(Y_t^1)^*$ and
Cauchy--Schwarz give
\[
\begin{aligned}
 \mathbb E_{\mathcal G}
   \|\Psi_\infty(Y_t^1)\Psi_\infty(Y_t^1)^*\|_{*;\,\mathsf E\to\mathsf E}
 &=\mathbb E_{\mathcal G}\|\Psi_\infty(Y_t^1)\|_{\mathrm{HS};\,\mathbb R^d\to\mathsf E}^2\\
 &\le\bigl(\mathbb E_{\mathcal G}
                  \|\Psi_\infty(Y_t^1)\|_{\mathrm{HS};\,\mathbb R^d\to\mathsf E}^4\bigr)^{1/2}
 \le C_{\Phi,4}(K_{\rm pop})^{1/2}<\infty.
\end{aligned}
\]
This establishes Bochner integrability in trace norm.
For $u\in\mathsf E$, Cauchy--Schwarz gives
\[
 \langle u,\Sigma_tu\rangle
 =\mathbb E_{\mathcal G}\|\Psi_\infty(Y_t^1)^*u\|_{\mathbb R^d}^2
 \le\|u\|_{\mathsf E}^2\mathbb E_{\mathcal G}\|\Psi_\infty(Y_t^1)\|_{\mathrm{HS};\,\mathbb R^d\to\mathsf E}^2
 \le C_{\Phi,4}(K_{\rm pop})^{1/2}\|u\|_{\mathsf E}^2.
\]
\end{proof}

\paragraph{The full residual as a feature covariance error.}
Recall the kernel covariance residual from \cref{lemma-kernel-mismatch-bound}:
\begin{equation}
\label{eq:tilde-gamma-full-definition-weighted}
 U_t(z):=[\nabla_W^2F(\mu_t;Y_t^i,z)]_{i=1}^N,
 \qquad
 \widetilde\Gamma_t^N
 =\frac1N\left\|\int U_t(z)U_t(z)^\top
                          (\mu_t-\rho_{Y_t})(dz)\right\|_{*;\,\mathbb R^{dN}\to\mathbb R^{dN}}.
\end{equation}
The matrix $U_t(z)$ has $N$ block rows and $d$ columns. Define
\[
 \Psi_t^{Y,N}(v_1,\ldots,v_N):=\sum_{i=1}^N\Psi_\infty(Y_t^i)v_i,
 \qquad
 \Psi_t^{Y,N}:\mathbb R^{dN}\longrightarrow\mathsf E,
\]
where the domain has the ordinary Euclidean inner product. Recall $\Sigma_t$
from \cref{lemma-weighted-feature-envelope-derived} and define
\[
 \widehat\Sigma_t^N:=\frac1N\sum_{i=1}^N
       \Psi_\infty(Y_t^i)\Psi_\infty(Y_t^i)^*
       =\frac1N\Psi_t^{Y,N}(\Psi_t^{Y,N})^*.
\]

\begin{lemma}[Full covariance representation and nuclear-norm bound]
\label{lem:tilde-gamma-weighted-representation}
The full residual satisfies
\begin{align}
 \widetilde\Gamma_t^N
 &=\frac1N\left\|(\Psi_t^{Y,N})^*\overline A_\infty(\mu_t)
       (\Sigma_t-\widehat\Sigma_t^N)
       \overline A_\infty(\mu_t)\Psi_t^{Y,N}\right\|_{*;\,\mathbb R^{dN}\to\mathbb R^{dN}},
 \label{eq:tilde-gamma-weighted-matrix-identity}\\
 \widetilde\Gamma_t^N
 &\le\mathrm L_D^2\operatorname{tr}(\widehat\Sigma_t^N)
                  \|\widehat\Sigma_t^N-\Sigma_t\|_{\mathrm{HS};\,\mathsf E\to\mathsf E}.
 \label{eq:tilde-gamma-deterministic-reduction}
\end{align}
Consequently,
\begin{equation}
\label{eq:tilde-gamma-full-expectation-bound}
 \mathbb E_{\mathcal G}\widetilde\Gamma_t^N
 \le\frac{\mathrm L_D^2 C_{\Phi,4}(K_{\rm pop})}{\sqrt N}.
\end{equation}
\end{lemma}
\begin{proof}
\noindent\textbf{Step 1: factor and bound the covariance residual.}
By \eqref{eq:full-weighted-hessian-factorization},
\[
 U_t(z)=(\Psi_t^{Y,N})^*\overline A_\infty(\mu_t)\Psi_\infty(z).
\]
Since $\overline A_\infty(\mu_t)$ is self-adjoint, the product is
\[
 U_t(z)U_t(z)^\top
 =(\Psi_t^{Y,N})^*\overline A_\infty(\mu_t)
 [\Psi_\infty(z)\Psi_\infty(z)^*]
 \overline A_\infty(\mu_t)\Psi_t^{Y,N}.
\]
Only the bracketed factor depends on $z$. Integration of this factor
$\Psi_\infty(z)\Psi_\infty(z)^*$ against $\mu_t$ gives $\Sigma_t$,
and integration against $\rho_{Y_t}=N^{-1}\sum_i\delta_{Y_t^i}$ gives
$\widehat\Sigma_t^N$. Subtracting proves
\eqref{eq:tilde-gamma-weighted-matrix-identity}.
Apply the Hilbert--Schmidt product bound
\eqref{eq:nuclear-hilbert-schmidt-product} to the two factors
$(\Psi_t^{Y,N})^*\overline A_\infty(\mu_t)$ and
$(\Sigma_t-\widehat\Sigma_t^N)\overline A_\infty(\mu_t)\Psi_t^{Y,N}$.
Then use submultiplicativity of the Hilbert--Schmidt norm with the
operator norm, also after taking adjoints:
\[
\begin{aligned}
 \widetilde\Gamma_t^N
 &\le\frac1N\|(\Psi_t^{Y,N})^*\overline A_\infty(\mu_t)\|_{\mathrm{HS};\,\mathsf E\to\mathbb R^{dN}}
 \|(\Sigma_t-\widehat\Sigma_t^N)\overline A_\infty(\mu_t)
                              \Psi_t^{Y,N}\|_{\mathrm{HS};\,\mathbb R^{dN}\to\mathsf E}\\
 &\le\frac1N\|\Psi_t^{Y,N}\|_{\mathrm{HS};\,\mathbb R^{dN}\to\mathsf E}^2
       \|\overline A_\infty(\mu_t)\|_{\mathsf E\to\mathsf E}^2
       \|\Sigma_t-\widehat\Sigma_t^N\|_{\mathsf E\to\mathsf E}\\
 &\le\mathrm L_D^2\operatorname{tr}(\widehat\Sigma_t^N)
       \|\Sigma_t-\widehat\Sigma_t^N\|_{\mathrm{HS};\,\mathsf E\to\mathsf E}.
\end{aligned}
\]
Here $N^{-1}\|\Psi_t^{Y,N}\|_{\mathrm{HS};\,\mathbb R^{dN}\to\mathsf E}^2
=N^{-1}\sum_i\|\Psi_\infty(Y_t^i)\|_{\mathrm{HS};\,\mathbb R^d\to\mathsf E}^2
=\operatorname{tr}(\widehat\Sigma_t^N)$.

\par\noindent\textbf{Step 2: bound its conditional expectation.}
For the expectation bound put
$W_i=\Psi_\infty(Y_t^i)\Psi_\infty(Y_t^i)^*$ and
$Z_i=\|\Psi_\infty(Y_t^i)\|_{\mathrm{HS};\,\mathbb R^d\to\mathsf E}^2$. Jensen's inequality gives
\[
 \mathbb E_{\mathcal G}\left(\operatorname{tr}\widehat\Sigma_t^N\right)^2
 =\mathbb E_{\mathcal G}(N^{-1}\sum_iZ_i)^2
 \le N^{-1}\sum_i\mathbb E_{\mathcal G}Z_i^2
 \le C_{\Phi,4}(K_{\rm pop}).
\]
In the Hilbert space $\mathcal S_2(\mathsf E)$, conditional independence
and $\mathbb E_{\mathcal G}(W_i-\Sigma_t)=0$ imply, for $i\ne j$,
\[
 \mathbb E_{\mathcal G}\langle W_i-\Sigma_t,W_j-\Sigma_t\rangle_{\mathrm{HS};\,\mathsf E\to\mathsf E}
 =\langle\mathbb E_{\mathcal G}(W_i-\Sigma_t),
          \mathbb E_{\mathcal G}(W_j-\Sigma_t)\rangle_{\mathrm{HS};\,\mathsf E\to\mathsf E}=0.
\]
In particular, the singular-value calculation in
\cref{lemma-weighted-feature-envelope-derived} gives
$\|W_1\|_{\mathrm{HS};\,\mathsf E\to\mathsf E}^2\le(\operatorname{tr}W_1)^2=Z_1^2$,
with an inequality rather than equality in general. Expanding gives
\begin{equation}
\label{eq:weighted-covariance-mean-square}
\begin{aligned}
 \mathbb E_{\mathcal G}\|\widehat\Sigma_t^N-\Sigma_t\|_{\mathrm{HS};\,\mathsf E\to\mathsf E}^2
 &=N^{-2}\sum_{i,j}\mathbb E_{\mathcal G}
        \langle W_i-\Sigma_t,W_j-\Sigma_t\rangle_{\mathrm{HS};\,\mathsf E\to\mathsf E}\\
 &=N^{-2}\sum_i\mathbb E_{\mathcal G}\|W_i-\Sigma_t\|_{\mathrm{HS};\,\mathsf E\to\mathsf E}^2\\
 &=N^{-1}(\mathbb E_{\mathcal G}\|W_1\|_{\mathrm{HS};\,\mathsf E\to\mathsf E}^2-\|\Sigma_t\|_{\mathrm{HS};\,\mathsf E\to\mathsf E}^2)\\
 &\le N^{-1}\mathbb E_{\mathcal G}Z_1^2
 \le C_{\Phi,4}(K_{\rm pop})/N.
\end{aligned}
\end{equation}
Conditional Cauchy--Schwarz in \eqref{eq:tilde-gamma-deterministic-reduction}
now gives
\[
\begin{aligned}
 \mathbb E_{\mathcal G}\widetilde\Gamma_t^N
 &\le\mathrm L_D^2
 [\mathbb E_{\mathcal G}(\operatorname{tr}\widehat\Sigma_t^N)^2]^{1/2}
 [\mathbb E_{\mathcal G}\|\widehat\Sigma_t^N-\Sigma_t\|_{\mathrm{HS};\,\mathsf E\to\mathsf E}^2]^{1/2}\\
 &\le\mathrm L_D^2 C_{\Phi,4}(K_{\rm pop})^{1/2}
          [C_{\Phi,4}(K_{\rm pop})/N]^{1/2}.
\end{aligned}
\]
The two factors use the same sample; their independence is not needed.
\end{proof}

\begin{lemma}[A deterministic envelope at the sampled particles]
\label{lem:weighted-feature-max-event}
Fix a time trace $t$, $N\ge2$, and $\delta\in(0,1)$. Define
\[
 R_{N,\delta}:=K_{\rm pop}
   [e(\alpha\vee1)\{1+\log(8N/\delta)\}]^{1/\alpha},
\]
\begin{equation}
\label{eq:weighted-feature-max-tau-N-delta}
 \tau_{N,\delta}:=1\vee
   \left[d\sum_{m\ge1}a_m^2L_m^2(1+R_{N,\delta}^{m-1})^2\right].
\end{equation}
Then
\begin{equation}
\label{eq:weighted-feature-max-event}
 \mathbb P_{\mathcal G}\left(
   \max_{i\le N}\|\Psi_\infty(Y_t^i)\|_{\mathrm{HS};\,\mathbb R^d\to\mathsf E}^2>\tau_{N,\delta}
                         \right)\le\delta/8,
\end{equation}
and the individual-particle probability is at most $\delta/(8N)$.
Moreover $\tau_{N,\delta}\le C\exp\{C\chi(R_{N,\delta})\}$.
If $\Xi\ge N/\delta$ and $K_{\rm pop}\le C_K(\log\Xi)^{s_K}$ for fixed
$C_K,s_K$, then, for every $\varepsilon>0$,
\begin{equation}
\label{eq:weighted-envelope-subpolynomial}
 \tau_{N,\delta}^2\sqrt{\log(8/\delta)}
       +C_{\Phi,4}(K_{\rm pop})\le C_\varepsilon \Xi^\varepsilon.
\end{equation}
\end{lemma}
\begin{proof}
Use the Markov calculation of \cref{lemma-beta-tail-bound},
\eqref{eq:moment-tail-markov-substitution}, with threshold $R_{N,\delta}$.
In the moment input \eqref{eq:weighted-feature-conditional-input}, take
$r=(\alpha\vee1)\{1+\log(8N/\delta)\}\ge1$. Then
\[
 \mathbb P_{\mathcal G}(\|Y_t^i\|_{\mathbb R^d}>R_{N,\delta})
 \le\left(\frac{K_{\rm pop}r^{1/\alpha}}
                   {K_{\rm pop}(er)^{1/\alpha}}\right)^r
 =e^{-r/\alpha}\le\frac{\delta}{8N}.
\]
By \eqref{eq:full-weighted-feature-norm} and
\eqref{eq:weighted-feature-max-tau-N-delta},
\[
 \{\|\Psi_\infty(Y_t^i)\|_{\mathrm{HS};\,\mathbb R^d\to\mathsf E}^2>\tau_{N,\delta}\}
 \subseteq\{\|Y_t^i\|_{\mathbb R^d}>R_{N,\delta}\}.
\]
The union over $N$ particles costs $N\delta/(8N)=\delta/8$, proving
\eqref{eq:weighted-feature-max-event}.
The calculation \eqref{eq:weighted-feature-growth-series} in
\cref{lemma-weighted-feature-envelope-derived}, with $\|x\|_{\mathbb R^d}$ replaced by
$R_{N,\delta}$, proves the growth bound.
Since $\Xi\ge N/\delta\ge2$,
$\log(8N/\delta)\le\log8+\log\Xi\le4\log\Xi$. Hence
\[
 R_{N,\delta}\le C(\log\Xi)^{s_K+1/\alpha},\qquad
 \log(e+R_{N,\delta})\le C+C\log\log\Xi.
\]
For the whole left-hand side of \eqref{eq:weighted-envelope-subpolynomial},
these bounds and \eqref{eq:derived-weighted-fourth-moment} give
\[
\begin{aligned}
 &\log\{\tau_{N,\delta}^2\sqrt{\log(8/\delta)}
                      +C_{\Phi,4}(K_{\rm pop})\}\\
 &\quad\le C+C[1+\log\log(e+\Xi)]^{\kappa_{\rm dec}/(\kappa_{\rm dec}-1)}
 \le C_\varepsilon+\varepsilon\log\Xi.
\end{aligned}
\]
The last inequality holds because any fixed power of $\log\log(e+\Xi)$
is $o(\log\Xi)$; a fixed constant covers bounded $\Xi\ge2$.
Exponentiation proves the claim. The constant may depend on $C_K,s_K$
and $\varepsilon$, but not on $\Xi$.
\end{proof}

\begin{lemma}[Full weighted covariance concentration in Hilbert--Schmidt norm]
\label{lem:tilde-gamma-weighted-covariance-input}
Fix a time trace $t$, $N\ge2$, and $\delta\in(0,1)$. Use the threshold
$\tau_{N,\delta}$ in \eqref{eq:weighted-feature-max-tau-N-delta}.
There is an event of conditional probability at least $1-\delta$ on which
\begin{align}
 \max_{i\le N}\|\Psi_\infty(Y_t^i)\|_{\mathrm{HS};\,\mathbb R^d\to\mathsf E}^2&\le\tau_{N,\delta},
 \label{eq:full-covariance-sample-envelope}\\
 \|\widehat\Sigma_t^N-\Sigma_t\|_{\mathrm{HS};\,\mathsf E\to\mathsf E}
 &\le\frac{C_{\Phi,4}(K_{\rm pop})^{1/2}}{\sqrt N}
          (1+\sqrt{\delta/8})
       +\tau_{N,\delta}\sqrt{\frac{2\log(8/\delta)}N}.
 \label{eq:tilde-gamma-covariance-good-event}
\end{align}
\end{lemma}
\begin{proof}
The repeated random objects needed in this proof are
\[
 W_i:=\Psi_\infty(Y_t^i)\Psi_\infty(Y_t^i)^*,\quad
 Z_i:=\|\Psi_\infty(Y_t^i)\|_{\mathrm{HS};\,\mathbb R^d\to\mathsf E}^2,\quad
 W_i^{(\tau)}:=W_i\mathbf1_{\{Z_i\le\tau_{N,\delta}\}}.
\]
The conditional mean of $W_i^{(\tau)}$ is
$\Sigma_t^{(\tau)}:=\mathbb E_{\mathcal G}W_1^{(\tau)}$.
The sample mean of $W_i^{(\tau)}$ is
$\widehat\Sigma_t^{N,(\tau)}:=N^{-1}\sum_iW_i^{(\tau)}$.
Conditional on $\mathcal G_{\rm pop}$, these clipped variables remain
independent. We apply concentration before intersecting with
$\{\max_iZ_i\le\tau_{N,\delta}\}$.

\par\noindent\textbf{Step 1: mean square and bounded differences.}
The singular-value inequality in
\cref{lemma-weighted-feature-envelope-derived} gives
$\|W_i^{(\tau)}\|_{\mathrm{HS};\,\mathsf E\to\mathsf E}\le Z_i\mathbf1_{\{Z_i\le\tau_{N,\delta}\}}
\le\tau_{N,\delta}$. These operators are conditionally i.i.d. and
square integrable. Apply the centered-sum calculation
\eqref{eq:weighted-covariance-mean-square} of
\cref{lem:tilde-gamma-weighted-representation}, replacing
$W_i,\Sigma_t$ by $W_i^{(\tau)},\Sigma_t^{(\tau)}$:
\[
 \mathbb E_{\mathcal G}
 \|\widehat\Sigma_t^{N,(\tau)}-\Sigma_t^{(\tau)}\|_{\mathrm{HS};\,\mathsf E\to\mathsf E}^2
 \le N^{-1}\mathbb E_{\mathcal G}\|W_1^{(\tau)}\|_{\mathrm{HS};\,\mathsf E\to\mathsf E}^2
 \le N^{-1}\mathbb E_{\mathcal G}Z_1^2
 \le C_{\Phi,4}(K_{\rm pop})/N.
\]
Cauchy--Schwarz therefore bounds the mean norm by
$\sqrt{C_{\Phi,4}(K_{\rm pop})/N}$.
If sample $i$ is replaced, denote the new sample mean by
$\widehat\Sigma_t^{N,(\tau)\prime}$ and its changed summand by
$W_i^{(\tau)\prime}$. The threshold and population mean are fixed, so
\[
\begin{aligned}
 &\left|\|\widehat\Sigma_t^{N,(\tau)}-\Sigma_t^{(\tau)}\|_{\mathrm{HS};\,\mathsf E\to\mathsf E}
 -\|\widehat\Sigma_t^{N,(\tau)\prime}-\Sigma_t^{(\tau)}\|_{\mathrm{HS};\,\mathsf E\to\mathsf E}\right|\\
 &\quad\le\|\widehat\Sigma_t^{N,(\tau)}-
                        \widehat\Sigma_t^{N,(\tau)\prime}\|_{\mathrm{HS};\,\mathsf E\to\mathsf E}
 =N^{-1}\|W_i^{(\tau)}-W_i^{(\tau)\prime}\|_{\mathrm{HS};\,\mathsf E\to\mathsf E}
 \le 2\tau_{N,\delta}/N.
\end{aligned}
\]
In \cref{lemma:mcdiarmid}, the sum of squared differences is
$N(2\tau_{N,\delta}/N)^2=4\tau_{N,\delta}^2/N$. Substituting
$s=\tau_{N,\delta}\sqrt{2\log(8/\delta)/N}$ gives failure probability
\[
 \exp\left(-\frac{2s^2}{4\tau_{N,\delta}^2/N}\right)
 =\exp(-\log(8/\delta))=\delta/8.
\]
Thus, except on that event,
\[
 \|\widehat\Sigma_t^{N,(\tau)}-\Sigma_t^{(\tau)}\|_{\mathrm{HS};\,\mathsf E\to\mathsf E}
 \le\sqrt{C_{\Phi,4}(K_{\rm pop})/N}
       +\tau_{N,\delta}\sqrt{2\log(8/\delta)/N}.
\]

\par\noindent\textbf{Step 2: restore the conditional mean and the sample.}
The mean difference is
$\Sigma_t-\Sigma_t^{(\tau)}=\mathbb E_{\mathcal G}[W_1\mathbf1_{\{Z_1>\tau_{N,\delta}\}}]$.
The individual tail bound in \cref{lem:weighted-feature-max-event}
and Cauchy--Schwarz give
\[
\begin{aligned}
 \|\Sigma_t-\Sigma_t^{(\tau)}\|_{\mathrm{HS};\,\mathsf E\to\mathsf E}
 &\le\mathbb E_{\mathcal G}
       [\|W_1\|_{\mathrm{HS};\,\mathsf E\to\mathsf E}\mathbf1_{\{Z_1>\tau_{N,\delta}\}}]\\
 &\le(\mathbb E_{\mathcal G}Z_1^2)^{1/2}
       \mathbb P_{\mathcal G}(Z_1>\tau_{N,\delta})^{1/2}
 \le C_{\Phi,4}(K_{\rm pop})^{1/2}\sqrt{\delta/(8N)}.
\end{aligned}
\]
The bound \eqref{eq:weighted-feature-max-event} in
\cref{lem:weighted-feature-max-event} costs another $\delta/8$ to ensure
$\max_iZ_i\le\tau_{N,\delta}$ and hence
$\widehat\Sigma_t^N=\widehat\Sigma_t^{N,(\tau)}$. On the intersection,
\[
\begin{aligned}
 \|\widehat\Sigma_t^N-\Sigma_t\|_{\mathrm{HS};\,\mathsf E\to\mathsf E}
 &\le\|\widehat\Sigma_t^{N,(\tau)}-\Sigma_t^{(\tau)}\|_{\mathrm{HS};\,\mathsf E\to\mathsf E}
          +\|\Sigma_t^{(\tau)}-\Sigma_t\|_{\mathrm{HS};\,\mathsf E\to\mathsf E}\\
 &\le\frac{C_{\Phi,4}(K_{\rm pop})^{1/2}}{\sqrt N}(1+\sqrt{\delta/8})
       +\tau_{N,\delta}\sqrt{\frac{2\log(8/\delta)}N}.
\end{aligned}
\]
The population bias bound is deterministic after conditioning and costs
no further probability. Thus the failure budget is
$\delta/8+\delta/8\le\delta$, proving
\eqref{eq:tilde-gamma-covariance-good-event}.
\end{proof}

\begin{corollary}[The full jump residual at a fixed time]
\label{cor:tilde-gamma-full-weighted-bound}
For every fixed time trace $t$, $N\ge2$, and $\delta\in(0,1)$, with
conditional probability at least $1-\delta$,
\begin{equation}
\label{eq:tilde-gamma-full-high-prob-bound}
 \widetilde\Gamma_t^N
 \le\frac{\mathrm L_D^2\tau_{N,\delta}}{\sqrt N}
       \left[2C_{\Phi,4}(K_{\rm pop})^{1/2}
                 +\tau_{N,\delta}\sqrt{2\log(8/\delta)}\right].
\end{equation}
\end{corollary}
\begin{proof}
On the joint event of \cref{lem:tilde-gamma-weighted-covariance-input},
\[
 \operatorname{tr}(\widehat\Sigma_t^N)
 =N^{-1}\sum_i\|\Psi_\infty(Y_t^i)\|_{\mathrm{HS};\,\mathbb R^d\to\mathsf E}^2\le\tau_{N,\delta}.
\]
Substitute this and \eqref{eq:tilde-gamma-covariance-good-event} into
\cref{lem:tilde-gamma-weighted-representation},
\eqref{eq:tilde-gamma-deterministic-reduction}:
\[
 \widetilde\Gamma_t^N\le\mathrm L_D^2\tau_{N,\delta}
 \left[\frac{C_{\Phi,4}(K_{\rm pop})^{1/2}}{\sqrt N}(1+\sqrt{\delta/8})
       +\tau_{N,\delta}\sqrt{\frac{2\log(8/\delta)}N}\right].
\]
Use $1+\sqrt{\delta/8}\le2$ to obtain
\eqref{eq:tilde-gamma-full-high-prob-bound} on the same event.
\end{proof}

\begin{corollary}[Union bound over prescribed jump times]
\label{cor:tilde-gamma-jump-time-union-bound}
Let $0<t_1<\cdots<t_{\NofJ}\le T$ be deterministic, and set
$q=p/(\NofJ+1)$ for $p\in(0,1)$. Define
\begin{equation}
\label{eq:phase1-HS-tilde-delta-envelope}
 \widetilde\delta_{N,T,p}^{\rm HS}
 :=\frac{\mathrm L_D^2\tau_{N,q}}{\sqrt N}
       \left[2C_{\Phi,4}(K_{\rm pop})^{1/2}
                    +\tau_{N,q}\sqrt{2\log(8/q)}\right].
\end{equation}
Then, with conditional probability at least $1-p$,
\begin{equation}
\label{eq:tilde-gamma-jump-time-union-bound}
 \max_{1\le\ell\le\NofJ}\widetilde\Gamma_{t_\ell-}^N
 \le\widetilde\delta_{N,T,p}^{\rm HS}.
\end{equation}
The maximum is zero when $\NofJ=0$. The parameters $K_{\rm pop}$ and
$\mathrm L_D$ retain their horizon and population-confidence dependence.
If $K_{\rm pop},D$ grow at most as fixed powers of $\log\Xi$,
$\Xi\ge N(\NofJ+1)/p$, and the structural parameters are fixed, then
for every $\varepsilon>0$,
\begin{equation}
\label{eq:tilde-gamma-jump-time-final-rate}
 \widetilde\delta_{N,T,p}^{\rm HS}
 \le C_\varepsilon N^{-1/2}\Xi^\varepsilon.
\end{equation}
In particular the residual has rate $N^{-1/2+o(1)}$ when $\Xi$ is polynomial
in $N$; its square root in the jump comparison has rate $N^{-1/4+o(1)}$.
\end{corollary}
\begin{proof}
\noindent\textbf{Step 1: combine the prescribed jump times.}
Apply \cref{cor:tilde-gamma-full-weighted-bound} at each deterministic
pre-jump trace with $\delta=q=p/(\NofJ+1)$ in
\eqref{eq:tilde-gamma-full-high-prob-bound}. The union costs
$\NofJ q=p\NofJ/(\NofJ+1)\le p$; different times need not be independent.

\par\noindent\textbf{Step 2: simplify the deterministic coefficient.}
For the rate, \eqref{eq:local-envelope-growth} bounds
$\mathrm L_D\le Ce^{C\chi(K_{\rm pop}+D)}$.
For polylogarithmic $K_{\rm pop},D$, this is at most
$C_\eta \Xi^\eta$ for every $\eta>0$, by the logarithm calculation proving
\eqref{eq:weighted-envelope-subpolynomial} in
\cref{lem:weighted-feature-max-event}. Apply
\eqref{eq:weighted-envelope-subpolynomial} with the same $\eta$.
Since $\log(8/q)>1$, it implies
\[
 \tau_{N,q}\le C_\eta \Xi^{\eta/2},\quad
 C_{\Phi,4}(K_{\rm pop})^{1/2}\le C_\eta \Xi^{\eta/2},\quad
 \tau_{N,q}^2\sqrt{\log(8/q)}\le C_\eta \Xi^\eta.
\]
The bracket after multiplication by $\tau_{N,q}$ is therefore at most
$C_\eta \Xi^\eta$. Together with $\mathrm L_D^2\le C_\eta \Xi^{2\eta}$,
this gives \eqref{eq:tilde-gamma-jump-time-final-rate} by taking
$\eta=\varepsilon/3$.

\par\noindent\textbf{Step 3: average the conditional failure probability.}
Finally, since $E_{\rm pop}\in\mathcal G_{\rm pop}$,
\[
\begin{aligned}
 &\mathbb P\left(E_{\rm pop}\cap
   \{\max_\ell\widetilde\Gamma_{t_\ell-}^N
                  >\widetilde\delta_{N,T,p}^{\rm HS}\}\right)\\
 &\quad=\mathbb E\left[\mathbf1_{E_{\rm pop}}
       \mathbb P_{\mathcal G}\left(
          \max_\ell\widetilde\Gamma_{t_\ell-}^N
                     >\widetilde\delta_{N,T,p}^{\rm HS}\right)\right]
 \le p.
\end{aligned}
\]
Adding $\mathbb P(E_{\rm pop}^c)$ gives the unconditional failure budget
when the population event is also required.
\end{proof}

\begin{remark}[Why no feature cutoff or nuclear-norm rank loss is needed]
\label{remark:tilde-gamma-no-rank-loss}
The finite $dN\times dN$ residual factors through the full Hilbert feature
space $\mathsf E$. The independent random operators are clipped by their
norm before scalar bounded differences is applied in $\mathcal S_2(\mathsf E)$.
Neither argument uses the dimension of a feature space. The nuclear norm
is bounded by the empirical feature trace times the Hilbert--Schmidt
covariance error on one event. This proves the bound for
$\widetilde\Gamma_t^N$ directly.
\end{remark}

%% file: appendix_final_general/02_fixed_schedule_statement.tex
\subsection{Fixed-schedule comparison with explicit high-probability rates}
\label{subsec:fixed-schedule-package}

We assemble the deterministic ODE--jump comparison and the two
stochastic concentration bounds into the final fixed-schedule tracking theorem.

This subsection packages the detailed estimates from
\cref{lemma-common-empirical-moment-residual-directional-rt},
\cref{lem:jump-bound-common-seed},
\cref{lem:piecewise-gronwall-forced-jumps},
\cref{lemma-uniform-sup-gamma-bound-convenient},
\cref{cor:tilde-gamma-jump-time-union-bound}
into a single fixed-schedule statement.  

We keep only the quantities that enter the final comparison bound.
\begin{itemize}
    \item the pathwise comparison variable is
    \[
    r_t=\frac{1}{\sqrt N}\norm{\bs X_t-\bs Y_t}_2,
    \]
    \item the continuous stochastic residual is
    \[
    \Gamma_t^N=\sum_{m\ge1}\mathrm L_Da_m
    \left\|\frac1N\sum_{i=1}^N\phi_m(Y_t^i)
                         -\int\phi_m(y)\,\mu_t(dy)\right\|_{\mathsf H_m},
    \]
    as in \eqref{eq:common-moment-residual-and-lipschitz-sum},
    \item the jump defect is controlled by $\widetilde\Gamma_{t_\ell-}^N$
    from \eqref{eq:tilde-gamma-full-definition-weighted} only at the prescribed jump times,
    \item curvature is retained through the propagator
    $G_\kappa^Y(t,s)=\exp(-\int_s^t\kappa_u^{Y,N}\,du)$, using the
    sampled full Hessian curvature in \eqref{eq:curvature-main}.
\end{itemize}

The first lemma solves the deterministic comparison inequality across ODE
intervals and multiplicative jump errors.

\begin{lemma}[Piecewise Gr\"onwall estimate with recursive jump amplification]
\label{lem:piecewise-gronwall-forced-jumps}
Fix $T>0$ and jump times
\[
0<t_1<\cdots<t_{\NofJ}\le T,
\qquad
t_0:=0,
\qquad
t_{\NofJ+1}:=T.
\]
Assume that $r_t:=\frac{1}{\sqrt N}\norm{\bs X_t-\bs Y_t}_2$
is absolutely continuous on each interval $(t_\ell,t_{\ell+1})$, $\ell=0,\dots,\NofJ$.
Assume right-continuity and absolute continuity up to each branch endpoint. Here $c_D$ is a deterministic upper bound on the ODE quadratic coefficient, and $q(p)=1+\sqrt{2\log(2/p)}$. For a given pathwise radius $R>0$, define
\[
\tau_R:=\inf\{t\in[0,T]: r_t\ge R\},
\]
with the convention $\inf\emptyset=\infty$.

Let
\[
\beta_t:=\kappa_t^{Y,N}-c_D R-\Gamma_t^N.
\]
Assume the following.

\begin{enumerate}
    \item For every $\ell=0,\dots,\NofJ$ and for a.e.
    $t\in (t_\ell,t_{\ell+1})\cap[0,\tau_R)$,
    \begin{equation}
    \label{eq:forced-ode-assumption-recursive}
    \dot r_t\le -\beta_t r_t+\Gamma_t^N.
    \end{equation}
    (corresponding to the conclusion in \cref{lemma-common-empirical-moment-residual-directional-rt})

    \item For each jump time $t_\ell\le\tau_R$, including a candidate first exit at that jump, $\ell=1,\dots,\NofJ$, the previous jump lemma yields
    \begin{equation}
    \label{eq:jump-recursion-assumption}
    r_{t_\ell}
    \le
    \bigl(1+a_\ell\bigr)r_{t_\ell-}+b_\ell,
    \end{equation}
    where
    \[
    a_\ell(\supXY,p):=c_J(D)\eta_\ell q(p),
    \qquad
    b_\ell:=a_\ell\Bigl(\sqrt{\Gamma_{t_\ell-}^N}+\sqrt{\widetilde{\Gamma}_{t_\ell-}^N}\Bigr).
    \]
    (corresponding to \cref{lem:jump-bound-common-seed} on the event $E_{T,\supXY}$).
\end{enumerate}

Then, for every $t\in[0,T]\cap[0,\tau_R)$,
\begin{align}
\label{eq:piecewise-recursive-bound}
r_t
&\le
\int_0^t
\exp\Bigl(-\int_s^t \beta_u\,du\Bigr)
\prod_{\ell:\,s<t_\ell\le t}(1+a_\ell)\,
\Gamma_s^N\,ds\\
&+
\sum_{\ell:\,t_\ell\le t}
\exp\Bigl(-\int_{t_\ell}^t \beta_u\,du\Bigr)
\prod_{j:\,t_\ell<t_j\le t}(1+a_j)\,
b_\ell
\end{align}
if $r_0=0$. The same bound holds at a finite candidate exit $\tau_R$, using the pre-jump or post-jump trace as appropriate. No adaptedness of $R$ is required.
\end{lemma}

\begin{proof}
\noindent\textbf{Step 1: integrate within one ODE interval.}
For $0\le s\le t\le T$, define
\[
G(t,s):=\exp\Bigl(-\int_s^t \beta_u\,du\Bigr).
\]
By the definition of $\beta$, this is
$G(t,s)=G_\kappa^Y(t,s)\exp\{c_DR(t-s)+\int_s^t\Gamma_u^N\,du\}$,
with the curvature propagator of \cref{def:curvature-responses}.
On each continuity interval $(t_\ell,t_{\ell+1})$, the differential inequality
\eqref{eq:forced-ode-assumption-recursive} implies
\begin{equation}
\label{eq:continuous-duhamel-recursive}
r_t
\le
G(t,s)\,r_s+\int_s^t G(t,u)\Gamma_u^N\,du,
\qquad
t_\ell\le s\le t<t_{\ell+1},
\quad t<\tau_R.
\end{equation}
Indeed,
\[
\dot r_t+\beta_t r_t\le \Gamma_t^N,
\]
and multiplying by the integrating factor
$\exp(\int_s^t\beta_u\,du)$ and integrating from $s$ to $t$
gives \eqref{eq:continuous-duhamel-recursive}.

\par\noindent\textbf{Step 2: propagate the bound through each jump and interval.}
We now prove \eqref{eq:piecewise-recursive-bound} by induction over the jump intervals.

First, if $t\in[0,t_1)\cap[0,\tau_R)$, then there are no jumps before time $t$, hence
\eqref{eq:piecewise-recursive-bound} reduces exactly to
\eqref{eq:continuous-duhamel-recursive} with $s=0$.

Next, let $m\in\{1,\dots,\NofJ\}$ and assume that
\eqref{eq:piecewise-recursive-bound} has been proved for all times in $[0,t_m)$.
Take $t\in[t_m,t_{m+1})\cap[0,\tau_R)$.
Applying \eqref{eq:continuous-duhamel-recursive} on $[t_m,t]$ yields
\begin{equation}
\label{eq:after-last-jump-recursive}
r_t
\le
G(t,t_m)\,r_{t_m}
+
\int_{t_m}^t G(t,s)\Gamma_s^N\,ds.
\end{equation}
Using the jump recursion \eqref{eq:jump-recursion-assumption},
\[
r_{t_m}\le (1+a_m)r_{t_m-}+b_m,
\]
we obtain
\begin{equation}
\label{eq:after-substitute-jump-recursive}
r_t
\le
G(t,t_m)(1+a_m)r_{t_m-}
+
G(t,t_m)b_m
+
\int_{t_m}^t G(t,s)\Gamma_s^N\,ds.
\end{equation}

Now apply the induction hypothesis at time $t_m-$:
\begin{align}
r_{t_m-}\le&
G(t_m,0)\prod_{\ell=1}^{m-1}(1+a_\ell)\,r_0\quad
+
\int_0^{t_m}
G(t_m,s)\prod_{\ell:\,s<t_\ell<t_m}(1+a_\ell)\Gamma_s^N\,ds\\
&+
\sum_{\ell=1}^{m-1}
G(t_m,t_\ell)\prod_{j:\,t_\ell<t_j<t_m}(1+a_j)\,b_\ell.
\label{eq:IH-at-tmminus}
\end{align}
Substituting \eqref{eq:IH-at-tmminus} into
\eqref{eq:after-substitute-jump-recursive}, we get
\begin{align*}
r_t
&\le
G(t,t_m)(1+a_m)G(t_m,0)\prod_{\ell=1}^{m-1}(1+a_\ell)\,r_0
\\
&\quad
+
G(t,t_m)(1+a_m)
\int_0^{t_m}
G(t_m,s)\prod_{\ell:\,s<t_\ell<t_m}(1+a_\ell)\Gamma_s^N\,ds
\\
&\quad
+
G(t,t_m)(1+a_m)
\sum_{\ell=1}^{m-1}
G(t_m,t_\ell)\prod_{j:\,t_\ell<t_j<t_m}(1+a_j)\,b_\ell
\\
&\quad
+
G(t,t_m)b_m
+
\int_{t_m}^t G(t,s)\Gamma_s^N\,ds.
\end{align*}
Using the semigroup property of $G$,
\[
G(t,t_m)G(t_m,s)=G(t,s),
\qquad 0\le s\le t_m\le t,
\]
we rewrite this as
\begin{align*}
r_t
&\le
G(t,0)\prod_{\ell=1}^{m}(1+a_\ell)\,r_0
+
\int_0^{t_m}
G(t,s)\prod_{\ell:\,s<t_\ell\le t_m}(1+a_\ell)\Gamma_s^N\,ds
\\
&\quad
+
\sum_{\ell=1}^{m-1}
G(t,t_\ell)\prod_{j:\,t_\ell<t_j\le t_m}(1+a_j)\,b_\ell
+
G(t,t_m)b_m
+
\int_{t_m}^t G(t,s)\Gamma_s^N\,ds.
\end{align*}
Since there is no jump in $(t_m,t)$, for $s\in(t_m,t)$ the product
$\prod_{\ell:\,s<t_\ell\le t}(1+a_\ell)$ is empty and equals $1$.
Also, because $t\in[t_m,t_{m+1})$, the sets
\[
\{\ell:\,s<t_\ell\le t\}
\quad\text{and}\quad
\{\ell:\,s<t_\ell\le t_m\}
\]
coincide when $s<t_m$, and similarly
\[
\{j:\,t_\ell<t_j\le t\}
=
\{j:\,t_\ell<t_j\le t_m\}
\qquad (\ell\le m-1),
\]
while for $\ell=m$ the product over $j:\,t_m<t_j\le t$ is empty and equals $1$.
Therefore,
\begin{align*}
r_t&\le
G(t,0)\prod_{\ell:\,t_\ell\le t}(1+a_\ell)\,r_0
+
\int_0^t
G(t,s)\prod_{\ell:\,s<t_\ell\le t}(1+a_\ell)\Gamma_s^N\,ds\\
&+
\sum_{\ell:\,t_\ell\le t}
G(t,t_\ell)\prod_{j:\,t_\ell<t_j\le t}(1+a_j)\,b_\ell,
\end{align*}
which is exactly \eqref{eq:piecewise-recursive-bound}. This completes the induction. Taking the left limit gives the bound at a continuous candidate exit; at a candidate exit that is a jump, apply the assumed recursion at that jump itself. Empty sums and products cover the case of no jumps.
\end{proof}

The next lemma packages the local coefficients without discarding their dependence on the population moment scale or anti-spikiness.

\begin{lemma}[Explicit local comparison coefficients]
\label{lem:comparison-coefficients-v15}
Fix the deterministic population scale $K=K_{\rm pop}$ and radius $D\ge1$.
Use $\mathrm L_D,\gamma_m(D)=\mathrm L_Da_m$ from
\cref{lemma-localized-moment-smoothness-consequences} and put
\[
\Lambda(D)=\sum_{m\ge1}\gamma_m(D)L_m(1+2D^{m-1}),\qquad
c_D=\Lambda(D)+(2+A_{\rm sp})\mathrm L_D,
\]
\[
c_J(D)=1+\sqrt d\{\Lambda(D)+2\mathrm L_D\}+\sqrt{2d\mathrm L_D}.
\]
On $E_{\rm pop}\cap E_{T,D}\cap E_{T,\rm sp}(A_{\rm sp})$,
$c_D$ and $c_J(D)$ are the coefficients in
\cref{lemma-common-empirical-moment-residual-directional-rt,lem:jump-bound-common-seed}.
With $p_0=p/8$, write $\Xi=8N(e+T)(\NofJ+1)(1+B)/p$ and choose
\begin{equation}
\label{eq:def-supXY-v15}
\supXY=D=C_D(1+\log\Xi)^{1/\alpha},
\end{equation}
where the fixed $C_D$ is sufficiently large for
\eqref{eq:gamma-radius-choice} with budget $p_0$ and for the clipping-bias
estimate in Lemma~\ref{lemma-uniform-sup-gamma-bound-convenient}.
For fixed model parameters and uniformly bounded $K_0,B,A_{\rm sp}$,
for every $\varepsilon>0$ there is $C_\varepsilon$ such that
\begin{equation}
\label{eq:Ccmp-localized-v15}
1+c_D+c_J(D)+\Lambda(D)\le C_\varepsilon\Xi^\varepsilon.
\end{equation}
The constants may depend on the fixed anchor derivatives and $C_3$.
\end{lemma}
\begin{proof}
The empirical moment displacement is
$L_m(1+2D^{m-1})r_t$ by \cref{lemma-directional-measure-slot-error},
\eqref{eq:paired-empirical-moment-displacement}. The sum of the two Hessian measure errors is bounded
with weights $\gamma_m(D)$ by \cref{eq:local-two-hessians-moment-bound},
so its coefficient is $\Lambda(D)$.
The two spatial kernel slots contribute $2\mathrm L_D$ by
\cref{eq:local-kernel-spatial-bound}; the diagonal term contributes
$A_{\rm sp}\mathrm L_D$ by \cref{eq:local-jacobian-spatial-bound} and
\cref{eq-cubic-antispike-bound}. Thus
the ODE coefficient is exactly $c_D$. The jump
coefficient is exactly the one in Lemma~\ref{lem:jump-bound-common-seed}.
For the common growth bound, $D\ge1$ and
\eqref{eq:conditional-comparison-envelope} give
\[
\begin{aligned}
 \Lambda(D)
 &\le2\mathrm L_D\sum_m a_mL_m
       (2+D^m+K_{\rm pop}^m m^{m/\alpha})
 \le\mathrm L_D^2.
\end{aligned}
\]
For the last inequality, use $2u\le4(1+u)^3$ for $u\ge0$ and
$1+g_0+h_0+C_3\ge1$ in the definition of $\mathrm L_D$.
Using $\mathrm L_D\ge1$ in the displayed definitions therefore yields
\begin{equation}
\label{eq:comparison-coefficients-common-envelope}
 1+c_D+c_J(D)+\Lambda(D)
 \le(6+A_{\rm sp}+3\sqrt d+\sqrt{2d})\mathrm L_D^2.
\end{equation}
By \cref{lem:population-moment-event},
\cref{eq:population-exponential-amplitude,eq:population-exponential-scale},
the grid count is at most $C(e+T)(\NofJ+1)(1+B)$, so
$\log A_0\le C+\log\Xi$ and
$K_{\rm pop}\le C(1+\log\Xi)^{1/\alpha}$ when $K_0,B$ are bounded.
Insert this and \eqref{eq:def-supXY-v15} into
\cref{lem:conditional-local-coefficient-growth}, \eqref{eq:local-envelope-growth}:
\[
 \mathrm L_D^2\le C\exp\{C[1+\log\log(e+\Xi)]^{\kappa_{\rm dec}/(\kappa_{\rm dec}-1)}\}
 \le C_\varepsilon\Xi^\varepsilon.
\]
The last step uses $[\log\log(e+\Xi)]^{\kappa_{\rm dec}/(\kappa_{\rm dec}-1)}
=o(\log\Xi)$. This proves \eqref{eq:Ccmp-localized-v15}.
\end{proof}

\paragraph{Setup and curvature-response factors for the tracking theorem.}
Fix $N\ge2$, $p\in(0,1)$, a deterministic schedule
$0<t_1<\cdots<t_{\NofJ}\le T$ with $\eta_\ell\ge0$, and $A_{\rm sp}\ge1$.
Set $p_0=p/8$ and use the population constants of
\cref{lem:population-moment-event} with budget $p_0$.
Choose $D$ satisfying \eqref{eq:gamma-radius-choice} with budget $p_0$;
\eqref{eq:def-supXY-v15} is one such choice. Use $c_D,c_J$ from
\cref{lem:comparison-coefficients-v15}.
Define the actual deterministic residual envelopes by
\begin{equation}
\label{eq:def-delta-v15}
\delta_{N,T,p}:=\delta_\Gamma(D)
\quad\text{from \eqref{eq:gamma-explicit-envelope}, with }p_\Gamma=p_0,
\end{equation}
\begin{equation}
\label{eq:def-tdelta-v15}
\widetilde\delta_{N,T,p}:=
\widetilde\delta_{N,T,p_0}^{\rm HS}
\quad\text{from \eqref{eq:phase1-HS-tilde-delta-envelope}},
\end{equation}
For readability write $\delta=\delta_{N,T,p}$ and
$\widetilde\delta=\widetilde\delta_{N,T,p}$ in the following statement and proof.

For the sampled full Hessian curvature $\kappa_t^{Y,N}$, on any path
where $\int_0^T|\kappa_t^{Y,N}|\,dt<\infty$, recall the curvature-response factors
of \cref{def:curvature-responses}:
\begin{equation}
\label{eq:tracking-curvature-responses}
G_\kappa^Y(t,s)=\exp\!\left(-\int_s^t\kappa_u^{Y,N}\,du\right),\qquad
\mathfrak C_t^Y=\int_0^tG_\kappa^Y(t,s)\,ds,\qquad
\mathfrak J_t^Y=\sum_{\ell:t_\ell\le t}\eta_\ell G_\kappa^Y(t,t_\ell).
\end{equation}
Set
\[
q_p=1+\sqrt{2\log(2(\NofJ+1)/p_0)},\qquad
a_\ell=c_J(D)q_p\eta_\ell,
\]
\begin{equation}
\label{eq:def-jump-amplification-v15}
J_{N,T,p}:=\prod_{\ell=1}^{\NofJ}(1+a_\ell),
\end{equation}
\begin{equation}
\label{eq:def-B-v15}
\mathfrak B_{N,T,p}(t):=
\delta\mathfrak C_t^Y+
c_J(D)q_p(\sqrt\delta+\sqrt{\widetilde\delta})\mathfrak J_t^Y,
\qquad \mathfrak B_*:=\sup_{t\le T}\mathfrak B_{N,T,p}(t).
\end{equation}
Empty sums and products are zero and one, respectively. The envelope
$\mathfrak B$ is generally random because the curvature-response factors are random.
Let $R=2J_{N,T,p}\mathfrak B_*$ and define the nonlinear stability event
\begin{equation}
\label{eq:smallness-v15}
\mathcal S_{N,T,p}:=
\{T(\delta+c_D R)\le1/4\}.
\end{equation}
\paragraph{Which quantities are fixed.}
The following convention applies to the asymptotic corollary; the explicit
theorem retains the actual deterministic envelopes.
\begin{center}
\begin{tabular}{p{0.25\linewidth}p{0.66\linewidth}}
Fixed structural bounds & $d,\alpha$, feature growth/decay parameters,
$g_0,h_0,C_3$, and the initial population law.\\
Uniform regularity assumptions & $K_0,B,A_{\rm sp}$; these are assumptions
along the growing horizon, not consequences of localization.\\
Deterministic growing quantities & $D,K_{\rm pop},\delta_{N,T,p},
\widetilde\delta_{N,T,p},c_D,c_J,\Lambda,q_p,J_{N,T,p}$.
The residual envelopes are defined in
\cref{eq:def-delta-v15,eq:def-tdelta-v15}; their dependence on $N,T,p$ is retained.\\
Random reference quantities & $\kappa^{Y,N},\mathfrak C^Y,\mathfrak J^Y,
\mathfrak B_*,R$ and the nonlinear stability event. Their bounds are pathwise conditions.
\end{tabular}
\end{center}

\begin{theorem}[Fixed-schedule comparison on the nonlinear stability event]
\label{thm:fixed-schedule-high-prob-r-v15}
Assume \cref{assumption-structured-moment-smoothness,assumption-common-seed,assumption-moment-growth,assumption-uniform-concentration-piecewise-lipschitz}
and use the deterministic schedule, $N\ge2$, $p\in(0,1)$, and radius
specified above, with $r_0=0$ and integrable curvature on $[0,T]$.
Use the residual envelopes \cref{eq:def-delta-v15,eq:def-tdelta-v15},
the amplification \eqref{eq:def-jump-amplification-v15}, the forcing
\eqref{eq:def-B-v15}, and the nonlinear stability event \eqref{eq:smallness-v15}.
There is an event $\Omega_p$ with $\mathbb P(\Omega_p)\ge1-p$ such that on
$\Omega_p\cap E_{T,\rm sp}(A_{\rm sp})\cap\mathcal S_{N,T,p}$,
simultaneously at all times and left limits,
\begin{equation}
\label{eq:high-prob-r-bound-v15}
r_t\le e^{1/4}J_{N,T,p}\mathfrak B_{N,T,p}(t)
<\tfrac32J_{N,T,p}\mathfrak B_{N,T,p}(t)
\quad\text{when }\mathfrak B_{N,T,p}(t)>0,
\end{equation}
and $r_t=0$ if that envelope is zero. In particular,
\[
\mathbb P\!\left(E_{T,\rm sp}(A_{\rm sp})\cap\mathcal S_{N,T,p}
\cap\{\exists t\le T:r_t>2J_{N,T,p}\mathfrak B_{N,T,p}(t)\}\right)\le p.
\]
This is a bound on failure within the stated events, not a bound on the
conditional failure probability given $\mathcal S_{N,T,p}$.
\end{theorem}
\par\noindent\textbf{Remark (Relation to the main-text growing-horizon theorem).}
The main-text result in \Cref{theorem-main-r_t} applies the preceding theorem
on paths satisfying the controlled-curvature path condition $\mathcal R_N$ and uses the simple sufficient horizon
$T\le N^{1/4-\upsilon-\zeta}$. The integrable-curvature condition imposes no separate
model assumption there: on the common good event used in the proof,
\Cref{eq:fixed-schedule-curvature-bound} gives
$|\kappa_t^{Y,N}|\le2\mathrm L_D$, and hence
\[
 \int_0^T|\kappa_t^{Y,N}|\,\dd t
 \le \int_0^T2\mathrm L_D\,\dd t
 =2T\mathrm L_D<\infty.
\]
The explicit stability requirement remains
$T(\delta_{N,T,p}+c_DR)\le1/4$ by
\Cref{eq:smallness-v15}; it can also hold on longer stable trajectories. The
strict margin $\zeta>0$ in the main-text sufficient regime absorbs the
subpolynomial concentration and localization factors. Finally, the probability
statement is a bound on failure within the controlled-curvature path condition
and the nonlinear stability event. It neither lower-bounds
$\mathbb P(\mathcal R_N)$ nor gives a conditional failure probability given
that condition.
\begin{proof}
\noindent\textbf{Step 1: a common good event.}
Recall $p_0=p/8$ and the residual envelopes
\eqref{eq:def-delta-v15}--\eqref{eq:def-tdelta-v15}. Define
\[
 E_\Gamma:=\{\sup_{t\le T}\Gamma_t^N\le\delta\},\qquad
 E_{\widetilde\Gamma}:=
 \{\max_{\ell\le\NofJ}\widetilde\Gamma_{t_\ell-}^N\le\widetilde\delta\}.
\]
Every supremum includes the relevant left and right traces. Use the
population and radius events with budgets $p_0$ specified in the statement.
For each jump, let $\Omega_{\ell,p_0/(\NofJ+1)}^{\rm jump}$ be the Gaussian
event in \cref{lem:jump-bound-common-seed}, and set
\[
 \Omega_p:=E_{\rm pop}\cap E_{T,D}\cap E_\Gamma\cap E_{\widetilde\Gamma}
       \cap\bigcap_{\ell=1}^{\NofJ}
                   \Omega_{\ell,p_0/(\NofJ+1)}^{\rm jump}.
\]
The failure bounds and their sources are:
\begin{center}
\begin{tabular}{p{0.38\linewidth}p{0.10\linewidth}p{0.42\linewidth}}
Failure event & Budget & Source\\\hline
$E_{\rm pop}^c$ & $p_0$ & \cref{lem:population-moment-event}\\
$E_{T,D}^c$ & $p_0$ & \cref{lemma:uniform-radius-bound-XY}, \eqref{eq:gamma-radius-choice}\\
$E_{\rm pop}\cap E_{T,D}\cap E_\Gamma^c$ & $p_0$ & \cref{lemma-uniform-sup-gamma-bound-localized-conditional}, \eqref{eq:gamma-localized-probability}\\
$E_{\rm pop}\cap E_{\widetilde\Gamma}^c$ & $p_0$ & \cref{cor:tilde-gamma-jump-time-union-bound}, \eqref{eq:tilde-gamma-jump-time-union-bound}\\
$(\Omega_{\ell,p_0/(\NofJ+1)}^{\rm jump})^c$ & $\frac{p_0}{\NofJ+1}$ & \cref{lem:jump-bound-common-seed}
\end{tabular}
\end{center}
The covariance estimate conditions on $\mathcal G_{\rm pop}$, while the
Gaussian jump estimate conditions on $\mathcal H_{\ell-}$. The tower
property converts these bounds to unconditional ones:
\[
\begin{aligned}
 \mathbb P(E_{\rm pop}\cap E_{\widetilde\Gamma}^c)
 &=\mathbb E[\mathbf1_{E_{\rm pop}}
      \mathbb P(E_{\widetilde\Gamma}^c\mid\mathcal G_{\rm pop})]\le p_0,\\
 \mathbb P((\Omega_{\ell,p_0/(\NofJ+1)}^{\rm jump})^c)
 &=\mathbb E[\mathbb P((\Omega_{\ell,p_0/(\NofJ+1)}^{\rm jump})^c
      \mid\mathcal H_{\ell-})]\le p_0/(\NofJ+1).
\end{aligned}
\]
Consequently,
\[
\begin{aligned}
 \mathbb P(\Omega_p^c)
 &\le\mathbb P(E_{\rm pop}^c)+\mathbb P(E_{T,D}^c)
      +\mathbb P(E_{\rm pop}\cap E_{T,D}\cap E_\Gamma^c)\\
 &\quad+\mathbb P(E_{\rm pop}\cap E_{\widetilde\Gamma}^c)
      +\sum_{\ell=1}^{\NofJ}
              \mathbb P((\Omega_{\ell,p_0/(\NofJ+1)}^{\rm jump})^c)\\
 &\le4p_0+\frac{\NofJ}{\NofJ+1}p_0<5p_0<p.
\end{aligned}
\]
This union bound does not require independence. The sample event $E_{T,D}$
is intersected after concentration; it is not part of the conditioning used
to establish the reference product law.

\par\noindent\textbf{Step 2: deterministic comparison on the good event.}
Fix a realization in $\Omega_p\cap E_{T,\rm sp}(A_{\rm sp})$.
The two deterministic inputs can now be tied to explicit parts of this event.
On
$E_{\rm pop}\cap E_{T,D}\cap E_{T,\rm sp}(A_{\rm sp})\cap E_\Gamma$,
\cref{lemma-common-empirical-moment-residual-directional-rt} gives
\[
 \dot r_t\le-\kappa_t^{Y,N}r_t+c_Dr_t^2+\delta r_t+\delta.
\]
On the other hand, on
$E_{\rm pop}\cap E_{T,D}\cap E_\Gamma\cap E_{\widetilde\Gamma}
\cap\Omega_{\ell,p_0/(\NofJ+1)}^{\rm jump}$,
\cref{lem:jump-bound-common-seed} gives
\[
 r_{t_\ell}\le(1+a_\ell)r_{t_\ell-}
                 +a_\ell(\sqrt\delta+\sqrt{\widetilde\delta}).
\]
Before applying these two inputs,
\cref{lemma-localized-moment-smoothness-consequences},
\eqref{eq:local-field-magnitude-bound}, and the block estimate in
\cref{lemma-operator-norm-block-matrix} give
\begin{equation}
\label{eq:fixed-schedule-curvature-bound}
 |\kappa_t^{Y,N}|\le\|\HessianMatrices_t^Y\|_{\mathbb R^{dN}\to\mathbb R^{dN}}
 \le\left\|(\mathrm L_D/N)\mathbf1\mathbf1^\top+\mathrm L_DI_N
        \right\|_{\mathbb R^N\to\mathbb R^N}=2\mathrm L_D.
\end{equation}
Here $\mathbf1=(1,\ldots,1)^\top\in\mathbb R^N$ and
$P=N^{-1}\mathbf1\mathbf1^\top$ satisfies $P^\top=P$, $P^2=P$,
and $P\mathbf1=\mathbf1$. Thus $\|P+I_N\|_{\mathbb R^N\to\mathbb R^N}=2$.
In particular $\int_0^T|\kappa_t^{Y,N}|\,dt\le2T\mathrm L_D<\infty$
on this good event, so the curvature-response factors in
\eqref{eq:tracking-curvature-responses} are well-defined.
The absolute-continuity premise of
\cref{lem:piecewise-gronwall-forced-jumps} is also automatic on each ODE
branch. Indeed, the branchwise $B$-Lipschitz bounds in
\cref{assumption-uniform-concentration-piecewise-lipschitz} give
\[
\begin{aligned}
 |r_t-r_s|
 &\le\frac1{\sqrt N}\|\Delta_t-\Delta_s\|_{\mathbb R^{dN}}\\
 &\le\frac1{\sqrt N}
       \left[\sum_{i=1}^N
       \bigl(\|X_t^i-X_s^i\|_{\mathbb R^d}
             +\|Y_t^i-Y_s^i\|_{\mathbb R^d}\bigr)^2\right]^{1/2}
 \le2B|t-s|.
\end{aligned}
\]
Suppose first $\mathfrak B_*>0$, and recall
$R=2J_{N,T,p}\mathfrak B_*$ from the statement. Before a first exit from
$r<R$, replace $c_Dr_t^2$ by $c_DRr_t$.
The exponent in \cref{lem:piecewise-gronwall-forced-jumps} satisfies
\[
\begin{aligned}
 \exp\left(-\int_s^t\beta_u\,du\right)
 &=G_\kappa^Y(t,s)
       \exp\left((t-s)c_DR+\int_s^t\Gamma_u^N\,du\right)\\
 &\le G_\kappa^Y(t,s)e^{T(c_DR+\delta)}.
\end{aligned}
\]
Every indicated subproduct of $(1+a_\ell)$ is at most
$J_{N,T,p}=\prod_{\ell=1}^{\NofJ}(1+a_\ell)$, since $a_\ell\ge0$.
Apply \cref{lem:piecewise-gronwall-forced-jumps},
\eqref{eq:piecewise-recursive-bound}, retaining the two forcing terms:
\begin{equation}
\label{eq:tracking-stopped-propagator-bound}
 r_t\le J_{N,T,p}e^{T(c_DR+\delta)}
 \left[\int_0^tG_\kappa^Y(t,s)\Gamma_s^N\,ds
       +\sum_{\ell:t_\ell\le t}G_\kappa^Y(t,t_\ell)b_\ell\right].
\end{equation}
The continuous and jump forcing terms satisfy
\[
\begin{aligned}
 \int_0^tG_\kappa^Y(t,s)\Gamma_s^N\,ds
 &\le\delta\mathfrak C_t^Y,\\
 \sum_{\ell:t_\ell\le t}G_\kappa^Y(t,t_\ell)b_\ell
 &\le c_J(D)q_p(\sqrt\delta+\sqrt{\widetilde\delta})
                \sum_{\ell:t_\ell\le t}\eta_\ell G_\kappa^Y(t,t_\ell)\\
 &=c_J(D)q_p(\sqrt\delta+\sqrt{\widetilde\delta})\mathfrak J_t^Y.
\end{aligned}
\]
Here $b_\ell$ is the additive forcing in
\eqref{eq:jump-recursion-assumption}. The sum of the two right-hand sides is
$\mathfrak B_{N,T,p}(t)$ from \eqref{eq:def-B-v15}. Substitution into
\eqref{eq:tracking-stopped-propagator-bound} gives
\begin{equation}
\label{eq:tracking-forcing-envelope-bound}
 r_t\le J_{N,T,p}e^{T(c_DR+\delta)}\mathfrak B_{N,T,p}(t).
\end{equation}

\par\noindent\textbf{Step 3: exclude all first exits.}
The candidate-exit conclusion of
\cref{lem:piecewise-gronwall-forced-jumps},
\eqref{eq:piecewise-recursive-bound}, applies at a continuous exit,
a pre-jump trace, or a post-jump exit, including a jump at $T$.
At a post-jump exit its partial products already include that jump.
Thus the same bounds on the propagator, products, and forcing give
\cref{eq:tracking-stopped-propagator-bound,eq:tracking-forcing-envelope-bound}
at the candidate exit itself; no extra factor $1+a_\ell$ is introduced.
On the nonlinear stability event \eqref{eq:smallness-v15}, the candidate value is
therefore at most
\[
 e^{1/4}J_{N,T,p}\mathfrak B_*
 =\frac{e^{1/4}}2 R<R.
\]
There are finitely many branches and jumps, so this strict margin excludes
an exit on $[0,T]$. The radius may depend on future reference curvature-response factors:
the argument is pathwise and uses no stopping-time theorem.

If $\mathfrak B_*=0$ and $T>0$, positivity of $G_\kappa^Y(t,s)$ implies
$\delta=0$, and every nonnegative jump forcing
$a_\ell(\sqrt\delta+\sqrt{\widetilde\delta})$ is zero. On a branch and
below any fixed positive radius $R'$, the ODE inequality becomes
$\dot r_t\le(|\kappa_t^{Y,N}|+c_DR')r_t$.
Multiplication by the integrating factor, starting from $r=0$, gives
$r_t\le0$ on that branch. The jump recursion sends zero to zero.
Induction proves $r\equiv0$; no zero radius is used in the first-exit lemma.
For $T=0$, the conclusion is just $r_0=0$.

\par\noindent\textbf{Step 4: the failure event.}
The preceding argument proves the pathwise bound on
$\Omega_p\cap E_{T,\rm sp}(A_{\rm sp})\cap\mathcal S_{N,T,p}$.
Consequently any violation within the regularity event and the nonlinear stability event
belongs to $\Omega_p^c$, whose probability is at most $p$. No conditional
failure probability given the nonlinear stability event is asserted.
\end{proof}

\begin{corollary}[A simple sufficient horizon and the curvature-dependent rate]
\label{cor:fixed-schedule-high-prob-r-rate-v15}
Use the same model as $N$ varies, with fixed structural and anchor bounds and
uniform $K_0,B,A_{\rm sp}$. Fix $\upsilon\ge0$, $\zeta>0$ with $\upsilon+\zeta<1/4$, and $C_p,C_0,L_0>0$.
Suppose
\[
T\le N^{1/4-\upsilon-\zeta},\qquad p\ge N^{-C_p},\qquad
\NofJ\le L_0,\qquad 0\le\eta_\ell\le C_0,
\]
and choose $D$ as in \eqref{eq:def-supXY-v15}.
Define the controlled-curvature path condition $\mathcal R_N$ by
\begin{equation}
\label{eq:reference-response-event}
\mathcal R_N:=\left\{\int_0^T|\kappa_t^{Y,N}|\,dt<\infty,\quad
\mathfrak C_*^Y:=\sup_{t\le T}\mathfrak C_t^Y
\le C_0N^\upsilon(1+T),\quad
\mathfrak J_*^Y:=\sup_{t\le T}\mathfrak J_t^Y\le C_0N^\upsilon\right\}.
\end{equation}
For all sufficiently large $N$, the condition $\mathcal R_N$ implies the
nonlinear stability event $\mathcal S_{N,T,p}$ of \eqref{eq:smallness-v15}.
For every fixed $\varepsilon>0$, on
$\Omega_p\cap E_{T,\rm sp}(A_{\rm sp})\cap\mathcal R_N$,
\begin{equation}
\label{eq:r-rate-simplified-v15}
r_t\le C_\varepsilon\Xi^\varepsilon
\left[(N^{-1/2}+T/N)\mathfrak C_t^Y
+(N^{-1/4}+\sqrt{T/N})\mathfrak J_t^Y\right],\qquad t\le T.
\end{equation}
Thus the same bound holds with failure probability at most $p$ within
$E_{T,\rm sp}(A_{\rm sp})\cap\mathcal R_N$.
\end{corollary}
\begin{proof}
\noindent\textbf{Step 1: recall the deterministic envelopes.}
In this proof, write $\delta=\delta_{N,T,p}$ and
$\widetilde\delta=\widetilde\delta_{N,T,p}$ as defined in
\cref{eq:def-delta-v15,eq:def-tdelta-v15}.
Use \cref{lemma-uniform-sup-gamma-bound-convenient},
\eqref{eq:tilde-gamma-jump-time-final-rate}, and
\eqref{eq:Ccmp-localized-v15}. Uniform $K_0,B,A_{\rm sp}$ imply that
$K_{\rm pop},D$ are bounded by fixed powers of $\log\Xi$.
For \cref{lemma-uniform-sup-gamma-bound-convenient}, use total budget
$3p_0=3p/8<1$. Its population, radius, and concentration budgets are
then each $p_0$, and its scale agrees exactly with the present one:
\[
 \frac{3N(e+T)(\NofJ+1)(1+B)}{3p_0}
 =\frac{N(e+T)(\NofJ+1)(1+B)}{p_0}=\Xi.
\]
The polynomial-amplitude premise of that lemma holds with $c_M=1$.
Indeed, \cref{eq:population-exponential-amplitude} and the grid bound in
\cref{lem:population-moment-event} give
\[
 M_{\rm pop}\le\frac{4eQ}{p_0}
 \le\frac{C(e+T)(\NofJ+1)(1+B)}{p_0}
 =\frac{C\Xi}{N}\le C\Xi.
\]
For \eqref{eq:tilde-gamma-jump-time-final-rate} with budget $p_0=p/8$,
the present $\Xi\ge N(\NofJ+1)/p_0$ satisfies its scale condition
directly. For every $\theta>0$, the cited bounds give
\begin{equation}
\label{eq:main-rate-internal-envelopes}
\begin{aligned}
 \delta&\le C_\theta\Xi^\theta(N^{-1/2}+T/N),&
 \widetilde\delta&\le C_\theta\Xi^\theta N^{-1/2},\\
 c_D+c_J(D)q_p+\Lambda(D)&\le C_\theta\Xi^\theta.
\end{aligned}
\end{equation}
For $c_Jq_p$, apply the individual bounds with exponent $\theta/2$ to
each factor: $q_p=1+\sqrt{2\log(2(\NofJ+1)/p_0)}$ also satisfies
$q_p\le C_\theta\Xi^{\theta/2}$.

\par\noindent\textbf{Step 2: count the powers in the jump and curvature-response bounds.}
The amplitude bound $0\le\eta_\ell\le C_0$,
\eqref{eq:def-jump-amplification-v15}, and $\NofJ\le L_0$ imply
\[
 J_{N,T,p}\le(1+C_0c_J(D)q_p)^{L_0}
             \le C_\theta\Xi^{L_0\theta}.
\]
Using $\sqrt{a+b}\le\sqrt a+\sqrt b$ in
\eqref{eq:main-rate-internal-envelopes},
\[
 \sqrt\delta+\sqrt{\widetilde\delta}
 \le C_\theta\Xi^{\theta/2}(N^{-1/4}+\sqrt{T/N}).
\]
By \eqref{eq:def-B-v15} and the controlled-curvature path condition
\eqref{eq:reference-response-event},
\[
 \mathfrak B_*\le\delta\mathfrak C_*^Y
       +c_J(D)q_p(\sqrt\delta+\sqrt{\widetilde\delta})\mathfrak J_*^Y.
\]
Substitute the preceding estimates and $T/N\le N^{-1/2}$ to get
\[
 \mathfrak B_*
 \le C_\theta\Xi^\theta N^\upsilon(1+T)N^{-1/2}
       +C_\theta\Xi^{3\theta/2}N^\upsilon N^{-1/4}.
\]
Multiplying by $2J_{N,T,p}$ and using $\Xi\ge1$ yields
\[
 R\le C_\theta\Xi^{(L_0+3/2)\theta}N^\upsilon
         \{(1+T)N^{-1/2}+N^{-1/4}\}.
\]

\par\noindent\textbf{Step 3: verify the nonlinear stability event.}
Recall $\Xi=8N(e+T)(\NofJ+1)(1+B)/p$. The horizon and probability
assumptions imply $\Xi\le C N^{5/4+C_p}$. For the desired
$\varepsilon>0$, choose a fixed $\theta>0$ sufficiently small that
\begin{equation}
\label{eq:main-rate-internal-exponent}
 (5/4+C_p)(L_0+5/2)\theta\le\zeta/2,
 \qquad (L_0+3/2)\theta<\varepsilon.
\end{equation}
This choice depends only on the fixed parameters and $\varepsilon,\zeta$,
not on $N,T,p$. The preceding bounds and $N^\upsilon\ge1$ give
\[
\begin{aligned}
 T(\delta+c_DR)
 &\le C_\theta\Xi^{(L_0+5/2)\theta}N^\upsilon
       \left\{\frac{T(1+T)}{\sqrt N}+\frac{T}{N^{1/4}}\right\}\\
 &\le C_\theta N^{\zeta/2}
       (N^{-1/4-\zeta}+N^{-\upsilon-2\zeta}+N^{-\zeta})
 \le3C_\theta N^{-\zeta/2}\longrightarrow0.
\end{aligned}
\]
The second line substitutes $T\le N^{1/4-\upsilon-\zeta}$ into each term:
\[
 N^\upsilon T/\sqrt N\le N^{-1/4-\zeta},\qquad
 N^\upsilon T^2/\sqrt N\le N^{-\upsilon-2\zeta},\qquad
 N^\upsilon T/N^{1/4}\le N^{-\zeta}.
\]
The choice \eqref{eq:main-rate-internal-exponent} ensures
$(5/4+C_p)(L_0+5/2)\theta\le\zeta/2$.
For all sufficiently large $N$, the last bound is at most $1/4$, proving
$\mathcal R_N\subset\mathcal S_{N,T,p}$.

\par\noindent\textbf{Step 4: substitute into the tracking bound.}
The exact theorem and Steps 1--2 give
\[
 r_t\le C_\theta\Xi^{(L_0+3/2)\theta}
 \left[(N^{-1/2}+T/N)\mathfrak C_t^Y
 +(N^{-1/4}+\sqrt{T/N})\mathfrak J_t^Y\right].
\]
Since $(L_0+3/2)\theta<\varepsilon$, this proves
\eqref{eq:r-rate-simplified-v15}. The exact theorem continues to apply
at longer horizons on its nonlinear stability event \eqref{eq:smallness-v15}.
\end{proof}

%% file: appendix_final_general/04_functional_gap.tex
\subsection{Functional-value comparison}
\label{subsec:functional-gap}

This subsection converts the particle tracking estimate into a functional-value
estimate.  The key point is that no new stochastic residual is needed.

We now turn the $r_t$-comparison into the final estimate for
\[
\bigl|F(\rho_{\bs X_t})-F(\mu_t)\bigr|.
\]

The comparison uses $\Gamma_t^N$ for the empirical moment fluctuation of $\rho_{\mathbf Y_t}$ and $r_t$ for the discrepancy between the two particle systems.

No additional error term is introduced.

The next lemma is the deterministic conversion step: the functional gap is
controlled by the pathwise particle discrepancy and the same moment residual
\(\Gamma_t^N\).
\begin{lemma}[Fixed-time functional discrepancy]
\label{lem:functional-gap-fixed-time-v15}
In this subsection we use the convention
\[
\gamma_m(D)=\mathrm L_Da_m\quad\text{from \cref{eq:common-local-moment-weights}}.
\]
Fix the population scale and work on $E_{\rm pop}$ as in the tracking theorem. Then, on the localization event $E_{T,\supXY}$, for every $t\in[0,T]$,
\begin{equation}
\label{eq:functional-gap-fixed-time-v15}
\bigl|F(\rho_{\bs X_t})-F(\mu_t)\bigr|
\le
\Lambda(\supXY)\,r_t+\Gamma_t^N,
\end{equation}
where
\[
\Lambda(\supXY)
\coloneq
\sum_{m\ge1}\gamma_m(\supXY)L_m(1+2\supXY^{m-1}).
\]
\end{lemma}

\begin{proof}
\noindent\textbf{Step 1: split the functional error through the reference sample.}
By the value bound in
\cref{lemma-localized-moment-smoothness-consequences,eq:local-value-gradient-moment-bound},
\[
\bigl|F(\rho_{\bs X_t})-F(\mu_t)\bigr|
\le
\sum_{m\ge1}\gamma_m(\supXY)\,
\|M_m(\rho_{\bs X_t})-M_m(\mu_t)\|_{\mathsf H_m}.
\]
Split the difference as
\[
M_m(\rho_{\bs X_t})-M_m(\mu_t)
=
\bigl(M_m(\rho_{\bs X_t})-M_m(\rho_{\bs Y_t})\bigr)
+
\bigl(M_m(\rho_{\bs Y_t})-M_m(\mu_t)\bigr).
\]

\par\noindent\textbf{Step 2: bound the paired empirical moments and recall the residual.}
On $E_{T,\supXY}$, both configurations lie in the ball of radius
$D=\supXY$. Set $s=1$ in the empirical-moment calculation
\cref{eq:paired-empirical-moment-displacement} in the proof of
\cref{lemma-directional-measure-slot-error}. Since
$\mathbf Z_t^{(1)}=\mathbf X_t$, it gives
\[
 \|M_m(\rho_{\bs X_t})-M_m(\rho_{\bs Y_t})\|_{\mathsf H_m}
 \le L_m(1+2D^{m-1})r_t.
\]
For the population-to-reference difference, recall
\cref{lemma-directional-measure-slot-error,eq:common-moment-residual-and-lipschitz-sum}:
\[
 \sum_m\gamma_m(D)\|M_m(\rho_{\mathbf Y_t})-M_m(\mu_t)\|_{\mathsf H_m}
 =\Gamma_t^N.
\]
Summing the two comparisons with weights $\gamma_m(D)$ yields
\[
\bigl|F(\rho_{\bs X_t})-F(\mu_t)\bigr|
\le
\Lambda(\supXY)r_t+\Gamma_t^N.
\]
This is \eqref{eq:functional-gap-fixed-time-v15}.
\end{proof}

Combining this deterministic conversion with the fixed-schedule tracking theorem
gives the final uniform functional-value comparison.

\begin{theorem}[Uniform functional comparison on the tracking event]
\label{thm:functional-gap-fixed-schedule-v15}
Under Theorem~\ref{thm:fixed-schedule-high-prob-r-v15}, on the same event
$\Omega_p\cap E_{T,\rm sp}(A_{\rm sp})\cap\mathcal S_{N,T,p}$,
\[
|F(\rho_{\mathbf X_t})-F(\mu_t)|
\le2\Lambda(D)J_{N,T,p}\mathfrak B_{N,T,p}(t)+\delta_{N,T,p},\qquad t\le T.
\]
Under all the assumptions of
\cref{cor:fixed-schedule-high-prob-r-rate-v15}, for sufficiently large $N$
the following rate holds on
$\Omega_p\cap E_{T,\rm sp}(A_{\rm sp})\cap\mathcal R_N$, where
\(\mathcal R_N\) is the controlled-curvature path condition in
\cref{eq:reference-response-event}. The curvature-response factors are those of
\cref{thm:fixed-schedule-high-prob-r-v15}. On this event,
\begin{equation}
\label{eq:functional-rate-simplified}
|F(\rho_{\mathbf X_t})-F(\mu_t)|
\le C_\varepsilon\Xi^\varepsilon
\left[(N^{-1/2}+T/N)(1+\mathfrak C_t^Y)
+(N^{-1/4}+\sqrt{T/N})\mathfrak J_t^Y\right].
\end{equation}
More precisely, with the same $\Omega_p$ and no additional event,
\[
 \mathbb P\left(E_{T,\rm sp}(A_{\rm sp})\cap\mathcal R_N
       \cap\{\eqref{eq:functional-rate-simplified}
                         \text{ fails for some }t\le T\}\right)\le p.
\]
This is a joint failure bound within the indicated sets, not a
conditional probability bound given those sets.
\end{theorem}
\begin{proof}
\noindent\textbf{Step 1: apply the deterministic value comparison.}
On the event in the first assertion,
\cref{thm:fixed-schedule-high-prob-r-v15,eq:high-prob-r-bound-v15}
gives $r_t\le e^{1/4}J_{N,T,p}\mathfrak B_{N,T,p}(t)
\le2J_{N,T,p}\mathfrak B_{N,T,p}(t)$.
Also $\Gamma_t^N\le\delta_{N,T,p}$ by the construction of $\Omega_p$
in that theorem. Substituting these two bounds into
\cref{lem:functional-gap-fixed-time-v15,eq:functional-gap-fixed-time-v15} gives
\[
 |F(\rho_{\mathbf X_t})-F(\mu_t)|
 \le\Lambda(D)r_t+\Gamma_t^N
 \le2\Lambda(D)J_{N,T,p}\mathfrak B_{N,T,p}(t)+\delta_{N,T,p}.
\]
\par\noindent\textbf{Step 2: substitute the envelopes for the sufficient-horizon rate.}
Fix $\varepsilon>0$. Use exponent $\varepsilon/2$ in the coefficient
bound \cref{lem:comparison-coefficients-v15,eq:Ccmp-localized-v15} and the
envelope bound \cref{cor:fixed-schedule-high-prob-r-rate-v15,eq:main-rate-internal-envelopes}:
\[
 \Lambda(D)\le C_\varepsilon\Xi^{\varepsilon/2},\qquad
 \delta_{N,T,p}\le C_\varepsilon\Xi^{\varepsilon/2}(N^{-1/2}+T/N).
\]
Apply the tracking rate in
\cref{cor:fixed-schedule-high-prob-r-rate-v15,eq:r-rate-simplified-v15}
with exponent $\varepsilon/2$ as well. Multiplying that rate by
$\Lambda(D)$ and adding $\delta_{N,T,p}$ yields
\[
\begin{aligned}
 \Lambda(D)r_t+\delta_{N,T,p}
 &\le C_\varepsilon\Xi^{\varepsilon/2}\Xi^{\varepsilon/2}
 \left[(N^{-1/2}+T/N)\mathfrak C_t^Y
 +(N^{-1/4}+\sqrt{T/N})\mathfrak J_t^Y\right]\\
 &\quad+C_\varepsilon\Xi^{\varepsilon/2}(N^{-1/2}+T/N)\\
 &\le C_\varepsilon\Xi^\varepsilon
 \left[(N^{-1/2}+T/N)(1+\mathfrak C_t^Y)
 +(N^{-1/4}+\sqrt{T/N})\mathfrak J_t^Y\right],
\end{aligned}
\]
where $C_\varepsilon$ is enlarged as needed. The last step uses
$\Xi^{\varepsilon/2}\Xi^{\varepsilon/2}=\Xi^\varepsilon$ and $\Xi\ge1$.
Together with \cref{lem:functional-gap-fixed-time-v15,eq:functional-gap-fixed-time-v15}
and $\Gamma_t^N\le\delta_{N,T,p}$ on the same $\Omega_p$, this proves
\eqref{eq:functional-rate-simplified}.
\end{proof}

%% file: appendix_final_general/canonical_variance_cosine.tex
\subsection{A growing-horizon example: escape and curvature recovery}
\label{subsec:canonical-variance-cosine}

This example follows a single Gaussian jump from a negative-curvature
mean to a positive-curvature region. Two lemmas verify the comparison
assumptions and bound the curvature-response factors; the proposition then
applies the general comparison theorem on a growing horizon.

\paragraph{Model and fixed parameters.}
In dimension one, let
\begin{equation}
 F(\mu)=\frac a2\operatorname{Var}(\mu)+\beta\cos(\pi m_\mu),
 \qquad m_\mu:=\int x\,\mu(dx),\qquad a>0,\quad\beta>0.
 \label{eq:canonical-objective}
\end{equation}
Fix $R_0\ge0$, a jump time $\tau>0$, and an amplitude $\eta>0$,
independently of $N$ and $T$. Let the deterministic initial law $\mu_0$
be supported in $[-R_0,R_0]$ with $m_{\mu_0}=0$, and set
$X_0^i=Y_0^i=U_i$ for an i.i.d. sample $(U_i)_{i=1}^N$ from $\mu_0$.
Use the population-first coupling of
\cref{lem:population-first-finite-coupling}, with exactly one prescribed
jump at $\tau$ of amplitude $\eta$, and no later jumps.
Write $m_t=m_{\mu_t}$ and $b:=\beta\pi^2$.
Below, $C$ depends only on $a,\beta,R_0,\tau,\eta$ and may increase
between inequalities; $C_\varepsilon$ may also depend on $\varepsilon$.
The minimum of $F$ over $\mathcal P_2(\mathbb R)$ is $-\beta$:
variance is nonnegative, $\cos\ge-1$, and a Dirac mass at an odd integer
attains equality.

\begin{proposition}[Tracking after escape on a growing horizon]
\label{prop:canonical-model-assumptions}
For every fixed $\varepsilon>0$, there are $C_\varepsilon<\infty$ and
$N_0$, depending only on the fixed model parameters and $\varepsilon$,
such that for $N\ge N_0$ and $T=N^{1/8}$, the following hold on the
same event of probability at least $1-N^{-1/16}$.
Here $r_t$ is the discrepancy in \eqref{eq:tracking-metric}, and
$\rho_{\mathbf X_t}=N^{-1}\sum_i\delta_{X_t^i}$.
\begin{enumerate}[label=(\roman*),leftmargin=*]
 \item Uniformly over the growing horizon,
 \begin{equation}
 \sup_{0\le t\le T}
 \left[r_t+|F(\rho_{\mathbf X_t})-F(\mu_t)|\right]
 \le C_\varepsilon N^{-3/16+\varepsilon}.
 \label{eq:canonical-growing-horizon-conclusion}
 \end{equation}
 \item After a time $C\log N<T$,
 \begin{equation}
 \sup_{C\log N\le t\le T}
 \left[r_t+|F(\rho_{\mathbf X_t})-F(\mu_t)|
             +F(\rho_{\mathbf X_t})+\beta\right]
 \le C_\varepsilon N^{-1/2+\varepsilon}.
 \label{eq:canonical-after-recovery-conclusion}
 \end{equation}
\end{enumerate}
\end{proposition}

\begin{lemma}[Verification of the comparison assumptions]
\label{lem:canonical-comparison-assumptions}
The model \eqref{eq:canonical-objective} with the initial law and
single jump specified above satisfies:
\begin{enumerate}[label=(\roman*),leftmargin=*]
 \item \cref{assumption-structured-moment-smoothness}, with structural
 parameters independent of $N,T$;
 \item \cref{assumption-common-seed}, with a spatially constant Gaussian field;
 \item \cref{assumption-moment-growth}, with $\alpha=2$ and $K_0\le C$;
 \item \cref{assumption-uniform-concentration-piecewise-lipschitz}, with $B\le C$;
 \item \cref{assumption-anti-spikiness-event}, with $A_{\rm sp}=1$ and
 $E_{T,\rm sp}(1)$ holding almost surely for every $T$.
\end{enumerate}
\end{lemma}
\begin{proof}
\noindent\textbf{Step 1: moment structure and weighted smoothness, for (i).}
Use scalar feature spaces with $\phi_1(x)=x$, $\phi_2(x)=x^2$, and
$\phi_j=0$ for $j\ge3$, and set
\[
 G(\theta)=\frac a2(\theta_2-\theta_1^2)+\beta\cos(\pi\theta_1).
\]
The nonzero features have derivative triples
$(\phi_1',\phi_1'',\phi_1''')=(1,0,0)$ and
$(\phi_2',\phi_2'',\phi_2''')=(2x,2,0)$.
They satisfy \eqref{eq:coordinate-feature-growth} with
$C_\phi=c_{\rm gr}=1$, $\kappa_{\rm gr}=0$, and $L_j=ej$.
Choose $a_j=e^{-j^2}$, so $c_a=1$ and $\kappa_{\rm dec}=2$.
The weighted norm in \eqref{eq:weighted-moment-space} gives
$|h_j|\le a_j^{-1}\|h\|_{\mathsf H_a}$; hence each coordinate projection
is continuous and the finite-coordinate composition $G$ is $C^3$ on
$\mathcal U=\mathsf H_a$. Its derivatives are
\begin{equation}
 \begin{aligned}
 \fD G(\theta)[h]
 &=(-a\theta_1-\beta\pi\sin(\pi\theta_1))h_1+\tfrac a2h_2,\\
 \fD^2G(\theta)[h,k]&=(-a-b\cos(\pi\theta_1))h_1k_1,\\
 \fD^3G(\theta)[h,k,r]&=\beta\pi^3\sin(\pi\theta_1)h_1k_1r_1.
 \end{aligned}
 \label{eq:canonical-moment-derivatives}
\end{equation}
In particular,
\[
 |\fD^3G(\theta)[h,k,r]|
 \le\beta\pi^3|h_1k_1r_1|
 \le\beta\pi^3a_1^{-3}
       \|h\|_{\mathsf H_a}\|k\|_{\mathsf H_a}\|r\|_{\mathsf H_a}.
\]
Thus \eqref{eq:structured-moment-open-neighbourhood} holds with
$C_3=\beta\pi^3a_1^{-3}$; the anchor $\theta_0=M(\delta_0)=0$ in
\eqref{eq:moment-anchor} gives $g_0\le a/(2a_2)$ and
$h_0\le(a+b)/a_1^2$.
For $\nu\in\mathcal P_2(\mathbb R)$, the moment sequence has at most two
nonzero coordinates and
\[
 \begin{aligned}
 \|M(\nu)\|_{\mathsf H_a}^2
 &=a_1^2\left|\int x\,\nu(dx)\right|^2
       +a_2^2\left|\int x^2\,\nu(dx)\right|^2<\infty,\\
 G(M(\nu))&=\tfrac a2\operatorname{Var}(\nu)+\beta\cos(\pi m_\nu).
 \end{aligned}
\]
This verifies the objective identity of \cref{def:moment-objective}.

\par\noindent\textbf{Step 2: Gaussian perturbations, for (ii).}
In the chain-rule formulas of
\cref{lemma-localized-moment-smoothness-consequences,eq:weighted-gradient-representation,eq:weighted-spatial-hessian-representation,eq:weighted-integral-hessian-representation},
substitute \eqref{eq:canonical-moment-derivatives} and the feature derivatives:
\[
 \begin{aligned}
 \nabla_WF(\mu,x)&=(-am_\mu-\beta\pi\sin(\pi m_\mu))\cdot1+\tfrac a2\cdot2x,\\
 \nabla_x\nabla_WF(\mu,x)&=\tfrac a2\cdot2,\qquad
 \nabla_W^2F(\mu;x,y)=(-a-b\cos(\pi m_\mu))\cdot1\cdot1.
 \end{aligned}
\]
Consequently,
\begin{equation}
 \begin{split}
 \nabla_WF(\mu,x)&=a(x-m_\mu)-\beta\pi\sin(\pi m_\mu),\\
 \nabla_x\nabla_WF(\mu,x)&=a,\qquad
 \nabla_W^2F(\mu;x,y)=c(m_\mu),\qquad c(u):=-a-b\cos(\pi u).
 \end{split}
 \label{eq:canonical-wasserstein-derivatives}
\end{equation}
The Hessian kernel is symmetric and $|c(u)|\le a+b$.
Its covariance kernel is $K_\nu(x,y)=\int c(m_\nu)^2\,\nu(dz)=c(m_\nu)^2$.
The corresponding operator sends $f$ to $c(m_\nu)^2\int f\,d\nu$;
it has eigenvalue $c(m_\nu)^2$ on the constant unit function and zero
on its orthogonal complement, so is positive and trace class.
Take a fresh $Z\sim N(0,1)$ independent of the full pre-jump history
$\mathcal H_{1-}$ of \cref{lem:population-first-finite-coupling}.
The spatially constant field $\xi_{1,\mu_{\tau-}}(x)=c(m_{\tau-})Z$
is jointly measurable and has the required conditional covariance.

For $e_N=N^{-1/2}\mathbf1\in\mathbb R^N$ and
$\overline x=N^{-1}\sum_i x_i$, the finite factor is
\begin{equation}
 \mathbf M^N(\rho_{\mathbf x},\mathbf x)=\sqrt N c(\overline x)e_Ne_N^\top,
 \qquad \mathbf M^N(\rho_{\mathbf x},\mathbf x)g
             =c(\overline x)\mathbf1(e_N^\top g).
 \label{eq:canonical-finite-factor}
\end{equation}
Here $\mathbf1$ has all entries equal to one.
Use the coordinate $\mathbf g_1^N$ constructed in
\cref{lem:population-first-finite-coupling,eq:population-whitening-covariances}:
conditional on $\mathcal H_{1-}$ it has law $N(0,I_N)$.
Thus $\zeta:=e_N^\top\mathbf g_1^N$ has conditional law $N(0,1)$,
since $e_N^\top e_N=1$.
Writing $\overline X_t=N^{-1}\sum_iX_t^i$, the jumps are
\begin{equation}
 Y_\tau^i=Y_{\tau-}^i+\eta c(m_{\tau-})Z,\qquad
 X_\tau^i=X_{\tau-}^i+\eta c(\overline X_{\tau-})\zeta.
 \label{eq:canonical-jumps}
\end{equation}
The population pushforward and conditional-i.i.d. property follow from
\cref{lemma-gp-jump-implies-conditional-iid} and
\cref{lem:population-first-finite-coupling,eq:population-pushforward-product-law}.

\par\noindent\textbf{Step 3: trajectories and unconditional moments, for (iii).}
Averaging the negative gradient in \eqref{eq:canonical-wasserstein-derivatives}
gives, on each ODE interval,
\begin{equation}
 \begin{aligned}
 \dot m_t&=-a\int(x-m_t)\,\mu_t(dx)+\beta\pi\sin(\pi m_t)
                =\beta\pi\sin(\pi m_t),\\
 \dot{\overline X}_t&=-\frac aN\sum_i(X_t^i-\overline X_t)
                +\beta\pi\sin(\pi\overline X_t)
                =\beta\pi\sin(\pi\overline X_t).
 \end{aligned}
 \label{eq:canonical-mean-equations}
\end{equation}
These scalar ODEs are globally Lipschitz. Since $m_0=0$, uniqueness
gives $m_t=0$ before the jump; hence \eqref{eq:canonical-jumps} yields
\begin{equation}
 m_\tau=-\eta(a+b)Z.
 \label{eq:canonical-population-jump-mean}
\end{equation}
Subtracting the mean equations from the particle equations gives
\[
 \frac d{dt}(Y_t^i-m_t)=-a(Y_t^i-m_t),\qquad
 \frac d{dt}(X_t^i-\overline X_t)=-a(X_t^i-\overline X_t).
\]
Each translation in \eqref{eq:canonical-jumps} also translates its mean,
so $Y_\tau^i-m_\tau=Y_{\tau-}^i-m_{\tau-}$ and
$X_\tau^i-\overline X_\tau=X_{\tau-}^i-\overline X_{\tau-}$.
Solving the centered equations therefore gives global trajectories
\begin{equation}
 \begin{split}
 Y_t^i&=m_t+e^{-at}U_i,\qquad
 X_t^i=\overline X_t+e^{-at}(U_i-\overline X_0),\\
 X_t^i-Y_t^i&=\overline X_t-m_t-e^{-at}\overline X_0.
 \end{split}
 \label{eq:canonical-trajectories}
\end{equation}
They hold also at both sides of the jump.
For $\dot u=\beta\pi\sin(\pi u)$, the drift is positive on
$(2k,2k+1)$ and negative on $(2k-1,2k)$.
Uniqueness prevents crossing an integer equilibrium, so a mean moves
at most one unit on each ODE interval. There are two intervals and one
jump. Using $|c|\le a+b$ and $|U_i|,|\overline X_0|\le R_0$ gives
\[
 \sup_{t\ge0}|Y_t^i|\le3R_0+2+\eta(a+b)|Z|,\qquad
 \sup_{t\ge0}|X_t^i|\le3R_0+2+\eta(a+b)|\zeta|.
\]
For a standard normal $Z$, integration by parts, with vanishing boundary
term, gives
\[
 \mathbb E Z^{2k}
 =-\frac1{\sqrt{2\pi}}\int z^{2k-1}\frac d{dz}e^{-z^2/2}\,dz
 =(2k-1)\mathbb E Z^{2k-2}
 =\prod_{j=1}^k(2j-1)\le(2k)^k.
\]
Taking $k=\lceil q/2\rceil$ for $q\ge1$ yields
$\|Z\|_{L^q}\le\|Z\|_{L^{2k}}\le\sqrt{2k}\le2\sqrt q$;
$\zeta$ has the same marginal bound.
Minkowski's inequality now gives
\[
 \left\|\sup_{t\ge0}|Y_t^i|\right\|_{L^q}
 \vee\left\|\sup_{t\ge0}|X_t^i|\right\|_{L^q}
 \le3R_0+2+2\eta(a+b)\sqrt q\le C\sqrt q.
\]
This proves (iii) with $\alpha=2$ and a fixed $K_0\le C$.

\par\noindent\textbf{Step 4: piecewise Lipschitz regularity, for (iv).}
By \eqref{eq:canonical-trajectories}, the centered coordinates have
absolute value at most $2R_0$.
The gradient formula \eqref{eq:canonical-wasserstein-derivatives} gives
$|\dot X_t^i|\vee|\dot Y_t^i|\le2aR_0+\beta\pi$.
Integrating on an ODE interval gives
$|X_t^i-X_s^i|\vee|Y_t^i-Y_s^i|\le C(t-s)$.
The continuous ODE branches and the translations supply the required traces.

\par\noindent\textbf{Step 5: anti-spikiness, for (v).}
The difference in \eqref{eq:canonical-trajectories} is independent of $i$.
Thus \eqref{eq:tracking-metric} gives
$r_t=|\overline X_t-m_t-e^{-at}\overline X_0|$, and
\[
 \frac1N\sum_i|X_t^i-Y_t^i|^3
 =|\overline X_t-m_t-e^{-at}\overline X_0|^3=r_t^3.
\]
This includes $r_t=0$ and both jump traces, proving
\eqref{eq-cubic-antispike-bound} with $A_{\rm sp}=1$ almost surely.
\end{proof}

\begin{remark}[Local intuition for the coupled jump]
\label{rem:canonical-jump-alignment}
Recall $c(u)=-a-b\cos(\pi u)$ from
\cref{lem:canonical-comparison-assumptions,eq:canonical-wasserstein-derivatives}
and write $\overline Y_t=N^{-1}\sum_iY_t^i$.
On the constant direction $e_N$ in
\eqref{eq:canonical-finite-factor}, the alignment in
\cref{lem:population-first-finite-coupling} gives
\[
 \zeta=-\operatorname{sign}_+(c(\overline Y_{\tau-}))Z,
 \qquad c(u)=-(a+b)+O(u^2)\quad(u\to0),
\]
where $\operatorname{sign}_+(v)=-1$ for $v<0$ and $+1$ otherwise.
When both empirical means are sufficiently near zero,
$c(\overline Y_{\tau-})<0$, so $\zeta=Z$ and the jumps in
\eqref{eq:canonical-jumps} satisfy
\[
 X_\tau^i-X_{\tau-}^i\simeq-\eta(a+b)Z=Y_\tau^i-Y_{\tau-}^i.
\]
This explains locally why the coupled translations agree to leading order;
the comparison proof below uses the general theorem, not this approximation.
\end{remark}

\begin{lemma}[Curvature-response and population-objective bounds after escape]
\label{lem:canonical-response-control}
For $N\ge2$ sufficiently large, there is a population-measurable event
$E_N^{\rm esc}$ with $\mathbb P(E_N^{\rm esc})\ge1-\tfrac12N^{-1/16}$
on which the curvature-response factors in
\cref{def:curvature-responses,eq:tracking-curvature-responses} satisfy
\begin{equation}
 \begin{aligned}
 \sup_{t\ge0}(\mathfrak C_t^Y+\mathfrak J_t^Y)&\le CN^{1/16},\\
 \mathfrak C_t^Y&\le C,\qquad
 \mathfrak J_t^Y\le CN^{-1/4},\qquad
 0\le F(\mu_t)+\beta\le CN^{-1/2}\quad(t\ge C\log N).
 \end{aligned}
 \label{eq:canonical-response-bounds}
\end{equation}
The constants depend only on the fixed model parameters.
\end{lemma}
\begin{proof}
\noindent\textbf{Step 1: exclude landings close to unstable equilibria.}
By \cref{lem:canonical-comparison-assumptions,eq:canonical-population-jump-mean},
$m_\tau=-\sigma Z$ with $\sigma=\eta(a+b)>0$.
Its density is $f_\sigma(x)=(2\pi\sigma^2)^{-1/2}e^{-x^2/(2\sigma^2)}$.
For $0<\rho<1/2$, symmetry and monotonicity on $[0,\infty)$ give
\[
 \begin{aligned}
 \mathbb P\{\operatorname{dist}(m_\tau,2\mathbb Z)<\rho\}
 &=\sum_{k\in\mathbb Z}\int_{2k-\rho}^{2k+\rho}f_\sigma(x)\,dx\\
 &\le2\rho f_\sigma(0)+4\rho\sum_{k\ge1}f_\sigma(2k-1)
 \le2\rho\{f_\sigma(0)+1\}.
 \end{aligned}
\]
For the last inequality,
$\sum_{k\ge1}f_\sigma(2k-1)
\le\sum_{k\ge1}\int_{2k-2}^{2k-1}f_\sigma(x)\,dx\le1/2$.
Here $\operatorname{dist}(u,2\mathbb Z)=\inf_{k\in\mathbb Z}|u-2k|$.
Choose $\rho_N$ to be a sufficiently small fixed positive multiple of
$N^{-1/16}$. Then
\begin{equation}
 E_N^{\rm esc}:=\{\operatorname{dist}(m_\tau,2\mathbb Z)\ge\rho_N\},
 \qquad \mathbb P((E_N^{\rm esc})^c)\le\tfrac12N^{-1/16}.
 \label{eq:canonical-escape-probability}
\end{equation}
This event depends only on the population jump. Work on it below.

\par\noindent\textbf{Step 2: compute the curvature and its recovery time.}
Insert \cref{lem:canonical-comparison-assumptions,eq:canonical-wasserstein-derivatives}
into the sampled full Hessian of \cref{def:curvature-responses,eq:curvature-main}.
With $e_N=N^{-1/2}\mathbf1$ as in \eqref{eq:canonical-finite-factor},
\begin{equation}
 \begin{split}
 \HessianMatrices_t^Y&=aI_N+c(m_t)e_Ne_N^\top,\\
 \HessianMatrices_t^Ye_N&=-b\cos(\pi m_t)e_N,\qquad
 \HessianMatrices_t^Yv=av\quad(e_N^\top v=0),\\
 \kappa_t^{Y,N}&=\min\{a,-b\cos(\pi m_t)\}\ge-b.
 \end{split}
 \label{eq:canonical-sampled-curvature}
\end{equation}
The constant direction and its orthogonal complement are nontrivial since $N\ge2$.
Also $|\kappa_t^{Y,N}|\le\max\{a,b\}$, so it is integrable on every
finite horizon.
Let $d_t=\operatorname{dist}(m_t,2\mathbb Z)\in[0,1]$ for $t\ge\tau$.
On either branch $m_t=2k\pm d_t$,
\cref{lem:canonical-comparison-assumptions,eq:canonical-mean-equations}
gives $\dot d_t=\beta\pi\sin(\pi d_t)\ge0$.
If $d_\tau=1$, the mean is an odd-integer equilibrium and remains there.
Otherwise $0<d_\tau<1$ on $E_N^{\rm esc}$.
We use the half-angle coordinate $\tan(\pi d_t/2)$ to turn the sine drift
into a linear equation and obtain an explicit curvature-recovery time.
Direct differentiation gives
\begin{equation}
 \begin{split}
 \frac d{dt}\tan\frac{\pi d_t}{2}
 &=\frac{\pi}{2\cos^2(\pi d_t/2)}\beta\pi\sin(\pi d_t)
   =b\tan\frac{\pi d_t}{2},\\
 \tan\frac{\pi d_t}{2}&=e^{b(t-\tau)}\tan\frac{\pi d_\tau}{2}.
 \end{split}
 \label{eq:canonical-mean-solution}
\end{equation}
Set $t_N=\tau+b^{-1}\log(2/\rho_N)$ and $\lambda=\min\{a,b/2\}>0$.
Since $\tan(\pi\rho_N/2)\ge\rho_N$,
$\tan(\pi d_{t_N}/2)\ge(2/\rho_N)\rho_N=2>\sqrt3=\tan(\pi/3)$.
Hence $d_t\ge2/3$ for $t\ge t_N$, including the equilibrium case,
and $\cos(\pi m_t)=\cos(\pi d_t)\le-1/2$.
Consequently \eqref{eq:canonical-sampled-curvature} gives
\begin{equation}
 \kappa_t^{Y,N}\ge\lambda\quad(t\ge t_N),\qquad
 e^{bt_N}=2e^{b\tau}/\rho_N\le CN^{1/16}.
 \label{eq:canonical-curvature-recovery}
\end{equation}

\par\noindent\textbf{Step 3: split the curvature-response factors at the recovery time.}
Recall from \eqref{eq:tracking-curvature-responses} that
$G_\kappa^Y(t,s)=e^{-\int_s^t\kappa_u^{Y,N}\,du}$,
$\mathfrak C_t^Y=\int_0^tG_\kappa^Y(t,s)\,ds$, and, for this one jump,
$\mathfrak J_t^Y=0$ before $\tau$ and
$\mathfrak J_t^Y=\eta G_\kappa^Y(t,\tau)$ after it.
For $t\le t_N$, \cref{eq:canonical-sampled-curvature,eq:canonical-curvature-recovery} give
\[
 \mathfrak C_t^Y\le\int_0^te^{b(t-s)}\,ds=(e^{bt}-1)/b\le C/\rho_N,
 \qquad
 \mathfrak J_t^Y\le\eta e^{b(t-\tau)}\le C/\rho_N\quad(t\ge\tau).
\]
For $t\ge t_N$, split the integral and use
$G_\kappa^Y(t,s)=G_\kappa^Y(t,t_N)G_\kappa^Y(t_N,s)$ when $s\le t_N$.
By \eqref{eq:canonical-curvature-recovery},
\begin{equation}
 \begin{aligned}
 \mathfrak C_t^Y
 &=G_\kappa^Y(t,t_N)\mathfrak C_{t_N}^Y+\int_{t_N}^tG_\kappa^Y(t,s)\,ds
 \le(C/\rho_N)e^{-\lambda(t-t_N)}+\lambda^{-1},\\
 \mathfrak J_t^Y
 &=G_\kappa^Y(t,t_N)\mathfrak J_{t_N}^Y
 \le(C/\rho_N)e^{-\lambda(t-t_N)}.
 \end{aligned}
 \label{eq:canonical-integrated-responses}
\end{equation}
The integral over $[t_N,t]$ satisfies
\[
 \int_{t_N}^tG_\kappa^Y(t,s)\,ds
 \le\int_{t_N}^te^{-\lambda(t-s)}\,ds
 =\frac{1-e^{-\lambda(t-t_N)}}{\lambda}\le\lambda^{-1}.
\]
These bounds imply the all-time estimate in \eqref{eq:canonical-response-bounds}.
We choose the waiting time so that the exponential decay cancels the factor
$\rho_N^{-1}$ in the bound for $\mathfrak J_t^Y$ in
\cref{eq:canonical-integrated-responses}, leaving a bound of order $N^{-1/4}$:
\begin{equation}
 t\ge t_N+\lambda^{-1}\log(\rho_N^{-1}N^{1/4})
 \quad\Longrightarrow\quad e^{-\lambda(t-t_N)}\le\rho_NN^{-1/4}.
 \label{eq:canonical-late-time}
\end{equation}
The threshold on the left is at most $C\log N$, because
$\rho_N$ is a fixed multiple of $N^{-1/16}$.
Substitution into \eqref{eq:canonical-integrated-responses} gives
$\mathfrak C_t^Y\le CN^{-1/4}+\lambda^{-1}\le C$ and
$\mathfrak J_t^Y\le CN^{-1/4}$ in this range.

\par\noindent\textbf{Step 4: decay of the population objective.}
The population trajectory in
\cref{lem:canonical-comparison-assumptions,eq:canonical-trajectories}
is the pushforward $u\mapsto m_t+e^{-at}u$, so
$\operatorname{Var}(\mu_t)=e^{-2at}\operatorname{Var}(\mu_0)$.
For $t\ge t_N$, we have $\cos(\pi m_t)\le-1/2$, whence
$\sin^2(\pi m_t)=(1-\cos(\pi m_t))(1+\cos(\pi m_t))
\ge1+\cos(\pi m_t)$.
Differentiate \eqref{eq:canonical-objective} using
\eqref{eq:canonical-mean-equations}, and recall $b=\beta\pi^2$:
\[
 \begin{aligned}
 \frac d{dt}[F(\mu_t)+\beta]
 &=\tfrac a2[-2a\operatorname{Var}(\mu_t)]
   -\beta\pi\sin(\pi m_t)[\beta\pi\sin(\pi m_t)]\\
 &=-a^2\operatorname{Var}(\mu_t)-b\beta\sin^2(\pi m_t)\\
 &\le-\lambda a\operatorname{Var}(\mu_t)
       -2\lambda\beta(1+\cos(\pi m_t))
 =-2\lambda[F(\mu_t)+\beta],
 \end{aligned}
\]
where $\lambda\le a$ and $2\lambda\le b$.
Multiplying by $e^{2\lambda(t-t_N)}$ and integrating, with
$F(\mu_{t_N})+\beta\le aR_0^2/2+2\beta\le C$, yields
\[
 0\le F(\mu_t)+\beta\le Ce^{-2\lambda(t-t_N)}
       \le C\rho_N^2N^{-1/2}\le CN^{-1/2}
       \qquad(t\ge C\log N),
\]
where the second inequality is \eqref{eq:canonical-late-time} squared.
\end{proof}

\begin{proof}[Proof of \cref{prop:canonical-model-assumptions}]
\noindent\textbf{Step 1: verify stability on the escape event.}
\Cref{lem:canonical-comparison-assumptions} verifies all comparison
assumptions with uniform constants.
Take $T=N^{1/8}$ and $p=\tfrac12N^{-1/16}$, with $N$ large enough that
$T\ge\tau$. For one jump, the scale in \eqref{eq:def-supXY-v15} satisfies
\[
 \Xi=16(1+B)N(e+T)/p\le CN^{1+1/8+1/16}=CN^{19/16}.
\]
The three powers used here have separate roles. The choice
$\rho_N\asymp N^{-1/16}$ pays the escape failure probability in
\cref{eq:canonical-escape-probability}; the recovery estimate
\cref{eq:canonical-curvature-recovery} then bounds the transient curvature-response factor
growth by $\rho_N^{-1}=O(N^{1/16})$; and the horizon $T=N^{1/8}$ leaves a
strictly negative power in the nonlinear stability check below.
Use the residual and coefficient bounds in the proof of
\cref{cor:fixed-schedule-high-prob-r-rate-v15}, specifically
\eqref{eq:main-rate-internal-envelopes}.
Since $T/N=N^{-7/8}\le N^{-1/2}$ and those bounds hold for every
positive exponent, applying them with exponent $16\theta/19$ gives,
for any $\theta>0$,
\begin{equation}
 \begin{split}
 \delta+\widetilde\delta&\le C_\theta N^{-1/2+\theta},\qquad
 \sqrt\delta+\sqrt{\widetilde\delta}\le C_\theta N^{-1/4+\theta},\\
 1+c_D+c_J(D)q_p+\Lambda(D)+J_{N,T,p}&\le C_\theta N^\theta.
 \end{split}
 \label{eq:canonical-residual-envelopes}
\end{equation}
Here the square-root bound uses
$\sqrt\delta+\sqrt{\widetilde\delta}\le\sqrt{2(\delta+\widetilde\delta)}$
and $N^{\theta/2}\le N^\theta$.
The one-jump product in \eqref{eq:def-jump-amplification-v15} is
$J_{N,T,p}=1+\eta c_J(D)q_p$, which proves its bound with the same exponent.
The other quantities retain the definitions in
\cref{eq:def-delta-v15,eq:def-tdelta-v15,eq:def-supXY-v15}.

Work on $E_N^{\rm esc}$ from
\cref{lem:canonical-response-control,eq:canonical-escape-probability}.
Insert its all-time curvature-response bound \eqref{eq:canonical-response-bounds}
and \eqref{eq:canonical-residual-envelopes} into the forcing
\eqref{eq:def-B-v15} and radius $R=2J_{N,T,p}\mathfrak B_*$:
\[
 \begin{aligned}
 \mathfrak B_*
 &\le\delta\sup_{t\le T}\mathfrak C_t^Y
       +c_J(D)q_p(\sqrt\delta+\sqrt{\widetilde\delta})
                          \sup_{t\le T}\mathfrak J_t^Y\\
 &\le C_\theta\left[N^{-1/2+\theta+1/16}
                    +N^{-1/4+2\theta+1/16}\right]
 \le C_\theta N^{-3/16+2\theta},\\
 R&\le C_\theta N^\theta N^{-3/16+2\theta}
       =C_\theta N^{-3/16+3\theta},\\
 T(\delta+c_DR)
 &\le C_\theta\left[N^{-3/8+\theta}+N^{-1/16+4\theta}\right]
 \le C_\theta N^{-1/16+4\theta}.
 \end{aligned}
\]
Fix $4\theta<\min\{1/16,\varepsilon\}$.
For sufficiently large $N$, the last bound is at most $1/4$, exactly
as required by the nonlinear stability event \eqref{eq:smallness-v15}.
Thus $E_N^{\rm esc}\subset\mathcal S_{N,T,p}$.

\par\noindent\textbf{Step 2: apply comparison and prove (i).}
By \cref{lem:canonical-comparison-assumptions}(v), anti-spikiness holds
almost surely. Curvature integrability follows from
\cref{lem:canonical-response-control,eq:canonical-sampled-curvature}.
Apply \cref{thm:fixed-schedule-high-prob-r-v15} with the above $p$.
Its proof constructs $\Omega_p$ so that $\sup_{t\le T}\Gamma_t^N\le\delta$
and the population and localization events hold there.
On $\Omega_p\cap E_N^{\rm esc}$,
\cref{eq:high-prob-r-bound-v15,eq:def-B-v15,eq:canonical-residual-envelopes}
give
\[
 \begin{aligned}
 r_t&\le2J_{N,T,p}
       [\delta\mathfrak C_t^Y
        +c_J(D)q_p(\sqrt\delta+\sqrt{\widetilde\delta})\mathfrak J_t^Y]\\
 &\le C_\theta N^{3\theta}
       [N^{-1/2}\mathfrak C_t^Y+N^{-1/4}\mathfrak J_t^Y].
 \end{aligned}
\]
Then \cref{lem:functional-gap-fixed-time-v15,eq:functional-gap-fixed-time-v15}
implies
\begin{equation}
 \begin{split}
 r_t+|F(\rho_{\mathbf X_t})-F(\mu_t)|
 &\le(1+\Lambda(D))r_t+\delta\\
 &\le C_\theta N^{4\theta}
       [N^{-1/2}(1+\mathfrak C_t^Y)+N^{-1/4}\mathfrak J_t^Y].
 \end{split}
 \label{eq:canonical-response-tracking}
\end{equation}
The all-time bound in
\cref{lem:canonical-response-control,eq:canonical-response-bounds}
makes the last line at most
$C_\theta N^{4\theta}[N^{-1/2}(1+N^{1/16})+N^{-1/4+1/16}]
\le C_\varepsilon N^{-3/16+\varepsilon}$, proving (i).
A union bound gives
$\mathbb P(\Omega_p\cap E_N^{\rm esc})
\ge1-p-\tfrac12N^{-1/16}=1-N^{-1/16}$.

\par\noindent\textbf{Step 3: substitute the late-time bounds to prove (ii).}
For $t\ge C\log N$, substitute
$\mathfrak C_t^Y\le C$ and $\mathfrak J_t^Y\le CN^{-1/4}$ from
\cref{lem:canonical-response-control,eq:canonical-response-bounds}
into \eqref{eq:canonical-response-tracking}. Its right-hand side is
at most $C_\theta N^{4\theta}[N^{-1/2}+N^{-1/4}N^{-1/4}]
\le C_\varepsilon N^{-1/2+\varepsilon}$.
Finally, on the same event,
\[
 0\le F(\rho_{\mathbf X_t})+\beta
 \le |F(\rho_{\mathbf X_t})-F(\mu_t)|+F(\mu_t)+\beta.
\]
The population bound in \eqref{eq:canonical-response-bounds} is
$CN^{-1/2}$, so adding this inequality proves (ii), after increasing
$C_\varepsilon$. Increase $N_0$ if needed so that $C\log N<N^{1/8}$.
\end{proof}
\pagebreak[3]

%% file: appendix_final_general/03_trigger_stability.tex
\subsection{A fixed-time comparison of squared gradients}
\label{subsec:trigger-stability}

We compare the squared-gradient statistics on paths already coupled under
the prescribed common schedule of \cref{thm:fixed-schedule-high-prob-r-v15}.
The observation time is deterministic and lies inside an ODE interval.
Define
\begin{equation}
\begin{split}
 \mathcal G_t&:=\int\|\nabla_WF(\mu_t,x)\|^2\,\mu_t(dx),\\
 \mathcal G_t^N&:=\frac1N\sum_{i=1}^N
       \|\nabla_WF(\rho_{\mathbf X_t},X_t^i)\|^2.
\end{split}
\label{eq:def-trigger-functional-squared}
\end{equation}

\paragraph{Recall of the comparison quantities.}
Let $E_{\rm pop}\in\mathcal G_{\rm pop}$ and $K_{\rm pop}$ be the
population moment event and its deterministic moment scale from
\cref{lem:population-moment-event}. Conditional on the population fields
$\mathcal G_{\rm pop}$, the variables $Y_t^i$ are i.i.d. with law $\mu_t$;
see \cref{lemma-gp-jump-implies-conditional-iid}.
For $D\ge1$, recall the sample localization event
\[
 E_{T,D}=\left\{\sup_{s\le T}\max_{i\le N}
          (\|X_s^i\|\vee\|Y_s^i\|)\le D\right\}
\]
from \cref{eq:sample-localization-event}.
The common local coefficients in
\cref{eq:conditional-comparison-envelope,eq:common-local-moment-weights}
and the drift comparison are
\[
 \mathrm L_D=\mathrm L_{K_{\rm pop},D},\qquad
 \gamma_m(D)=\mathrm L_Da_m,\qquad
 \Lambda(D)=\sum_{m\ge1}\gamma_m(D)L_m(1+2D^{m-1}).
\]
Here $a_m$ are the moment weights and $L_m$ are the feature derivative
bounds in \cref{assumption-structured-moment-smoothness}. The existing
discrepancy and moment residual are
\[
 r_t=\left(\frac1N\sum_i\|X_t^i-Y_t^i\|^2\right)^{1/2},\qquad
 \Gamma_t^N=\sum_{m\ge1}\gamma_m(D)
 \left\|\frac1N\sum_i\phi_m(Y_t^i)-\int\phi_m\,d\mu_t\right\|;
\]
see \cref{subsec:tracking-metric,lemma-directional-measure-slot-error}.
Let $B$ be the deterministic ODE velocity bound in
\cref{assumption-uniform-concentration-piecewise-lipschitz}.

\begin{lemma}[Fixed-time discrepancy of the squared-gradient statistic]
\label{lem:fixed-time-trigger-squared-gap}
Assume
\cref{assumption-structured-moment-smoothness,assumption-common-seed,assumption-moment-growth,assumption-uniform-concentration-piecewise-lipschitz},
with the prescribed-schedule coupling in
\cref{definition-ode-and-jump,lem:population-first-finite-coupling}.
The reference conditional product law is
$\operatorname{Law}((Y_t^i)_i\mid\mathcal G_{\rm pop})=\mu_t^{\otimes N}$
by \cref{lemma-gp-jump-implies-conditional-iid}.
Conditional moments are bounded on $E_{\rm pop}$ by
\eqref{eq:population-uniform-conditional-moments};
$E_{T,D}$ is the sample event \eqref{eq:sample-localization-event}.
The existing $B$-Lipschitz ODE condition holds almost surely.
Fix a deterministic
$t\in(0,T)\setminus\{t_1,\ldots,t_{\NofJ}\}$ and $\delta\in(0,1)$.
Except on an event whose intersection with $E_{\rm pop}\cap E_{T,D}$
has probability at most $\delta$, one has
\begin{equation}
 |\mathcal G_t^N-\mathcal G_t|
 \le2B\{(\Lambda(D)+\mathrm L_D)r_t+\Gamma_t^N\}
       +B^2\sqrt{\frac{\log(2/\delta)}{2N}}.
 \label{eq:fixed-time-trigger-squared-gap}
\end{equation}
Equivalently,
\[
 \mathbb P\!\left(
 E_{\rm pop}\cap E_{T,D}\cap
 \left\{|\mathcal G_t^N-\mathcal G_t|>
 2B\{(\Lambda(D)+\mathrm L_D)r_t+\Gamma_t^N\}
 +B^2\sqrt{\frac{\log(2/\delta)}{2N}}\right\}\right)\le\delta.
\]
\end{lemma}
\begin{proof}
\noindent\textbf{Step 1: the velocity bound controls the actual gradients.}
We first work on $E_{\rm pop}$, without imposing $E_{T,D}$.
Recall the ODE in \cref{definition-ode-and-jump} and the population
moment continuity calculation in the proof of
\cref{lemma-common-empirical-moment-residual-directional-rt}.
For two nearby times $s,t$ in the same ODE interval, integrate
$D\phi_m$ along the segment joining the two particle positions.
The first-derivative and conditional moment bounds give
\[
 \sum_m a_m\|M_m(\mu_s)-M_m(\mu_t)\|
 \le B|s-t|\sum_m a_mL_m
       \{1+2(K_{\rm pop}m^{1/\alpha})^{m-1}\}
 \longrightarrow0.
\]
The series is finite by the decay estimate
\cref{lem:conditional-local-coefficient-growth}.
Choose a finite $R\ge1$ containing the finitely many continuous particle
paths in a neighborhood of $t$. On $E_{\rm pop}$,
$\mu_s\in\mathcal P_{\alpha,K_{\rm pop}}$ and
$\rho_{\mathbf X_s}\in\mathcal P_R$, so their moment vectors belong to
$\mathcal V_{K_{\rm pop},R}$ by \cref{eq:localized-feasible-moment-region}.
The local moment and spatial estimates
\cref{eq:local-value-gradient-moment-bound,eq:local-gradient-spatial-bound}
then imply
\begin{align*}
 &\|\nabla_WF(\mu_s,Y_s^i)-\nabla_WF(\mu_t,Y_t^i)\|\\
 &\quad\le \mathrm L_{K_{\rm pop},R}
 \left\{\sum_m a_m\|M_m(\mu_s)-M_m(\mu_t)\|
                      +\|Y_s^i-Y_t^i\|\right\}\longrightarrow0.
\end{align*}
For the empirical moments the corresponding estimate is
\[
 \sum_m a_m\|M_m(\rho_{\mathbf X_s})-M_m(\rho_{\mathbf X_t})\|
 \le B|s-t|\sum_m a_mL_m(1+2R^{m-1})\longrightarrow0.
\]
The same local estimates therefore give continuity of
$\nabla_WF(\rho_{\mathbf X_s},X_s^i)$ at $t$.
The radius $R$ is used only to establish this local continuity; it
imposes no additional sample event.

Using the integral form of the reference ODE, for $h\downarrow0$,
\[
 \frac{Y_{t+h}^i-Y_t^i}{h}
 =-\frac1h\int_t^{t+h}\nabla_WF(\mu_s,Y_s^i)\,ds
 \longrightarrow-\nabla_WF(\mu_t,Y_t^i).
\]
Every difference quotient has norm at most $B$, so its limit does too.
Applying the identical argument to $X$ gives, on $E_{\rm pop}$,
\[
 \|\nabla_WF(\mu_t,Y_t^i)\|\le B,\qquad
 \|\nabla_WF(\rho_{\mathbf X_t},X_t^i)\|\le B.
\]
The population bound extends to $\mu_t$-almost every point. Indeed,
conditional-i.i.d. sampling and $E_{\rm pop}\in\mathcal G_{\rm pop}$ give
\begin{align*}
 0
 &=\mathbb E\!\left[\mathbf1_{E_{\rm pop}}
   \mathbf1_{\{\|\nabla_WF(\mu_t,Y_t^1)\|>B\}}\right]\\
 &=\mathbb E\!\left[\mathbf1_{E_{\rm pop}}
   \int\mathbf1_{\{\|\nabla_WF(\mu_t,x)\|>B\}}\,\mu_t(dx)\right].
\end{align*}
The integrand is nonnegative. Its integral is therefore zero for almost
every population realization in $E_{\rm pop}$, proving
$\|\nabla_WF(\mu_t,x)\|\le B$ for $\mu_t$-almost every $x$.

\par\noindent\textbf{Step 2: compare the gradients at paired particles.}
On $E_{\rm pop}\cap E_{T,D}$, insert the reference empirical moments and use
\cref{eq:local-value-gradient-moment-bound,eq:local-gradient-spatial-bound}:
\begin{align*}
 &\|\nabla_WF(\rho_{\mathbf X_t},X_t^i)
                -\nabla_WF(\mu_t,Y_t^i)\|\\
 &\le\sum_m\gamma_m(D)
       \left\|\frac1N\sum_j\phi_m(X_t^j)-\int\phi_m\,d\mu_t\right\|
        +\mathrm L_D\|X_t^i-Y_t^i\|\\
 &\le\sum_m\gamma_m(D)\frac1N\sum_j
         \|\phi_m(X_t^j)-\phi_m(Y_t^j)\|
        +\Gamma_t^N+\mathrm L_D\|X_t^i-Y_t^i\|\\
 &\le\Lambda(D)r_t+\Gamma_t^N+\mathrm L_D\|X_t^i-Y_t^i\|.
\end{align*}
For the last step, the line segment lies in the radius-$D$ ball, so
the first-derivative bound gives
\[
 \begin{aligned}
 \|\phi_m(X_t^j)-\phi_m(Y_t^j)\|
 &\le\int_0^1
 \|D\phi_m(Y_t^j+s(X_t^j-Y_t^j))\|\,\|X_t^j-Y_t^j\|\,ds
 \\
 &\le L_m(1+2D^{m-1})\|X_t^j-Y_t^j\|.
 \end{aligned}
\]
Here $1+D^{m-1}\le1+2D^{m-1}$ allows reuse of $\Lambda(D)$.
Cauchy--Schwarz gives
\[
 \frac1N\sum_j\|X_t^j-Y_t^j\|
 \le\left(\frac1N\sum_j\|X_t^j-Y_t^j\|^2\right)^{1/2}=r_t.
\]
For two vectors of norm at most $B$,
$|\|u\|^2-\|v\|^2|=|\langle u-v,u+v\rangle|\le2B\|u-v\|$.
Apply this inequality to the two actual gradients bounded in Step 1
and average the preceding estimate:
\begin{equation}
 \left|\mathcal G_t^N-\frac1N\sum_i
          \|\nabla_WF(\mu_t,Y_t^i)\|^2\right|
 \le2B\{(\Lambda(D)+\mathrm L_D)r_t+\Gamma_t^N\}.
 \label{eq:trigger-paired-gradient-comparison}
\end{equation}
No bound on a gradient evaluated at an intermediate measure--point pair
is needed.

\par\noindent\textbf{Step 3: concentration of the reference average.}
For almost every population realization in $E_{\rm pop}$, the scalars
$\|\nabla_WF(\mu_t,Y_t^i)\|^2$ are conditionally i.i.d., lie in $[0,B^2]$,
and have conditional mean
\[
 \mathbb E[\|\nabla_WF(\mu_t,Y_t^i)\|^2\mid\mathcal G_{\rm pop}]
 =\int\|\nabla_WF(\mu_t,x)\|^2\,\mu_t(dx)=\mathcal G_t.
\]
For $B>0$, the bounded-differences inequality
\cref{lemma:mcdiarmid} applied to their average has coordinate
sensitivities $B^2/N$. Consequently
\begin{align*}
 &\mathbb P\!\left(
 \left|\frac1N\sum_i\|\nabla_WF(\mu_t,Y_t^i)\|^2-\mathcal G_t\right|
 >B^2\sqrt{\frac{\log(2/\delta)}{2N}}
 \,\middle|\,\mathcal G_{\rm pop}\right)\\
 &\quad\le
 2\exp\left\{-\frac{2N}{B^4}
       \left(B^2\sqrt{\frac{\log(2/\delta)}{2N}}\right)^2\right\}
 =\delta.
\end{align*}
For $B=0$, both statistics are zero on $E_{\rm pop}$ and the result
holds directly. For $B>0$, the triangle inequality and
\eqref{eq:trigger-paired-gradient-comparison} show that failure of
\eqref{eq:fixed-time-trigger-squared-gap} inside $E_{\rm pop}\cap E_{T,D}$
requires failure of the reference-average bound. Thus its probability
is at most
\[
 \mathbb E\!\left[\mathbf1_{E_{\rm pop}}\,
 \mathbb P\!\left(
 \left|\frac1N\sum_i\|\nabla_WF(\mu_t,Y_t^i)\|^2-\mathcal G_t\right|
 >B^2\sqrt{\frac{\log(2/\delta)}{2N}}
 \,\middle|\,\mathcal G_{\rm pop}\right)\right]
 \le\delta\,\mathbb P(E_{\rm pop})\le\delta.
\]
The conditioning throughout is on the population fields, not on
$E_{T,D}$.
\end{proof}

\begin{remark}[A threshold decision at the same fixed time]
Fix a deterministic threshold $\theta\ge0$. Suppose that
\eqref{eq:fixed-time-trigger-squared-gap} holds and
\begin{equation}
 |\mathcal G_t-\theta|>
 2B\{(\Lambda(D)+\mathrm L_D)r_t+\Gamma_t^N\}
       +B^2\sqrt{\frac{\log(2/\delta)}{2N}}.
 \label{eq:trigger-margin-agreement}
\end{equation}
If $\mathcal G_t>\theta$, then
$\mathcal G_t^N\ge\mathcal G_t-|\mathcal G_t^N-\mathcal G_t|>\theta$.
If $\mathcal G_t<\theta$, then
$\mathcal G_t^N\le\mathcal G_t+|\mathcal G_t^N-\mathcal G_t|<\theta$.
The decisions agree at this prescribed observation time. This comparison
does not identify or synchronize adaptive jump times.
\end{remark}

%% file: appendix_final_general/kappa_concentration_on_saddle_point.tex
\subsection{Finite-feature transfer of negative curvature at a fixed time}
\label{subsection:concentration-kappa}

This fixed-time supplement is not an input to the tracking theorem,
objective comparison, or fixed-time gradient comparison.
Its finite-feature conclusion does not control whole-path curvature-response
factors or adaptive jump times.

We give a sufficient sample size for a negative population Hessian
direction to remain negative in the reference sample. The statement
uses an exact finite-feature representation and a spatial-Hessian
condition. We first fix a deterministic law; the conditional
interpretation for a fixed time of PWGF follows after the proof.

\paragraph{Finite features and the two Hessian terms.}
Consider the finite-feature case of
\cref{assumption-structured-moment-smoothness}: for admissible laws $\nu$,
\begin{equation}
 F(\nu)=G\left(\int\phi\,d\nu\right),\qquad
 \phi:\mathbb R^d\longrightarrow\mathbb R^q,\qquad q\in\mathbb N.
 \label{eq:finite-feature-objective}
\end{equation}
The dimension $q$ is fixed independently of $N$; no feature cutoff or
approximation of $F$ is taken. At the law $\mu$ under consideration,
define
\begin{equation}
 \Phi(x):=\fD\phi(x)\in\mathbb R^{q\times d},\qquad
 A:=\fD^2G\left(\int\phi\,d\mu\right)=A^\top\in\mathbb R^{q\times q}.
 \label{eq:finite-feature-derivatives}
\end{equation}
The moment chain rule in \cref{eq:weighted-integral-hessian-representation}
reduces to the finite sum
\[
 \nabla_W^2F(\mu;x,y)
 =\sum_{j,k=1}^q A_{jk}\nabla\phi_j(x)\nabla\phi_k(y)^\top
 =\Phi(x)^\top A\Phi(y).
\]
Recall the integral and spatial Hessian operators from
\cref{subsec:population-pwgf}. On $L^2(\mu;\mathbb R^d)$ they are
\begin{equation}
 \begin{split}
 (H_\mu v)(x)&:=
   \int\Phi(x)^\top A\Phi(y)v(y)\,\mu(dy),\\
 (H'_\mu v)(x)&:=
   \nabla_x\nabla_WF(\mu,x)v(x).
 \end{split}
 \label{eq:finite-feature-hessian-operators}
\end{equation}
Their sum is the full operator $\mathcal H_\mu=H_\mu+H'_\mu$.
For independent $Y_1,\ldots,Y_N\sim\mu$, its sampled full Hessian is
\begin{equation}
 \begin{split}
 \mathcal H^{Y,N}(i,j)
 &:=\frac1N\Phi(Y_i)^\top A\Phi(Y_j)
      +\delta_{ij}\nabla_x\nabla_WF(\mu,Y_i),\\
 \kappa^{Y,N}&:=\lambda_{\min}(\mathcal H^{Y,N}).
 \end{split}
 \label{eq:finite-feature-sampled-hessian}
\end{equation}
The matrix acts on $(\mathbb R^d)^N$ with its Euclidean inner product.
At $\mu=\mu_t$, this is precisely the reference-sample curvature
$\kappa_t^{Y,N}$ of \eqref{eq:curvature-main}: derivatives are evaluated
at $\mu_t$, not at the empirical measure.

\Needspace{8\baselineskip}
\begin{proposition}[Transfer of negative curvature to the reference sample]
\label{prop:finite-feature-negative-curvature}
Fix $\mu$ as above, and suppose that the following hold for deterministic
constants $K\ge1$, $C_s\ge0$, $\varepsilon\ge0$, and $\gamma>0$.
\begin{enumerate}
 \item The coefficient and feature moments satisfy
 \begin{equation}
 \|A\|_{\mathbb R^q\to\mathbb R^q}\le K,\qquad
 \mathbb E\|\Phi(Y_1)\|_F^4\le K^4.
 \label{eq:finite-feature-common-bound}
 \end{equation}
 Here $\|\cdot\|_F$ is the Frobenius norm, the square root of the sum of
 squared entries.
 \item The spatial Hessian is controlled by the population gradient:
 \begin{equation}
 \mathop{\rm ess\,sup}_{x\sim\mu}
 \|\nabla_x\nabla_WF(\mu,x)\|_{\mathbb R^d\to\mathbb R^d}
 \le C_s\|\nabla_WF(\mu,\cdot)\|_{L^2(\mu;\mathbb R^d)}.
 \label{eq:finite-feature-spatial-condition}
 \end{equation}
 \item Near-stationarity and negative full curvature hold separately:
 \begin{equation}
 \begin{split}
 \|\nabla_WF(\mu,\cdot)\|_{L^2(\mu;\mathbb R^d)}&\le\varepsilon,
       \qquad C_s\varepsilon\le\gamma/8,\\
 \inf\operatorname{Spec}(H_\mu+H'_\mu)&\le-\gamma.
 \end{split}
 \label{eq:finite-feature-negative-gap}
 \end{equation}
\end{enumerate}
Then, for every $\delta\in(0,1)$,
\begin{equation}
 N\ge\frac1\delta\max\left\{1,\frac{12^4K^{12}}{\gamma^4}\right\}
 \quad\Longrightarrow\quad
 \mathbb P(\kappa^{Y,N}\le-\gamma/2)\ge1-\delta.
 \label{eq:finite-feature-negative-transfer}
\end{equation}
\end{proposition}

Condition \eqref{eq:finite-feature-spatial-condition} has the form of
Assumption~2, equation~(8), in \citet{yamamoto2025hessianguided}.
That work imposes the condition for every law in its setting; here it is
required only at the law in the proposition. It does not follow merely
from finite features or from a small gradient.

\begin{proof}
\noindent\textbf{Step 1: a single covariance event.}
Define the population and sample feature covariances
\begin{equation}
 \Sigma:=\mathbb E[\Phi(Y_1)\Phi(Y_1)^\top],\qquad
 \widehat\Sigma:=\frac1N\sum_{i=1}^N\Phi(Y_i)\Phi(Y_i)^\top.
 \label{eq:finite-feature-covariances}
\end{equation}
They are positive semidefinite $q\times q$ matrices. If $s_j$ are the
singular values of a matrix $\Phi$, then
\[
 \|\Phi\Phi^\top\|_F^2=\sum_js_j^4
 \le\left(\sum_js_j^2\right)^2=\|\Phi\|_F^4.
\]
The centered matrices
$Z_i:=\Phi(Y_i)\Phi(Y_i)^\top-\Sigma$ are independent and have mean zero.
Expanding the squared Frobenius norm gives
\[
 \mathbb E\|Z_i\|_F^2
 =\mathbb E\|\Phi(Y_i)\Phi(Y_i)^\top\|_F^2-\|\Sigma\|_F^2
 \le K^4.
\]
For $i\ne j$, independence gives
$\mathbb E\langle Z_i,Z_j\rangle_F
=\langle\mathbb EZ_i,\mathbb EZ_j\rangle_F=0$.
Consequently
\begin{align*}
 \mathbb E\|\widehat\Sigma-\Sigma\|_F^2
 &=\frac1{N^2}\sum_{i,j}\mathbb E\langle Z_i,Z_j\rangle_F\\
 &=\frac1{N^2}\sum_i\mathbb E\|Z_i\|_F^2\le\frac{K^4}{N}.
\end{align*}
Markov's inequality, applied to this squared norm, yields
\[
 \mathbb P\left(\|\widehat\Sigma-\Sigma\|_F>
                   \frac{K^2}{\sqrt{N\delta}}\right)
 \le\frac{K^4/N}{K^4/(N\delta)}=\delta.
\]
Thus, with probability at least $1-\delta$,
\begin{equation}
\label{eq:finite-feature-covariance-success}
 \|\widehat\Sigma-\Sigma\|_{\mathbb R^q\to\mathbb R^q}
 \le\|\widehat\Sigma-\Sigma\|_F
 \le\frac{K^2}{\sqrt{N\delta}}.
\end{equation}
All subsequent sample calculations use
\eqref{eq:finite-feature-covariance-success}.
Since $\|\Phi\Phi^\top\|_{\mathbb R^q\to\mathbb R^q}\le\|\Phi\|_F^2$,
\[
 \|\Sigma\|_{\mathbb R^q\to\mathbb R^q}\le\mathbb E\|\Phi(Y_1)\|_F^2
 \le(\mathbb E\|\Phi(Y_1)\|_F^4)^{1/2}\le K^2.
\]
The sample-size condition gives $N\delta\ge1$, and hence, on
\eqref{eq:finite-feature-covariance-success},
\[
 \|\widehat\Sigma\|_{\mathbb R^q\to\mathbb R^q}
 \le\|\Sigma\|_{\mathbb R^q\to\mathbb R^q}+\|\widehat\Sigma-\Sigma\|_{\mathbb R^q\to\mathbb R^q}
 \le K^2+\frac{K^2}{\sqrt{N\delta}}\le2K^2.
\]

\par\noindent\textbf{Step 2: compare the square roots and Hessian matrices.}
Continue on the covariance bound \eqref{eq:finite-feature-covariance-success}.
Take a unit eigenvector $v$ of the symmetric matrix
$\widehat\Sigma^{1/2}-\Sigma^{1/2}$ with eigenvalue $\lambda$ such that
$|\lambda|=\|\widehat\Sigma^{1/2}-\Sigma^{1/2}\|_{\mathbb R^q\to\mathbb R^q}$.
Here each square root is the positive semidefinite square root.
The identity
\[
 \widehat\Sigma-\Sigma
 =\widehat\Sigma^{1/2}(\widehat\Sigma^{1/2}-\Sigma^{1/2})
  +(\widehat\Sigma^{1/2}-\Sigma^{1/2})\Sigma^{1/2}
\]
holds even when the two matrices do not commute.
Positive semidefiniteness gives
\[
 v^\top(\widehat\Sigma^{1/2}+\Sigma^{1/2})v
 \ge|v^\top(\widehat\Sigma^{1/2}-\Sigma^{1/2})v|=|\lambda|.
\]
Multiplying the preceding identity by $v^\top$ and $v$, and using the
eigenvector equation on its two sides, gives
\begin{align*}
 \|\widehat\Sigma-\Sigma\|_{\mathbb R^q\to\mathbb R^q}
 &\ge|v^\top(\widehat\Sigma-\Sigma)v|\\
 &=|\lambda|\,v^\top(\widehat\Sigma^{1/2}+\Sigma^{1/2})v\\
 &\ge|\lambda|^2
   =\|\widehat\Sigma^{1/2}-\Sigma^{1/2}\|_{\mathbb R^q\to\mathbb R^q}^2.
\end{align*}
Thus Step 1 implies
$\|\widehat\Sigma^{1/2}-\Sigma^{1/2}\|_{\mathbb R^q\to\mathbb R^q}
\le K(N\delta)^{-1/4}$.
Define the symmetric matrices reused in the eigenvalue comparison by
\[
 \mathcal T:=\Sigma^{1/2}A\Sigma^{1/2},\qquad
 \widehat{\mathcal T}:=\widehat\Sigma^{1/2}A\widehat\Sigma^{1/2}.
\]
Inserting and subtracting $\Sigma^{1/2}A\widehat\Sigma^{1/2}$ gives
\begin{align*}
 \widehat{\mathcal T}-\mathcal T
 &=(\widehat\Sigma^{1/2}-\Sigma^{1/2})A\widehat\Sigma^{1/2}
       +\Sigma^{1/2}A(\widehat\Sigma^{1/2}-\Sigma^{1/2}),\\
 \|\widehat{\mathcal T}-\mathcal T\|_{\mathbb R^q\to\mathbb R^q}
 &\le K(N\delta)^{-1/4}\,K\,(\sqrt2K+K)\\
 &\le3K^3(N\delta)^{-1/4}\le\gamma/4.
\end{align*}
The last inequality follows directly from
$(N\delta)^{1/4}\ge12K^3/\gamma$ in
\eqref{eq:finite-feature-negative-transfer}.

\par\noindent\textbf{Step 3: identify the negative eigenvalues.}
Define the bounded feature operator and its adjoint by
\[
 \begin{aligned}
 \mathcal B &:L^2(\mu;\mathbb R^d)\to\mathbb R^q,
 &\mathcal Bv&=\int\Phi(x)v(x)\,\mu(dx),\\
 \mathcal B^* &:\mathbb R^q\to L^2(\mu;\mathbb R^d),
 &(\mathcal B^*u)(x)&=\Phi(x)^\top u.
 \end{aligned}
\]
Indeed, Cauchy--Schwarz gives
\begin{align*}
 \|\mathcal Bv\|_{\mathbb R^q}
 &\le\int\|\Phi(x)\|_F\|v(x)\|_{\mathbb R^d}\,\mu(dx)\\
 &\le(\mathbb E\|\Phi(Y_1)\|_F^2)^{1/2}\|v\|_{L^2(\mu;\mathbb R^d)}
 \le K\|v\|_{L^2(\mu;\mathbb R^d)}.
\end{align*}
Moreover
$\langle\mathcal Bv,u\rangle_{\mathbb R^q}
=\int\langle v(x),\Phi(x)^\top u\rangle_{\mathbb R^d}\,\mu(dx)$,
which verifies the adjoint formula. By
\cref{eq:finite-feature-covariances,eq:finite-feature-hessian-operators},
for $u\in\mathbb R^q$ and $v\in L^2(\mu;\mathbb R^d)$, substitution gives
\begin{align*}
 \mathcal B\mathcal B^*u
 &=\int\Phi(x)\Phi(x)^\top u\,\mu(dx)=\Sigma u,\\
 (\mathcal B^*A\mathcal Bv)(x)
 &=\Phi(x)^\top A\int\Phi(y)v(y)\,\mu(dy)=(H_\mu v)(x).
\end{align*}
Thus $\mathcal B\mathcal B^*=\Sigma$ and
$H_\mu=\mathcal B^*A\mathcal B$.
In particular $H_\mu$ is self-adjoint and has finite rank.

We use the following elementary eigenvector calculation for products of
two bounded maps $U,V$ between two spaces. For $\lambda\ne0$,
\begin{align*}
 UVx=\lambda x,\quad x\ne0
 &\ \Longrightarrow\
 Vx\ne0,\quad VU(Vx)=\lambda Vx,\\
 VUy=\lambda y,\quad y\ne0
 &\ \Longrightarrow\
 Uy\ne0,\quad UV(Uy)=\lambda Uy.
\end{align*}
For the first implication, $Vx=0$ would give $\lambda x=UVx=0$,
a contradiction; the second is proved in the same way.
Apply these identities first with $U=\mathcal B^*$, $V=A\mathcal B$,
then with $U=A\Sigma^{1/2}$, $V=\Sigma^{1/2}$:
\[
 \mathcal B^*A\mathcal B
 \ \longleftrightarrow\ A\Sigma
 \ \longleftrightarrow\ \Sigma^{1/2}A\Sigma^{1/2}=\mathcal T
 \quad\hbox{for nonzero eigenvalues}.
\]
To check why only eigenvalues matter here, take
$v\perp\operatorname{ran}H_\mu$. Self-adjointness implies, for every $w$,
\[
 \langle H_\mu v,w\rangle_{L^2(\mu;\mathbb R^d)}
 =\langle v,H_\mu w\rangle_{L^2(\mu;\mathbb R^d)}=0,
 \qquad\text{hence }H_\mu v=0.
\]
The range is finite-dimensional and thus closed, so
\[
 L^2(\mu;\mathbb R^d)
 =\operatorname{ran}H_\mu\oplus(\operatorname{ran}H_\mu)^\perp,
 \qquad
 H_\mu=(H_\mu|_{\operatorname{ran}H_\mu})\oplus0.
\]
The first restriction is a finite-dimensional self-adjoint matrix.
Consequently any negative spectral bottom is one of its eigenvalues.
This justifies
using the displayed eigenvector correspondence for its spectral bottom
whenever that bottom is negative.

For the sample, set
$P=N^{-1/2}[\Phi(Y_1)\ \cdots\ \Phi(Y_N)]\in\mathbb R^{q\times dN}$.
Matrix multiplication gives
\[
 PP^\top=\widehat\Sigma,\qquad
 (P^\top AP)(i,j)=N^{-1}\Phi(Y_i)^\top A\Phi(Y_j).
\]
The same two product calculations, replacing $\mathcal B$ by $P$, show
that the negative eigenvalues of $P^\top AP$ and
$\widehat{\mathcal T}$ coincide. No assertion about the multiplicity
of zero eigenvalues is needed.

\par\noindent\textbf{Step 4: restore both spatial-Hessian terms.}
By \cref{eq:finite-feature-spatial-condition,eq:finite-feature-negative-gap},
\begin{gather*}
 \|H'_\mu v\|_{L^2(\mu;\mathbb R^d)}^2
 \le(C_s\varepsilon)^2\|v\|_{L^2(\mu;\mathbb R^d)}^2,\\
 \|H'_\mu\|_{L^2(\mu;\mathbb R^d)\to L^2(\mu;\mathbb R^d)}\le C_s\varepsilon\le\gamma/8.
\end{gather*}
For every unit $v\in L^2(\mu;\mathbb R^d)$, subtracting the spatial term gives
\begin{align*}
 \langle v,H_\mu v\rangle_{L^2(\mu;\mathbb R^d)}
 &=\langle v,(H_\mu+H'_\mu)v\rangle_{L^2(\mu;\mathbb R^d)}
      -\langle v,H'_\mu v\rangle_{L^2(\mu;\mathbb R^d)}\\
 &\le\langle v,(H_\mu+H'_\mu)v\rangle_{L^2(\mu;\mathbb R^d)}+\gamma/8.
\end{align*}
Taking infima of the Rayleigh quotients yields
\[
 \inf\operatorname{Spec}(H_\mu)
 \le\inf\operatorname{Spec}(H_\mu+H'_\mu)+\gamma/8
 \le-7\gamma/8.
\]
The negative spectral bottom of $H_\mu$ is an eigenvalue.
Step 3 therefore gives $\lambda_{\min}(\mathcal T)\le-7\gamma/8$.
For a unit minimum-eigenvalue vector $u$ of $\mathcal T$,
\[
 \lambda_{\min}(\widehat{\mathcal T})
 \le u^\top\widehat{\mathcal T}u
 \le u^\top\mathcal Tu+\|\widehat{\mathcal T}-\mathcal T\|_{\mathbb R^q\to\mathbb R^q}
 \le-7\gamma/8+\gamma/4=-5\gamma/8.
\]
Its negative eigenvalue corresponds to one of $P^\top AP$ by Step 3.

The essential bound in \eqref{eq:finite-feature-spatial-condition}
holds at every $Y_i$ almost surely, since a finite union of $\mu$-null
sets still has sampling probability zero. The block-diagonal spatial
matrix has norm at most
$\max_i\|\nabla_x\nabla_WF(\mu,Y_i)\|_{\mathbb R^d\to\mathbb R^d}\le\gamma/8$.
Evaluating the sampled full Hessian on a minimum-eigenvalue vector of
$P^\top AP$ now gives
\[
 \kappa^{Y,N}
 \le\lambda_{\min}(P^\top AP)
       +\max_i\|\nabla_x\nabla_WF(\mu,Y_i)\|_{\mathbb R^d\to\mathbb R^d}
 \le-5\gamma/8+\gamma/8=-\gamma/2.
\]
The only event with nonzero failure probability is the failure of
\eqref{eq:finite-feature-covariance-success}, with probability at most $\delta$.
\end{proof}

\paragraph{Fixed-time conditional interpretation.}
At a deterministic time $t$, condition on the population fields
$\mathcal G_{\rm pop}$ from \cref{lemma-gp-jump-implies-conditional-iid}
and put $\mu=\mu_t$. Let $E\in\mathcal G_{\rm pop}$ be any event on
which
\cref{eq:finite-feature-common-bound,eq:finite-feature-spatial-condition,eq:finite-feature-negative-gap}
hold with the same deterministic $q,K,C_s,\varepsilon,\gamma$.
Here $\Phi=\fD\phi$ is measure-independent, while
$A=\fD^2G(\int\phi\,d\mu_t)$ and all Hessian terms are evaluated at
the population law $\mu_t$.
For $N$ as in \eqref{eq:finite-feature-negative-transfer}, the
conditional-i.i.d. property and the tower rule give
\[
 \mathbb P(E\cap\{\kappa_t^{Y,N}>-\gamma/2\})
 =\mathbb E\!\left[\mathbf1_E
    \mathbb P(\kappa_t^{Y,N}>-\gamma/2\mid\mathcal G_{\rm pop})\right]
 \le\delta\,\mathbb P(E)\le\delta.
\]
This statement does not estimate $\mathbb P(E)$ or a whole-path curvature-response factor,
and it concerns the reference sample $Y$, not the empirical dynamics $X$.

\paragraph{A simple instance of the assumptions.}
In \eqref{eq:canonical-objective}, take $a=0$ and a law of mean zero.
The one-feature representation is $\phi(x)=x$ and
$G(z)=\beta\cos(\pi z)$. Thus $\Phi=1$, $A=-\beta\pi^2$,
$\nabla_WF(\mu,x)=0$, and $\nabla_x\nabla_WF(\mu,x)=0$.
The constant unit function is an eigenvector of $H_\mu$ with
eigenvalue $-\beta\pi^2$. Hence the conditions hold with
$K=\max\{1,\beta\pi^2\}$, $C_s=\varepsilon=0$, and
$\gamma=\beta\pi^2$. This observation does not impose the additional
spatial-Hessian condition on the variance-plus-cosine example with general $a$.

%% file: appendix_mf_dirac/examples_matrix_nonlinear_rewrite_v15.tex
\section{Product-measure WGF and Matrix Factorization}

\input{appendix_mf_dirac/EMN_tensor_notations}

\input{appendix_mf_dirac/product_WGF}

\input{appendix_mf_dirac/EMN_rewrite_overview_v3}

\input{appendix_mf_dirac/EMN_problem_settings_v4}

\input{appendix_mf_dirac/EMN_derivatives_ambient_v5}

\input{appendix_mf_dirac/EMN_rewrite_active_ambient_reduction_v4}

\input{appendix_mf_dirac/EMN_rewrite_antisymmetric_branch_v10}

\input{appendix_mf_dirac/EMN_rewrite_curvature_v8}

\input{appendix_mf_dirac/EMN_rewrite_descent_v8}

\input{appendix_mf_dirac/EMN_rewrite_post_escape_convexity_v9}

%% file: appendix_mf_dirac/EMN_tensor_notations.tex
\subsection{Tensor notation and Hermite tensors}
\label{subsection:tensor-preliminaries}

We collect the tensor conventions used below.  They are standard; see, for example,
\citet{Hackbusch2012} for tensor spaces.
For Hermite polynomials, Gaussian chaos, and the Wiener chaos product formula, see \citet[Chapter~1-7]  {peccati2011wiener}.

\paragraph{Tensor notation.}
For an integer $r\ge0$, write $(\R^d)^{\otimes r}$ for order-$r$ tensors, with
$(\R^d)^{\otimes0}=\R$.  For vectors $u^{(1)},\dots,u^{(r)}\in\R^d$,
\[
\bigl(u^{(1)}\otimes\cdots\otimes u^{(r)}\bigr)_{i_1\dots i_r}
:=\prod_{s=1}^r u^{(s)}_{i_s},
\qquad
x^{\otimes r}:=x\otimes\cdots\otimes x.
\]
For $A,B\in(\R^d)^{\otimes r}$, the Frobenius inner product and norm are
\[
\innerProd{A}{B}_{\HS}
:=\sum_{i_1,\dots,i_r=1}^d A_{i_1\dots i_r}B_{i_1\dots i_r},
\qquad
\norm{A}_{\HS}:=\sqrt{\innerProd{A}{A}}.
\]
If $A\in(\R^d)^{\otimes p}$, $B\in(\R^d)^{\otimes q}$, and
$0\le a\le\min(p,q)$, we write $A\contr{a}B\in(\R^d)^{\otimes(p+q-2a)}$ for contraction over
$a$ matched indices:
\[
\bigl(A\contr{a}B\bigr)_{i_1\dots i_{p-a}j_1\dots j_{q-a}}
:=
\sum_{\ell_1,\dots,\ell_a=1}^d
A_{i_1\dots i_{p-a}\ell_1\dots\ell_a}
B_{\ell_1\dots\ell_a j_1\dots j_{q-a}} .
\]
The case $a=0$ is the tensor product $A\otimes B$, and a full contraction gives the Frobenius
inner product.  When $A$ and $B$ are symmetric, the particular choice of contracted modes is
immaterial after symmetrization.

\paragraph{Symmetrization.}
For $T\in(\R^d)^{\otimes r}$, define
\[
\Sym(T):=\frac1{r!}\sum_{\pi\in\Sg_r}T^\pi,
\qquad
(T^\pi)_{i_1\dots i_r}:=T_{i_{\pi(1)}\dots i_{\pi(r)}}.
\]
The symmetric subspace is
\[
\Sym^r(\R^d):=
\{T\in(\R^d)^{\otimes r}:T^\pi=T\ \text{for all }\pi\in\Sg_r\}.
\]
The map $\Sym$ is the orthogonal projection onto $\Sym^r(\R^d)$ under the Frobenius inner product.
Since $x^{\otimes r}$ is symmetric,
\[
\innerProd{T}{x^{\otimes r}}=
\innerProd{\Sym(T)}{x^{\otimes r}}.
\]

\paragraph{Normalized Hermite basis and moment tensors.}
Let $\mathrm{He}_n$ be the probabilists' Hermite polynomial and set
\[
    \psi_n(t):=\frac{\mathrm{He}_n(t)}{\sqrt{n!}},
    \qquad n\ge0.
\]
For $\sigma\in L^2(\mathcal N(0,1);\R^{d_h})$, we use the normalized Hermite expansion
\[
    \sigma(t)=\sum_{n\ge0}a_n\psi_n(t),
    \qquad
    a_n:=\E_{g\sim\mathcal N(0,1)}[\sigma(g)\psi_n(g)]\in\R^{d_h}.
\]
When the weights are supported on the unit sphere, define
\[
    T_{i,n}:=\int w^{\otimes n}\,\dd\mu_i(w)
    \in\Sym^n(\R^{d_w}),
    \qquad
    T_{i,n}^o:=\int w^{\otimes n}\,\dd\mu_i^o(w),
    \qquad i=1,2.
\]
The unit-norm convention is important: $w^\top z$ is standard normal only when
$z\sim\mathcal N(0,I_{d_w})$ and $\norm{w}=1$.

\paragraph{Hermite tensors.}
Let $\gamma_d(x):=(2\pi)^{-d/2}\exp(-\norm{x}^2/2)$ be the standard Gaussian density on $\R^d$.
We use the normalized order-$r$ Hermite tensor
\[
\Hertensor_r(x)
:=
\frac1{\sqrt{r!}}
(-1)^r\frac{\nabla^{\otimes r}\gamma_d(x)}{\gamma_d(x)}
=
\frac1{\sqrt{r!}}
\exp(\norm{x}^2/2)(-\nabla)^{\otimes r}\exp(-\norm{x}^2/2)
\in\Sym^r(\R^d).
\]
Thus
\[
\Hertensor_0(x)=1,
\qquad
\Hertensor_1(x)=x,
\qquad
\Hertensor_2(x)=\frac{x^{\otimes2}-I_d}{\sqrt2}.
\]
For $u\ne0$,
\begin{equation}
\label{eq:directional-hermite-tensor-normalized}
    \innerProd{\Hertensor_r(x)}{u^{\otimes r}}
    =
    \norm{u}^r\psi_r\!\left(\frac{u^\top x}{\norm{u}}\right)
    =
    \frac{\norm{u}^r}{\sqrt{r!}}
    \mathrm{He}_r\!\left(\frac{u^\top x}{\norm{u}}\right).
\end{equation}
In particular, if $\norm{u}=1$, then
\[
    \innerProd{\Hertensor_r(x)}{u^{\otimes r}}=\psi_r(u^\top x).
\]
Thus, under the unit-sphere convention,
\begin{equation}
\label{eq:activation-moment-expansion}
    \int \sigma(w^\top z)\,\dd\mu_i(w)
    =
    \sum_{n\ge0}a_n\innerProd{\Hertensor_n(z)}{T_{i,n}}.
\end{equation}
If the weights are not constrained to the sphere, write $w=\rho\bar w$ with $\rho=\norm{w}$ and
$\norm{\bar w}=1$.  Then
\[
    \sigma(w^\top z)
    =
    \sum_{n\ge0}a_n(\rho)\innerProd{\Hertensor_n(z)}{\bar w^{\otimes n}},
    \qquad
    a_n(\rho):=
    \E_{g\sim\mathcal N(0,1)}[\sigma(\rho g)\psi_n(g)].
\]
Equivalently, using $w^{\otimes n}$ directly introduces the factor $\rho^{-n}a_n(\rho)$.

If $G\sim\mathcal N(0,I_d)$, then for $A\in(\R^d)^{\otimes r}$ and
$B\in(\R^d)^{\otimes s}$,
\begin{equation}
\label{eq:normalized-hermite-tensor-orthogonality}
\E\bigl[
\innerProd{A}{\Hertensor_r(G)}
\innerProd{B}{\Hertensor_s(G)}
\bigr]
=
\mathbf{1}_{\{r=s\}}
\innerProd{\Sym(A)}{\Sym(B)}.
\end{equation}
In particular, for symmetric $A\in\Sym^r(\R^d)$ and $B\in\Sym^s(\R^d)$, the right-hand side is
$\mathbf{1}_{\{r=s\}}\innerProd{A}{B}$.

\paragraph{Wiener chaos product formula.}
For symmetric tensors $A\in\Sym^n(\R^{d_w})$ and $B\in\Sym^m(\R^{d_w})$,
\begin{equation}
\label{eq:normalized-hermite-tensor-product-formula}
    \innerProd{\Hertensor_n(z)}{A}
    \innerProd{\Hertensor_m(z)}{B}
    =
    \sum_{r=0}^{\min(n,m)}
    c_{n,m,r}
    \innerProd{
        \Hertensor_{n+m-2r}(z)
    }{
        \sym{A\contr{r}B}
    },
\end{equation}
where
\begin{equation}
\label{eq:normalized-wiener-chaos-product-coefficient}
    c_{n,m,r}
    :=
    r!\binom nr\binom mr
    \sqrt{\frac{(n+m-2r)!}{n!\,m!}} .
\end{equation}
The square-root factor in \cref{eq:normalized-wiener-chaos-product-coefficient} is exactly the
conversion from the unnormalized Hermite tensors to the normalized ones.  If
$\widetilde{\Hertensor}_q:=\sqrt{q!}\,\Hertensor_q$ is used for every order $q$, then the coefficient
is $r!\binom nr\binom mr$.

Gaussian integration by parts gives the normalized tensor Stein identity
\[
\E\bigl[f(G)\Hertensor_r(G)\bigr]
=
\frac1{\sqrt{r!}}\E\bigl[\nabla^{\otimes r}f(G)\bigr],
\]
whenever the expectations are finite.  Equivalently, the degree-$r$ normalized Hermite coefficient
tensor of $f\in L^2(\gamma_d)$ is
\[
\widehat f_r:=\E\bigl[f(G)\Hertensor_r(G)\bigr]
\in\Sym^r(\R^d).
\]
A useful special case for a one-dimensional activation $\sigma$ is
\[
\E\bigl[\sigma(w^\top G)\Hertensor_r(G)\bigr]
=
\frac1{\sqrt{r!}}\E\bigl[\sigma^{(r)}(w^\top G)\bigr]w^{\otimes r},
\qquad
w\in\R^d,
\]
under the same integrability assumptions.

%% file: appendix_mf_dirac/product_WGF.tex
\subsection{Wasserstein gradient flow on product measures}
\label{app:product-measure-wgf}

Before analyzing the matrix-decomposition functional, we record a simple form of
Wasserstein gradient flow restricted to product measures.  This is useful when the
optimization variable is constrained to remain factorized across parameter blocks,
so that admissible perturbations move each marginal separately.
Related Wasserstein product-space and multi-distribution gradient-flow
structures are studied by
\citet{chen2016wassersteinProduct,conger2025coupled}.
For a two-species nonlocal PDE with symmetric interactions and an associated
energy principle, see \citet{giunta2024weaklynonlinear}.

\paragraph{Product family and product metric.}
Let $\Omega_i\subset\mathbb R^{d_i}$ and $\Omega=\prod_{i=1}^m\Omega_i$.  We consider the
product family
\begin{equation}
    \mathcal M
    \coloneqq
    \left\{
        \mu=\bigotimes_{i=1}^m\mu_i
        \;\middle|\;
        \mu_i\in\mathcal P_2(\Omega_i)
    \right\}.
\end{equation}
For $\mu=\bigotimes_i\mu_i$ and $\nu=\bigotimes_i\nu_i$, we equip $\mathcal M$ with the
product Wasserstein metric
\begin{equation}
    \label{eq:product-Wasserstein-metric}
    W_{2,\mathcal M}(\mu,\nu)^2
    \coloneqq
    \sum_{i=1}^m W_2(\mu_i,\nu_i)^2 .
\end{equation}
An admissible tangent vector on $\mathcal M$ is a coordinate-wise vector field
$v=(v_1,\ldots,v_m)$ with $v_i\in L^2(\mu_i;\mathbb R^{d_i})$, identified with
$v(x)=(v_1(x_1),\ldots,v_m(x_m))$ for $x=(x_1,\ldots,x_m)$.  The product inner product is
\begin{equation}
    \langle u,v\rangle_{\mu,\mathcal M}
    \coloneqq
    \sum_{i=1}^m \int_{\Omega_i} \langle u_i(x_i),v_i(x_i)\rangle\,\mu_i(\dd x_i)
    =
    \int_{\Omega}\sum_{i=1}^m \langle u_i(x_i),v_i(x_i)\rangle\,\mu(\dd x).
\end{equation}
The corresponding product perturbation keeps the measure in $\mathcal M$:
\begin{equation}
    (\Id+h v)_\sharp\mu
    =
    \bigotimes_{i=1}^m (\Id_i+h v_i)_\sharp\mu_i .
\end{equation}

\paragraph{Restricted Wasserstein gradient.}
Let $\overline F(\mu_1,\ldots,\mu_m)\coloneqq F(\bigotimes_i\mu_i)$.  For
$\mu=\bigotimes_i\mu_i$, write $\mu_{-i}=\bigotimes_{j\neq i}\mu_j$ and
$x_{-i}=(x_j)_{j\neq i}$.  If $F$ admits a first variation, then the partial first
variation of $\overline F$ with respect to $\mu_i$ is, up to an additive constant,
\begin{equation}
    \frac{\delta \overline F}{\delta \mu_i}(\mu_1,\ldots,\mu_m;x_i)
    =
    \int_{\Omega_{-i}}
    \frac{\delta F}{\delta\mu}(\mu;x_i,x_{-i})\,\mu_{-i}(\dd x_{-i}).
\end{equation}
Consequently, the product Wasserstein gradient is
\begin{equation}
    g_i^\mu(x_i)
    \coloneqq
    \nabla_{x_i}\frac{\delta \overline F}{\delta \mu_i}(\mu_1,\ldots,\mu_m;x_i)
    =
    \int_{\Omega_{-i}}
    \left(\nabla_\mu F(\mu;x_i,x_{-i})\right)_i\,\mu_{-i}(\dd x_{-i}),
    \qquad i=1,\ldots,m,
\end{equation}
whenever differentiation and integration can be interchanged.  Equivalently,
$g^\mu=(g_1^\mu,\ldots,g_m^\mu)$ is the orthogonal projection of the full Wasserstein
gradient $\nabla_\mu F(\mu)$ onto coordinate-wise product tangent directions.
Indeed, for every admissible $v=(v_1,\ldots,v_m)$,
\begin{equation}
    \frac{\dd}{\dd h}\Big|_{h=0}F\bigl((\Id+h v)_\sharp\mu\bigr)
    =
    \sum_{i=1}^m \int_{\Omega_i}\langle g_i^\mu(x_i),v_i(x_i)\rangle\,\mu_i(\dd x_i).
\end{equation}

\paragraph{Restricted WGF.}
The Wasserstein gradient flow of $F$ restricted to $\mathcal M$ is the product curve
$\mu_t=\bigotimes_i\mu_{i,t}$ satisfying
\begin{equation}
    \partial_t\mu_{i,t}
    + \nabla_{x_i}\cdot\left(v_{i,t}\mu_{i,t}\right)=0,
    \qquad
    v_{i,t}(x_i)=-g_i^{\mu_t}(x_i),
    \qquad i=1,\ldots,m .
\end{equation}
Equivalently, as a continuity equation on $\Omega$, this reads
\begin{equation}
    \partial_t\mu_t
    + \sum_{i=1}^m \nabla_{x_i}\cdot\left(v_{i,t}(x_i)\mu_t\right)=0.
\end{equation}
Formally, the characteristic chain rule gives the dissipation identity
\begin{equation}
    \frac{\dd}{\dd t}F(\mu_t)
    =
    -\sum_{i=1}^m \int_{\Omega_i}\|g_i^{\mu_t}(x_i)\|^2\,\mu_{i,t}(\dd x_i)
    \le 0.
\end{equation}
In the local matrix-factorization result where this identity is used, its
regularity premises are imposed on the characteristic solution and the
identity is justified for almost every time in
\cref{prop:reduced-dissipation-antisymmetric-branch,eq:mf-local-energy-dissipation}.

%% file: appendix_mf_dirac/EMN_rewrite_overview_v3.tex
\subsection{Overview of the nonlinear-moment landscape}
\label{subsec:nonlinear-landscape-overview}

\paragraph{Why matrix factorization is the testbed.}
The example in \cref{sec:matrix-factorization} should be read as a compact
matrix-factorization instance of a broader tensor-decomposition theme.  Classical
low-rank matrix models represent an object by a small number of latent factors,
while tensor decompositions replace a two-way array by multiway rank-one
components and their variants
\citep{kolda2009tensor,sidiropoulos2017tensor}.  Such tensor decompositions are not
only algebraic objects: in latent-variable learning, observable second- and
third-order moments often reduce parameter recovery to a structured symmetric
tensor decomposition problem
\citep{anandkumar2014tensor}.  This makes matrix and tensor
factorization a natural source of non-convex objectives whose geometry is simple
enough to analyze but rich enough to contain saddles, symmetries, and multiple
factor branches.

The same family of examples has also played a central role in non-convex
optimization theory.  Low-rank matrix recovery and matrix factorization provide
canonical settings where, under appropriate measurement and rank assumptions,
local minima are global or nearly global and saddle points carry negative
curvature
\citep{BhojanapalliNeyshaburSrebro2016,GeJinZheng2017}.  Orthogonal
tensor decomposition gives a parallel strict-saddle model with many symmetry
related optima and saddles
\citep{GeHuangJinYuan2015}.  These works motivate looking not only at first-order
stationarity, but also at the second-variation geometry that decides whether a
stationary point is stable or escapable.

Our population model transfers this matrix/tensor-factorization geometry to a
mean-field product-measure setting.  The two finite factor vectors are replaced
by two mean-field feature maps
\[
    h_{\mu_i}(z)
    =
    \int \sigma(w^\top z)\,\dd\mu_i(w),
    \qquad i=1,2,
\]
and the fitting term compares the matrix-valued products
\[
    h_{\mu_1}(z)h_{\mu_2}(z)^\top
    \quad\text{and}\quad
    h_{\mu^o_1}(z)h_{\mu^o_2}(z)^\top .
\]
Thus the non-convexity is not introduced by a finite-dimensional factor matrix,
but by a nonlinear product map on probability measures.  In the Gaussian-input
spherical specialization, the Hermite expansion in
\cref{lemma:EMN-expansion-F} rewrites this product map in terms of moment tensors
\(T_{i,n}=\int w^{\otimes n}\,\dd\mu_i(w)\).  The fitting loss becomes a squared
residual system for the tensor blocks \(\Delta\mathcal M_p\), while the metric
regularizer controls the Wasserstein distance to the antipodal anchor.  This is
the precise sense in which the appendix studies a nonlinear moment landscape:
the active coordinates are tensor moments, but the actual dynamics remain
Wasserstein dynamics of the underlying measures.

This perspective is close to the mean-field two-layer neural-network viewpoint,
where wide-network training is described by the evolution of a parameter
measure, often through Wasserstein gradient flow
\citep{songmei2018,chizat2018,nitanda2022convex}. It is also close to the
matrix-decomposition example in \citet{yamamoto2025hessianguided}, where a
single mean-field feature map is used through the symmetric self-product
\(h_{\mu}(z)h_{\mu}(z)^\top\).  The present product version separates the two
factor measures and studies the product WGF on
\(\mu_1\otimes\mu_2\).  This separation is useful because the two natural
branches
\[
    \mathcal M_-=\{\mu_2=(-\Id)_\#\mu_1\},
    \qquad
    \mathcal M_+=\{\mu_1=\mu_2\}
\]
become visible in the active odd moments. For
\(n\in\mathcal O\), where \(\mathcal O\) is the finite set of active odd
Hermite degrees, recall
\[
 T_{i,n}=\int w^{\otimes n}\,\dd\mu_i(w),\qquad
 S_n=\frac{T_{1,n}+T_{2,n}}2,\qquad
 A_n=\frac{T_{1,n}-T_{2,n}}2
\]
from \cref{assumption:odd-symmetric-teacher,eq:mf-symmetric-antisymmetric-active-blocks}.
Thus
membership in \(\mathcal M_-\) implies \(\mathbf S=0\), while
membership in \(\mathcal M_+\) implies \(\mathbf A=0\).
These finite moment implications are not converses; see
\eqref{eq:mf-finite-active-moment-map}.
The appendix establishes four local statements: attraction, transverse
descent, and positive second variation in two separate regions. Each statement
has its own hypotheses.

\subsubsection{Summary of the section}

This subsection summarizes the separate population-level statements behind
the diagram in \cref{sec:matrix-factorization}. The local incoming branch,
transverse descent direction, and positive-curvature regions are proved below.
A trajectory connecting these regions, or a finite-sample escape guarantee,
is not established by these geometric statements.
Two conventions apply throughout.
First, throughout this appendix the dynamics are the product Wasserstein gradient
flow, namely the WGF restricted to product measures and product tangent fields, as
recorded in \cref{app:product-measure-wgf}.  This should not be read as a claim
that the unrestricted WGF on all joint measures preserves product structure.
Second, $D^2_{W_2}F(\mu)[v,v]$ is the spherical product pushforward second
variation, defined as the full EXP-pushforward derivative. It equals the
quadratic form of the full
spherical product Hessian $\mathcal H_{\Sphere,\mu}$ in
\cref{lem:mf-full-spherical-Hessian,eq:mf-full-Hessian-second-variation}.
This includes both the measure derivative and the spatial Hessian,
also at stationary states. When the prescribed pushforward path is a product
Wasserstein geodesic, this quantity is its geodesic second derivative; general
admissible fields are not assumed to generate such a geodesic.

The model is fixed by the spherical product objective
\cref{eq:EMN-objective-squared-chi2}.  Under the cap and summability assumptions
in \cref{assumption:on-sphere}, and the finite odd-block symmetric-teacher
assumption in \cref{assumption:odd-symmetric-teacher}, the fitting term admits the
Hermite tensor expansion of \cref{lemma:EMN-expansion-F}. Our Hermite-basis analysis follows a now-standard Gaussian-input toolkit in which activation coefficients directly parameterize optimization\citep{arous2021online,abbe23a, bietti2023} and kernel phenomena~\citep{daniely2016toward,murray2023characterizing}, rather than merely serving as a formal expansion.  The metric regularizer
is
\[
    F_{\mathrm{reg}}(\mu_1,\mu_2)
    =
    \frac{\lambdaReg}{p_W}
    Z_\pm(\mu_1,\mu_2)^{p_W/2},
    \qquad
    p_W=2+\delta_W>2,
\]
where \(Z_\pm\) is the squared product distance to the antipodal Dirac pair
\(\nu_\star^\pm=\frac12\delta_{a_\star}+\frac12\delta_{-a_\star}\).  The
fixed-assignment geometry in \cref{lemma-fixed-assignment} and
\cref{definition-spherical-logarithm-exponential} is used locally on the two caps.
The choice \(p_W>2\) is deliberate: the regularizer gives positive curvature on
metric annuli with \(Z_\pm>0\), but it is Hessian-flat at the antipodal Dirac target
\[
    \mu^\dagger
    \coloneq
    \mu_\star^\pm
    =
    \nu_\star^\pm\otimes\nu_\star^\pm .
\]
Thus the quadratic saddle structure at \(\mu^\dagger\) is governed by the fitting
term.

The finite odd Hermite block gives the active coordinates used to describe the
landscape; their full definition is in
\cref{subsec:introduce-symmetric-antisymmetric-branches-and-saddle-point,eq:mf-symmetric-antisymmetric-active-blocks}.
The measure-level anti-symmetric branch
\(\submanifoldforsaddlepoint\) implies \(\mathbf S=0\) on the active odd block,
whereas the same-density branch \(\mathcal M_+\) implies \(\mathbf A=0\).
These coordinates are diagnostic coordinates for the finite active fitting term,
not a complete parametrization of the full metric-regularized objective.  The
antipodal target belongs to the active-origin intersection and is first-order
stationary, as stated in \cref{prop:intersection-zero-output-fiber}.  The same
subsection also identifies the reference zero-fit teacher point on the
same-density branch in \cref{prop:reference-fit-minimum-on-Mplus}.

\paragraph{Local attraction on the anti-symmetric branch.}
Reflection covariance of the full first variation, proved in
\cref{lem:gradient-symmetry-antisymmetric-branch}, yields branch invariance
up to cap exit in \cref{lem:invariance-antisymmetric-branch}.
The attraction proof in
\cref{subsec:antisymmetric-branch-and-dissipation} then has three steps.
Recall the assigned-pole cost \(c_\star(w)=d_\Sphere(w,\pi_\star(w))^2/2\)
from \cref{eq:mf-fixed-assignment-cost} and its outward gradient
\(q_\star(w)=\nabla_\Sphere c_\star(w)=-\LOG_w\pi_\star(w)\) from
\cref{eq:mf-cost-gradient-LOG}.
The constants \(C_\star\) and \(L_{\mathrm{fit}}\) below depend only on the
fixed activation and teacher; they are defined immediately before the pole
contraction lemma in
\cref{eq:mf-gaussian-derivative-constants,eq:mf-branch-estimate-constants}.
\begin{enumerate}
 \item \textbf{An admissible pole contraction.}
 \Cref{lem:mf-pole-contraction-estimates} uses the spherical transport
 \cref{eq:mf-pole-contraction-path} to show
 \[
 DF_{\mathrm{fit}}(\mu)[(q_\star,q_\star)]\ge-C_\star Z_\pm(\mu)^2,
 \qquad \sup_w\|g_{i,\mathrm{fit}}(\mu,w)\|
                 \le L_{\mathrm{fit}}\sqrt{Z_\pm(\mu)}.
 \]
 Here \(g_{i,\mathrm{fit}}=\nabla_\Sphere(\delta F_{\mathrm{fit}}/\delta\mu_i)\),
 as defined in \cref{eq:mf-branch-radial-notation}.
 \item \textbf{A local radial inequality.}
 Adding the regularizer derivative gives
 \[
 DF(\mu)[(q_\star,q_\star)]
 \ge\lambdaReg Z_\pm(\mu)^{p_W/2}-C_\star Z_\pm(\mu)^2.
 \]
 For \(2<p_W<4\), if
 \(C_\star Z_\pm(\mu)^{(4-p_W)/2}\le\lambdaReg/2\), then
 \cref{lem:radial-positivity-antisymmetric-branch-full-density,eq:mf-full-radial-lower-bound}
 gives
 \[
 DF(\mu)[(q_\star,q_\star)]
 \ge\frac{\lambdaReg}{2}Z_\pm(\mu)^{p_W/2}.
 \]
 The identity
 \[
 \frac{\dd}{\dd t}Z_\pm(\mu_t)=-2DF(\mu_t)[(q_\star,q_\star)]
 \]
 then controls distance while \(\mu_t\in\mathcal M_-\cap
 \mathcal X_\pm(r_{\mathrm{cap}})\).
 \item \textbf{Confinement and convergence from initial data.}
 For \(2<p_W<3\), choose \(z_\star\) by
 \cref{eq:mf-local-basin-threshold}. For every globally well-posed
 characteristic product WGF starting from
 \(\mu_0\in\mathcal M_-\cap\mathcal X_\pm(r_{\mathrm{cap}})\) with
 \(Z_\pm(\mu_0)\le z_\star\), the mean-distance barrier and the pointwise
 fitting-field estimate give cap confinement, dissipation while away from the
 target, and \(\mu_t\to\mu^\dagger\), by
 \cref{prop:reduced-dissipation-antisymmetric-branch}.
 Here globally well posed means existence for all times and uniqueness in the
 characteristic-solution class.
 The distance decay and support barrier are
 \cref{eq:mf-local-distance-decay,eq:mf-local-support-barrier}.
\end{enumerate}
The stationary-state conclusion uses the same direct smallness condition in
\cref{eq:mf-local-radial-smallness}. The distance and support comparisons do
not require a density and apply equally to atomic initial measures.

\paragraph{Spherical product pushforward second variation at anti-symmetric states.}
The positive-curvature estimate is proved in
\cref{subsec:curvature-away-from-saddle-on-antisymmetric}.  The fitting term may
have negative curvature, but on an anti-symmetric active ball
\(\|\mathbf A\|_{\mathcal H_{\mathcal O}}\le R_{\mathrm{act}}\) its negative
second variation is uniformly bounded below by
\cref{lem:fitting-curvature-lower-bound-Mminus-smoothness}.  The metric part gives
the compensating positive curvature: \cref{lem:spherical-squared-distance-Hessian-on-caps}
lower bounds the spherical Hessian of the fixed squared-distance cost on the caps,
and \cref{lem:W2-power-regularizer-Hessian-on-fixed-caps} lifts this to the
regularizer.  Combining the fitting lower bound with the metric lower bound gives
the ambient second-variation bound
\cref{prop:sqchi2-away-from-origin-Wasserstein-convexity}.  Its branchwise
specialization is \cref{cor:branchwise-positivity-Wasserstein-Mminus}.  This is the
appendix-level estimate underlying the annular label in the main-text diagram:
at fixed-assignment states with \(Z_\pm\ge\zeta_{\min}^2\), every admissible
product tangent field has positive spherical product pushforward second variation
with an explicit domination condition
\[
    \lambdaReg\kappa_{\mathrm{cap}}\zeta_{\min}^{\delta_W}
    >
    C_{\mathrm{fit}}^-(R_{\mathrm{act}}) .
\]
Here \(C_{\mathrm{fit}}^-\) is defined in
\eqref{eq:def-Cfit-Mminus-derived}. The field may point outside
\(\mathcal M_-\); global geodesic convexity of the annulus is not asserted.

\paragraph{The escape direction at the saddle.}
The escape direction at the antipodal saddle is constructed in
\cref{subsec:descent-direction-to-escape}.  The teacher imbalance condition is
formulated as a two-sided transverse moment condition in
\cref{assumption:quantitative-unweighted-two-sided-transverse-imbalance}; the
associated transverse moment measure is defined in
\cref{def:two-sided-transverse-teacher-moment-measure}.  The tangential visibility
operator at an antipodal Dirac pair is introduced in
\cref{def:tangential-visibility-matrix-dirac-pair}, and
\cref{lem:active-weighted-transverse-moment-implies-visibility} turns the
transverse teacher signal into a strictly positive escape coefficient.  The
resulting same-density perturbation is summarized in
\cref{lemma:descent-direction-antipodal-dirac-from-unweighted-imbalance}: for a
suitable spherical pushforward path starting at
\(\mu^\dagger=\nu_{\star}^\pm\otimes\nu_{\star}^\pm\) at the fixed anchor
\(a=a_\star\),
\[
    \frac{\dd}{\dd h}\bigg|_{h=0}F(\mu_h)=0,
    \qquad
    F(\mu_h)
    =
    F(\mu^\dagger)
    -
    h^2\gamma_{\mathrm{esc}}
    +
    o(h^2),
    \qquad
    \gamma_{\mathrm{esc}}>0 .
\]
At the same time, because \(Z_\pm(\mu_h)=O(h^2)\) and \(p_W=2+\delta_W\), the
metric regularizer contributes only \(O(h^{2+\delta_W})=o(h^2)\).  Hence the
quadratic escape curvature is not masked by the regularizer. This verifies an available same-density escape direction, not that every
perturbation necessarily follows it.

Our escape argument adapts the zero-output saddle mechanism of
\citet{yamamoto2025hessianguided} to the product-spherical setting by replacing
affine interpolation with a teacher-visible same-density tangent perturbation.

\paragraph{Ambient curvature near the same-density teacher.}
Finally, \cref{subsec:post-escape-convexity-on-the-same-density} proves the
local teacher-side curvature estimate. The branch \(\mathcal M_+\) is
invariant by exchange symmetry, as shown in
\cref{lem:invariance-same-density-branch}. At the same-density teacher point
\(\mu_+^o=\nu^o\otimes\nu^o\), the fitting residual vanishes and its second
variation is a nonnegative square. If
\(Z_o:=Z_\pm(\mu_+^o)>0\), the metric regularizer supplies a strictly positive
margin. The proof of \cref{prop:Mplus-teacher-local-positive-curvature} keeps
this margin positive in a sufficiently small product Wasserstein neighborhood
relative to the fixed-assignment cap class. Its key nearby-state estimate is
\[
 D_{W_2}^2F(\mu)[v,v]
 \ge
 \left[
 \lambdaReg\kappa_{\mathrm{cap}}Z_\pm(\mu)^{(p_W-2)/2}
 -C\|h_{\mu_1}h_{\mu_2}^{\top}
          -h_{\nu^o}h_{\nu^o}^{\top}\|_{L^2_z(\mathrm F)}
 \right]\|v\|_{L^2(\mu;\mathbb R^{d_w})}^2,
\]
where the proof-local \(C<\infty\) depends only on the fixed activation.
The result does not imply that the teacher is stationary for the regularized
objective or that a trajectory escaping the saddle reaches this neighborhood.

%% file: appendix_mf_dirac/EMN_problem_settings_v4.tex
\subsection{Product spherical model and Hermite coordinates}
\label{subsec:product-spherical-model}

\paragraph{Role of this subsection.}
This subsection fixes the product-measure model and the coordinate system used in the geometry argument below. We restrict the dynamics to a product ansatz
\(\mu_t=\mu_{1,t}\otimes\mu_{2,t}\). Thus the flow considered here is the product Wasserstein gradient flow on the two marginals, rather than the unrestricted WGF on all couplings. This convention is natural for two interacting mean-field networks and keeps each marginal on the sphere.

\subsubsection{Product-measure objective and product WGF}

We first define the product state space, the fitting loss, and the metric
regularizer used in the curvature example.

We consider a product mean-field neural network whose two marginals live on the sphere. The fitting term compares the matrix-valued product of two mean-field neural networks, while a metric regularizer softly anchors each marginal to an antipodal Dirac pair.

The cap constraint fixes the mass split and makes the antipodal assignment unambiguous.

Define the cap class
\begin{equation}
\label{eq:mf-spherical-caps}
0<r_{\mathrm{cap}}<\frac\pi2,
\qquad
K_\pm(r_{\mathrm{cap}})
:=
\{w\in\Sphere^{d_w-1}: d_\Sphere(w,\pm a_\star)<r_{\mathrm{cap}}\},
\end{equation}
\begin{equation}
\label{eq:mf-cap-class}
\mathcal P_\pm(r_{\mathrm{cap}})
:=
\left\{
\mu\in\mathcal P(\Sphere^{d_w-1}):
\operatorname{supp}\mu\subset K_+(r_{\mathrm{cap}})\cup K_-(r_{\mathrm{cap}}),
\ \mu(K_+)=\mu(K_-)=\frac12
\right\}.
\end{equation}

The product ansatz treats the two factor distributions as separate Wasserstein
variables. Define the product state space:
\begin{equation}
\label{eq:mf-product-cap-class}
   \mu=\mu_1\otimes\mu_2\in  \mathcal X_\pm(r_{\mathrm{cap}})
\coloneq 
\mathcal P_\pm(r_{\mathrm{cap}})\times\mathcal P_\pm(r_{\mathrm{cap}}).
\end{equation}
We will use the above product measure space to define the objective.

Let \(\sigma:\R\to\R^{d_h}\) be the activation, applied componentwise in the mean-field neural network
\begin{equation}
    h_\mu(z)
    \;=\; \int \sigma(w^\top z)\,\dd \mu(w) \in \R^{d_h}.
\end{equation}

The regularizer is built from spherical Wasserstein distance to a fixed antipodal Dirac pair.
Define $W_2$ metric on the sphere as
\[
W_{2,\Sphere}^2(\mu,\nu)
:=
\inf_{\gamma\in\Gamma(\mu,\nu)}
\int d_\Sphere(w,y)^2\,\dd\gamma(w,y).
\]
Fix a unit vector $a_\star\in\Sphere^{d_w-1}, $
and define the antipodal Dirac target
\[
    \nu_\star^\pm
    \coloneq
    \frac12\delta_{a_\star}+\frac12\delta_{-a_\star}.
\]
Let $p_W\coloneq 2+\delta_W,
    \;
    \delta_W>0.$
For product measures $\mu=\mu_1\otimes\mu_2,\;
    \mu^o=\mu_1^o\otimes\mu_2^o,$
we set
\[
    Z_{\pm,i}(\mu_i)
    \coloneq
    W_{2,\Sphere}^2(\mu_i,\nu_\star^\pm),
    \qquad
    Z_\pm(\mu_1,
    \mu_2)
    \coloneq
    Z_{\pm,1}(\mu_1)+Z_{\pm,2}(\mu_2).
\]

The objective combines the matrix-factorization fitting loss with this metric regularizer:
\begin{align}
\label{eq:EMN-objective-squared-chi2}
F(\mu_1\otimes\mu_2)
:=&
F_{\mathrm{fit}}(\mu_1,\mu_2)
+
F_{\mathrm{reg}}(\mu_1,\mu_2),\\
F_{\mathrm{fit}}
:=&
\frac12\E_z\Bigl[
\|h_{\mu_1^o}(z)h_{\mu_2^o}(z)^\top
-h_{\mu_1}(z)h_{\mu_2}(z)^\top\|_{\mathrm F}^2
\Bigr],\\
F_{\mathrm{reg}}
:=&
\frac{\lambdaReg}{p_W}Z_\pm(\mu_1,\mu_2)^{p_W/2}.
\end{align}
Because the regularizer is metric rather than density-based, it is well-defined at singular targets. In particular, at
\(\mu_\star^\pm:=\nu_\star^\pm\otimes\nu_\star^\pm\), one has \(Z_\pm=0\). Since \(p_W>2\), this regularizer is Hessian-flat at the antipodal Dirac target, so the local saddle curvature there comes from the fitting term.

\paragraph{Product WGF dynamics.}
When discussing dynamics, we use the product WGF associated with this objective. Namely, for a curve
\(\mu_t=\mu_{1,t}\otimes\mu_{2,t}\), each marginal evolves on the sphere by
\begin{equation}
    \partial_t\mu_{i,t}
    +\nabla_\Sphere\cdot(v_{i,t}\mu_{i,t})=0,
    \qquad
    v_{i,t}(w)
    =
    -\grad_\Sphere
    \frac{\delta F}{\delta \mu_i}(\mu_{1,t},\mu_{2,t})(w),
    \qquad i=1,2.
\end{equation}
Here \(\grad_\Sphere\) denotes the spherical gradient, equivalently the Euclidean gradient projected by \(I-ww^\top\).Equivalently, infinitesimal steps are spherical pushforwards
\(\mu_{i,t+h}\approx (w\mapsto \EXP_w(hv_{i,t}(w)))_\#\mu_{i,t}\).
Thus the dynamics remain on the sphere by construction. This should be viewed as WGF on the product ansatz, not as a claim that the unrestricted WGF on all measures over the product space preserves product form. In discrete or perturbed steps we use spherical pushforwards
\(w\mapsto \EXP_w(\eta v_i(w))\), so each marginal remains supported on \(\Sphere^{d_w-1}\).
For a product tangent vector \(v=(v_1,v_2)\), we use
\[
    \|v\|_{L^2(\mu)}^2
    :=
    \|v_1\|_{L^2(\mu_1)}^2+
    \|v_2\|_{L^2(\mu_2)}^2.
\]
We use product WGF, so each marginal follows its own spherical Wasserstein gradient flow.

\begin{definition}[Characteristic solution class]
\label{def:mf-characteristic-solution}
A characteristic solution consists of deterministic maps
\(\Phi_i:[0,\infty)\times\Sphere^{d_w-1}\to\Sphere^{d_w-1}\)
and gradient representatives \(g_{i,t}\), with
\[
 \mu_{i,t}=(\Phi_i(t,\cdot))_\#\mu_{i,0},
 \qquad
 \Phi_i(0,w)=w
 \quad(w\in\operatorname{supp}\mu_{i,0}).
\]
For each fixed \(t\ge0\), the map \(\Phi_i(t,\cdot)\) is Borel
and continuous on \(\operatorname{supp}\mu_{i,0}\).
The pullback \((t,w)\mapsto g_{i,t}(\Phi_i(t,w))\) is jointly Borel.
On every compact time interval, the ambient paths
\(t\mapsto\Phi_i(t,w)\) are absolutely continuous for
\(\mu_{i,0}\)-almost every \(w\), and
\[
 \frac{d}{dt}\Phi_i(t,w)=-g_{i,t}(\Phi_i(t,w))
\]
holds for \(\mu_{i,0}\)-almost every label and Lebesgue-almost every
nonnegative time.

At almost every time, \(g_{i,t}\) is a Borel representative of a tangent
\(L^2(\mu_{i,t})\) field representing the intrinsic first variation of the
full objective. Specifically, if \(F_{i,t}(\nu)\) denotes \(F\) with its
\(i\)-th marginal replaced by \(\nu\), then, for every tangent
\(v\in L^2(\mu_{i,t})\),
\[
 \left.\frac{d}{dh}\right|_{h=0}
 F_{i,t}\!\left((w\mapsto\EXP_w(hv(w)))_\#\mu_{i,t}\right)
 =\int\langle g_{i,t}(w),v(w)\rangle\,d\mu_{i,t}(w).
\]
The representatives are identified only \(\mu_{i,t}\)-almost everywhere.
Global well-posedness in this class means existence for all nonnegative times
and uniqueness of the marginal laws among all global solutions in this class
with the same initial laws.
\end{definition}

\subsubsection{Hermite and moment coordinates}

The Gaussian input allows us to rewrite the fitting term in orthogonal Hermite coordinates.
Throughout, we use the normalized Hermite basis
\begin{equation}
    \psi_n(t) \coloneq \frac{\mathrm{He}_n(t)}{\sqrt{n!}},
    \qquad
    \sigma(t) = \sum_{n\ge 0} a_n\psi_n(t),
    \qquad
    a_n \coloneq \E_{g\sim \mathcal N(0,1)}[\sigma(g)\psi_n(g)] \in \R^{d_h}.
\end{equation}

We also define the moment tensors
\begin{equation}
    T_{i,n} \coloneq \int w^{\otimes n}\,\dd\mu_i(w) \in \Sym^n(\R^{d_w}),
    \qquad
    T_{i,n}^o \coloneq \int w^{\otimes n}\,\dd\mu_i^o(w),
    \qquad i=1,2.
\end{equation}

For each $n\ge 0$, we let $\Hertensor_n(z)\in \Sym^n(\R^{d_w})$ be the multivariate Hermite tensor.

The Hermite coordinates diagonalize the Gaussian-input \(L^2\) geometry of the fitting term.

\subsubsection{Standing assumptions}
\label{subsubsection:spherical-model-assumptions}

We now list the model-specific assumptions used only for this geometric verification.

The first assumption keeps the Hermite expansion controlled and all measures in the fixed-cap regime. 
\begin{assumption}[Product spherical cap class and Hermite summability]
\label{assumption:on-sphere}
We assume:
\begin{enumerate}
    \item The activation has fourth-order weighted square-summable Hermite coefficients:
    \[
        \sum_{n\ge0} n^4\|a_n\|_2^2<\infty.
    \]
    \item For \(i=1,2\), the student and teacher marginals satisfy
    \[
        \mu_i,\mu_i^o\in\mathcal P_\pm(r_{\mathrm{cap}}).
    \]
\end{enumerate}
\end{assumption}
The cap condition places the model in a fixed-assignment regime where
spherical logarithm/exponential maps and the metric regularizer have
explicit second variations.  In the calculations below we also use the
finite odd-block assumption, stated next, to justify differentiating the
polynomial Gaussian-input outputs and their products.

The second assumption isolates a finite odd Hermite block and imposes symmetry between the two teacher factors.
\begin{assumption}[Finite odd-block activation and symmetric teacher setting]
\label{assumption:odd-symmetric-teacher}
There exists a finite set \(\mathcal O\subset 2\mathbb Z_{\ge0}+1\) such that
\[
    a_n\neq0 \quad (n\in\mathcal O),
    \qquad
    a_n=0 \quad (n\notin\mathcal O).
\]
Moreover, the teacher is symmetric across the two factors:
\[
    \mu_1^o=\mu_2^o=:\nu^o,
    \qquad
    T_n^o:=T_{1,n}^o=T_{2,n}^o
    \quad(n\in\mathcal O).
\]
\end{assumption}
The finite odd block is a simplifying active-coordinate assumption.  Odd degrees
separate the anti-symmetric and same-density branches because antipodal
reflection sends \(T_n(\nu)\) to \((-1)^nT_n(\nu)\), while the symmetric teacher
assumption isolates the product symmetry responsible for the saddle mechanism.

The escape direction at the antipodal point is generated by a transverse first-order imbalance of the teacher.  To express this imbalance without
choosing one of the two poles \(+a\) or \(-a\), we slice the sphere by the
two-sided latitude variable
\[
    r_a(w)=|a^\top w|\in[0,1].
\]
On each slice \(\{w:|a^\top w|=r\}\), we measure the average transverse
component \(\Pi_{a^\perp}w\).  The following vector-valued measure records
these sliced transverse first moments.
\begin{definition}[Two-sided transverse teacher moment measure]
\label{def:two-sided-transverse-teacher-moment-measure}
Fix \(a_\star\in\Sphere^{d_w-1}\), and let
\[
    r_{a_\star}(w)\coloneq |a_\star^\top w|,
    \qquad
    \Pi_{a_\star^\perp}\coloneq I-a_\star a_\star^\top .
\]
For every Borel set \(E\subset[0,1]\), define the \(a_\star^\perp\)-valued finite
measure
\[
    \mathfrak m_{a_\star}^o(E)
    \coloneq
    \int_{\{w:\ r_{a_\star}(w)\in E\}}
    \Pi_{{a_\star}^\perp}w
    \,\dd\nu^o(w)
    \in {a_\star}^\perp .
\]
We call \(\mathfrak m_{a_\star}^o\) the two-sided transverse teacher moment measure
around the antipodal axis \(\{\pm {a_\star}\}\).

Equivalently, if
\[
    \bar\nu_{a_\star}^o\coloneq (r_{a_\star})_\#\nu^o
\]
and
\[
    \nu^o=\int_0^1\nu_{{a_\star},r}^o\,\dd\bar\nu_{a_\star}^o(r)
\]
is a disintegration over the two-sided latitude sets
\[
    \Sigma_{a_\star}^\pm(r)
    \coloneq
    \{w\in\Sphere^{d_w-1}: |{a_\star}^\top w|=r\},
\]
then
\[
    \mathfrak m_{a_\star}^o(E)
    =
    \int_E m_{a_\star}^o(r)\,\dd\bar\nu_{a_\star}^o(r),
\]
where
\[
    m_{a_\star}^o(r)
    \coloneq
    \int_{\Sigma_{a_\star}^\pm(r)}
    \Pi_{{a_\star}^\perp}w\,\dd\nu_{{a_\star},r}^o(w)
    \in {a_\star}^\perp .
\]
\end{definition}
Thus \(\mathfrak m_{a_\star}^o(E)\) is the transverse barycenter of the teacher mass whose two-sided latitude lies in \(E\).  When \(d_w=3\), the sets
\(\{|{a_\star}^\top w|=r\}\) are pairs of latitude circles; in higher dimension they are two-sided latitude spheres.  The measure \(\mathfrak m_{a_\star}^o\) records whether, on these slices, the teacher has a coherent first-order bias in some direction
orthogonal to the antipodal axis.

The final assumption is a nondegeneracy condition for the escape direction.
It requires the sliced transverse teacher moment to have a coherent component
along one visible direction \(e_\star\): after projecting each two-sided latitude
slice onto \(a_\star^\perp\), the teacher signal does not cancel across slices
and is quantitatively positive away from the equator.  This positive transverse
margin is what drives the same-density perturbation escaping the antipodal
saddle.
\begin{assumption}[Two-sided transverse imbalance]
\label{assumption:quantitative-unweighted-two-sided-transverse-imbalance}
Let $\nu^o \coloneq \mu_1^o = \mu_2^o$.
There exist a unit vector \(e_\star\in {a_\star}^\perp\), a radius
\(r_{\mathrm{vis}}\in(0,1)\), and a constant \(\gamma_\perp>0\), such that
the scalar signed measure
\[
    \mathfrak m_{a_\star,e_\star}^o(E)
    \coloneq
    \left\langle e_\star,\mathfrak m_{a_\star}^o(E)\right\rangle
\]
is a positive finite measure on \([0,1]\).  Equivalently,
\[
    \left\langle
        e_\star,
        \int_{\{w:\ |a_\star^\top w|\in E\}}
        \Pi_{{a_\star}^\perp}w\,\dd\nu^o(w)
    \right\rangle
    \ge0
\]
for every Borel set \(E\subset[0,1]\).  Moreover, the imbalance is
quantitatively strict on the visible latitudes:
\[
    \mathfrak m_{{a_\star},e_\star}^o([r_{\mathrm{vis}},1])
    =
    \left\langle
        e_\star,
        \int_{\{w:\ |{a_\star}^\top w|\in [r_{\mathrm{vis}},1]\}}
        \Pi_{{a_\star}^\perp}w\,\dd\nu^o(w)
    \right\rangle
    \ge \gamma_\perp.
\]
\end{assumption}

\begin{remark}[Interpretation of the transverse imbalance condition]
The condition does not require pointwise asymmetry of the teacher density,
but it requires \(d_w\ge2\) and excludes global point-reflection symmetry.
Indeed, if \(\nu^o=(-\Id)_\#\nu^o\), then the integration domain is invariant
under \(w\mapsto-w\), whereas \(\Pi_{a_\star^\perp}w\) is odd, and hence
\(\mathfrak m_{a_\star}^o(E)=0\) for every Borel set \(E\subset[0,1]\),
contradicting \(\gamma_\perp>0\).
It only asks for a one-dimensional signed projection of the sliced transverse
moment to be nonnegative across all two-sided latitudes and nonzero on a visible
range.  Equivalently, the transverse slice barycenters
\[
    m_{a_\star}^o(r)
    =
    \int_{\{|{a_\star}^\top w|=r\}}
    \Pi_{{a_\star}^\perp}w\,\dd\nu_{{a_\star},r}^o(w)
\]
have a coherent component along \(e_\star\).  This prevents cancellations of
the transverse teacher signal across latitudes and supplies the direction used
later to construct a same-density escape perturbation from the antipodal saddle.
\end{remark}

\subsubsection{Fixed-assignment metric geometry}

We next record the local spherical geometry of the antipodal metric
regularizer.

All fixed-assignment formulas below are used locally, along spherical pushforward perturbations whose supports remain in the same two caps.

The cap separation makes the optimal assignment to the two Dirac atoms fixed.
\begin{lemma}[Fixed-assignment formula for the antipodal Dirac target]
\label{lemma-fixed-assignment}
\textbf{Assumptions used.} \cref{assumption:on-sphere}.

Let \(\mu\in\mathcal P_\pm(r_{\mathrm{cap}})\).  Then the optimal transport from \(\mu\) to
\[
    \nu_\star^\pm=\frac12\delta_{a_\star}+\frac12\delta_{-a_\star}
\]
is the fixed map \(\pi_\star\)                                                               
defined as
\begin{equation}
\label{eq:mf-assigned-pole}
    \pi_\star(w)
    \coloneq
    \begin{cases}
        a_\star, & w\in K_+(r_{\mathrm{cap}}),\\
        -a_\star, & w\in K_-(r_{\mathrm{cap}}),
    \end{cases}
\end{equation}
Consequently,
\begin{equation}
\label{eq:fixed-assignment-W2-square-formula}
    W_{2,\Sphere}^2(\mu,\nu_\star^\pm)
    =
    \int_{\Sphere} d_\Sphere(w,\pi_\star(w))^2\,\dd\mu(w)
    =
    2\int_{\Sphere}c_\star(w)\,\dd\mu(w),
\end{equation}
where the fixed-assignment cost is
\begin{equation}
\label{eq:mf-fixed-assignment-cost}
c_\star(w):=\frac12 d_\Sphere(w,\pi_\star(w))^2.
\end{equation}
The same identity remains valid for sufficiently small smooth pushforward perturbations of \(\mu\) whose support stays in the same two caps.
\end{lemma}

\begin{proof}
For \(w\in K_+(r_{\mathrm{cap}})\), one has
\[
    d_\Sphere(w,a_\star)<r_{\mathrm{cap}}<\frac\pi2,
    \qquad
    d_\Sphere(w,-a_\star)=\pi-d_\Sphere(w,a_\star)>\frac\pi2.
\]
Thus every point in the positive cap is strictly closer to \(a_\star\) than to \(-a_\star\).  The analogous statement holds on the negative cap.  Since the mass of each cap is \(1/2\), the fixed assignment sends exactly the correct mass to each atom.  Any alternative assignment would move some positive-cap mass to \(-a_\star\), or some negative-cap mass to \(a_\star\), increasing the transport cost.  This proves optimality and the displayed formula.
\end{proof}

The following notation describes geodesic contraction toward the assigned antipodal atom.
\begin{definition}
    \label{definition-spherical-logarithm-exponential}
The tangent space at \(y\in\Sphere^{d_w-1}\) is
\(T_y\Sphere^{d_w-1}=\{\xi\in\R^{d_w}:y^\top\xi=0\}\).
The logarithm encodes a shortest spherical arc as a tangent vector; the
exponential follows the arc specified by a tangent vector. If
\(y,w\in\Sphere^{d_w-1}\) and \(d_\Sphere(y,w)<\pi\), we write
\[
    \theta\coloneq d_\Sphere(y,w)=\arccos(y^\top w),
\]
and define
\begin{equation}
\label{eq:mf-spherical-LOG}
    \LOG_y w
    \coloneq
    \begin{cases}
    \displaystyle
    \theta\,
    \frac{w-\cos\theta\,y}{\sin\theta},
    & w\neq y,\\[1.2em]
    0, & w=y.
    \end{cases}
\end{equation}
Conversely, for \(\xi\in T_y\Sphere^{d_w-1}\),
\begin{equation}
\label{eq:mf-spherical-EXP}
    \EXP_y(\xi)
    \coloneq
    \begin{cases}
    \displaystyle
    \cos(\|\xi\|)y+
    \sin(\|\xi\|)\frac{\xi}{\|\xi\|},
    & \xi\neq0,\\[1.2em]
    y, & \xi=0.
    \end{cases}
\end{equation}
The vector \(\LOG_yw\) has length \(\theta\), and
\[
 \EXP_y(\LOG_yw)=w\quad(d_\Sphere(y,w)<\pi),\qquad
 \LOG_y(\EXP_y\xi)=\xi\quad(\|\xi\|<\pi).
\]
At the antipode \(-y\), there is no unique shortest direction, so the
logarithm is not defined there. The exponential is defined for all tangent
vectors, but the inverse identity just stated requires \(\|\xi\|<\pi\).
Since \(r_{\mathrm{cap}}<\pi/2\), for every
\(w\in K_+(r_{\mathrm{cap}})\cup K_-(r_{\mathrm{cap}})\), the vector
\[
    \xi_\star(w)
    \coloneq
    \LOG_{\pi_\star(w)}w
\]
is uniquely defined and satisfies
\[
    \|\xi_\star(w)\|
    =
    d_\Sphere(w,\pi_\star(w))
    <r_{\mathrm{cap}}.
\]
We define the fixed-assignment spherical contraction toward the antipodal Dirac target by
\[
    C_s(w)
    \coloneq
    \EXP_{\pi_\star(w)}\bigl(s\xi_\star(w)\bigr),
    \qquad
    0\le s\le1.
\]
Thus
\[
    C_1(w)=w,
    \qquad
    C_0(w)=\pi_\star(w),
    \qquad
    d_\Sphere(C_s(w),\pi_\star(w))
    =s\,d_\Sphere(w,\pi_\star(w)).
\]
The corresponding tangent vector at the current point is
\[
    q_\star(w)
    \coloneq
    \frac{\partial}{\partial s}\Big|_{s=1}C_s(w)
    \in T_w\Sphere^{d_w-1},
\]
with
\[
    \|q_\star(w)\|=d_\Sphere(w,\pi_\star(w)).
\]
The direction \(-q_\star\) is the infinitesimal contraction toward the antipodal Dirac target.
Moreover, because \(\pi_\star(-w)=-\pi_\star(w)\), one has
\[
    C_s(-w)=-C_s(w),
    \qquad
    q_\star(-w)=-q_\star(w).
\]
\end{definition}

\begin{lemma}[Fixed-assignment contraction geometry]
\label{lem:fixed-assignment-contraction-geometry}
Let \(w\in K_+(r_{\mathrm{cap}})\cup K_-(r_{\mathrm{cap}})\), set
\(y=\pi_\star(w)\), and write \(r=d_\Sphere(w,y)\).  If \(r>0\), define
\[
 u:=\frac{w-\cos r\,y}{\sin r}.
\]
Then \(y^\top u=0\), \(\|u\|_{\mathbb R^{d_w}}=1\), and
\(w=\cos r\,y+\sin r\,u\).  The contraction in
\cref{definition-spherical-logarithm-exponential} satisfies
\begin{equation}
\label{eq:mf-contraction-geodesic-derivatives}
\begin{aligned}
 C_s(w)&=\cos(sr)y+\sin(sr)u,\\
 \partial_s C_s(w)&=r[-\sin(sr)y+\cos(sr)u],
 &\|\partial_s C_s(w)\|_{\mathbb R^{d_w}}&=r,\\
 \partial_s^2 C_s(w)&=-r^2C_s(w),
 &(I-C_s(w)C_s(w)^\top)\partial_s^2 C_s(w)&=0.
\end{aligned}
\end{equation}
Moreover,
\begin{equation}
\label{eq:mf-cost-gradient-LOG}
 q_\star(w)=\nabla_\Sphere c_\star(w)=-\LOG_w\pi_\star(w),
\end{equation}
and \(q_\star(-w)=-q_\star(w)\).  At \(r=0\), both sides of
\cref{eq:mf-cost-gradient-LOG} are zero, and the identity is understood by
continuous extension.
\end{lemma}

\begin{proof}
For \(r>0\), using \(w^\top y=\cos r\) gives
\[
 y^\top u=\frac{\cos r-\cos r}{\sin r}=0,
 \qquad
 \|u\|_{\mathbb R^{d_w}}^2
 =\frac{1-2\cos^2r+\cos^2r}{\sin^2r}=1.
\]
Thus \(w=\cos r\,y+\sin r\,u\).  Substituting this decomposition into
\cref{eq:mf-spherical-LOG,eq:mf-spherical-EXP} yields the first line of
\cref{eq:mf-contraction-geodesic-derivatives}; differentiating it gives the
remaining lines.  In particular, the ambient acceleration is
\(-r^2C_s(w)\), whereas its tangent projection, the covariant acceleration
on the sphere, is zero.

On either cap, \(y=\pi_\star(w)\) is fixed and
\(r=\arccos(w^\top y)\).  Therefore
\[
 \nabla_\Sphere r
 =-\frac{(I-ww^\top)y}{\sin r}
 =\frac{\cos r\,w-y}{\sin r},
 \qquad
 \nabla_\Sphere c_\star(w)
 =\frac r{\sin r}(\cos r\,w-y).
\]
Substituting \(w=\cos r\,y+\sin r\,u\) into the last expression gives
\[
 \frac r{\sin r}\bigl[\cos r(\cos r\,y+\sin r\,u)-y\bigr]
 =r[-\sin r\,y+\cos r\,u]
 =\left.\partial_sC_s(w)\right|_{s=1}.
\]
The formula for \(\LOG_wy\) in \cref{eq:mf-spherical-LOG} gives the same
vector with the opposite sign, proving \cref{eq:mf-cost-gradient-LOG}.
Finally, \(\pi_\star(-w)=-\pi_\star(w)\) and the explicit formulas give
\(C_s(-w)=-C_s(w)\) and \(q_\star(-w)=-q_\star(w)\).  If \(r=0\), the
contraction is constant, while the smooth squared-distance cost has zero
gradient; no differentiability of the unsquared distance at the pole is used.
\end{proof}

\begin{figure}[tbp]
\centering
\includegraphics[width=\linewidth]{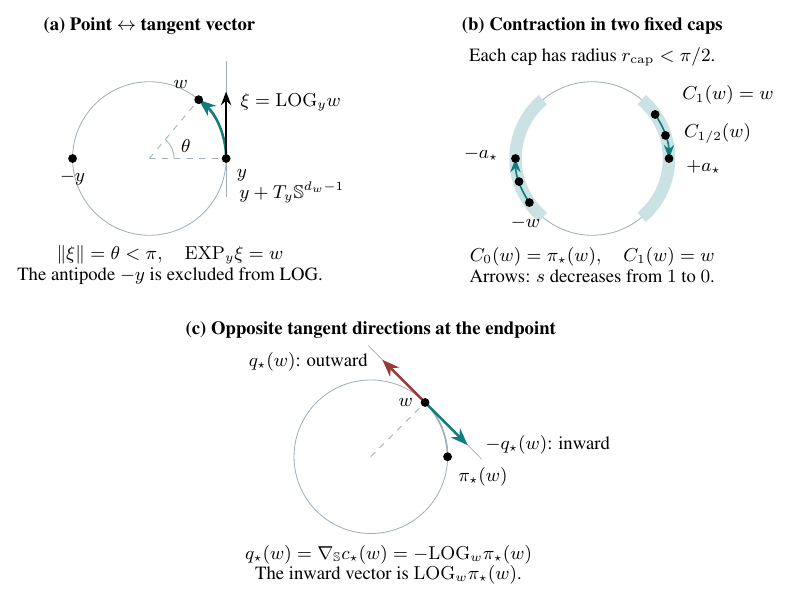}
\caption{Spherical logarithms, exponentials, and pole contraction
(\cref{definition-spherical-logarithm-exponential,lem:fixed-assignment-contraction-geometry}).
(a) A point \(w\) at distance \(\theta<\pi\) from \(y\) corresponds to
the tangent vector \(\LOG_yw\) of length \(\theta\); \(\EXP_y\) maps it
back to \(w\). Vectors are drawn with their tails at their base points.
(b) Decreasing \(s\) from \(1\) to \(0\) in \(C_s(w)\) moves each point
to its assigned pole and preserves its cap.
(c) At the endpoint \(w\), \(q_\star(w)=\nabla_\Sphere c_\star(w)
=-\LOG_w\pi_\star(w)\) points outward; its negative points toward the
pole. The cost is \(c_\star(w)=d_\Sphere(w,\pi_\star(w))^2/2\).
Each circle is a great-circle cross section of the unit sphere; these
geometric paths do not represent a WGF trajectory.}
\label{fig:mf-spherical-geometry}
\end{figure}

\subsubsection{Tensorized objective}
We now rewrite the fitting loss as a sum of squared tensor residuals.

The sphere constraint gives a uniform bound on all moment tensors.
\begin{remark}
Under \cref{assumption:on-sphere}, the sphere constraint implies
\[
    \|T_{i,n}\|_{\mathrm{F}} \le 1,
    \qquad
    \|T_{i,n}^o\|_{\mathrm{F}} \le 1,
    \qquad i=1,2,
\]
for every $n\ge 0$. Under \cref{assumption:odd-symmetric-teacher},
the Hermite-moment expansions used below are finite sums.
\end{remark}

The next lemma is the algebraic bridge from the neural-network fitting loss to finite tensor residuals.

\begin{lemma}[Hermite expansion of the metric-regularized objective]
\label{lemma:EMN-expansion-F}
\textbf{Assumptions used.} \cref{assumption:on-sphere,assumption:odd-symmetric-teacher}.

Define
\begin{align}
    \mathcal M_p
    &\coloneq
    \sum_{\substack{n,m\ge 0\\ 0\le r\le \min(n,m)\\ p=n+m-2r}}
    c_{n,m,r}(a_n a_m^\top)\otimes \sym{T_{1,n}\contr{r} T_{2,m}},
    \\
    \mathcal M_p^o
    &\coloneq
    \sum_{\substack{n,m\ge 0\\ 0\le r\le \min(n,m)\\ p=n+m-2r}}
    c_{n,m,r}(a_n a_m^\top)\otimes \sym{T_{1,n}^o\contr{r} T_{2,m}^o},
    \\
    c_{n,m,r}
    \coloneq&
    r!\binom nr\binom mr
    \sqrt{\frac{(n+m-2r)!}{n!\,m!}},
    \qquad
    \Delta\mathcal M_p\coloneq \mathcal M_p-\mathcal M_p^o.
\end{align}
These residuals have both output-matrix indices and spatial tensor indices:
$\mathcal M_p,\mathcal M_p^o,\Delta\mathcal M_p
\in\mathbb R^{d_h\times d_h}\otimes\Sym^p(\mathbb R^{d_w})$.
For such a tensor $B$, write $B_{\alpha\beta}$ for its spatial tensor
at output indices $(\alpha,\beta)$. For $U\in\Sym^p(\mathbb R^{d_w})$
and $A\in\mathbb R^{d_h\times d_h}$, the two partial pairings mean
\begin{equation}
\label{eq:mf-matrix-tensor-partial-pairings}
\begin{aligned}
 \langle U,B\rangle_{\alpha\beta}
 &:=\langle U,B_{\alpha\beta}\rangle_{\rm HS},
 &\langle U,B\rangle&\in\mathbb R^{d_h\times d_h},\\
 \langle B,A\rangle_{\rm F}
 &:=\sum_{\alpha,\beta=1}^{d_h}A_{\alpha\beta}B_{\alpha\beta},
 &\langle B,A\rangle_{\rm F}&\in\Sym^p(\mathbb R^{d_w}),\\
 \|B\|_{\rm HS}^2
 &:=\sum_{\alpha,\beta=1}^{d_h}\|B_{\alpha\beta}\|_{\rm HS}^2.
\end{aligned}
\end{equation}
Thus pairing with $\Hertensor_p(z)$ contracts the spatial indices,
whereas pairing with $a_na_m^\top$ contracts the output indices.
Then
\begin{equation}
\label{eq:EMN-expanded-squared-chi2-objective}
    F(\mu)
    =
    \frac12\sum_{p\ge 0}\|\Delta\mathcal M_p\|_{\mathrm{HS}}^2
    +
    \frac{\lambdaReg}{p_W}Z_\pm(\mu_1,\mu_2)^{p_W/2}.
\end{equation}
The first term is the Hermite expansion of the fitting loss, and the second term is the antipodal Dirac metric regularizer from \eqref{eq:EMN-objective-squared-chi2}.
\end{lemma}

\begin{remark}
    Under \cref{assumption:odd-symmetric-teacher}, this tensor residual system becomes finite dimensional. The sums defining \(\mathcal M_p\) and \(\mathcal M_p^o\) reduce to indices \(n,m\in\mathcal O\).
\end{remark}
\begin{proof}
\noindent\textbf{Step 1: multiply the finite Hermite expansions.}
Throughout the proof, all
Hermite tensors are understood with the normalization used in
\cref{eq:normalized-hermite-tensor-orthogonality,eq:normalized-hermite-tensor-product-formula}, so that different homogeneous chaoses are
orthogonal in \(L^2\) under the law of \(z\).

For \(i=1,2\), the Hermite expansion of the feature map gives
\begin{equation}
    h_{\mu_i}(z)
    =
    \sum_{n\ge0}
    a_n\,
    \big\langle \Hertensor_n(z),T_{i,n}\big\rangle,
    \qquad
    T_{i,n}
    =
    \int_{\Sphere} w^{\otimes n}\,\dd\mu_i(w).
\end{equation}
Hence the matrix-valued product can be written as
\begin{align}
    h_{\mu_1}(z)h_{\mu_2}(z)^\top
    &=
    \sum_{n,m\ge0}
    a_n a_m^\top\,
    \big\langle \Hertensor_n(z),T_{1,n}\big\rangle
    \big\langle \Hertensor_m(z),T_{2,m}\big\rangle .
\end{align}
We now apply the Wiener-chaos product formula.  For symmetric tensors
\(A\in (\mathbb R^{d_w})^{\otimes n}\) and
\(B\in(\mathbb R^{d_w})^{\otimes m}\), one has
\begin{equation}
    \big\langle \Hertensor_n(z),A\big\rangle
    \big\langle \Hertensor_m(z),B\big\rangle
    =
    \sum_{r=0}^{\min(n,m)}
    c_{n,m,r}
    \Big\langle
        \Hertensor_{n+m-2r}(z),
        \sym{A\contr{r}B}
    \Big\rangle,
\end{equation}
where
\[
    c_{n,m,r}
    =
    r!\binom nr\binom mr \sqrt\frac{(n+m-2r)!}{n!m!}.
\]
Substituting \(A=T_{1,n}\) and \(B=T_{2,m}\), we obtain
\begin{align}
    h_{\mu_1}(z)h_{\mu_2}(z)^\top
    &=
    \sum_{n,m\ge0}
    \sum_{r=0}^{\min(n,m)}
    c_{n,m,r}
    (a_n a_m^\top)
    \Big\langle
        \Hertensor_{n+m-2r}(z),
        \sym{T_{1,n}\contr{r}T_{2,m}}
    \Big\rangle .
\end{align}
Collecting all terms with the same Hermite degree $p=n+m-2r,$
we can rewrite this as
\begin{equation}
    h_{\mu_1}(z)h_{\mu_2}(z)^\top
    =
    \sum_{p\ge0}
    \big\langle \Hertensor_p(z),\mathcal M_p\big\rangle,
\end{equation}
where the coefficient tensor is precisely
\begin{equation}
    \mathcal M_p
    =
    \sum_{\substack{n,m\ge0\\0\le r\le \min(n,m)\\p=n+m-2r}}
    c_{n,m,r}
    (a_na_m^\top)\otimes
    \sym{T_{1,n}\contr{r}T_{2,m}}.
\end{equation}
The same calculation applied to the teacher measures gives
\begin{equation}
    h_{\mu_1^o}(z)h_{\mu_2^o}(z)^\top
    =
    \sum_{p\ge0}
    \big\langle \Hertensor_p(z),\mathcal M_p^o\big\rangle,
\end{equation}
with
\begin{equation}
    \mathcal M_p^o
    =
    \sum_{\substack{n,m\ge0\\0\le r\le \min(n,m)\\p=n+m-2r}}
    c_{n,m,r}
    (a_na_m^\top)\otimes
    \sym{T_{1,n}^o\contr{r}T_{2,m}^o}.
\end{equation}
Therefore
\begin{equation}
    h_{\mu_1}(z)h_{\mu_2}(z)^\top
    -
    h_{\mu_1^o}(z)h_{\mu_2^o}(z)^\top
    =
    \sum_{p\ge0}
    \big\langle \Hertensor_p(z),\Delta\mathcal M_p\big\rangle,
    \qquad
    \Delta\mathcal M_p=\mathcal M_p-\mathcal M_p^o .
\end{equation}

\par\noindent\textbf{Step 2: use orthogonality to sum the residual squares.}
We now take the \(L^2\)-norm in \(z\), with Frobenius norm in the
matrix output.  Since distinct Hermite chaoses are orthogonal, all
cross terms with \(p\neq q\) vanish:
\begin{align}
    &\E_z
    \Big[
        \big\|
        h_{\mu_1}(z)h_{\mu_2}(z)^\top
        -
        h_{\mu_1^o}(z)h_{\mu_2^o}(z)^\top
        \big\|_{\mathrm F}^2
    \Big]
    \notag\\
    &\qquad =
    \sum_{p,q\ge0}
    \E_z
    \Big[
        \Big\langle
            \big\langle \Hertensor_p(z),\Delta\mathcal M_p\big\rangle,
            \big\langle \Hertensor_q(z),\Delta\mathcal M_q\big\rangle
        \Big\rangle_{\mathrm F}
    \Big]
    \notag\\
    &\qquad =
    \sum_{p\ge0}
    \|\Delta\mathcal M_p\|_{\mathrm{HS}}^2 .
\end{align}
Consequently,
\begin{equation}
    \frac12
    \E_z
    \Big[
        \big\|
        h_{\mu_1^o}(z)h_{\mu_2^o}(z)^\top
        -
        h_{\mu_1}(z)h_{\mu_2}(z)^\top
        \big\|_{\mathrm F}^2
    \Big]
    =
    \frac12
    \sum_{p\ge0}
    \|\Delta\mathcal M_p\|_{\mathrm{HS}}^2 .
\end{equation}

It remains only to identify the regularization term.  By definition of
\(Z_\pm\), the metric regularization term in \eqref{eq:EMN-objective-squared-chi2} is exactly
\[
    \frac{\lambdaReg}{p_W}Z_\pm(\mu_1,\mu_2)^{p_W/2}.
\]
This proves \eqref{eq:EMN-expanded-squared-chi2-objective}.
\end{proof}

\paragraph{Bridge to the geometry overview.}
This tensorized form is the coordinate system used in the curvature arguments below.
Thus the model is reduced to a tensor residual system: the fitting loss is the squared Hilbert-Schmidt norm of the residuals \(\Delta\mathcal M_p\), while the metric term controls distance to the antipodal pair. In the next subsection, the active odd moments are recombined into symmetric and anti-symmetric coordinates to describe the curvature transition.

%% file: appendix_mf_dirac/EMN_derivatives_ambient_v5.tex
\subsection{Ambient first-order formulas}
\label{subsec:useful-first-order-formulas}

This subsection records the first-order ambient formulas used to analyze branch invariance, descent directions, and spherical Wasserstein gradients.

We first differentiate the tensorized fitting loss with respect to the active
moment tensors.
\begin{lemma}[Moment derivatives of the fitting part and first variation of $F$]
\label{lemma:EMN-moment-gradients}
\textbf{Assumptions used.} \cref{assumption:on-sphere,assumption:odd-symmetric-teacher}.

Let $\Delta\mathcal M_p$ be as in \cref{lemma:EMN-expansion-F}.
The output contraction in
\eqref{eq:mf-matrix-tensor-partial-pairings} is the spatial tensor
\[
 \langle\Delta\mathcal M_p,a_na_m^\top\rangle_{\rm F}
 =\sum_{\alpha,\beta=1}^{d_h}
        (a_n)_\alpha(a_m)_\beta(\Delta\mathcal M_p)_{\alpha\beta}.
\]
Define the active fitting coefficients
\begin{equation}
    G_{1,n}^{\mathrm{fit}} \coloneq \frac{\partial F_{\mathrm{fit}}}{\partial T_{1,n}},
    \qquad
    G_{2,m}^{\mathrm{fit}} \coloneq \frac{\partial F_{\mathrm{fit}}}{\partial T_{2,m}},
\end{equation}
where
\begin{equation}
    F_{\mathrm{fit}}
    \coloneq
    \frac12\sum_{p\ge0}\|\Delta\mathcal M_p\|_{\HS}^2.
\end{equation}
Then
\begin{align}
    G_{1,n}^{\mathrm{fit}}
    &=
    \sum_{m\ge 0}\sum_{r\le \min(n,m)}
    c_{n,m,r}
    \sym{\langle \Delta\mathcal M_{n+m-2r}, a_n a_m^\top\rangle_{\mathrm{F}} \contr{m-r} T_{2,m}},
    \\
    G_{2,m}^{\mathrm{fit}}
    &=
    \sum_{n\ge 0}\sum_{r\le \min(n,m)}
    c_{n,m,r}
    \sym{\langle \Delta\mathcal M_{n+m-2r}, a_n a_m^\top\rangle_{\mathrm{F}} \contr{n-r} T_{1,n}}.
\end{align}

The first variations of the fitting terms are
\begin{equation}
\label{eq:first-variation-squared-chi2-objective}
    \fv{F_\mathrm{fit}}{\mu_1}(w)
    =
    \sum_{n\ge0}\langle G_{1,n}^{\mathrm{fit}},w^{\otimes n}\rangle,
\end{equation}
\begin{equation}
    \fv{F_\mathrm{fit}}{\mu_2}(w)
    =
    \sum_{m\ge0}\langle G_{2,m}^{\mathrm{fit}},w^{\otimes m}\rangle.
\end{equation}
\end{lemma}

\begin{proof}
\noindent\textbf{Step 1: differentiate the tensor residuals.}
For the data-fit term,
\begin{equation}
    F_{\mathrm{fit}} \coloneq \frac12\sum_{p\ge 0}\|\Delta\mathcal M_p\|_{\mathrm{HS}}^2,
\end{equation}
we have
\begin{equation}
    \delta F_{\mathrm{fit}}
    = \sum_{p\ge 0}\langle \Delta\mathcal M_p, \delta\mathcal M_p\rangle_{\mathrm{HS}}.
\end{equation}
If only $T_{1,n}$ is varied, then
\begin{equation}
    \delta\mathcal M_p
    =
    \sum_{\substack{m\ge 0\\ r\le \min(n,m)\\ p=n+m-2r}}
    c_{n,m,r}(a_n a_m^\top)\otimes \sym{\delta T_{1,n}\contr{r} T_{2,m}}.
\end{equation}
Therefore
\begin{align}
    \delta F_{\mathrm{fit}}
    &=
    \sum_{\substack{p,m,r\\ p=n+m-2r}}
    c_{n,m,r}
    \Big\langle
        \Delta\mathcal M_p,
        (a_n a_m^\top)\otimes \sym{\delta T_{1,n}\contr{r} T_{2,m}}
    \Big\rangle_{\mathrm{HS}}
    \\
    &=
    \sum_{\substack{p,m,r\\ p=n+m-2r}}
    c_{n,m,r}
    \Big\langle
        \langle \Delta\mathcal M_p,a_n a_m^\top\rangle_{\mathrm F},
        \sym{\delta T_{1,n}\contr{r} T_{2,m}}
    \Big\rangle_{\mathrm{HS}}.
\end{align}
\par\noindent\textbf{Step 2: contract the output and moment indices.}
Set
\begin{equation}
    B_{p,n,m,r}
    \coloneq
    \langle \Delta\mathcal M_p,a_n a_m^\top\rangle_{\mathrm F}
    \in \Sym^{n+m-2r}(\R^{d_w}).
\end{equation}
Then, componentwise,
\begin{align}
    \Big\langle B_{p,n,m,r}, \delta T_{1,n}\contr{r} T_{2,m}\Big\rangle_{\mathrm{HS}}
    &=
    \sum_{I,J,K}
    (B_{p,n,m,r})_{I,J}
    (\delta T_{1,n})_{I,K}
    (T_{2,m})_{K,J}
    \\
    &=
    \Big\langle B_{p,n,m,r}\contr{m-r}T_{2,m},\delta T_{1,n}\Big\rangle_{\mathrm{HS}},
\end{align}
where $I$ has length $n-r$, $J$ has length $m-r$, and $K$ has length $r$. Since $\delta T_{1,n}$ is symmetric, pairing against it only depends on the symmetric part of the tensor in the first slot, so
\begin{equation}
    \Big\langle B_{p,n,m,r}, \sym{\delta T_{1,n}\contr{r} T_{2,m}}\Big\rangle_{\mathrm{HS}}
    =
    \Big\langle \sym{B_{p,n,m,r}\contr{m-r}T_{2,m}}, \delta T_{1,n}\Big\rangle_{\mathrm{HS}}.
\end{equation}
Substituting this into the previous display yields
\begin{equation}
    \delta F_{\mathrm{fit}}
    =
    \Bigg\langle
        \sum_{m\ge 0}\sum_{r\le \min(n,m)}
        c_{n,m,r}
        \sym{\langle \Delta\mathcal M_{n+m-2r}, a_n a_m^\top\rangle_{\mathrm F}\contr{m-r}T_{2,m}},
        \delta T_{1,n}
    \Bigg\rangle_{\mathrm{HS}},
\end{equation}
which is exactly the stated formula for the fit part of $G_{1,n}^{\mathrm{fit}}$. The formula for \(G_{2,m}^{\mathrm{fit}}\) follows by the same argument with the two factors exchanged.

\par\noindent\textbf{Step 3: recover the first measure variations.}
For a signed marginal perturbation $\delta\mu_i$,
$\delta T_{i,n}=\int w^{\otimes n}\,\delta\mu_i(dw)$, so
\[
 \delta F_{\mathrm{fit}}
 =\sum_n\langle G_{i,n}^{\mathrm{fit}},\delta T_{i,n}\rangle
 =\int\sum_n\langle G_{i,n}^{\mathrm{fit}},w^{\otimes n}\rangle
                    \,\delta\mu_i(dw).
\]
This proves the stated first variations, up to constants that do not
change the spherical gradients.
\end{proof}

The next corollary converts the moment derivatives into ambient and spherical
particle-wise gradients.

\begin{corollary}[Ambient and spherical gradients]
\label{corollary:EMN-first-order-gradients}
\textbf{Assumptions used.} \cref{assumption:on-sphere,assumption:odd-symmetric-teacher}.

Under the notation of \cref{lemma:EMN-moment-gradients}, the ambient gradients on the sphere are
\begin{align}
    \nabla_w\fv{F_\mathrm{fit}}{\mu_1}(w)
    &=
    \sum_{n\ge1} n\,G_{1,n}^{\mathrm{fit}}\contr{n-1}w^{\otimes(n-1)},
    \\
    \nabla_w\fv{F_\mathrm{fit}}{\mu_2}(w)
    &=
    \sum_{m\ge1} m\,G_{2,m}^{\mathrm{fit}}\contr{m-1}w^{\otimes(m-1)}.
\end{align}
Their spherical projections are
\begin{align}
\label{eq:spherical-gradient-factor1-squared-chi2}
    \nabla_{\Sphere}\fv{F_\mathrm{fit}}{\mu_1}(w)
    &=
    \sum_{n\ge1} n\Big(
        G_{1,n}^{\mathrm{fit}}\contr{n-1}w^{\otimes(n-1)}
        -\big(G_{1,n}^{\mathrm{fit}}\contr{n}w^{\otimes n}\big)w
    \Big)
    \\
\label{eq:spherical-gradient-factor2-squared-chi2}
    \nabla_{\Sphere}\fv{F_\mathrm{fit}}{\mu_2}(w)
    &=
    \sum_{m\ge1} m\Big(
        G_{2,m}^{\mathrm{fit}}\contr{m-1}w^{\otimes(m-1)}
        -\big(G_{2,m}^{\mathrm{fit}}\contr{m}w^{\otimes m}\big)w
    \Big).
\end{align}
\end{corollary}

\begin{proof}
For a symmetric tensor $G\in\Sym^n(\mathbb R^{d_w})$ and direction $u$,
the product rule and symmetry give
\[
\begin{aligned}
 D_w\langle G,w^{\otimes n}\rangle[u]
 &=\sum_{j=1}^n
 \langle G,w^{\otimes(j-1)}\otimes u\otimes w^{\otimes(n-j)}\rangle\\
 &=n\langle G\contr{n-1}w^{\otimes(n-1)},u\rangle.
\end{aligned}
\]
Moreover,
$w^\top(G\contr{n-1}w^{\otimes(n-1)})=G\contr{n}w^{\otimes n}$.
Apply these identities to each $G_{i,n}^{\mathrm{fit}}$ in
\cref{lemma:EMN-moment-gradients}, and project the ambient gradient
with $I-ww^\top$. This gives both displayed spherical formulas.
The active sums are finite by \cref{assumption:odd-symmetric-teacher}.
\end{proof}

We then compute the first and second variations of the metric regularizer on the fixed-assignment cap class.

\begin{lemma}[First and second variations of the fixed-assignment metric regularizer]
\label{lem:first-variation-flat-antipodal-potential}
\textbf{Assumptions used.} \cref{assumption:on-sphere}.

Let
\[
    F_{\mathrm{reg}}(\mu_1,\mu_2)
    =
    \frac{\lambdaReg}{p_W}Z_\pm(\mu_1,\mu_2)^{p_W/2},
    \qquad
    p_W=2+\delta_W,
\]
where
\[
    Z_\pm(\mu_1,\mu_2)
    =
    W_2^2(\mu_1,\nu_\star^\pm)+W_2^2(\mu_2,\nu_\star^\pm).
\]
On the fixed-assignment class of \cref{lemma-fixed-assignment}, define
\[
    c_\star(w)=\frac12d_\Sphere(w,\pi_\star(w))^2,
    \qquad
    D_\pm(\mu_1,
    \mu_2)
    \coloneq
    \sum_{i=1}^2\int c_\star(w)\,\dd\mu_i(w).
\]
Then
\[
    Z_\pm=2D_\pm.
\]
The first variation of \(F_{\mathrm{reg}}\), up to an irrelevant additive constant, is
\begin{equation}
\label{eq:first-variation-W2-dirac-metric-regularizer}
    \frac{\delta F_{\mathrm{reg}}}{\delta\mu_i}(w)
    =
    \lambdaReg Z_\pm^{p_W/2-1}c_\star(w),
    \qquad
    i=1,2.
\end{equation}
Consequently, the Wasserstein gradient on the sphere is
\begin{equation}
\label{eq:W2-dirac-metric-gradient}
    \nabla_\Sphere\frac{\delta F_{\mathrm{reg}}}{\delta\mu_i}(w)
    =
    \lambdaReg Z_\pm^{p_W/2-1}\nabla_\Sphere c_\star(w).
\end{equation}

The second variation formula below is the curvature contribution used in the
pre-saddle and post-escape convexity arguments.
For \(Z_\pm>0\) and a smooth spherical pushforward path generated by tangent
fields \(v=(v_1,v_2)\), as long as the fixed assignment remains valid for small \(|h|\),
\begin{align}
\label{eq:second-variation-W2-dirac-metric-regularizer}
    \frac{\dd^2}{\dd h^2}\Big|_{h=0}
    F_{\mathrm{reg}}(\mu_{1,h},\mu_{2,h})
    =&
    \lambdaReg Z_\pm^{p_W/2-1}
    \sum_{i=1}^2
    \int
        \nabla_\Sphere^2c_\star(w)[v_i(w),v_i(w)]
    \,\dd\mu_i(w)
    \\
    &+
    \lambdaReg(p_W-2)Z_\pm^{p_W/2-2}
    \left(
        \sum_{i=1}^2
        \int
            \langle\nabla_\Sphere c_\star(w),v_i(w)\rangle
        \,\dd\mu_i(w)
    \right)^2,
\end{align}
where $\mu_{i,h}=(\Phi_{i,h})_\#\mu_i,\;
    \Phi_{i,h}(w)=\EXP_w(hv_i(w))$.
The second term is nonnegative.  If \(Z_\pm=0\), i.e. at
\(\mu_1=\mu_2=\nu_\star^\pm\), then its EXP-pushforward second derivative exists and equals zero;
the exact central calculation is \eqref{eq:mf-center-EXP-regularizer}.
\end{lemma}

\begin{proof}
\noindent\textbf{Step 1: differentiate the fixed-assignment cost.}
For a signed perturbation of \(\mu_i\), the variation of \(Z_\pm\) is
\[
    \delta Z_\pm
    =
    2\int c_\star(w)\,\delta\mu_i(\dd w).
\]
Therefore
\[
    \frac{\delta F_{\mathrm{reg}}}{\delta\mu_i}(w)
    =
    \lambdaReg Z_\pm^{p_W/2-1}c_\star(w).
\]
Taking the spherical gradient in \(w\) gives
\[
    \nabla_\Sphere\frac{\delta F_{\mathrm{reg}}}{\delta\mu_i}(w)
    =
    \lambdaReg Z_\pm^{p_W/2-1}\nabla_\Sphere c_\star(w).
\]

\par\noindent\textbf{Step 2: compute the second derivative at positive distance.}
Let
\[
    \mu_{i,h}=(\Phi_{i,h})_\#\mu_i,
    \qquad
    \Phi_{i,h}(w)=\EXP_w(hv_i(w)),
\]
where $v_i(w)\in T_w\Sphere^{d_w-1}$. By \eqref{eq:mf-spherical-EXP},
\[
 \Phi_{i,h}(w)=\cos(h\|v_i(w)\|)w+
       \frac{\sin(h\|v_i(w)\|)}{\|v_i(w)\|}v_i(w).
\]
The quotient is interpreted by continuity at $v_i(w)=0$.
Differentiating the sine and cosine gives
\begin{equation}
\label{eq:mf-EXP-second-jet}
 \dot\Phi_{i,0}(w)=v_i(w),\qquad
 \ddot\Phi_{i,0}(w)=-\|v_i(w)\|^2w,\qquad
 \nabla_h\dot\Phi_{i,h}(w)|_{h=0}
 =(I-ww^\top)\ddot\Phi_{i,0}(w)=0.
\end{equation}
Thus the covariant acceleration, namely the tangential part of the
ambient acceleration, vanishes.

Define
\[
    D_\pm(h)
    \coloneq
    \sum_{i=1}^2
    \int c_\star(\Phi_{i,h}(w))\,\dd\mu_i(w),
    \qquad
    Z_\pm(h)=2D_\pm(h).
\]
By differentiating under the integral,
\[
    D_\pm'(0)
    =
    \sum_{i=1}^2
    \int
        \langle \nabla_\Sphere c_\star(w),v_i(w)\rangle
    \,\dd\mu_i(w).
\]
Moreover, since the paths \(h\mapsto \Phi_{i,h}(w)\) are geodesics,
\[
    \frac{\dd^2}{\dd h^2}\Big|_{h=0}
    c_\star(\Phi_{i,h}(w))
    =
    \nabla_\Sphere^2 c_\star(w)[v_i(w),v_i(w)].
\]
Thus
\[
    D_\pm''(0)
    =
    \sum_{i=1}^2
    \int
        \nabla_\Sphere^2 c_\star(w)[v_i(w),v_i(w)]
    \,\dd\mu_i(w).
\]
Consequently, $Z_\pm'(0)=2D_\pm'(0),\;
    Z_\pm''(0)=2D_\pm''(0).$
For \(Z_\pm>0\), differentiate
\[
    h\longmapsto
    F_{\mathrm{reg}}(\mu_{1,h},\mu_{2,h})
    =
    \frac{\lambdaReg}{p_W}Z_\pm(h)^{p_W/2}.
\]
The second derivative is
\[
\begin{aligned}
    \frac{\dd^2}{\dd h^2}\Big|_{h=0}
    F_{\mathrm{reg}}(\mu_{1,h},\mu_{2,h})
    &=
    \frac{\lambdaReg}{2}
    Z_\pm^{p_W/2-1}Z_\pm''(0)
    \\
    &\quad+
    \frac{\lambdaReg}{2}
    \left(\frac{p_W}{2}-1\right)
    Z_\pm^{p_W/2-2}
    \bigl(Z_\pm'(0)\bigr)^2.
\end{aligned}
\]
Substituting \(Z_\pm'(0)=2D_\pm'(0)\) and \(Z_\pm''(0)=2D_\pm''(0)\), we obtain
\[
\begin{aligned}
    \frac{\dd^2}{\dd h^2}\Big|_{h=0}
    F_{\mathrm{reg}}(\mu_{1,h},\mu_{2,h})
    &=
    \lambdaReg Z_\pm^{p_W/2-1}
    D_\pm''(0)
    \\
    &\quad+
    \lambdaReg(p_W-2)Z_\pm^{p_W/2-2}
    \bigl(D_\pm'(0)\bigr)^2.
\end{aligned}
\]
Using the formulas for \(D_\pm'(0)\) and \(D_\pm''(0)\), this is exactly
\[
\begin{aligned}
    \frac{\dd^2}{\dd h^2}\Big|_{h=0}
    F_{\mathrm{reg}}(\mu_{1,h},\mu_{2,h})
    =&
    \lambdaReg Z_\pm^{p_W/2-1}
    \sum_{i=1}^2
    \int
        \nabla_\Sphere^2c_\star(w)[v_i(w),v_i(w)]
    \,\dd\mu_i(w)
    \\
    &+
    \lambdaReg(p_W-2)Z_\pm^{p_W/2-2}
    \left(
        \sum_{i=1}^2
        \int
            \langle\nabla_\Sphere c_\star(w),v_i(w)\rangle
        \,\dd\mu_i(w)
    \right)^2.
\end{aligned}
\]
The second term is nonnegative because \(p_W=2+\delta_W>2\).

\par\noindent\textbf{Step 3: differentiate directly at the antipodal target.}
If $Z_\pm=0$, then $\mu_0=\nu_\star^\pm\otimes\nu_\star^\pm$.
Each starting atom $y$ is its own assigned pole. For sufficiently small
$|h|$, \cref{eq:mf-spherical-EXP,eq:fixed-assignment-W2-square-formula}
give, including zero velocities,
\[
 d_\Sphere(\EXP_y(hv_i(y)),y)^2=h^2\|v_i(y)\|^2,
 \qquad
 Z_\pm(\mu_h)=h^2\sum_i\int\|v_i(y)\|^2\,d\mu_{0,i}(y).
\]
Thus, with the product norm
$\|v\|_{L^2(\mu_0)}^2=\sum_i\int\|v_i\|^2\,d\mu_{0,i}$,
\begin{equation}
\label{eq:mf-center-EXP-regularizer}
 F_{\mathrm{reg}}(\mu_h)
 =\frac{\lambdaReg}{p_W}|h|^{p_W}\|v\|_{L^2(\mu_0)}^{p_W}.
\end{equation}
For $h\ne0$, direct differentiation yields
\[
\begin{aligned}
 \frac d{dh}F_{\mathrm{reg}}(\mu_h)
 &=\lambdaReg\operatorname{sgn}(h)|h|^{p_W-1}\|v\|_{L^2(\mu_0)}^{p_W},\\
 \frac{d^2}{dh^2}F_{\mathrm{reg}}(\mu_h)
 &=\lambdaReg(p_W-1)|h|^{p_W-2}\|v\|_{L^2(\mu_0)}^{p_W}.
\end{aligned}
\]
At $h=0$, the value difference quotient tends to zero. Dividing the
first derivative at $h\ne0$ by $h$ gives
$\lambdaReg|h|^{p_W-2}\|v\|_{L^2(\mu_0)}^{p_W}\to0$.
Therefore the second derivative exists at zero and equals zero, since
$p_W>2$; the second derivative displayed above is also continuous there.
\end{proof}

\begin{lemma}[Full spherical product Hessian and spherical product pushforward second variation]
\label{lem:mf-full-spherical-Hessian}
Assume \cref{assumption:on-sphere,assumption:odd-symmetric-teacher}.
Let $\mu=\mu_1\otimes\mu_2\in\mathcal X_\pm(r_{\mathrm{cap}})$, and let
$v=(v_1,v_2)$ be a smooth spherical product field. Use the product norm
$\|v\|_{L^2(\mu)}^2=\sum_i\int\|v_i(w)\|^2\,d\mu_i(w)$.
For sufficiently small $|h|$, require the paths
\[
 \Phi_{i,h}(w)=\EXP_w(hv_i(w)),\qquad
 \mu_h=\bigotimes_{i=1}^2(\Phi_{i,h})_\#\mu_i
\]
to remain in the fixed-assignment caps of \eqref{eq:mf-product-cap-class}.
The partial measure derivatives of $F_{\mathrm{fit}}$ are those of
its finite polynomial in the moments. In terms of
$G_{i,n}^{\mathrm{fit}}=\partial F_{\mathrm{fit}}/\partial T_{i,n}$
from \cref{lemma:EMN-moment-gradients}, this means
\begin{equation}
\label{eq:mf-fitting-second-measure-variation}
 \frac{\delta^2F_{\mathrm{fit}}}{\delta\mu_i\delta\mu_j}(\mu;w,u)
 :=\sum_{n,m\in\mathcal O}
 \left\langle
 D_{T_{j,m}}G_{i,n}^{\mathrm{fit}}[u^{\otimes m}],
 w^{\otimes n}\right\rangle_{\rm HS}.
\end{equation}
For $Z_\pm(\mu)>0$, define the full spherical product Hessian by
\begin{equation}
\label{eq:mf-full-spherical-product-Hessian}
\begin{aligned}
 (\mathcal H_{\Sphere,\mu}v)_i(w)
 &=\nabla_\Sphere^2\frac{\delta F}{\delta\mu_i}(\mu;w)v_i(w)\\
 &\quad+\sum_{j=1}^2\int
 \nabla_{\Sphere,w}\nabla_{\Sphere,u}
 \frac{\delta^2F}{\delta\mu_i\delta\mu_j}(\mu;w,u)
                  v_j(u)\,d\mu_j(u).
\end{aligned}
\end{equation}
The first term differentiates the location with the measures fixed.
The integral term differentiates the measures; its mixed derivative maps
$T_u\Sphere$ to $T_w\Sphere$. At $Z_\pm(\mu)=0$, the direct calculation
\cref{eq:mf-center-EXP-regularizer} shows that the regularizer has zero
pushforward second variation. Hence the full operator there is
$\mathcal H_{\Sphere,\mu}=\mathcal H_{\Sphere,\mu}^{\mathrm{fit}}$, obtained
from \cref{eq:mf-full-spherical-product-Hessian} by using
$F_{\mathrm{fit}}$ in both terms.
Then the spherical product pushforward second variation exists and satisfies
\begin{equation}
\label{eq:mf-full-Hessian-second-variation}
\begin{aligned}
 D^2_{W_2}F(\mu)[v,v]
 &:=\left.\frac{d^2}{dh^2}\right|_0 F(\mu_h)\\
 &=\sum_{i=1}^2\int
       \langle v_i(w),(\mathcal H_{\Sphere,\mu}v)_i(w)\rangle\,d\mu_i(w)
 =\langle v,\mathcal H_{\Sphere,\mu}v\rangle_{L^2(\mu)}.
\end{aligned}
\end{equation}
This notation refers to the stated product pushforward; it does not
require every such path to be an optimal Wasserstein geodesic. When the
displayed pushforward path is a product Wasserstein geodesic, the same quantity
is its usual geodesic second derivative. Thus the estimates below apply, in
particular, along every admissible path of that form.
\end{lemma}
\begin{proof}
\noindent\textbf{Step 1: differentiate the fitting term in both its arguments.}
The explicit EXP calculation \eqref{eq:mf-EXP-second-jet} gives
$\dot\Phi_{i,0}=v_i$ and zero covariant acceleration.
The product first-variation formula in \cref{app:product-measure-wgf} gives
\[
 \frac d{dh}F_{\mathrm{fit}}(\mu_h)
 =\sum_i\int\left\langle
   \nabla_\Sphere\frac{\delta F_{\mathrm{fit}}}{\delta\mu_i}
                   (\mu_h;\Phi_{i,h}(w)),\dot\Phi_{i,h}(w)
                 \right\rangle\,d\mu_i(w).
\]
The moment derivative along the path is the finite product rule
\[
 \left.\frac d{dh}\right|_0T_{j,m}(\mu_h)
 =\int\sum_{\ell=1}^m
 u^{\otimes(\ell-1)}\otimes v_j(u)\otimes u^{\otimes(m-\ell)}
 \,d\mu_j(u).
\]
Equivalently, the integrand is
\(D_u(u^{\otimes m})[v_j(u)]\).  Hence the coefficient derivative is
the finite-dimensional chain rule
\begin{equation}
\label{eq:mf-moment-kernel-chain-rule}
\begin{aligned}
 \left.\frac d{dh}\right|_0G_{i,n}^{\mathrm{fit}}(\mu_h)
 &=\sum_{j=1}^2\sum_{m\in\mathcal O}
 D_{T_{j,m}}G_{i,n}^{\mathrm{fit}}
       \left[\left.\frac d{dh}\right|_0T_{j,m}(\mu_h)\right]\\
 &=\sum_{j=1}^2\sum_{m\in\mathcal O}\int
 D_{T_{j,m}}G_{i,n}^{\mathrm{fit}}
       \bigl[D_u(u^{\otimes m})[v_j(u)]\bigr],d\mu_j(u).
\end{aligned}
\end{equation}
Insert this into the derivative of
$\sum_n\langle G_{i,n}^{\mathrm{fit}},w^{\otimes n}\rangle$ from
\cref{eq:first-variation-squared-chi2-objective}.  Pairing
\cref{eq:mf-moment-kernel-chain-rule} with \(w^{\otimes n}\), summing over
\(n\), and differentiating \(u^{\otimes m}\) in the direction
\(v_j(u)\) is exactly the \(u\)-gradient of the kernel in
\cref{eq:mf-fitting-second-measure-variation}.  Thus, for fixed \(w\),
\[
 \left.\frac d{dh}\right|_0
       \frac{\delta F_{\mathrm{fit}}}{\delta\mu_i}(\mu_h;w)
 =\sum_j\int\left\langle
       \nabla_{\Sphere,u}
       \frac{\delta^2F_{\mathrm{fit}}}{\delta\mu_i\delta\mu_j}(\mu;w,u),
       v_j(u)\right\rangle\,d\mu_j(u).
\]
All sums are finite by \cref{assumption:odd-symmetric-teacher}, so the
sum--integral interchange in \cref{eq:mf-moment-kernel-chain-rule} requires
no limiting argument.  This is the finite polynomial chain rule for the
moments in \cref{lemma:EMN-expansion-F,lemma:EMN-moment-gradients}.
Taking its spherical gradient in $w$ gives the integral term of
\eqref{eq:mf-full-spherical-product-Hessian} for $F_{\mathrm{fit}}$.
Differentiating the evaluation point gives its spatial-Hessian term.
The remaining derivative of $\dot\Phi_{i,h}$ contributes
$\langle\nabla_\Sphere(\delta F_{\mathrm{fit}}/\delta\mu_i),
        \nabla_h\dot\Phi_{i,h}\rangle=0$ at $h=0$.
Thus differentiating the first display proves
\eqref{eq:mf-full-Hessian-second-variation} for $F_{\mathrm{fit}}$.
All differentiations under the integrals are valid for atomic as well
as diffuse measures: the Hermite sums are finite and the supports are
compact subsets of the smooth cap region.

\par\noindent\textbf{Step 2: identify the regularizer contribution.}
Recall
$c_\star(w)=d_\Sphere(w,\pi_\star(w))^2/2$ from
\eqref{eq:mf-fixed-assignment-cost}, and
$Z_\pm=2\sum_i\int c_\star\,d\mu_i$ from
\eqref{eq:fixed-assignment-W2-square-formula}.
For $Z_\pm>0$, differentiating
\eqref{eq:first-variation-W2-dirac-metric-regularizer} gives
\[
\begin{aligned}
 \frac{\delta F_{\mathrm{reg}}}{\delta\mu_i}(\mu;w)
 &=\lambdaReg Z_\pm^{(p_W-2)/2}c_\star(w),\\
 \frac{\delta^2F_{\mathrm{reg}}}{\delta\mu_i\delta\mu_j}(\mu;w,u)
 &=\lambdaReg(p_W-2)Z_\pm^{(p_W-4)/2}c_\star(w)c_\star(u).
\end{aligned}
\]
The factor $p_W-2$ is $2\cdot(p_W-2)/2$, since a variation of
$\mu_j$ changes $Z_\pm$ by $2\int c_\star\,d(\delta\mu_j)$.
Their spatial derivatives in
\eqref{eq:mf-full-spherical-product-Hessian}, paired with $v$, give
\[
\begin{aligned}
 &\lambdaReg Z_\pm^{(p_W-2)/2}
       \sum_i\int\nabla_\Sphere^2c_\star(w)[v_i(w),v_i(w)]\,d\mu_i(w)\\
 &\quad+\lambdaReg(p_W-2)Z_\pm^{(p_W-4)/2}
       \left(\sum_i\int\langle\nabla_\Sphere c_\star(w),v_i(w)\rangle
                     \,d\mu_i(w)\right)^2.
\end{aligned}
\]
This equals the directly computed second derivative
\eqref{eq:second-variation-W2-dirac-metric-regularizer}.
For $Z_\pm=0$, use \eqref{eq:mf-center-EXP-regularizer} and its
difference-quotient calculation instead: the second derivative is zero.
No second measure variation of $Z_\pm^{p_W/2}$ at zero is used.
Adding this contribution to Step 1 proves the identity.
\end{proof}

\begin{remark}[Relation to the full Hessian and the auxiliary comparison]
The identity \eqref{eq:mf-full-Hessian-second-variation} is the spherical
product counterpart of the full Hessian in \cref{subsec:population-pwgf}.
Stationarity alone does not remove its spatial term.
The normalized retractions used below have the same first and second
derivatives as EXP at zero by \eqref{eq:mf-retraction-second-jet};
the finite-moment fitting chain rule therefore gives the same second
variation. For the regularizer at the center, the corresponding direct
calculation is \eqref{eq:mf-zero-distance-retraction}.
This suggests a possible finite-sample curvature comparison of the kind
studied in \cref{subsection:concentration-kappa}.
Such a transfer theorem for this spherical product model remains open
in the present work: the hypotheses of that auxiliary result are not
verified here. The central calculation is a second-variation identity,
not an application of the general $C^3$ theorem at the MF target.
\end{remark}

%% file: appendix_mf_dirac/EMN_rewrite_active_ambient_reduction_v4.tex
\subsection{Symmetric and Antisymmetric Sub-Manifolds}
\label{subsec:introduce-symmetric-antisymmetric-branches-and-saddle-point}
This subsection introduces the symmetric and antisymmetric coordinates that separate the teacher branch from the antipodal saddle branch.

We first collect the active odd moments into a single finite-dimensional tensor space.

Define the active odd raw-moment space
\begin{equation}
    \mathcal H_{\mathcal O}
    \coloneq
    \mathcal H_{\mathcal O}^{\mathrm{raw}}
    =
    \bigoplus_{n\in\mathcal O}\Sym^n(\R^{d_w}),
    \qquad
    \|\mathbf X\|_{\mathcal H_{\mathcal O}}^2
    \coloneq
    \|\mathbf X\|_{\mathrm{raw}}^2
    =
    \sum_{n\in\mathcal O}\|X_n\|_{\HS}^2.
\end{equation}

The weighted norm records the size of active moments as seen through the
activation coefficients:
\begin{equation}
\label{eq:def-weighted-active-norm}
    \|\mathbf X\|_{a}^2
    \coloneq
    \sum_{n\in\mathcal O}\|a_n\|_2^2\|X_n\|_{\HS}^2.
\end{equation}

We now pass from the two raw moment sequences to symmetric and antisymmetric coordinates.

\begin{definition}
Introduce the two active student blocks and the teacher block
\begin{equation}
    \mathbf T_i
    \coloneq
    (T_{i,n})_{n\in \mathcal O}
    \in \mathcal H_{\mathcal O},\quad i=1,2,
    \quad 
    \mathbf T^o
    \coloneq
    (T_n^o)_{n\in \mathcal O}
    \in \mathcal H_{\mathcal O},
\end{equation}
where $T_{i,n}=\int w^{\otimes n}\,d\mu_i(w)$ and
$T_n^o=\int w^{\otimes n}\,d\nu^o(w)$ as in
\cref{assumption:odd-symmetric-teacher}. Recall the symmetric and
antisymmetric blocks
\begin{equation}
\label{eq:mf-symmetric-antisymmetric-active-blocks}
    \mathbf S
    \coloneq
    (S_n)_{n\in \mathcal O},
    \qquad
    \mathbf A
    \coloneq
    (A_n)_{n\in \mathcal O}.
\end{equation}
with $S_n=(T_{1,n}+T_{2,n})/2$ and
$A_n=(T_{1,n}-T_{2,n})/2$. Thus
\begin{equation}
    T_{1,n}=S_n+A_n,
    \qquad
    T_{2,n}=S_n-A_n,
    \qquad
    n\in \mathcal O.
\end{equation}

The finite active moment map used in the branch geometry is
\begin{equation}
\label{eq:mf-finite-active-moment-map}
    \Theta(\mu_1,\mu_2)
    \coloneq
    \bigl(\mathbf S(\mu_1,\mu_2),\mathbf A(\mu_1,\mu_2)\bigr).
\end{equation}
Only the moments with $n\in\mathcal O$ enter $\Theta$; it is not asserted
to be an injective coordinate chart of probability measures.

For each $p\ge 0$, we define the active ambient moment polynomial. In these coordinates, the active fitting residual becomes an ambient polynomial
in \((\mathbf S,\mathbf A)\).
\begin{equation}
\label{eq:ambient-active-reduced-M}
    \widetilde{\mathcal M}_p(\mathbf S,\mathbf A)
    \coloneq
    \sum_{\substack{n,m\in \mathcal O\\ 0\le r\le \min(n,m)\\ p=n+m-2r}}
    c_{n,m,r}(a_n a_m^\top)\otimes
    \sym{(S_n+A_n)\contr{r}(S_m-A_m)}.
\end{equation}
The corresponding ambient active fitting functional is
\begin{equation}
\label{eq:ambient-active-fit-functional}
    \widetilde{\mathcal J}_{\mathrm{fit}}(\mathbf S,\mathbf A)
    \coloneq
    \frac12\sum_{p\ge 0}
    \big\|
        \widetilde{\mathcal M}_p(\mathbf S,\mathbf A)-\mathcal M_p^o
    \big\|_{\HS}^2.
\end{equation}
\end{definition}

Membership in either branch implies that its finite active moments lie
on the corresponding coordinate axis.
\begin{definition}
For any measurable set $A$, let $-A \coloneq \{-w \mid w\in A\}$.
Consider the following two branches:
\begin{enumerate}
\item \textbf{Antisymmetric branch}: The antisymmetric branch is the candidate path toward the antipodal saddle.
\begin{equation}
    \submanifoldforsaddlepoint
    \coloneq
    \left\{
        \mu_1\otimes \mu_2
        \mid
        \mu_1(-A)=\mu_2(A)
        \text{ for every measurable }A
    \right\},
\end{equation}
which is equivalent to
\begin{equation}
    T_{1,n}=(-1)^nT_{2,n}
    \qquad
    (n\ge 0)
\end{equation}
and this implies $\mathbf S=0$.
\item \textbf{Symmetric branch}: The symmetric branch contains the same-density teacher configuration.
\begin{equation}
    \mathcal M_+
    \coloneq
    \left\{
        \mu_1\otimes \mu_2
        \mid
        \mu_1=\mu_2
    \right\},
\end{equation}
which is equivalent to
\begin{equation}
    T_{1,n}=T_{2,n}
    \qquad
    (n\ge 0)
\end{equation}
and this implies $\mathbf A=0$.
\end{enumerate}

We write the fitting loss restricted to each branch as a one-coordinate functional.
\begin{equation}
\label{eq:branch-reduced-functional-Mplus-block-v3}
    \mathcal J_+(\mathbf S)
    \coloneq
    \widetilde{\mathcal J}_{\mathrm{fit}}(\mathbf S,0),
    \quad 
    \mathcal J_-(\mathbf A)
    \coloneq
    \widetilde{\mathcal J}_{\mathrm{fit}}(0,\mathbf A).
\end{equation}
\end{definition}

The equivalences above use all raw orders $n\ge0$, not only
$n\in\mathcal O$. Here is the justification on the compact sphere.
Equality of all tensor moments gives equality of every monomial integral
and hence every polynomial integral. Restricted polynomials form an
algebra containing constants and separating points of the sphere.
By Stone--Weierstrass, for each continuous $f$ and $\epsilon>0$
there is a polynomial $P$ with $\sup_\Sphere|f-P|<\epsilon$. Thus,
for two measures with the same raw moments,
\[
 \left|\int f\,d\mu_1-\int f\,d\mu_2\right|
 \le\int|f-P|\,d\mu_1+\int|f-P|\,d\mu_2<2\epsilon.
\]
Letting $\epsilon\downarrow0$ gives equality of the measures.
For reflection $\mathsf R(w)=-w$, the identity
\[
 T_n(\mathsf R_\#\mu_1)
 =\int(-w)^{\otimes n}\,d\mu_1(w)=(-1)^nT_n(\mu_1)
\]
gives the antisymmetric equivalence by the same argument.
Consequently $\mu\in\mathcal M_-$ implies $\mathbf S=0$ and
$\mu\in\mathcal M_+$ implies $\mathbf A=0$, but neither converse is
inferred from the finite active blocks.

The first proposition identifies the distinguished active-origin point where the two branches meet.
\begin{proposition}[Active origin fiber and the distinguished Dirac saddle]
\label{prop:intersection-zero-output-fiber}
\textbf{Assumptions used.} \cref{assumption:on-sphere,assumption:odd-symmetric-teacher}.

Define the active-coordinate zero fiber
\begin{equation}
    \mathcal C_0
    \coloneq
    \left\{
        \mu_1\otimes \mu_2
        \mid
        \Theta(\mu_1,\mu_2)=(0,0)
    \right\}.
\end{equation}
Then
\begin{equation}
    \mu_{\dag}
    \coloneq
    \nu_\star^\pm \otimes \nu_\star^\pm
    \in
    \mathcal M_+\cap\submanifoldforsaddlepoint\cap\mathcal C_0.
\end{equation}
In addition, if
\begin{equation}
    \mu_1=\mu_2=\nu_\star^\pm,
\end{equation}
then the Wasserstein gradients vanish. In other words,
$\mu_{\dag}=\nu_\star^\pm\otimes\nu_\star^\pm$ is a first-order stationary point.
\end{proposition}

\begin{proof}
The proof only uses antipodal symmetry, oddness of the active moments, and the flatness of the metric regularizer at the antipodal target.

$\mu_\dag$ is antipodally symmetric, equal in both factors, and has zero odd moments; hence it belongs to the displayed intersection.

If $\mu\in\mathcal C_0$, then $T_{1,n}=T_{2,n}=0$ for all $n\in\mathcal O$. Therefore, on $\mathcal C_0$, the fitting first variation vanishes because of the definition of $G_{i,n}^{\mathrm{fit}}$.

The full spherical gradient is only the regularization contribution
\begin{equation}
    \nabla_{\Sphere}\fv{F}{\mu_i}
    =
    \lambdaReg Z_{\pm}^{p_W/2-1} \nabla_{\Sphere} c_\star (w),
    \qquad i=1,2.
\end{equation}
If $\mu_1$ and $\mu_2$ are $\nu_\star^{\pm}$, then $Z_{\pm}=0$ and the spherical gradient vanishes.
\end{proof}

The next proposition fixes the opposite endpoint of the cartoon: the exact-fit teacher lies on the same-density branch.
\begin{proposition}[The reference exact-fit point lies on the same-density branch]
\label{prop:reference-fit-minimum-on-Mplus}
\textbf{Assumptions used.} \cref{assumption:on-sphere,assumption:odd-symmetric-teacher}.

The fitting part satisfies
\begin{equation}
    \widetilde{\mathcal J}_{\mathrm{fit}}(\mathbf T^o,0)=0
    =
    \inf_{(\mathbf S,\mathbf A)\in \mathcal H_{\mathcal O}\times \mathcal H_{\mathcal O}}
    \widetilde{\mathcal J}_{\mathrm{fit}}(\mathbf S,\mathbf A).
\end{equation}
Moreover, by the symmetric-teacher assumption, the reference teacher
product measure belongs to the same-density branch:
\begin{equation}
    \mu^o=\mu_1^o\otimes\mu_2^o\in \mathcal M_+,
    \qquad
    \mathbf S(\mu^o)=\mathbf T^o,
    \qquad
    \mathbf A(\mu^o)=0.
\end{equation}
In particular, the distinguished zero-fit global minimum of the data term lies on $\mathcal M_+$.
\end{proposition}

\begin{proof}
By \cref{assumption:odd-symmetric-teacher}, the teacher is symmetric across the two factors, so $\mu_1^o=\mu_2^o$ and therefore $\mu^o\in \mathcal M_+$. In the active coordinates this is exactly the identity $\mathbf S(\mu^o)=\mathbf T^o$ and $\mathbf A(\mu^o)=0$.

Next, by the definition of $\mathcal M_p^o$ and of the ambient moment map,
\begin{equation}
    \widetilde{\mathcal M}_p(\mathbf T^o,0)=\mathcal M_p^o
    \qquad
    \forall p\ge 0.
\end{equation}
Hence $\widetilde{\mathcal J}_{\mathrm{fit}}(\mathbf T^o,0)=0$. Since $\widetilde{\mathcal J}_{\mathrm{fit}}$ is a sum of squared Hilbert norms, it is nonnegative everywhere.
\end{proof}

%% file: appendix_mf_dirac/EMN_rewrite_antisymmetric_branch_v10.tex
\subsection{The anti-symmetric branch and descent towards the saddle}
\label{subsec:antisymmetric-branch-and-dissipation}

This subsection proves invariance of the anti-symmetric branch and, under a local initial condition, cap confinement and convergence to the antipodal Dirac target.

\subsubsection{Invariance of the anti-symmetric branch}
\label{subsubsec:invariance}

We first show that the first variation respects the antipodal reflection symmetry.

\begin{lemma}[Reflection--exchange covariance of the full first variation]
\label{lem:gradient-symmetry-antisymmetric-branch}
\textbf{Assumptions used.} \cref{assumption:on-sphere,assumption:odd-symmetric-teacher}.

Let \(\mathsf R(w)=-w\) and define the reflection--exchange involution
\[
 \mathfrak R(\mu_1,\mu_2)
 :=(\mathsf R_\#\mu_2,\mathsf R_\#\mu_1).
\]
For every admissible product state \(\mu=(\mu_1,\mu_2)\),
\begin{equation}
\label{eq:mf-reflection-swap-objective}
 F(\mathfrak R\mu)=F(\mu),
\end{equation}
and its spherical gradient fields satisfy
\begin{equation}
\label{eq:mf-reflection-swap-gradient}
\begin{aligned}
 g_1(\mathfrak R\mu;-w)&=-g_2(\mu;w),\\
 g_2(\mathfrak R\mu;-w)&=-g_1(\mu;w),
\end{aligned}
\end{equation}
where \(g_i=\nabla_\Sphere(\delta F/\delta\mu_i)\).
If \(\mu\in\submanifoldforsaddlepoint\), then \(\mathfrak R\mu=\mu\), so
\begin{equation}
\label{eq:full-first-variation-reflection-covariance}
 \fv{F}{\mu_2}(-w)=\fv{F}{\mu_1}(w)+\mathrm{const},
 \qquad g_2(\mu;-w)=-g_1(\mu;w).
\end{equation}
\end{lemma}

\begin{proof}
\noindent\textbf{Step 1: transform the objective at an arbitrary state.}
Oddness of the activation gives, for every marginal \(\nu\),
\[
 h_{\mathsf R_\#\nu}(z)
 =\int\sigma((-w)^\top z)\,d\nu(w)=-h_\nu(z).
\]
Consequently,
\[
 h_{\mathsf R_\#\mu_2}h_{\mathsf R_\#\mu_1}^\top
 =h_{\mu_2}h_{\mu_1}^\top
 =(h_{\mu_1}h_{\mu_2}^\top)^\top.
\]
The teacher matrix \(h_{\nu^o}h_{\nu^o}^\top\) is symmetric by
\cref{assumption:odd-symmetric-teacher}, and the Frobenius norm is invariant
under transposition.  Thus the fitting loss is unchanged.  Reflection is a
spherical isometry and \(\mathsf R_\#\nu_\star^\pm=\nu_\star^\pm\).  Pushing
any transport plan forward by \((\mathsf R,\mathsf R)\), and then applying
this involution once more, gives
\[
 W_2(\mathsf R_\#\nu,\nu_\star^\pm)=W_2(\nu,\nu_\star^\pm).
\]
The reflection exchanges the two summands in \(Z_\pm\), which proves
\cref{eq:mf-reflection-swap-objective} for every admissible state.

\par\noindent\textbf{Step 2: differentiate the symmetry and specialize to the branch.}
Vary the second marginal of \(\mu\) in
\cref{eq:mf-reflection-swap-objective}.  The corresponding variation of the
first marginal of \(\mathfrak R\mu\) is reflected, and
the first-variation identity is
\[
 \fv{F}{\mu_1}(\mathfrak R\mu;-w)
 =\fv{F}{\mu_2}(\mu;w)+\mathrm{const}.
\]
Here the additive constant is immaterial because first-variation potentials
are defined only up to constants.  Since \(D\mathsf R=-I\), differentiating
this identity gives, for every \(v\in T_w\Sphere^{d_w-1}\),
\[
 \langle g_1(\mathfrak R\mu;-w),-v\rangle_{\mathbb R^{d_w}}
 =\langle g_2(\mu;w),v\rangle_{\mathbb R^{d_w}}.
\]
Since this holds for every \(v\), it proves the first identity in
\cref{eq:mf-reflection-swap-gradient}; exchanging the factors proves the
second.  A state lies in \(\mathcal M_-\) precisely when
\(\mathfrak R\mu=\mu\), so the gradient and potential identities in
\cref{eq:full-first-variation-reflection-covariance} follow.
\end{proof}

We prove branch invariance by applying reflection and factor exchange
to an arbitrary admissible product state before using uniqueness.
\begin{proposition}[Invariance of $\submanifoldforsaddlepoint$]
\label{lem:invariance-antisymmetric-branch}
Assume \cref{assumption:on-sphere,assumption:odd-symmetric-teacher}.
Let \(\mu_0\in\submanifoldforsaddlepoint\). On every interval on which
the characteristic product WGF is uniquely defined and remains in the
fixed-assignment cap class \(\mu_{i,t}\in\mathcal P_\pm(r_{\mathrm{cap}})\), \(i=1,2\),
\[
 \mu_{2,t}=(-\Id)_\#\mu_{1,t},
 \qquad \mu_t\in\submanifoldforsaddlepoint.
\]
\end{proposition}

\begin{proof}
Let \(\mathfrak R\) be the reflection--exchange involution from
\cref{lem:gradient-symmetry-antisymmetric-branch}.  If
\(\Phi_{i,t}\) are the original characteristics, define
\[
 \widetilde\Phi_{1,t}(w):=-\Phi_{2,t}(-w),\qquad
 \widetilde\Phi_{2,t}(w):=-\Phi_{1,t}(-w).
\]
Using the characteristic equation
\(\dot\Phi_{i,t}=-g_i(\mu_t;\Phi_{i,t})\) and the first identity in
\cref{eq:mf-reflection-swap-gradient},
\[
\begin{aligned}
 \frac d{dt}\widetilde\Phi_{1,t}(w)
 &=g_2(\mu_t;\Phi_{2,t}(-w))\\
 &=-g_1(\mathfrak R\mu_t;\widetilde\Phi_{1,t}(w)).
\end{aligned}
\]
The second factor satisfies the analogous equation.  Their marginal laws are
\(\mathfrak R\mu_t=(\mathsf R_\#\mu_{2,t},\mathsf R_\#\mu_{1,t})\), and
reflection exchanges the two equal-mass caps, so the transformed pair remains
in the same admissible class.

Since \(\mu_0\in\mathcal M_-\), one has \(\mathfrak R\mu_0=\mu_0\); hence
the transformed and original characteristic systems have the same initial
state.  Uniqueness on the stated interval gives
\(\mathfrak R\mu_t=\mu_t\), or equivalently
\(\mu_{2,t}=\mathsf R_\#\mu_{1,t}\), throughout that interval.
\end{proof}

\subsubsection{Dissipation towards the antipodal saddle point on the anti-symmetric branch}
\label{subsubsec:diddipation}

We establish quantitative transport estimates and use them to prove local attraction to the antipodal target.

\paragraph{Fixed-assignment geometry and first variations.}
Recall the cap class \(\mathcal P_\pm(r_{\mathrm{cap}})\) and product
class \(\mathcal X_\pm(r_{\mathrm{cap}})=\mathcal P_\pm(r_{\mathrm{cap}})^2\)
from \cref{eq:mf-cap-class,eq:mf-product-cap-class}: each marginal has
mass \(1/2\) in each of the two open caps. The assigned pole and its cost
from \cref{eq:mf-assigned-pole,eq:mf-fixed-assignment-cost} are
\[
 \pi_\star(w)=\begin{cases}a_\star,&w\in K_+(r_{\mathrm{cap}}),\\
                         -a_\star,&w\in K_-(r_{\mathrm{cap}}),\end{cases}
 \qquad c_\star(w)=\frac12 r(w)^2,\qquad
 r(w):=d_\Sphere(w,\pi_\star(w)).
\]
The endpoint vector \(q_\star\) of
\cref{definition-spherical-logarithm-exponential} is illustrated in
\cref{fig:mf-spherical-geometry}.  The fixed-assignment calculation in
\cref{lem:fixed-assignment-contraction-geometry,eq:mf-contraction-geodesic-derivatives,eq:mf-cost-gradient-LOG}
shows that \(C_s\) is a constant-speed spherical geodesic with zero
covariant acceleration and that
\[
 q_\star(w)=\nabla_\Sphere c_\star(w)=-\LOG_w\pi_\star(w).
\]

For subsequent derivative formulas define the gradient fields
\begin{equation}
\label{eq:mf-branch-radial-notation}
 g_i(\mu,w):=\nabla_\Sphere\frac{\delta F}{\delta\mu_i}(\mu,w),
 \qquad
 g_{i,\mathrm{fit}}(\mu,w):=\nabla_\Sphere
                  \frac{\delta F_{\mathrm{fit}}}{\delta\mu_i}(\mu,w).
\end{equation}
By the fixed-assignment formula \cref{eq:fixed-assignment-W2-square-formula},
\begin{equation}
\label{eq:mf-branch-distance-gradient}
\begin{aligned}
 Z_\pm(\mu)&=\sum_{i=1}^2\int r(w)^2\,\dd\mu_i(w),\\
 \|(q_\star,q_\star)\|_{L^2(\mu)}^2&=Z_\pm(\mu),
 &\nabla_{W_2}Z_\pm(\mu)&=2(q_\star,q_\star).
\end{aligned}
\end{equation}
Here the product norm is
\(\|V\|_{L^2(\mu)}^2=\sum_{i=1}^2\int\|V_i\|^2\,\dd\mu_i\).
For a smooth product field \(V=(V_1,V_2)\), write
\(DF(\mu)[V]:=\sum_{i=1}^2\int\langle g_i,V_i\rangle\,\dd\mu_i\);
the same convention applies to \(F_{\mathrm{fit}}\).

\paragraph{Constants for the local estimates.}
Under \cref{assumption:odd-symmetric-teacher}, write
\(b(z):=h_{\nu^o}(z)\). Let \(G_1,G_2\) be independent standard real
Gaussians. For a vector-valued random variable \(X\), use
\(\|X\|_{L^4}:=(\E\|X\|_2^4)^{1/4}\), and define
\begin{equation}
\label{eq:mf-gaussian-derivative-constants}
\begin{aligned}
 M_4&:=\|\sigma(G_1)\|_{L^4},
 &B_4&:=\|b(z)\|_{L^4_z},\\
 K_1&:=\|\sigma'(G_1)G_2\|_{L^4},
 &K_2&:=\|\sigma''(G_1)G_2^2-\sigma'(G_1)G_1\|_{L^4}.
\end{aligned}
\end{equation}
The constants in the branch estimates are
\begin{equation}
\label{eq:mf-branch-estimate-constants}
 C_\star:=\frac{K_2^2(M_4^2+B_4^2)}{32},
 \qquad
 L_{\mathrm{fit}}:=\frac{K_1^2(M_4^2+B_4^2)}{\sqrt{2}}.
\end{equation}
All four Gaussian moments are finite because \(\sigma\) is a finite
polynomial. By Gaussian rotational invariance and Minkowski's inequality,
\begin{equation}
\label{eq:mf-output-fourth-moment-bound}
 \|h_\nu\|_{L^4_z}\le M_4
 \quad\text{for every probability measure \(\nu\) on \(\Sphere^{d_w-1}\)},
 \qquad B_4\le M_4.
\end{equation}
These constants depend only on the fixed activation and teacher, and not on
time, the student initialization, or particle number. The parameters
\(p_W,\lambdaReg,r_{\mathrm{cap}}\) are fixed separately.

\begin{lemma}[Pole-contraction and fitting-field estimates]
\label{lem:mf-pole-contraction-estimates}
Assume \cref{assumption:on-sphere,assumption:odd-symmetric-teacher}.
For every \(\mu\in\submanifoldforsaddlepoint\cap
\mathcal X_\pm(r_{\mathrm{cap}})\), the following two bounds hold:
\begin{equation}
\label{eq:mf-pole-contraction-estimates}
\begin{aligned}
 \textup{(i)}\quad &DF_{\mathrm{fit}}(\mu)[(q_\star,q_\star)]
                  \ge-C_\star Z_\pm(\mu)^2,\\
 \textup{(ii)}\quad &\sup_{w\in\Sphere^{d_w-1}}
       \|g_{i,\mathrm{fit}}(\mu,w)\|_{\mathbb R^{d_w}}
       \le L_{\mathrm{fit}}\sqrt{Z_\pm(\mu)},\qquad i=1,2.
\end{aligned}
\end{equation}
The constants are those in
\cref{eq:mf-gaussian-derivative-constants,eq:mf-branch-estimate-constants}.
The simultaneous pole contraction is an admissible smooth transport
perturbation at \(\mu\).
\end{lemma}

\begin{proof}
Within this proof only, write \(Z=Z_\pm(\mu)\) and
\(h=h_{\mu_1}=-h_{\mu_2}\). The teacher output is
\(h_{\nu^o}(z)=\int\sigma(w^\top z)\,\dd\nu^o(w)\), where
\(\nu^o=\mu_1^o=\mu_2^o\) by \cref{assumption:odd-symmetric-teacher}.

\par\noindent\textbf{Step 1: contraction, Gaussian derivatives, and the remainder.}
Recall the contraction from \cref{definition-spherical-logarithm-exponential}
and define its output by
\begin{equation}
\label{eq:mf-pole-contraction-path}
\begin{aligned}
 C_s(w)&=\EXP_{\pi_\star(w)}\bigl(s\LOG_{\pi_\star(w)}w\bigr),
 &\mu_{i,s}&=(C_s)_\#\mu_i,\\
 h_s(z)&=\int\sigma(C_s(w)^\top z)\,\dd\mu_1(w),
 &0&\le s\le1.
\end{aligned}
\end{equation}
Since \(C_s(-w)=-C_s(w)\), reflection of the first pushed-forward
marginal equals the second. Each point stays in its assigned cap, with
\(r(C_s(w))=s r(w)\). The mass \(1/2\) in each cap and oddness give
\(h_0=\tfrac12\sigma(a_\star^\top z)+
\tfrac12\sigma(-a_\star^\top z)=0\). Reflection also gives
\(\int r^2\,\dd\mu_2=\int r(-w)^2\,\dd\mu_1=\int r^2\,\dd\mu_1\).
Consequently,
\begin{equation}
\label{eq:mf-contraction-endpoint-identities}
 h_0=0,\qquad h_1=h,\qquad
 \int r^2\,\dd\mu_1=\frac Z2,\qquad
 Z_\pm(\mu_{1,s},\mu_{2,s})=s^2Z.
\end{equation}
The closed supports are compact subsets of the open caps; hence
\(s r(w)<r_{\mathrm{cap}}\) uniformly for \(s\) in a small two-sided
neighborhood of \(1\). This proves admissibility, with product velocity
\((q_\star,q_\star)\) at \(s=1\).

For \(r(w)>0\), \cref{eq:mf-contraction-geodesic-derivatives} gives
\begin{equation}
\label{eq:mf-contraction-orthogonal-unit-pair}
 \|C_s(w)\|_{\mathbb R^{d_w}}=1,\qquad
 \left\|\frac{\partial_sC_s(w)}{r(w)}\right\|_{\mathbb R^{d_w}}=1,
 \qquad
 \left\langle C_s(w),\frac{\partial_sC_s(w)}{r(w)}
 \right\rangle_{\mathbb R^{d_w}}=0.
\end{equation}
Thus the geometry separates the first derivative's factor \(r(w)\) from
second-order terms carrying \(r(w)^2\).  Only for the next calculation, set
\[
 X_s:=C_s(w)^\top z,
 \qquad
 Y_s:=\frac{\partial_sC_s(w)^\top z}{r(w)}.
\]
For \(z\sim\mathcal N(0,I_{d_w})\),
\cref{eq:mf-contraction-orthogonal-unit-pair} gives
\[
 \mathbb E_zX_s^2=1,\qquad \mathbb E_zY_s^2=1,\qquad
 \mathbb E_z[X_sY_s]=0.
\]
The pair \((X_s,Y_s)\) is jointly Gaussian, so it has the law of independent
standard normals \((G_1,G_2)\).  Recall from
\cref{eq:mf-gaussian-derivative-constants} that
\[
 K_1=\|\sigma'(G_1)G_2\|_{L^4},\qquad
 K_2=\|\sigma''(G_1)G_2^2-\sigma'(G_1)G_1\|_{L^4}.
\]
For \(f_w(s,z):=\sigma(C_s(w)^\top z)\), the chain rule and
\(\partial_s^2C_s=-r(w)^2C_s\) from
\cref{eq:mf-contraction-geodesic-derivatives} give
\begin{equation}
\label{eq:mf-spherical-feature-derivatives}
\begin{aligned}
 \partial_s f_w
 &=r(w)\sigma'(X_s)Y_s,\\
 \partial_s^2 f_w
 &=r(w)^2\bigl[\sigma''(X_s)Y_s^2-\sigma'(X_s)X_s\bigr],\\
 \|\partial_s f_w\|_{L^4_z}
 &=r(w)\|\sigma'(G_1)G_2\|_{L^4}=K_1r(w),\\
 \|\partial_s^2 f_w\|_{L^4_z}
 &=r(w)^2\|\sigma''(G_1)G_2^2-\sigma'(G_1)G_1\|_{L^4}
 =K_2r(w)^2.
\end{aligned}
\end{equation}
When \(r(w)=0\), \(C_s(w)=w\) is constant, so both derivatives vanish and
the same norm identities hold without defining \(Y_s\).

The endpoint identities \(h_0=0\) and \(h_1=h\) in
\cref{eq:mf-contraction-endpoint-identities} show that a linear output path
would satisfy \(h'_1=h_1-h_0=h\).  We therefore measure the failure of this
linearity by \(e:=h'_1-h\).  Step 2 substitutes \(h'_1=h+e\) into the
fitting derivative, so the next calculation bounds \(\|e\|_{L^4_z}\).
For each fixed \(w,z\), integration by parts gives
\[
 \int_0^1s f_w''(s)\,\dd s
 =[s f_w'(s)]_0^1-\int_0^1f_w'(s)\,\dd s
 =f_w'(1)-f_w(1)+f_w(0).
\]
Integrating this identity in \(w\) gives
\(\int f_w'(1)\,d\mu_1=h'_1\),
\(\int f_w(1)\,d\mu_1=h_1=h\), and
\(\int f_w(0)\,d\mu_1=h_0=0\).  Hence
\begin{equation}
\label{eq:mf-contraction-feature-remainder}
\begin{aligned}
 e:=h'_1-h
 &=\int\!\int_0^1s f_w''(s,\cdot)\,\dd s\,\dd\mu_1(w),\\
 \|e\|_{L^4_z}
 &\le\int\!\int_0^1s\|f_w''(s,\cdot)\|_{L^4_z}\,\dd s\,\dd\mu_1(w)\\
 &\le K_2\left(\int_0^1s\,\dd s\right)\int r(w)^2\,\dd\mu_1(w)\\
 &=\frac{K_2}{2}\frac Z2=\frac{K_2Z}{4},
 \qquad h'_1=h+e,
\end{aligned}
\end{equation}
where the last equality uses
\cref{eq:mf-contraction-endpoint-identities}.  The first inequality is
Minkowski for the Bochner integral over \(s\) and \(w\).
The activation is a finite polynomial and the sphere is compact, so its
Gaussian polynomial bounds justify these differentiations and integrals.

\par\noindent\textbf{Step 2: the fitting derivative.}
The product path in \cref{eq:mf-pole-contraction-path} has endpoint velocity
\((q_\star,q_\star)\), so by definition of the directional derivative,
\begin{equation}
\label{eq:mf-contraction-directional-derivative}
 DF_{\mathrm{fit}}(\mu)[(q_\star,q_\star)]
 =\left.\frac d{ds}\right|_{s=1}
   F_{\mathrm{fit}}(\mu_{1,s},\mu_{2,s}).
\end{equation}
Recall \(M_4=\|\sigma(G_1)\|_{L^4}\) and
\(B_4=\|h_{\nu^o}\|_{L^4_z}\le M_4\) from
\cref{eq:mf-gaussian-derivative-constants,eq:mf-output-fourth-moment-bound}.
In particular,
\(\|h\|_{L^4_z}\le\int\|\sigma(w^\top z)\|_{L^4_z}\,\dd\mu_1(w)=M_4\).
On the anti-symmetric branch,
\(h_{\mu_1}=h\) and \(h_{\mu_2}=-h\), so the model output matrix is
\(-hh^\top\).  Hence
\[
 F_{\mathrm{fit}}(\mu)
 =\frac12\mathbb E_z
   \|-hh^\top-h_{\nu^o}h_{\nu^o}^\top\|_{\mathrm F}^2.
\]
The Frobenius identities
\(\|hh^\top\|_{\mathrm F}^2=\|h\|_2^4\) and
\(\langle hh^\top,h_{\nu^o}h_{\nu^o}^\top\rangle_{\mathrm F}
=\langle h,h_{\nu^o}\rangle^2\) expand the branch objective as
\begin{equation}
\label{eq:mf-branch-energy-expansion}
 F_{\mathrm{fit}}(\mu)
 =\frac12\E_z\|h\|_2^4
  +\E_z\langle h,h_{\nu^o}\rangle^2
  +\frac12\E_z\|h_{\nu^o}\|_2^4.
\end{equation}
For the first two summands, before evaluating at \(s=1\),
\[
 \frac{\dd}{\dd s}\frac12\|h_s\|_2^4
 =2\|h_s\|_2^2\langle h_s,h_s'\rangle,\qquad
 \frac{\dd}{\dd s}\langle h_s,h_{\nu^o}\rangle^2
 =2\langle h_s,h_{\nu^o}\rangle\langle h_s',h_{\nu^o}\rangle.
\]
The teacher term is constant.  Evaluating at \(s=1\), using \(h_1=h\),
and substituting \(h'_1=h+e\) from
\cref{eq:mf-contraction-feature-remainder} into
\cref{eq:mf-contraction-directional-derivative} gives
\begin{align}
 DF_{\mathrm{fit}}(\mu)[(q_\star,q_\star)]
 =2\E_z\bigl[&\|h\|_2^4+\|h\|_2^2\langle h,e\rangle
   +\langle h,h_{\nu^o}\rangle^2\notag\\
   &+\langle h,h_{\nu^o}\rangle\langle e,h_{\nu^o}\rangle\bigr].
\end{align}
The two pointwise square completions are
\[
 2\|h\|_2^2(\|h\|_2^2+\langle h,e\rangle)
 =2\|h\|_2^2\|h+e/2\|_2^2-\tfrac12\|h\|_2^2\|e\|_2^2,
\]
\[
\begin{aligned}
 2\langle h,h_{\nu^o}\rangle^2
 +2\langle h,h_{\nu^o}\rangle\langle e,h_{\nu^o}\rangle
 &=2(\langle h,h_{\nu^o}\rangle+\tfrac12\langle e,h_{\nu^o}\rangle)^2
       -\tfrac12\langle e,h_{\nu^o}\rangle^2\\
 &\ge-\tfrac12\|e\|_2^2\|h_{\nu^o}\|_2^2.
\end{aligned}
\]
Discarding the nonnegative squares and applying H\"older with exponents
\(2,2\) to each product of squared norms yields
\begin{align}
 DF_{\mathrm{fit}}(\mu)[(q_\star,q_\star)]
 &\ge-\tfrac12\E_z[(\|h\|_2^2+\|h_{\nu^o}\|_2^2)\|e\|_2^2]\notag\\
 &\ge-\tfrac12(\|h\|_{L^4_z}^2+\|h_{\nu^o}\|_{L^4_z}^2)
                          \|e\|_{L^4_z}^2\notag\\
 &\ge-\tfrac12(M_4^2+B_4^2)(K_2Z/4)^2=-C_\star Z^2,
 \label{eq:mf-contraction-square-completion}
\end{align}
where \(C_\star=K_2^2(M_4^2+B_4^2)/32\), exactly as in
\cref{eq:mf-branch-estimate-constants}.

\par\noindent\textbf{Step 3: the pointwise fitting field and output norm.}
For vectors \(x,y\in\R^{d_h}\) and an increment \(a\),
\[
 \left.\frac{\dd}{\dd t}\frac12
 \|(x+ta)y^\top-h_{\nu^o}h_{\nu^o}^\top\|_{\mathrm F}^2\right|_{t=0}
 =\langle (xy^\top-h_{\nu^o}h_{\nu^o}^\top)y,a\rangle.
\]
Thus, at \((x,y)=(h,-h)\), the first output gradient is
\(\|h\|_2^2h+\langle h,h_{\nu^o}\rangle h_{\nu^o}\); the second is
its negative. This differentiates one output at a time, unlike Step 2,
which moves both marginals.
For a unit tangent \(u\in T_w\Sphere^{d_w-1}\), the vectors \(w,u\) are
orthogonal unit vectors.  Hence \((w^\top z,u^\top z)\) has the law of the
independent standard Gaussian pair \((G_1,G_2)\), and
\(\|\sigma'(w^\top z)(u^\top z)\|_{L^4_z}=K_1\) by
\cref{eq:mf-gaussian-derivative-constants}.  Euclidean Cauchy--Schwarz,
followed by H\"older with \(1/(4/3)+1/4=1\), gives
\[
\begin{aligned}
 |\langle g_{i,\mathrm{fit}}(\mu,w),u\rangle|
 &\le\mathbb E_z\!\left[
 \bigl\|\|h\|_2^2h+\langle h,h_{\nu^o}\rangle h_{\nu^o}\bigr\|_2
 \bigl\|\sigma'(w^\top z)(u^\top z)\bigr\|_2\right]\\
 &\le K_1\bigl\|\|h\|_2^2h
              +\langle h,h_{\nu^o}\rangle h_{\nu^o}\bigr\|_{L^{4/3}_z}\\
 &\le K_1\bigl(\|h\|_{L^4_z}^3
                     +\|h\|_{L^4_z}\|h_{\nu^o}\|_{L^4_z}^2\bigr).
\end{aligned}
\]
For the last line, the triangle inequality and
\[
 \|\|h\|_2^3\|_{L^{4/3}_z}
 =(\mathbb E_z\|h\|_2^4)^{3/4}=\|h\|_{L^4_z}^3
\]
handle the first term.  For the second, pointwise Cauchy--Schwarz and
H\"older with \(1/(4/3)=1/4+1/2\) give
\[
 \bigl\||\langle h,h_{\nu^o}\rangle|\,\|h_{\nu^o}\|_2
 \bigr\|_{L^{4/3}_z}
 \le\|\|h\|_2\|h_{\nu^o}\|_2^2\|_{L^{4/3}_z}
 \le\|h\|_{L^4_z}\|h_{\nu^o}\|_{L^4_z}^2.
\]
Taking the supremum over unit tangents gives the same bound for
\(\|g_{i,\mathrm{fit}}(\mu,w)\|_{\mathbb R^{d_w}}\), because the fitting
field belongs to \(T_w\Sphere^{d_w-1}\).
Finally, \(h_0=0\) and the fundamental theorem of calculus give
\[
 h(z)=\int\int_0^1\sigma'(C_s(w)^\top z)
                      (\partial_sC_s(w)^\top z)\,\dd s\,\dd\mu_1(w).
\]
Minkowski, \cref{eq:mf-spherical-feature-derivatives}, and
Cauchy--Schwarz against the probability measure \(\mu_1\) yield
\[
\begin{aligned}
 \|h\|_{L^4_z}
 &\le\int\int_0^1 K_1 r(w)\,\dd s\,\dd\mu_1(w)\\
 &\le K_1\left(\int r(w)^2\,\dd\mu_1(w)\right)^{1/2}
          \left(\int1\,\dd\mu_1\right)^{1/2}
 =K_1\sqrt{Z/2}.
\end{aligned}
\]
Combining this estimate with \(\|h\|_{L^4_z}\le M_4\) proves
\[
 \|g_{i,\mathrm{fit}}(\mu,w)\|_{\mathbb R^{d_w}}
 \le K_1(M_4^2+B_4^2)K_1\sqrt{Z/2}
 =L_{\mathrm{fit}}\sqrt Z,
 \qquad L_{\mathrm{fit}}=\frac{K_1^2(M_4^2+B_4^2)}{\sqrt2},
\]
which is the constant in \cref{eq:mf-branch-estimate-constants}.
\end{proof}

We now pass from the comparison transport to the gradient of the full
objective and the actual product WGF.

\begin{proposition}[Quantitative radial dissipation near the target]
\label{lem:radial-positivity-antisymmetric-branch-full-density}
Assume \cref{assumption:on-sphere,assumption:odd-symmetric-teacher} and let
\(2<p_W<4\). Every state
\(\mu\in\mathcal M_-\cap\mathcal X_\pm(r_{\mathrm{cap}})\) satisfying
\begin{equation}
\label{eq:mf-local-radial-smallness}
 0<Z_\pm(\mu),\qquad
 C_\star Z_\pm(\mu)^{(4-p_W)/2}\le\frac{\lambdaReg}{2}
\end{equation}
obeys
\begin{equation}
\label{eq:mf-local-radial-dissipation}
\begin{aligned}
 DF(\mu)[(q_\star,q_\star)]
 &\ge\frac{\lambdaReg}{2}Z_\pm(\mu)^{p_W/2},\\
 \sum_{i=1}^2\int
 \|g_i(\mu,w)\|_{\mathbb R^{d_w}}^2\,\dd\mu_i(w)
 &\ge\frac{\lambdaReg^2}{4}Z_\pm(\mu)^{p_W-1}.
\end{aligned}
\end{equation}
Consequently, on every time interval on which the product WGF remains in
\(\mathcal M_-\cap\mathcal X_\pm(r_{\mathrm{cap}})\), has positive distance,
and satisfies the smallness condition in
\cref{eq:mf-local-radial-smallness},
\begin{equation}
\label{eq:mf-local-distance-differential}
 \frac{\dd}{\dd t}Z_\pm(\mu_t)
 =-2DF(\mu_t)[(q_\star,q_\star)]
 \le-\lambdaReg Z_\pm(\mu_t)^{p_W/2}.
\end{equation}
Among the states in
\(\mathcal M_-\cap\mathcal X_\pm(r_{\mathrm{cap}})\) satisfying the second
inequality in \cref{eq:mf-local-radial-smallness}, the only stationary state
is \(\mu^\dagger:=\nu_\star^\pm\otimes\nu_\star^\pm\).
\end{proposition}

\begin{proof}
\noindent\textbf{Step 1: fix the radial margin and bound the squared gradient.}
The contraction in \cref{eq:mf-pole-contraction-path} has endpoint field
\((q_\star,q_\star)\). By
\cref{eq:mf-contraction-endpoint-identities}, its distance is
\(s^2Z_\pm(\mu)\), so
\[
 F_{\mathrm{reg}}(\mu_s)
 =\frac{\lambdaReg}{p_W}s^{p_W}Z_\pm(\mu)^{p_W/2},
 \qquad
 \left.\frac{\dd}{\dd s}F_{\mathrm{reg}}(\mu_s)\right|_{s=1}
 =\lambdaReg Z_\pm(\mu)^{p_W/2}.
\]
Adding the fitting bound in
\cref{lem:mf-pole-contraction-estimates,eq:mf-pole-contraction-estimates}
and using
\[
Z_\pm(\mu)^2
 =Z_\pm(\mu)^{(4-p_W)/2}Z_\pm(\mu)^{p_W/2}
\]
gives
\begin{align}
 DF(\mu)[(q_\star,q_\star)]
 &\ge \lambdaReg Z_\pm(\mu)^{p_W/2}-C_\star Z_\pm(\mu)^2\notag\\
 &=\left[\lambdaReg-C_\star Z_\pm(\mu)^{(4-p_W)/2}\right]
                  Z_\pm(\mu)^{p_W/2}\notag\\
 &\ge\frac{\lambdaReg}{2}Z_\pm(\mu)^{p_W/2}.
 \label{eq:mf-full-radial-lower-bound}
\end{align}
The exponent \((4-p_W)/2\) is positive because \(p_W<4\); hence the
\(O(Z_\pm^{p_W/2})\) regularizer term dominates the
\(O(Z_\pm^2)\) fitting remainder sufficiently near the target.

Recall from \cref{eq:mf-branch-distance-gradient} that
\(\|(q_\star,q_\star)\|_{L^2(\mu;\mathbb R^{d_w})}
=Z_\pm(\mu)^{1/2}\). Cauchy--Schwarz and
\cref{eq:mf-full-radial-lower-bound} therefore give
\[
 \frac{\lambdaReg}{2}Z_\pm(\mu)^{p_W/2}
 \le DF(\mu)[(q_\star,q_\star)]
 \le
 \left(\sum_{i=1}^2\int
       \|g_i(\mu,w)\|_{\mathbb R^{d_w}}^2\,\dd\mu_i(w)\right)^{1/2}
 Z_\pm(\mu)^{1/2}.
\]
Since \(Z_\pm(\mu)>0\), division by its square root and then squaring yield
the second inequality in \cref{eq:mf-local-radial-dissipation}.

\par\noindent\textbf{Step 2: differentiate the distance along WGF.}
The WGF velocity is \(-g_i\), whereas the outward cost gradient is
\(\nabla_\Sphere c_\star=q_\star\) by
\cref{eq:mf-fixed-assignment-cost,eq:mf-cost-gradient-LOG}. Thus the
fixed-assignment formula gives
\[
 \frac{\dd}{\dd t}Z_\pm(\mu_t)
 =2\sum_{i=1}^2\int
   \langle q_\star(w),-g_i(\mu_t,w)\rangle_{\mathbb R^{d_w}}
   \,\dd\mu_{i,t}(w)
 =-2DF(\mu_t)[(q_\star,q_\star)].
\]
Substitution of \cref{eq:mf-full-radial-lower-bound} proves
\cref{eq:mf-local-distance-differential}. The minus sign explains why a
positive outward directional derivative makes the negative gradient flow
move toward the assigned poles.

\par\noindent\textbf{Step 3: identify the stationary state.}
If \(Z_\pm(\mu)>0\), the squared-gradient lower bound in
\cref{eq:mf-local-radial-dissipation} rules out stationarity. If
\(Z_\pm(\mu)=0\), both marginals equal \(\nu_\star^\pm\). Their output is
\[
 \tfrac12\sigma(a_\star^\top z)+
 \tfrac12\sigma((-a_\star)^\top z)=0
\]
by oddness of \(\sigma\), so the fitting fields computed in Step 3 of
\cref{lem:mf-pole-contraction-estimates} vanish. The regularization field
\(\lambdaReg Z_\pm(\mu)^{(p_W-2)/2}q_\star\) also vanishes by
\cref{eq:W2-dirac-metric-gradient}. Hence \(\mu^\dagger\) is stationary.
\end{proof}

\paragraph{Second-order behavior at the target.}
For every \(p_W>2\), a smooth atom transport from \(\mu^\dagger\)
satisfies \(Z_\pm(\mu_h)=O(h^2)\). Consequently,
\[
 F_{\mathrm{reg}}(\mu_h)
 =\frac{\lambdaReg}{p_W}Z_\pm(\mu_h)^{p_W/2}
 =O(|h|^{p_W})=o(h^2).
\]
The regularizer has zero second variation at the target; the existence
of this derivative for the normalized retractions is verified explicitly
in \cref{eq:mf-zero-distance-retraction}. The exponent condition here is
\(p_W>2\); the next local-attraction result additionally requires \(p_W<3\).

\begin{proposition}[Cap confinement and convergence for a global characteristic flow]
\label{prop:reduced-dissipation-antisymmetric-branch}
Assume \cref{assumption:on-sphere,assumption:odd-symmetric-teacher}.
Fix \(2<p_W<3\) and \(\lambdaReg>0\). Choose \(z_\star>0\) such that
\begin{equation}
\label{eq:mf-local-basin-threshold}
 C_\star z_\star^{(4-p_W)/2}\le\frac{\lambdaReg}{2},\qquad
 \frac{L_{\mathrm{fit}}}{\lambdaReg}z_\star^{(3-p_W)/2}
 \le\frac{r_{\mathrm{cap}}}{2}.
\end{equation}
Let \(\mu_0\in\mathcal M_-\cap\mathcal X_\pm(r_{\mathrm{cap}})\), and define
\begin{equation}
\label{eq:mf-local-initial-data}
 Z_0:=Z_\pm(\mu_0),\qquad
 R_0:=\max_{i=1,2}\sup_{w\in\operatorname{supp}\mu_{i,0}}r(w)
 <r_{\mathrm{cap}}.
\end{equation}
Let \((\mu_t)_{t\ge0}\) be the globally well-posed characteristic product WGF
of \Cref{def:mf-characteristic-solution}
starting from \(\mu_0\): it exists for every \(t\ge0\) and is unique in the
class of characteristic solutions. If \(0\le Z_0\le z_\star\), then:
\begin{enumerate}[label=(\roman*),leftmargin=*]
\item The flow satisfies
\(\mu_t\in\mathcal M_-\cap\mathcal X_\pm(r_{\mathrm{cap}})\) for all
\(t\ge0\), and
\begin{equation}
\label{eq:mf-local-distance-decay}
 0\le Z_\pm(\mu_t)\le
 \frac{Z_0}{\bigl[1+\tfrac{\lambdaReg(p_W-2)}2
                   Z_0^{(p_W-2)/2}t\bigr]^{2/(p_W-2)}}
 \le Z_0.
\end{equation}
\item Its support satisfies
\begin{equation}
\label{eq:mf-local-support-barrier}
 \max_{i=1,2}\sup_{w\in\operatorname{supp}\mu_{i,t}}r(w)
 \le\overline R:=\max\left\{R_0,
        \frac{L_{\mathrm{fit}}}{\lambdaReg}Z_0^{(3-p_W)/2}\right\}
 <r_{\mathrm{cap}}.
\end{equation}
\item For almost every \(t\ge0\),
\begin{equation}
\label{eq:mf-local-energy-dissipation}
 \frac{\dd}{\dd t}F(\mu_t)
 =-\sum_{i=1}^2\int
       \|g_i(\mu_t,w)\|_{\mathbb R^{d_w}}^2\,\dd\mu_{i,t}(w)
 \le-\frac{\lambdaReg^2}{4}Z_\pm(\mu_t)^{p_W-1}\le0.
\end{equation}
The inequality is strict at every differentiability time for which
\(Z_\pm(\mu_t)>0\). Moreover, \(\mu_t\to\mu^\dagger\) in product
Wasserstein distance, \(\mathbf S(\mu_t)=0\), and
\(\mathbf A(\mu_t)\to0\).
\end{enumerate}
If \(Z_0=0\), the unique flow is the stationary target \(\mu^\dagger\).
No density or lower bound on atom masses is required.
\end{proposition}

The strict initial support bound in \cref{eq:mf-local-initial-data} follows
from \cref{eq:mf-cap-class}: the closed support is compact and contained in
the two open caps. Both exponents in \cref{eq:mf-local-basin-threshold} are
positive, so such a \(z_\star>0\) exists for every fixed
\(\lambdaReg>0\). Global well-posedness specifies the time domain and
uniqueness of the trajectory; cap confinement in
\cref{eq:mf-local-support-barrier} is a conclusion proved below.

\begin{proof}
\par\noindent\textbf{Step 1: branch invariance and the mean-distance barrier.}
If \(Z_0=0\), then \(\mu_0=\mu^\dagger\). By
\cref{lem:radial-positivity-antisymmetric-branch-full-density}, the full field
vanishes there, so the constant curve is a characteristic product WGF.
The assumed uniqueness makes it the trajectory in the proposition, and all
claims follow. Hence assume \(0<Z_0\le z_\star\).

Let $\Phi_{i,t}$ be the characteristic maps of the assumed global flow and
define the first cap-exit time
\[
 \tau_{\rm cap}:=\inf\left\{t\ge0:
  \max_{i=1,2}\sup_{w\in\operatorname{supp}\mu_{i,0}}
  r(\Phi_{i,t}(w))\ge r_{\rm cap}\right\},
 \qquad \inf\varnothing:=\infty.
\]
The strict initial support bound in \cref{eq:mf-local-initial-data} and
continuity of the characteristics give $\tau_{\rm cap}>0$.
Write \(Z(t)=Z_\pm(\mu_t)\). On $[0,\tau_{\rm cap})$,
\cref{lem:invariance-antisymmetric-branch} gives
\(\mu_t\in\mathcal M_-\). The fixed-assignment distance identity in
\cref{eq:mf-local-distance-differential} shows that \(Z\) is continuously
differentiable there: the characteristic maps, the finitely many active
moments, and both gradient fields are continuous in time.
Whenever \(0<Z(t)\le Z_0\), the basin condition
\cref{eq:mf-local-basin-threshold} and
\cref{eq:mf-local-distance-differential} give
\begin{equation}
\label{eq:mf-distance-decay-differential}
 \dot Z(t)\le-\lambdaReg Z(t)^{p_W/2}\le0.
\end{equation}
In particular, \(\dot Z(0)<0\). If \(Z\) later exceeded \(Z_0\), continuity
would give a first upward return time \(s>0\) with \(Z(s)=Z_0\) and
\(\dot Z(s)\ge0\). This contradicts
\(\dot Z(s)\le-\lambdaReg Z_0^{p_W/2}<0\). Thus, up to a possible cap exit,
\[
 Z(t)\le Z_0,
 \qquad
 \dot Z(t)\le-\lambdaReg Z(t)^{p_W/2}
 \quad\text{whenever }Z(t)>0.
\]

On every interval on which \(Z(t)>0\), the chain rule gives
\[
 \frac{\dd}{\dd t}Z(t)^{-(p_W-2)/2}
 =-\frac{p_W-2}{2}Z(t)^{-p_W/2}\dot Z(t)
 \ge\frac{\lambdaReg(p_W-2)}2.
\]
Integrating from \(0\) to \(t\), and then taking the negative power
\(-2/(p_W-2)\), gives the first upper bound in
\cref{eq:mf-local-distance-decay}. If the distance reaches zero at a finite
time, the state is \(\mu^\dagger\); the stationary solution and the assumed
uniqueness keep it there thereafter. Thus the same bound remains valid without
substituting \(Z=0\) into the preceding negative power.

\par\noindent\textbf{Step 2: pointwise support confinement.}
For a characteristic \(w_i(t)\), let
\(r_i(t)=d_\Sphere(w_i(t),\pi_\star(w_i(t)))\). This is a pointwise pole
distance, whereas \(Z(t)\) is the sum of the two mean squared distances.
Recall from
\cref{definition-spherical-logarithm-exponential,lem:fixed-assignment-contraction-geometry,eq:mf-cost-gradient-LOG}
that, with \(r(w)=d_\Sphere(w,\pi_\star(w))\),
\[
 q_\star(w)
 =\left.\partial_sC_s(w)\right|_{s=1}
 =\nabla_\Sphere c_\star(w)
 =-\LOG_w\pi_\star(w),
 \qquad
 \|q_\star(w)\|_{\mathbb R^{d_w}}=r(w).
\]
Since \(c_\star=r^2/2\), at every point with \(r(w)>0\),
\[
 q_\star(w)=r(w)\nabla_\Sphere r(w),
 \qquad
 \nabla_\Sphere r(w)=\frac{q_\star(w)}{r(w)},
 \qquad
 \left\|\frac{q_\star(w)}{r(w)}\right\|_{\mathbb R^{d_w}}=1.
\]
Thus \(q_\star/r\) is the outward unit radial tangent. By
\cref{eq:W2-dirac-metric-gradient,eq:mf-cost-gradient-LOG}, the
regularization field is
\[
 g_{i,\mathrm{reg}}(\mu,w)
 =\lambdaReg Z_\pm(\mu)^{(p_W-2)/2}q_\star(w).
\]
The fitting-field estimate in
\cref{lem:mf-pole-contraction-estimates,eq:mf-pole-contraction-estimates}
is uniform in the support point:
\[
 \|g_{i,\mathrm{fit}}(\mu_t,w_i(t))\|_{\mathbb R^{d_w}}
 \le L_{\mathrm{fit}}\sqrt{Z(t)}.
\]
This is the aggregate output estimate proved in that lemma; it does not use
the generally invalid pointwise comparison \(r_i(t)\le\sqrt{Z(t)}\).
For \(r_i(t)>0\), the characteristic equation
\(\dot w_i=-g_{i,\mathrm{fit}}-g_{i,\mathrm{reg}}\), the chain rule, and
Cauchy--Schwarz therefore give
\begin{align}
 \dot r_i(t)
 &=\left\langle\frac{q_\star(w_i(t))}{r_i(t)},\dot w_i(t)
          \right\rangle_{\mathbb R^{d_w}}\notag\\
 &=-\left\langle\frac{q_\star(w_i(t))}{r_i(t)},
                 g_{i,\mathrm{fit}}(\mu_t,w_i(t))
          \right\rangle_{\mathbb R^{d_w}}
       -\lambdaReg Z(t)^{(p_W-2)/2}r_i(t)\notag\\
 &\le L_{\mathrm{fit}}\sqrt{Z(t)}
       -\lambdaReg Z(t)^{(p_W-2)/2}r_i(t)\notag\\
 &=\lambdaReg Z(t)^{(p_W-2)/2}
       \left[\frac{L_{\mathrm{fit}}}{\lambdaReg}
                    Z(t)^{(3-p_W)/2}-r_i(t)\right]\notag\\
 &\le\lambdaReg Z(t)^{(p_W-2)/2}(\overline R-r_i(t)).
 \label{eq:mf-pointwise-radius-differential}
\end{align}
The last line uses \(p_W<3\), \(Z(t)\le Z_0\), and the definition of
\(\overline R\) in \cref{eq:mf-local-support-barrier}. At a pole, the upper
right derivative satisfies
\[
 D^+r_i(t)\le\|\dot w_i(t)\|_{\mathbb R^{d_w}}
 =\|g_{i,\mathrm{fit}}(\mu_t,w_i(t))\|_{\mathbb R^{d_w}}
 \le L_{\mathrm{fit}}\sqrt{Z(t)},
\]
because the regularizer field vanishes there; hence the same differential
bound holds. Multiplying by the integrating factor and using
\(r_i(0)\le R_0\le\overline R\) yields
\[
\begin{aligned}
 r_i(t)
 &\le\overline R+(r_i(0)-\overline R)
 \exp\!\left[-\lambdaReg\int_0^t
       Z(s)^{(p_W-2)/2}\,\dd s\right]\\
 &\le\overline R<r_{\mathrm{cap}}.
\end{aligned}
\]
Here \(\overline R<r_{\mathrm{cap}}\) follows from
\cref{eq:mf-local-basin-threshold,eq:mf-local-initial-data}. The estimate is
uniform over the compact initial supports for every $t<\tau_{\rm cap}$.
If $\tau_{\rm cap}<\infty$, continuity of the characteristic maps and the
preceding inequality, followed by $t\uparrow\tau_{\rm cap}$, would give
\[
 \max_{i=1,2}\sup_{w\in\operatorname{supp}\mu_{i,0}}
 r(\Phi_{i,\tau_{\rm cap}}(w))
 \le\overline R<r_{\rm cap},
\]
contradicting the definition of $\tau_{\rm cap}$. Hence
$\tau_{\rm cap}=\infty$.
Since the characteristic flow is globally defined by assumption,
\cref{lem:invariance-antisymmetric-branch} now gives branch invariance for all
\(t\ge0\), and the distance estimate from Step 1 also holds for all time.

\par\noindent\textbf{Step 3: dissipation and convergence.}
For \(Z(t)>0\), the basin condition and \(Z(t)\le Z_0\le z_\star\) allow us
to apply the squared-gradient estimate in
\cref{eq:mf-local-radial-dissipation}. The characteristic chain rule gives
\[
\begin{aligned}
 \frac{\dd}{\dd t}F(\mu_t)
 &=\sum_{i=1}^2\int
   \langle g_i(\mu_t,w),-g_i(\mu_t,w)\rangle_{\mathbb R^{d_w}}
   \,\dd\mu_{i,t}(w)\\
 &=-\sum_{i=1}^2\int
   \|g_i(\mu_t,w)\|_{\mathbb R^{d_w}}^2\,\dd\mu_{i,t}(w)\\
 &\le-\frac{\lambdaReg^2}{4}Z(t)^{p_W-1}<0.
\end{aligned}
\]
If \(Z(t)=0\), uniqueness keeps the flow at the stationary target, so the
energy derivative is zero. This proves
\cref{eq:mf-local-energy-dissipation} with the stated strictness.

The right-hand side of \cref{eq:mf-local-distance-decay} tends to zero, and
the product Wasserstein distance from \(\mu_t\) to \(\mu^\dagger\) is
\(\sqrt{Z(t)}\). Hence \(\mu_t\to\mu^\dagger\). Branch invariance gives
\(\mathbf S(\mu_t)=0\). For each \(n\in\mathcal O\), continuity of
\(w\mapsto w^{\otimes n}\) on the compact sphere gives
\[
 T_{i,n}(\mu_{i,t})\longrightarrow
 \tfrac12a_\star^{\otimes n}+\tfrac12(-a_\star)^{\otimes n}=0.
\]
Since \(\mathcal O\) is finite, \(\mathbf A(\mu_t)\to0\).
\end{proof}

%% file: appendix_mf_dirac/EMN_rewrite_curvature_v8.tex
\subsection{Curvature away from the saddle on the antisymmetric branch}
\label{subsec:curvature-away-from-saddle-on-antisymmetric}
This subsection bounds the spherical product pushforward second variation at
anti-symmetric branch points away from the antipodal target.  On a fixed
metric annulus, sufficiently strong regularization dominates the fitting curvature.

We first note that the antipodal Dirac target lies exactly at the active
odd-moment origin by \cref{prop:intersection-zero-output-fiber}:
\[
    \mathbf S(\mu_\star^\pm)=0,
    \qquad
    \mathbf A(\mu_\star^\pm)=0.
\]

The fitting term may have negative curvature, but on bounded anti-symmetric
active sets this negative part is uniformly controlled; no metric annulus is
needed for the following fitting estimate.

\begin{lemma}[Fitting-curvature lower bound on anti-symmetric active balls]
\label{lem:fitting-curvature-lower-bound-Mminus-smoothness}
\textbf{Assumptions used.}
\cref{assumption:on-sphere,assumption:odd-symmetric-teacher},
together with the finite active odd-block representation of the fitting term.

Fix \(R_{\mathrm{act}}<\infty\).  Consider anti-symmetric branch points
\[
    \mu=\mu_1\otimes\mu_2\in\submanifoldforsaddlepoint,
    \qquad
    \Theta(\mu)=(0,\mathbf A),
    \qquad
    \|\mathbf A\|_{\mathcal H_{\mathcal O}}\le R_{\mathrm{act}}.
\]
This active-radius condition can always be met by taking
\(R_{\mathrm{act}}=\sqrt{|\mathcal O|}\).  Indeed,
\(\|T_{i,n}\|_{\mathrm{HS}}\le
\int\|w^{\otimes n}\|_{\mathrm{HS}}\,\dd\mu_i(w)=1\), so
\[
 \|A_n\|_{\mathrm{HS}}
 \le\frac{\|T_{1,n}\|_{\mathrm{HS}}+\|T_{2,n}\|_{\mathrm{HS}}}{2}\le1,
 \qquad
 \|\mathbf A\|_{\mathcal H_{\mathcal O}}^2
 =\sum_{n\in\mathcal O}\|A_n\|_{\mathrm{HS}}^2\le|\mathcal O|.
\]
Let \(v=(v_1,v_2)\) be arbitrary smooth spherical tangent perturbations,
\[
    v_i(w)\in T_w\Sphere^{d_w-1},
    \qquad
    w\cdot v_i(w)=0.
\]
The base point is on the anti-symmetric branch, but \(v\) is an arbitrary
ambient product tangent field and need not be tangent to that branch.
Let
\[
    R_{i,h}(w)
    \coloneq
    \frac{w+h v_i(w)}{\|w+h v_i(w)\|},
    \qquad
    \mu_{i,h}\coloneq (R_{i,h})_\#\mu_i,
    \qquad
    \mu_h\coloneq \mu_{1,h}\otimes\mu_{2,h}.
\]
Then there exists a constant
\[
    C_{\mathrm{fit}}^{-}(R_{\mathrm{act}})<\infty
\]
depending only on \(R_{\mathrm{act}}\), the finite active block \(\mathcal O\), the Hermite coefficients, and the teacher block, such that
\begin{equation}
\label{eq:fitting-curvature-lower-bound-Mminus-derived}
    \frac{\dd^2}{\dd h^2}\Big|_{h=0}
    F_{\mathrm{fit}}(\mu_h)
    \ge
    -C_{\mathrm{fit}}^{-}(R_{\mathrm{act}})
    \|v\|_{L^2(\mu)}^2,
\end{equation}
where
\[
    \|v\|_{L^2(\mu)}^2
    \coloneq
    \|v_1\|_{L^2(\mu_1)}^2
    +
    \|v_2\|_{L^2(\mu_2)}^2.
\]
Consequently, the abstract lower bound
\[
    \frac{\dd^2}{\dd h^2}\Big|_{h=0}
    F_{\mathrm{fit}}\bigl((R_{1,h})_\#\mu_1\otimes(R_{2,h})_\#\mu_2\bigr)
    \ge
    -C_{\mathrm{fit}}\|v\|_{L^2(\mu;\mathbb R^{d_w})}^2
\]
holds on the anti-symmetric active ball
\(\|\mathbf A\|_{\mathcal H_{\mathcal O}}\le R_{\mathrm{act}}\), with
\[
    C_{\mathrm{fit}}
    =
    C_{\mathrm{fit}}^{-}(R_{\mathrm{act}}).
\]
\end{lemma}

\begin{proof}
\noindent\textbf{Step 1: write the finite-moment second-derivative formula.}
The proof reduces the second variation to a finite-dimensional chain rule in
\(\Theta=(\mathbf S,\mathbf A)\).  Recall the finite active moment map from
\cref{eq:mf-finite-active-moment-map} and the ambient fitting polynomial from
\cref{eq:ambient-active-fit-functional}.
Since the fitting term is represented by the finite active odd block, we have
\[
    F_{\mathrm{fit}}(\mu_h)
    =
    \widetilde{\mathcal J}_{\mathrm{fit}}\bigl(\Theta(\mu_h)\bigr),
\]
where
\[
    \Theta(\mu_h)
    =
    \bigl(\mathbf S(h),\mathbf A(h)\bigr)
    \in
    \mathcal H_{\mathcal O}\times\mathcal H_{\mathcal O}.
\]
At \(h=0\), since \(\mu\in\submanifoldforsaddlepoint\),
\[
    \Theta(\mu)=(0,\mathbf A).
\]
For \(h\ne0\), an ambient perturbation may leave the branch, so both
components \(\mathbf S(h)\) and \(\mathbf A(h)\) are retained.
Set
\[
    \dot\Theta[v]
    \coloneq
    \frac{\dd}{\dd h}\Big|_{h=0}\Theta(\mu_h),
    \qquad
    \ddot\Theta[v]
    \coloneq
    \frac{\dd^2}{\dd h^2}\Big|_{h=0}\Theta(\mu_h).
\]
The ordinary chain rule gives
\begin{align}
\label{eq:fit-second-variation-chain-Mminus}
    \frac{\dd^2}{\dd h^2}\Big|_{h=0}
    F_{\mathrm{fit}}(\mu_h)
    =
    &\,
    D^2\widetilde{\mathcal J}_{\mathrm{fit}}(0,\mathbf A)
    [\dot\Theta[v],\dot\Theta[v]]
    \\
    &+
    \innerProd{
        \nabla\widetilde{\mathcal J}_{\mathrm{fit}}(0,\mathbf A)
    }{
        \ddot\Theta[v]
    }_{\mathcal H_{\mathcal O}\times\mathcal H_{\mathcal O}}.
\end{align}
The second term is required because the moment-coordinate path
\(h\mapsto\Theta(\mu_h)\) generally has the nonzero acceleration
\(\ddot\Theta[v]\).

\par\noindent\textbf{Step 2: bound the first and second moment derivatives.}
The fields use the norms of \(L^2(\mu_i;\mathbb R^{d_w})\); moment tensors
use their Hilbert--Schmidt norms; and \(\dot\Theta,\ddot\Theta\), and the
fitting gradient use the product norm of
\(\mathcal H_{\mathcal O}\times\mathcal H_{\mathcal O}\).
For each \(n\in\mathcal O\),
\[
    T_{i,n}(h)
    =
    \int R_{i,h}(w)^{\otimes n}\,\dd\mu_i(w).
\]
Tangency \(w^\top v_i(w)=0\) gives
\(\|w+hv_i(w)\|^2=1+h^2\|v_i(w)\|^2\). Expanding its inverse square
root in the normalized retraction yields
\begin{equation}
\label{eq:mf-retraction-second-jet}
\begin{aligned}
 R_{i,h}(w)&=\frac{w+hv_i(w)}{\sqrt{1+h^2\|v_i(w)\|^2}}
 =w+hv_i(w)-\tfrac12h^2\|v_i(w)\|^2w+O(h^3),\\
 \dot R_{i,0}(w)&=v_i(w),\qquad
 \ddot R_{i,0}(w)=-\|v_i(w)\|^2w,\qquad
 (I-ww^\top)\ddot R_{i,0}(w)=0.
\end{aligned}
\end{equation}
These derivatives agree with those of \(h\mapsto\EXP_w(hv_i(w))\)
from \cref{eq:mf-spherical-EXP}; the two curves need not agree beyond
second order. Applying the product rule to \(R_{i,h}(w)^{\otimes n}\),
we obtain
\begin{align}
    \dot T_{i,n}[v_i]
    &=
    n\int
    \sym{w^{\otimes(n-1)}\otimes v_i(w)}
    \,\dd\mu_i(w),
    \\
    \ddot T_{i,n}[v_i]
    &=
    n(n-1)\int
    \sym{w^{\otimes(n-2)}
        \otimes v_i(w)\otimes v_i(w)}
    \,\dd\mu_i(w)
    \\
    &\qquad
    -
    n\int
    \|v_i(w)\|^2 w^{\otimes n}
    \,\dd\mu_i(w).
\end{align}
The term containing \(n(n-1)\) is omitted when \(n=1\).
Because \(\|w\|=1\), Cauchy--Schwarz yields
\begin{equation}
\label{eq:dot-T-bound-fit-smoothness}
    \|\dot T_{i,n}[v_i]\|_{\HS}
    \le
    n\|v_i\|_{L^2(\mu_i)},
\end{equation}
and
\begin{equation}
\label{eq:ddot-T-bound-fit-smoothness}
    \|\ddot T_{i,n}[v_i]\|_{\HS}
    \le
    n^2\|v_i\|_{L^2(\mu_i)}^2.
\end{equation}
Recall the active coordinates \(\Theta=(\mathbf S,\mathbf A)\), where
\(S_n=(T_{1,n}+T_{2,n})/2\) and \(A_n=(T_{1,n}-T_{2,n})/2\).
The parallelogram identity gives
\[
 \|\dot\Theta\|^2=\frac12\sum_{n\in\mathcal O}
       (\|\dot T_{1,n}\|_{\HS}^2+\|\dot T_{2,n}\|_{\HS}^2),
 \qquad
 \|\ddot\Theta\|^2=\frac12\sum_{n\in\mathcal O}
       (\|\ddot T_{1,n}\|_{\HS}^2+\|\ddot T_{2,n}\|_{\HS}^2).
\]
Here both norms on \(\Theta\) are in
\(\mathcal H_{\mathcal O}\times\mathcal H_{\mathcal O}\).
Using \cref{eq:dot-T-bound-fit-smoothness,eq:ddot-T-bound-fit-smoothness}
and \(a^2+b^2\le(a+b)^2\) for nonnegative \(a,b\), we obtain
\begin{equation}
\label{eq:dot-Theta-bound-fit-smoothness}
 \|\dot\Theta[v]\|\le
 \left(\sum_{n\in\mathcal O}n^2\right)^{1/2}\|v\|_{L^2(\mu)},
\end{equation}
\begin{equation}
\label{eq:ddot-Theta-bound-fit-smoothness}
 \|\ddot\Theta[v]\|\le
 \left(\sum_{n\in\mathcal O}n^4\right)^{1/2}\|v\|_{L^2(\mu)}^2.
\end{equation}
For example, the square of the second left-hand side is at most
\(\tfrac12\sum_n n^4(\|v_1\|_{L^2(\mu_1)}^4+
\|v_2\|_{L^2(\mu_2)}^4)\), which is at most
\(\sum_n n^4\|v\|_{L^2(\mu)}^4\).

\par\noindent\textbf{Step 3: choose the curvature bound and apply the chain rule.}
The set \(\mathcal O\) is finite and each tensor block is finite-dimensional.
Hence
\(\{(0,\mathbf A):\|\mathbf A\|_{\mathcal H_{\mathcal O}}
\le R_{\mathrm{act}}\}\) is compact.  The fitting functional
\(\widetilde{\mathcal J}_{\mathrm{fit}}\) is a polynomial, so its first two
derivatives are continuous and have finite suprema on this set.
Define the fitting curvature constant directly by
\begin{equation}
\label{eq:def-Cfit-Mminus-derived}
\begin{aligned}
 C_{\mathrm{fit}}^-(R_{\mathrm{act}}):={}&
 \left(\sum_{n\in\mathcal O}n^2\right)
 \sup_{\|\mathbf A\|\le R_{\mathrm{act}}}
       \|D^2\widetilde{\mathcal J}_{\mathrm{fit}}(0,\mathbf A)\|_{\mathrm{op}}\\
 &+\left(\sum_{n\in\mathcal O}n^4\right)^{1/2}
 \sup_{\|\mathbf A\|\le R_{\mathrm{act}}}
       \|\nabla\widetilde{\mathcal J}_{\mathrm{fit}}(0,\mathbf A)\|.
\end{aligned}
\end{equation}
The gradient norm is the product \(\mathcal H_{\mathcal O}\) norm.
The Hessian norm in \cref{eq:def-Cfit-Mminus-derived} is the operator norm of
\[
 D^2\widetilde{\mathcal J}_{\mathrm{fit}}(0,\mathbf A):
 \mathcal H_{\mathcal O}\times\mathcal H_{\mathcal O}
 \longrightarrow
 \mathcal H_{\mathcal O}\times\mathcal H_{\mathcal O}.
\]
The chain rule \cref{eq:fit-second-variation-chain-Mminus} and
Cauchy--Schwarz now give
\[
\begin{aligned}
 \left.\frac{\dd^2}{\dd h^2}F_{\mathrm{fit}}(\mu_h)\right|_0
 &\ge-\|D^2\widetilde{\mathcal J}_{\mathrm{fit}}(0,\mathbf A)\|_{\mathrm{op}}
                   \|\dot\Theta\|^2
       -\|\nabla\widetilde{\mathcal J}_{\mathrm{fit}}(0,\mathbf A)\|
                   \|\ddot\Theta\|\\
 &\ge-C_{\mathrm{fit}}^-(R_{\mathrm{act}})\|v\|_{L^2(\mu)}^2.
\end{aligned}
\]
This proves \cref{eq:fitting-curvature-lower-bound-Mminus-derived} with
the constant defined in \cref{eq:def-Cfit-Mminus-derived}.
\end{proof}

The metric regularizer relies on the positive Hessian of squared spherical distance inside each fixed cap.
\begin{lemma}[Spherical Hessian of the fixed squared-distance cost]
\label{lem:spherical-squared-distance-Hessian-on-caps}
\textbf{Assumptions used.} \cref{assumption:on-sphere}.

Let
\[
    c_y(w)=\frac12d_\Sphere(w,y)^2,
    \qquad
    y\in\Sphere^{d_w-1}.
\]
On the geodesic ball \(d_\Sphere(w,y)\le r_{\mathrm{cap}}<\pi/2\), the spherical Hessian satisfies
\begin{equation}
\label{eq:spherical-distance-Hessian-lower-bound}
    \nabla_\Sphere^2c_y(w)[v,v]
    \ge
    \kappa_{\mathrm{cap}}\|v\|^2,
    \qquad
    \kappa_{\mathrm{cap}}
    \coloneq
    r_{\mathrm{cap}}\cot r_{\mathrm{cap}}>0.
\end{equation}
Consequently, for \(c_\star(w)=\frac12d_\Sphere(w,\pi_\star(w))^2\),
\begin{equation}
\label{eq:cstar-Hessian-lower-bound}
    \nabla_\Sphere^2c_\star(w)[v,v]
    \ge
    \kappa_{\mathrm{cap}}
    \|v\|^2
\end{equation}
for every \(w\in K_+(r_{\mathrm{cap}})\cup K_-(r_{\mathrm{cap}})\) and every
\(v\in T_w\Sphere^{d_w-1}\).
\end{lemma}

\begin{proof}
For \(0<r=d_\Sphere(w,y)<\pi/2\), define
\(e=(y-\cos r\,w)/\sin r\).  Since \(w^\top y=\cos r\),
\[
 w^\top e=\frac{w^\top y-\cos r}{\sin r}=0,
 \qquad
 \|e\|_{\mathbb R^{d_w}}^2
 =\frac{1-2\cos r\,y^\top w+\cos^2r}{\sin^2r}=1.
\]
Thus \(e\in T_w\Sphere^{d_w-1}\) is the unit tangent pointing from \(w\)
toward \(y\).
Put \(x(t)=y^\top\EXP_w(tv)\) and \(f(x)=\tfrac12(\arccos x)^2\).
By \cref{eq:mf-spherical-EXP},
\[
 x(0)=\cos r,\quad x'(0)=y^\top v=\sin r\,\langle e,v\rangle,
 \quad x''(0)=-\|v\|^2\cos r,
\]
and direct differentiation gives
\[
 f'(x)=-\frac{\arccos x}{\sqrt{1-x^2}},\qquad
 f''(x)=\frac1{1-x^2}
        -\frac{x\arccos x}{(1-x^2)^{3/2}}.
\]
Substituting \(x=\cos r\) yields
\[
 f'(\cos r)=-\frac r{\sin r},\qquad
 f''(\cos r)=\frac{\sin r-r\cos r}{\sin^3r}.
\]
For sufficiently small \(|t|\), the EXP path does not reach the antipode of
\(y\), so the principal \(\arccos\) gives the spherical distance.  Hence
\[
 c_y(\EXP_w(tv))
 =\frac12d_\Sphere(\EXP_w(tv),y)^2
 =\frac12[\arccos(y^\top\EXP_w(tv))]^2
 =f(x(t)).
\]
The one-variable chain rule now gives
\begin{equation}
\label{eq:mf-spherical-cost-Hessian-formula}
\begin{aligned}
 \nabla_\Sphere^2c_y(w)[v,v]
 &=f''(\cos r)(x'(0))^2+f'(\cos r)x''(0)\\
 &=r\cot r\,\|v\|^2+(1-r\cot r)\langle v,e\rangle^2.
\end{aligned}
\end{equation}
To display the eigendirections, set
\(v_\parallel=\langle v,e\rangle e\) and
\(v_\perp=v-v_\parallel\).  Since
\(\|v\|_{\mathbb R^{d_w}}^2=\|v_\parallel\|_{\mathbb R^{d_w}}^2+
\|v_\perp\|_{\mathbb R^{d_w}}^2\),
\[
 \nabla_\Sphere^2c_y(w)[v,v]
 =\|v_\parallel\|_{\mathbb R^{d_w}}^2
  +r\cot r\,\|v_\perp\|_{\mathbb R^{d_w}}^2.
\]
The radial eigenvalue is therefore \(1\), and the orthogonal eigenvalue is
\(r\cot r\). Moreover,
\[
 (r\cot r)'=\frac{\sin r\cos r-r}{\sin^2r}\le0,\qquad
 r-\sin r\cos r=\int_0^r2\sin^2s\,\dd s\ge0,
 \qquad \lim_{r\to0}r\cot r=1.
\]
The integral identity shows that \((r\cot r)'\le0\).  Therefore
\(1\ge r\cot r\ge r_{\mathrm{cap}}\cot r_{\mathrm{cap}}>0\) for
\(0<r\le r_{\mathrm{cap}}\), so both eigenvalues are at least
\(\kappa_{\mathrm{cap}}\).
At \(w=y\), the identity
\(c_y(\EXP_y(tv))=t^2\|v\|_{\mathbb R^{d_w}}^2/2\) for small \(t\)
gives Hessian \(\|v\|_{\mathbb R^{d_w}}^2\) without introducing \(e\).
On each cap, \(\pi_\star(w)\) is the fixed point \(a_\star\) or
\(-a_\star\), so the calculation does not differentiate across a cap
boundary.  Substituting
\(y=a_\star\) or \(y=-a_\star\) gives the claim for the fixed-assignment
cost \(c_\star(w)=d_\Sphere(w,\pi_\star(w))^2/2\)
from \cref{eq:mf-fixed-assignment-cost,eq:mf-cap-class}.
\end{proof}

Combining fixed assignment with the spherical distance Hessian gives a positive spherical product pushforward second variation for the metric regularizer.
\begin{lemma}[Metric-power regularizer Hessian on fixed-assignment caps]
\label{lem:W2-power-regularizer-Hessian-on-fixed-caps}
\textbf{Assumptions used.} \cref{assumption:on-sphere,assumption:odd-symmetric-teacher}, together with \cref{lemma-fixed-assignment,lem:spherical-squared-distance-Hessian-on-caps}.

Let \(\mu_i\in\mathcal P_\pm(r_{\mathrm{cap}})\), and let \(v=(v_1,v_2)\) be smooth tangent perturbations whose spherical retractions preserve the fixed-assignment caps for sufficiently small \(|h|\).  Then
\begin{align}
\label{eq:W2-power-Hessian-lower-bound-fixed-caps}
    D^2_{W_2}F_{\mathrm{reg}}(\mu_1,
    \mu_2)[v,v]
    \ge
    \lambdaReg\,
    \kappa_{\mathrm{cap}}
    Z_\pm(\mu_1,
    \mu_2)^{\delta_W/2}
    \|v\|_{L^2(\mu)}^2.
\end{align}
In particular, if \(Z_\pm(\mu_1,
\mu_2)\ge\zeta_{\min}^2\), then
\begin{equation}
\label{eq:W2-power-Hessian-lower-bound-annulus}
    D^2_{W_2}F_{\mathrm{reg}}(
    \mu_1,
    \mu_2)[v,v]
    \ge
    \lambdaReg\,
    \kappa_{\mathrm{cap}}
    \zeta_{\min}^{\delta_W}
    \|v\|_{L^2(\mu)}^2.
\end{equation}
At \(Z_\pm=0\), the regularizer has zero quadratic variation because \(p_W=2+\delta_W>2\).
\end{lemma}

\begin{proof}
\noindent\textbf{Step 1: bound the positive-distance second derivative.}
By \cref{eq:mf-retraction-second-jet}, the normalized retraction used here
and the spherical EXP path have the same first and second derivatives at
\(h=0\).  Since the regularizer is smooth when \(Z_\pm>0\), its second
derivative along the retraction is therefore the one computed for the EXP
path in \cref{eq:second-variation-W2-dirac-metric-regularizer}.  In that
formula, the second term is nonnegative because
\(p_W-2=\delta_W>0\).  Substituting the pointwise Hessian bound
\cref{eq:cstar-Hessian-lower-bound} into the first term gives
\begin{align*}
 D^2_{W_2}F_{\mathrm{reg}}(\mu_1,\mu_2)[v,v]
 &\ge
 \lambdaReg\kappa_{\mathrm{cap}}Z_\pm^{p_W/2-1}
 \sum_{i=1}^2\int
 \|v_i(w)\|_{\mathbb R^{d_w}}^2\,\dd\mu_i(w)\\
 &=
 \lambdaReg\kappa_{\mathrm{cap}}Z_\pm^{\delta_W/2}
 \|v\|_{L^2(\mu;\mathbb R^{d_w})}^2,
\end{align*}
which is \cref{eq:W2-power-Hessian-lower-bound-fixed-caps} for
\(Z_\pm>0\).  If \(Z_\pm\ge\zeta_{\min}^2\), then \(\delta_W>0\) implies
\[
 Z_\pm^{\delta_W/2}
 \ge (\zeta_{\min}^2)^{\delta_W/2}
 =\zeta_{\min}^{\delta_W},
\]
and \cref{eq:W2-power-Hessian-lower-bound-annulus} follows.

\par\noindent\textbf{Step 2: compute the ordinary second derivative at the target.}
For \(Z_\pm=0\), both marginals equal
\(\nu_\star^\pm=(\delta_{a_\star}+\delta_{-a_\star})/2\).
At a pole \(y=\varepsilon a_\star\), tangency gives
\(y^\top R_{i,h}(y)=(1+h^2\|v_i(y)\|^2)^{-1/2}\), so its distance
from the pole is \(\arctan(|h|\|v_i(y)\|)\). The fixed-assignment
formula \cref{eq:fixed-assignment-W2-square-formula} consequently gives
\begin{equation}
\label{eq:mf-zero-distance-retraction}
 Z_\pm(\mu_h)=\frac12\sum_{i=1}^2\sum_{\varepsilon\in\{-1,1\}}
 \arctan^2(h\|v_i(\varepsilon a_\star)\|)
 =h^2\|v\|_{L^2(\mu)}^2+O(h^4).
\end{equation}
If \(\|v\|_{L^2(\mu)}=0\), all four atoms stay fixed. Otherwise
\(a(h):=Z_\pm(\mu_h)/h^2\) extends to a smooth positive function at
zero with \(a(0)=\|v\|_{L^2(\mu)}^2\). Set
\(b(h):=(\lambdaReg/p_W)a(h)^{p_W/2}\), so the regularizer is
\(|h|^{p_W}b(h)\). For \(h\ne0\), differentiating twice gives
\[
\begin{aligned}
 F_{\mathrm{reg}}'(h)
 &=p_W\operatorname{sgn}(h)|h|^{p_W-1}b(h)+|h|^{p_W}b'(h),\\
 F_{\mathrm{reg}}''(h)
 &=p_W(p_W-1)|h|^{p_W-2}b(h)
 +2p_W\operatorname{sgn}(h)|h|^{p_W-1}b'(h)+|h|^{p_W}b''(h).
\end{aligned}
\]
The difference quotients at zero and these formulas show that the
regularizer is \(C^2\), with both derivatives zero there, because
\(p_W>2\). This proves the bound at the target as well.
\end{proof}

The next proposition is an abstract combination statement: at a fixed state,
it adds any available fitting lower bound to the metric lower bound.  On an
anti-symmetric active ball, the fitting input has already been verified in
\cref{lem:fitting-curvature-lower-bound-Mminus-smoothness}; the subsequent
corollary makes that substitution.

\begin{proposition}[Ambient second-variation bound on fixed metric annuli]
\label{prop:sqchi2-away-from-origin-Wasserstein-convexity}
\textbf{Assumptions used.} \cref{assumption:on-sphere,assumption:odd-symmetric-teacher}, together with \cref{lem:W2-power-regularizer-Hessian-on-fixed-caps}.

Fix \(0<\zeta_{\min}<\infty\), and consider a state in a fixed-assignment
cap region satisfying
\[
    \mu_i\in\mathcal P_\pm(r_{\mathrm{cap}}),
    \qquad
    Z_\pm(\mu_1,
    \mu_2)
    \ge
    \zeta_{\min}^2.
\]
Suppose that, on the region under consideration, the fitting lower bound
\begin{equation}
\label{eq:fit-curvature-lower-bound-abstract}
    \frac{\dd^2}{\dd h^2}\Big|_{h=0}
    F_{\mathrm{fit}}\bigl((R_{1,h})_\#\mu_1\otimes(R_{2,h})_\#\mu_2\bigr)
    \ge
    -C_{\mathrm{fit}}\|v\|_{L^2(\mu)}^2
\end{equation}
is available for every smooth tangent perturbation \(v=(v_1,v_2)\) whose
retractions stay in the fixed-assignment caps for small \(|h|\).  Then the
full objective satisfies the following pointwise bound at that state:
\begin{equation}
\label{eq:sqchi2-away-from-origin-W2-convexity-bound}
    \innerProd{v}{\mathcal H_{\Sphere,\mu}v}_{L^2(\mu;\mathbb R^{d_w})}
    \ge
    \left(
        \lambdaReg\kappa_{\mathrm{cap}}\zeta_{\min}^{\delta_W}
        -C_{\mathrm{fit}}
    \right)
    \|v\|_{L^2(\mu;\mathbb R^{d_w})}^2.
\end{equation}
In particular, if
\[
    \lambdaReg>
    \frac{C_{\mathrm{fit}}}{\kappa_{\mathrm{cap}}\zeta_{\min}^{\delta_W}},
\]
then the spherical product pushforward second variation is strictly positive in
every nonzero admissible tangent direction at each point of the region.
\end{proposition}

\begin{proof}
By \cref{lem:mf-full-spherical-Hessian}, the spherical product pushforward
second variation is
\[
 D_{W_2}^2F(\mu)[v,v]
 =\innerProd{v}{\mathcal H_{\Sphere,\mu}v}_{L^2(\mu;\mathbb R^{d_w})}.
\]
The two inputs give
\begin{align*}
 D_{W_2}^2F_{\mathrm{fit}}(\mu)[v,v]
 &\ge -C_{\mathrm{fit}}\|v\|_{L^2(\mu;\mathbb R^{d_w})}^2,
 &&\text{by \cref{eq:fit-curvature-lower-bound-abstract},}\\
 D_{W_2}^2F_{\mathrm{reg}}(\mu)[v,v]
 &\ge \lambdaReg\kappa_{\mathrm{cap}}\zeta_{\min}^{\delta_W}
        \|v\|_{L^2(\mu;\mathbb R^{d_w})}^2,
 &&\text{by \cref{eq:W2-power-Hessian-lower-bound-annulus}.}
\end{align*}
Adding these inequalities proves
\cref{eq:sqchi2-away-from-origin-W2-convexity-bound}.
\end{proof}

\begin{remark}[Pointwise scope]
The annulus need not be geodesically convex.  A strong-convexity inequality
between two distinct states follows only along a comparison geodesic that
remains in a region where
\cref{eq:sqchi2-away-from-origin-W2-convexity-bound} holds at every point.
\end{remark}

Specializing this bound to the anti-symmetric branch gives positive ambient
curvature at the prescribed fixed-assignment annular states.

\begin{corollary}[Ambient curvature at anti-symmetric annular states]
\label{cor:branchwise-positivity-Wasserstein-Mminus}
\textbf{Assumptions used.}
\cref{assumption:on-sphere,assumption:odd-symmetric-teacher}, together with
\cref{prop:sqchi2-away-from-origin-Wasserstein-convexity}.

Fix \(R_{\mathrm{act}}<\infty\).  Let \(\mu_-\in\submanifoldforsaddlepoint\)
satisfy \(\|\mathbf A(\mu_-)\|_{\mathcal H_{\mathcal O}}\le R_{\mathrm{act}}\) and
\[
    \mu_{-,i}\in\mathcal P_\pm(r_{\mathrm{cap}}),
    \qquad
    Z_\pm(\mu_{-,1},\mu_{-,2})
    \ge
    \zeta_{\min}^2.
\]
Then \cref{eq:sqchi2-away-from-origin-W2-convexity-bound} holds with
\(C_{\mathrm{fit}}=C_{\mathrm{fit}}^{-}(R_{\mathrm{act}})\) from
\cref{eq:fitting-curvature-lower-bound-Mminus-derived,eq:def-Cfit-Mminus-derived}.
In particular, every nonzero
admissible product tangent field has positive spherical product pushforward
second variation when
\(\lambdaReg\kappa_{\mathrm{cap}}\zeta_{\min}^{\delta_W}>
C_{\mathrm{fit}}^{-}(R_{\mathrm{act}})\).  The field need not be tangent
to \(\submanifoldforsaddlepoint\).
\end{corollary}

%% file: appendix_mf_dirac/EMN_rewrite_descent_v8.tex
\subsection{Descent direction to escape the saddle point on the antisymmetric branch}
\label{subsec:descent-direction-to-escape}

This subsection uses the transverse teacher imbalance (\cref{assumption:quantitative-unweighted-two-sided-transverse-imbalance}) to construct a
same-density perturbation that escapes from the antipodal saddle.

The imbalance assumption says that the teacher has a coherent first-order
transverse bias across two-sided latitude slices around the antipodal axis. Here we show that this bias is visible to the active odd Hermite features, and
therefore produces a same-density tangent direction that decreases the fitting
loss to second order at the antipodal Dirac point.

\begin{remark}[Comparison with the zero-output saddle argument of \citet{yamamoto2025hessianguided}]
    The algebraic mechanism is the same as the known zero-output saddle calculation in
the matrix-decomposition example (Proposition G.1 in \citet{yamamoto2025hessianguided}): when the
student output vanishes, a perturbation that creates a first-order feature
\(g\) changes the product output only at order \(h^2\), producing the quadratic
cross term
\[
    -\E_z[(h^o(z)^\top g(z))^2].
\]
However, the affine mixture curve used there does not preserve our product
spherical ansatz.  We therefore rebuild the argument
with a Dirac-supported spherical tangent perturbation at the antipodal pair.
The transverse imbalance assumption is precisely the condition ensuring that one such tangent feature is visible to the teacher.
\end{remark}

The first step turns the sliced transverse imbalance assumption into a nonzero active-weighted transverse teacher moment.

\begin{lemma}[Quantitative control of the active-weighted transverse moment]
\label{lem:quantitative-control-weighted-transverse-moment}
\textbf{Assumptions used.}
\cref{def:two-sided-transverse-teacher-moment-measure} and 
\cref{assumption:quantitative-unweighted-two-sided-transverse-imbalance}.

Set \(a=a_\star\), the anchor in that imbalance assumption.  Define
\[
    \kappa_{\mathcal O}(r)
    \coloneq
    \frac12
    \sum_{n\in\mathcal O}
    n\|a_n\|_2^2 r^{n-1},
    \qquad r\in[0,1].
\]
Assume at least one active coefficient \(\|a_n\|_2\) is nonzero.  Since every
\(n\in\mathcal O\) is odd,
\[
    \kappa_{\mathcal O}(r)>0
    \qquad
    \forall r\in(0,1].
\]
Let
\[
    b_a^o
    \coloneq
    \int_{\Sphere^{d_w-1}}
    \kappa_{\mathcal O}(|a^\top w|)
    \Pi_{a^\perp}w
    \,\dd\nu^o(w)
    \in a^\perp .
\]
Equivalently,
\[
    b_a^o
    =
    \int_{[0,1]}
    \kappa_{\mathcal O}(r)\,\dd\mathfrak m_a^o(r).
\]
Define
\[
    \kappa_{\min}(r_{\mathrm{vis}})
    \coloneq
    \inf_{r\in[r_{\mathrm{vis}},1]}
    \kappa_{\mathcal O}(r).
\]
Then
\[
    \kappa_{\min}(r_{\mathrm{vis}})>0
\]
and
\[
    \|b_a^o\|
    \ge
    \langle e_\star,b_a^o\rangle
    \ge
    \kappa_{\min}(r_{\mathrm{vis}})\gamma_\perp.
\]
In particular,
\[
    b_a^o\neq0.
\]
\end{lemma}

\begin{proof}
We first rewrite the weighted transverse moment using the latitude
disintegration.  By
\cref{def:two-sided-transverse-teacher-moment-measure}, for every bounded
Borel function \(\psi:[0,1]\to\mathbb R\),
\[
    \int_{\Sphere^{d_w-1}}
    \psi(r_a(w))\Pi_{a^\perp}w\,\dd\nu^o(w)
    =
    \int_{[0,1]}
    \psi(r)\,\dd\mathfrak m_a^o(r).
\]
Taking
\[
    \psi(r)=\kappa_{\mathcal O}(r)
\]
gives
\[
    b_a^o
    =
    \int_{[0,1]}
    \kappa_{\mathcal O}(r)\,\dd\mathfrak m_a^o(r).
\]
Therefore,
\[
    \langle e_\star,b_a^o\rangle
    =
    \int_{[0,1]}
    \kappa_{\mathcal O}(r)\,
    \dd\mathfrak m_{a,e_\star}^o(r).
\]

Next, since every \(n\in\mathcal O\) is odd, \(n-1\) is even.  Hence
\[
    r^{n-1}>0
    \qquad
    \forall r\in(0,1]
\]
for every active degree \(n\).  Because at least one active coefficient is
nonzero,
\[
    \kappa_{\mathcal O}(r)>0
    \qquad
    \forall r\in(0,1].
\]
Since \(r_{\mathrm{vis}}\in(0,1)\) and \(\kappa_{\mathcal O}\) is continuous,
\[
    \kappa_{\min}(r_{\mathrm{vis}})
    =
    \inf_{r\in[r_{\mathrm{vis}},1]}
    \kappa_{\mathcal O}(r)
    >0.
\]

By
\cref{assumption:quantitative-unweighted-two-sided-transverse-imbalance},
\(\mathfrak m_{a,e_\star}^o\) is a positive finite measure on \([0,1]\).
Thus
\[
    \langle e_\star,b_a^o\rangle
    =
    \int_{[0,1]}
    \kappa_{\mathcal O}(r)\,
    \dd\mathfrak m_{a,e_\star}^o(r)
    \ge
    \int_{[r_{\mathrm{vis}},1]}
    \kappa_{\mathcal O}(r)\,
    \dd\mathfrak m_{a,e_\star}^o(r).
\]
Using the definition of \(\kappa_{\min}(r_{\mathrm{vis}})\), we obtain
\[
    \int_{[r_{\mathrm{vis}},1]}
    \kappa_{\mathcal O}(r)\,
    \dd\mathfrak m_{a,e_\star}^o(r)
    \ge
    \kappa_{\min}(r_{\mathrm{vis}})
    \mathfrak m_{a,e_\star}^o([r_{\mathrm{vis}},1]).
\]
The quantitative strict imbalance gives
\[
    \mathfrak m_{a,e_\star}^o([r_{\mathrm{vis}},1])
    \ge
    \gamma_\perp.
\]
Consequently,
\[
    \langle e_\star,b_a^o\rangle
    \ge
    \kappa_{\min}(r_{\mathrm{vis}})\gamma_\perp.
\]
Finally, since \(\|e_\star\|=1\),
\[
    \|b_a^o\|
    \ge
    \langle e_\star,b_a^o\rangle.
\]
This proves the claim.
\end{proof}

We next encode how a tangent displacement of the antipodal atoms is seen by the
teacher feature.

\begin{definition}[Tangential visibility matrix at an antipodal Dirac pair]
\label{def:tangential-visibility-matrix-dirac-pair}
Let
\[
    \nu_a^\pm
    \coloneq
    \frac12\delta_a+\frac12\delta_{-a}.
\]
For \(s\in a^\perp\), define the Dirac-supported active differential
\[
    \mathcal U_a(s)_n
    \coloneq
    \frac n2
    \sym{a^{\otimes(n-1)}\otimes s},
    \qquad n\in\mathcal O.
\]
Let \(g_{a,s}\) be the corresponding first-order student feature,
\begin{equation}
\label{eq:mf-visibility-first-order-feature}
 g_{a,s}(z):=\sum_{n\in\mathcal O}a_n
       \langle\Hertensor_n(z),\mathcal U_a(s)_n\rangle_{\rm HS}.
\end{equation}
This uses the finite expansion in \eqref{eq:activation-moment-expansion}.
Recall the teacher from \cref{assumption:odd-symmetric-teacher}:
\[
 h^o(z)=h_{\nu^o}(z)
 =\sum_{n\in\mathcal O}a_n\langle\Hertensor_n(z),T_n^o\rangle_{\rm HS},
 \qquad T_n^o=\int w^{\otimes n}\,d\nu^o(w).
\]
Equivalently, define the linear map
\[
    J_a(z):a^\perp\to\mathbb R^{d_h},
    \qquad
    J_a(z)s=g_{a,s}(z).
\]
For the teacher feature \(h^o\), define the positive semidefinite operator
on \(a^\perp\)
\[
    \mathsf B_a^o
    \coloneq
    \E_z
    \left[
        J_a(z)^\top h^o(z)h^o(z)^\top J_a(z)
    \right].
\]
We say that the teacher is tangentially visible at the antipodal pair
\(\{\pm a\}\) if
\[
    \lambda_{\max}(\mathsf B_a^o)>0.
\]
\end{definition}

The nonzero transverse teacher moment implies that some tangent direction has
nonzero correlation with the teacher output.

\begin{lemma}[Active-weighted transverse moment implies tangential visibility]
\label{lem:active-weighted-transverse-moment-implies-visibility}
\textbf{Assumptions used.}
\cref{assumption:on-sphere,assumption:odd-symmetric-teacher},
together with
\cref{def:tangential-visibility-matrix-dirac-pair}.

Let
\[
    b_a^o
    \coloneq
    \int_{\Sphere^{d_w-1}}
    \kappa_{\mathcal O}(|a^\top w|)
    \Pi_{a^\perp}w
    \,\dd\nu^o(w).
\]
Then
\[
    \mathsf B_a^o
    \succeq
    b_a^o(b_a^o)^\top.
\]
Consequently,
\[
    \lambda_{\max}(\mathsf B_a^o)
    \ge
    \|b_a^o\|^2.
\]
In particular, if \(b_a^o\neq0\), then the teacher is tangentially visible at
\(\{\pm a\}\).
\end{lemma}

\begin{proof}
For \(s\in a^\perp\), Hermite orthogonality gives
\[
    \E_z\bigl[g_{a,s}(z)^\top h^o(z)\bigr]
    =
    \sum_{n\in\mathcal O}
    \|a_n\|_2^2
    \left\langle
        \mathcal U_a(s)_n,T_n^o
    \right\rangle_{\HS}.
\]
By the definition of \(\mathcal U_a(s)\),
\[
    \left\langle
        \mathcal U_a(s)_n,T_n^o
    \right\rangle_{\HS}
    =
    \frac n2
    \left\langle
        \sym{a^{\otimes(n-1)}\otimes s},
        T_n^o
    \right\rangle_{\HS}.
\]
Since \(T_n^o\) is symmetric,
\[
    \left\langle
        \sym{a^{\otimes(n-1)}\otimes s},
        T_n^o
    \right\rangle_{\HS}
    =
    \left\langle
        s,
        T_n^o\contr{n-1}a^{\otimes(n-1)}
    \right\rangle.
\]
Using
\[
    T_n^o
    =
    \int_{\Sphere^{d_w-1}} w^{\otimes n}\,\dd\nu^o(w),
\]
and \(s\in a^\perp\), we obtain
\[
    \left\langle
        s,
        T_n^o\contr{n-1}a^{\otimes(n-1)}
    \right\rangle
    =
    \int_{\Sphere^{d_w-1}}
    (a^\top w)^{n-1}
    \langle s,\Pi_{a^\perp}w\rangle
    \,\dd\nu^o(w).
\]
Since \(n\) is odd, \(n-1\) is even, hence
\[
    (a^\top w)^{n-1}
    =
    |a^\top w|^{n-1}.
\]
Therefore
\[
    \E_z\bigl[g_{a,s}(z)^\top h^o(z)\bigr]
    =
    \left\langle s,b_a^o\right\rangle.
\]

Set
\[
    X(z)\coloneq J_a(z)^\top h^o(z)\in a^\perp.
\]
Then
\[
    \mathsf B_a^o=\E_z[X(z)X(z)^\top],
    \qquad
    \E_z[X(z)]=b_a^o.
\]
By Jensen's inequality in the Loewner order,
\[
    \mathsf B_a^o
    =
    \E_z[X(z)X(z)^\top]
    \succeq
    \E_z[X(z)]\E_z[X(z)]^\top
    =
    b_a^o(b_a^o)^\top.
\]
Thus
\[
    \lambda_{\max}(\mathsf B_a^o)\ge\|b_a^o\|^2.
\]
\end{proof}

Finally, the visible tangent direction yields a same-density perturbation with negative second-order fitting curvature at the antipodal saddle.
The proposition is an existence statement for a negative-curvature direction.
It does not by itself prove that the Hessian-guided perturbation selects this
particular same-density direction; that selection issue belongs to the
perturbation analysis rather than to the geometric verification.

\begin{proposition}[Dirac-supported transverse descent from unweighted transverse imbalance]
\label{lemma:descent-direction-antipodal-dirac-from-unweighted-imbalance}
\textbf{Assumptions used.}
\cref{assumption:on-sphere,assumption:odd-symmetric-teacher,assumption:quantitative-unweighted-two-sided-transverse-imbalance}.

Assume \(\mathcal O\ne\varnothing\).  Set \(a=a_\star\), so that the
perturbed Dirac pair is the fixed regularizer target.  Let
\[
    \nu_a^\pm
    \coloneq
    \frac12\delta_a+\frac12\delta_{-a},
    \qquad
    \mu^\dagger
    \coloneq
    \nu_a^\pm\otimes\nu_a^\pm.
\]
By                                                                      
\cref{lem:quantitative-control-weighted-transverse-moment},
\cref{lem:active-weighted-transverse-moment-implies-visibility},
\[
    \gamma_{\mathrm{esc}}
    \coloneq
    \lambda_{\max}(\mathsf B_a^o)
    \ge
    \kappa_{\min}(r_{\rm vis})^2\gamma_{\perp}^2
    >0.
\]
Let \(s_\star\in a^\perp\) be a unit eigenvector associated with
\(\gamma_{\mathrm{esc}}\).
Then there exists a same-density spherical
pushforward perturbation
\[
    \mu_h=\mu_{1,h}\otimes\mu_{2,h},
    \qquad
    \mu_{1,h}=\mu_{2,h},
    \qquad
    \mu_{i,0}=\nu_a^\pm,
\]
such that
\[
    \frac{\dd}{\dd h}\Big|_{h=0}F(\mu_h)=0
\]
and
\[
    F(\mu_h)
    =
    F(\mu^\dagger)
    -
    h^2\gamma_{\mathrm{esc}}
    +
    o(h^2).
\]
In particular,
\[
    \frac{\dd^2}{\dd h^2}\Big|_{h=0}F(\mu_h)
    =
    -2\gamma_{\mathrm{esc}}<0.
\]
Moreover, for the metric regularizer
\[
    F_{\mathrm{reg}}(\mu_1,\mu_2)
    =
    \frac{\lambdaReg}{p_W}
    Z_\pm(\mu_1,\mu_2)^{p_W/2},
    \qquad
    p_W=2+\delta_W,
\]
one has
\[
    F_{\mathrm{reg}}(\mu_h)=O(|h|^{2+\delta_W})=o(h^2).
\]
Thus the metric regularizer does not affect the quadratic descent term at
the antipodal Dirac saddle.
\end{proposition}

\begin{proof}
\noindent\textbf{Step 1: compute the moments at the antipodal pair.}
We first compute the active odd moments at the antipodal Dirac pair.  For
every odd \(n\in\mathcal O\),
\[
    T_n(\nu_a^\pm)
    =
    \frac12 a^{\otimes n}
    +
    \frac12(-a)^{\otimes n}
    =
    0.
\]
Therefore
\[
    \Theta(\mu^\dagger)=(0,0).
\]

\par\noindent\textbf{Step 2: construct the visible spherical perturbation.}

Recall that
\[
 a^\perp
 =\{s\in\mathbb R^{d_w}:\langle s,a\rangle_{\mathbb R^{d_w}}=0\}
 =T_a\Sphere^{d_w-1}=T_{-a}\Sphere^{d_w-1},
 \qquad
 \Pi_{a^\perp}=I-aa^\top,
\]
where the projection is the one used in
\cref{def:two-sided-transverse-teacher-moment-measure}.  Thus the visible
unit vector \(s_\star\in a^\perp\) from the statement is a common tangent
direction at both poles.  Set
\begin{equation}
\label{eq:mf-visible-tangent-field}
 v(w):=\frac12(I-ww^\top)s_\star,\qquad
 u_+:=v(a)=\frac{s_\star}{2},\qquad
 u_-:=v(-a)=\frac{s_\star}{2}.
\end{equation}
Indeed, \(w^\top v(w)=0\), so \(v(w)\in T_w\Sphere^{d_w-1}\).
Moreover, substituting \(w=a\) and \(w=-a\) and using
\(a^\top s_\star=0\) gives
\[
 v(a)=\tfrac12(I-aa^\top)s_\star=\tfrac12s_\star,
 \qquad
 v(-a)=\tfrac12(I-aa^\top)s_\star=\tfrac12s_\star.
\]
Hence the two atoms move in the same ambient transverse direction and
\(u_++u_-=s_\star\).  The factor \(1/2\) in \(v\) compensates for the
two atom masses \(1/2\), so that their summed first-order moment change is
exactly the differential \(\mathcal U_a(s_\star)\) defined in
\cref{def:tangential-visibility-matrix-dirac-pair}.
Use it in both marginals and define
\(\mu_{1,h}=\mu_{2,h}:=(R_h)_\#\nu_a^\pm\), where, for
\(w\in\{\pm a\}\),
\[
    R_h(w)=\frac{w+h v(w)}{\|w+h v(w)\|}.
\]
Since \(v(w)\in T_w\Sphere^{d_w-1}\), we have \(w^\top v(w)=0\), and hence
\[
    \frac{\dd}{\dd h}\Big|_{h=0}R_h(w)=v(w).
\]
Thus
\[
    \frac{\dd}{\dd h}\Big|_{h=0}R_h(w)^{\otimes n}
    =
    n\,\sym{w^{\otimes(n-1)}\otimes v(w)}.
\]
Consequently,
\begin{align}
    \dot T_n[v]
    &\coloneq
    \frac{\dd}{\dd h}\Big|_{h=0}T_n((R_h)_\#\nu_a^\pm)
    \\
    &=
    \frac n2
    \sym{a^{\otimes(n-1)}\otimes u_+}
    +
    \frac n2
    \sym{(-a)^{\otimes(n-1)}\otimes u_-}.
\end{align}
Because \(n\) is odd, \(n-1\) is even, and therefore
\[
    (-a)^{\otimes(n-1)}=a^{\otimes(n-1)}.
\]
Hence
\[
    \dot T_n[v]
    =
    \frac n2
    \sym{a^{\otimes(n-1)}\otimes(u_++u_-)}
    =
    \frac n2
    \sym{a^{\otimes(n-1)}\otimes s_\star}
    =
    \mathcal U_a(s_\star)_n.
\]
Since \(\mu_{1,h}=\mu_{2,h}\), we have
\[
    \dot{\mathbf S}[v]
    =
    \frac{\dot{\mathbf T}_1[v]+\dot{\mathbf T}_2[v]}{2}
    =
    \mathcal U_a(s_\star),
    \qquad
    \dot{\mathbf A}[v]
    =
    \frac{\dot{\mathbf T}_1[v]-\dot{\mathbf T}_2[v]}{2}
    =
    0.
\]
Thus this path leaves the active origin in the symmetric coordinate
\(\mathbf S\) with velocity \(\mathcal U_a(s_\star)\), while its
anti-symmetric coordinate \(\mathbf A\) remains zero.

\par\noindent\textbf{Step 3: expand the fitting term.}
Now we compute the model block.  By the active Hermite product formula,
\[
    \widetilde{\mathcal M}_p(\mathbf S,\mathbf A)
    =
    \sum_{\substack{
        n,m\in\mathcal O\\
        0\le r\le \min(n,m)\\
        p=n+m-2r
    }}
    c_{n,m,r}(a_na_m^\top)
    \otimes
    \sym{
        (S_n+A_n)\contr{r}(S_m-A_m)
    }.
\]
Along the same-density perturbation above,
\[
    \mathbf A(h)=0
\]
for every \(h\). The normalized atom maps constructed from the field in
\cref{eq:mf-visible-tangent-field} are smooth near $h=0$, so their finitely
many tensor moments are $C^2$ in $h$. Step 1 gives
$\mathbf S(0)=0$, and the derivative calculation in Step 2 gives
$\dot{\mathbf S}(0)=\mathcal U_a(s_\star)$. Taylor's formula therefore gives
\[
    \mathbf S(h)
    =
    h\,\mathcal U_a(s_\star)+O(h^2).
\]
Therefore
\begin{align}
    \widetilde{\mathcal M}_p(\Theta(\mu_h))
    &=
    \widetilde{\mathcal M}_p(\mathbf S(h),0)
    \\
    &=
    \sum_{\substack{
        n,m\in\mathcal O\\
        0\le r\le \min(n,m)\\
        p=n+m-2r
    }}
    c_{n,m,r}(a_na_m^\top)
    \otimes
    \sym{S_n(h)\contr{r}S_m(h)}
    \\
    &=
    h^2
    \underbrace{\sum_{\substack{
        n,m\in\mathcal O\\
        0\le r\le \min(n,m)\\
        p=n+m-2r
    }}
    c_{n,m,r}(a_na_m^\top)
    \otimes
    \sym{
        \mathcal U_a(s_\star)_n
        \contr{r}
        \mathcal U_a(s_\star)_m
    }}_{\eqqcolon  \mathcal Q_{p,+}(\mathcal U_a(s_\star))}
    +
    o(h^2).
\end{align}

We now expand the fitting term.  Recall that
\[
    F_{\mathrm{fit}}(\mu_h)
    =
    \frac12
    \sum_{p\ge0}
    \left\|
        \widetilde{\mathcal M}_p(\Theta(\mu_h))
        -
        \mathcal M_p^o
    \right\|_{\HS}^2.
\]
Using
\[
    \widetilde{\mathcal M}_p(\Theta(\mu_h))
    =
    h^2\mathcal Q_{p,+}(\mathcal U_a(s_\star))
    +
    o(h^2),
\]
we get
\begin{align}
    \frac12
    \left\|
        \widetilde{\mathcal M}_p(\Theta(\mu_h))
        -
        \mathcal M_p^o
    \right\|_{\HS}^2
    &=
    \frac12
    \left\|
        -\mathcal M_p^o
        +
        h^2\mathcal Q_{p,+}(\mathcal U_a(s_\star))
        +
        o(h^2)
    \right\|_{\HS}^2
    \\
    &=
    \frac12\|\mathcal M_p^o\|_{\HS}^2
    -
    h^2
    \left\langle
        \mathcal Q_{p,+}(\mathcal U_a(s_\star)),
        \mathcal M_p^o
    \right\rangle_{\HS}
    +
    o(h^2).
\end{align}
Summing over \(p\ge0\) proves
\[
    F_{\mathrm{fit}}(\mu_h)
    =
    F_{\mathrm{fit}}(\mu^\dagger)
    -
    h^2
    \sum_{p\ge0}
    \left\langle
        \mathcal Q_{p,+}(\mathcal U_a(s_\star)),
        \mathcal M_p^o
    \right\rangle_{\HS}
    +
    o(h^2).
\]

\par\noindent\textbf{Step 4: identify the visibility coefficient.}
It remains to identify the escape coefficient with the visibility matrix.
Use the first-order student feature $g_{a,s}(z)$ defined by the finite
Hermite sum in \eqref{eq:mf-visibility-first-order-feature}.
By the definition of the Hermite active blocks,
\[
    g_{a,s}(z)g_{a,s}(z)^\top
\]
has \(p\)-th active Hermite block
\[
    \mathcal Q_{p,+}(\mathcal U_a(s)).
\]
Similarly,
\[
    h^o(z)h^o(z)^\top
\]
has \(p\)-th active Hermite block \(\mathcal M_p^o\).  Therefore, by
orthogonality of the Hermite blocks,
\begin{align}
    \sum_{p\ge0}
    \left\langle
        \mathcal Q_{p,+}(\mathcal U_a(s)),
        \mathcal M_p^o
    \right\rangle_{\HS}
    &=
    \mathbb E_z
    \left[
        \left\langle
            g_{a,s}(z)g_{a,s}(z)^\top,
            h^o(z)h^o(z)^\top
        \right\rangle_{\mathrm F}
    \right]
    \\
    &=
    \mathbb E_z
    \left[
        \bigl(h^o(z)^\top g_{a,s}(z)\bigr)^2
    \right].
\end{align}
Now write \(g_{a,s}(z)=J_a(z)s\).  Then
\begin{align}
    \mathbb E_z
    \left[
        \bigl(h^o(z)^\top g_{a,s}(z)\bigr)^2
    \right]
    &=
    \mathbb E_z
    \left[
        s^\top J_a(z)^\top h^o(z)h^o(z)^\top J_a(z)s
    \right]
    \\
    &=
    s^\top
    \mathbb E_z
    \left[
        J_a(z)^\top h^o(z)h^o(z)^\top J_a(z)
    \right]
    s
    \\
    &=
    s^\top\mathsf B_a^o s.
\end{align}
Taking \(s=s_\star\) gives
\[
    \sum_{p\ge0}
    \left\langle
        \mathcal Q_{p,+}(\mathcal U_a(s_\star)),
        \mathcal M_p^o
    \right\rangle_{\HS}
    =
    s_\star^\top\mathsf B_a^o s_\star
    =
    \gamma_{\mathrm{esc}}.
\]
Hence
\[
    F_{\mathrm{fit}}(\mu_h)
    =
    F_{\mathrm{fit}}(\mu^\dagger)
    -
    h^2\gamma_{\mathrm{esc}}
    +
    o(h^2).
\]

\par\noindent\textbf{Step 5: compute the regularizer contribution.}

Recall that \(s_\star\) is the chosen unit transverse direction,
\(s_\star^\top a_\star=0\), and \(u_\pm=s_\star/2\). At the assigned
pole \(y=\pm a_\star\), the normalized perturbation satisfies
\[
 R_h(y)=\frac{y+(h/2)s_\star}{\sqrt{1+h^2/4}},\qquad
 y^\top R_h(y)=\frac1{\sqrt{1+h^2/4}}.
\]
The angle lies in \([0,\pi/2)\). Because
\(\cos(\arctan x)=1/\sqrt{1+x^2}\) for \(x\ge0\),
\[
 d_\Sphere(R_h(y),y)
 =\arccos\frac1{\sqrt{1+h^2/4}}=\arctan(|h|/2).
\]
For sufficiently small \(|h|\) both atoms retain their cap assignments.
Each marginal has two atoms of mass \(1/2\), and there are two marginals;
therefore \cref{eq:fixed-assignment-W2-square-formula} gives
\begin{equation}
\label{eq:mf-visible-path-metric-distance}
 Z_\pm(\mu_h)
 =\sum_{i=1}^2\sum_{\varepsilon\in\{-1,1\}}\frac12\arctan^2(|h|/2)
 =2\arctan^2(h/2)=O(h^2).
\end{equation}
For \(h\ne0\), the regularizer has the explicit factorization
\[
 F_{\mathrm{reg}}(\mu_h)=\frac{\lambdaReg}{p_W}|h|^{p_W}
      \left(\frac{2\arctan^2(h/2)}{h^2}\right)^{p_W/2}.
\]
The factor in parentheses extends to a smooth positive function at zero,
with value \(1/2\), because \(\arctan(h/2)=h/2+O(h^3)\).
The derivative calculation following \cref{eq:mf-zero-distance-retraction}
then gives \(F_{\mathrm{reg}}'(h)=O(|h|^{p_W-1})\) and
\(F_{\mathrm{reg}}''(h)=O(|h|^{p_W-2})\).
Since \(p_W=2+\delta_W>2\), the regularizer is \(C^2\) at zero, both
derivatives there are zero, and \(F_{\mathrm{reg}}(\mu_h)=o(h^2)\).
Thus the ordinary second derivative of the full objective exists and equals
\(-2\gamma_{\mathrm{esc}}\), as well as satisfying the displayed expansion.
Combining the fitting expansion and the regularizer estimate proves the claim.
\end{proof}

%% file: appendix_mf_dirac/EMN_rewrite_post_escape_convexity_v9.tex
\subsection{Ambient positive curvature near the same-density teacher}
\label{subsec:post-escape-convexity-on-the-same-density}

This subsection proves positive spherical product pushforward second variation
at, and near, the exact-fit same-density teacher.  It does not assert that an
escaping gradient-flow trajectory reaches this neighborhood, or that the
teacher is stationary for the regularized objective.

The descent direction in \cref{lemma:descent-direction-antipodal-dirac-from-unweighted-imbalance} creates a same-sign odd mode. This motivates the same-density branch
\begin{equation}
    \mathcal M_+
    \coloneq
    \left\{
        \mu_1\otimes \mu_2
        \mid
        \mu_1=\mu_2
    \right\}.
\end{equation}
On $\mathcal M_+$ we have $A_n=0$ for every $n\in \mathcal O$ and $S_n=T_{1,n}=T_{2,n}$. By \cref{prop:reference-fit-minimum-on-Mplus}, the reference zero-fit point of the data term lies on this branch.

\subsubsection{Invariance of the same-density branch}

We first check that the same-density branch is dynamically consistent with the product WGF.
\begin{proposition}[Invariance of $\mathcal M_+$ under the antipodal metric objective]
\label{lem:invariance-same-density-branch}
\textbf{Assumptions used.} \cref{assumption:on-sphere,assumption:odd-symmetric-teacher}, together with the exchange symmetry of $F$ in $(\mu_1,\mu_2)$.

If \(\mu_0=\mu_{1,0}\otimes \mu_{2,0}\in \mathcal M_+\) and \(\mu_t\) is the
unique corresponding product WGF solution on an interval \([0,T)\) in the
fixed-assignment cap class \(\mu_{i,t}\in\mathcal P_\pm(r_{\mathrm{cap}})\), \(i=1,2\), then
\begin{equation}
    \mu_t\in \mathcal M_+
    \qquad
    \forall t\in[0,T).
\end{equation}
\end{proposition}

\begin{proof}
\noindent\textbf{Step 1: exchange the two factors in the objective.}
If $\mu_1=\mu_2$, then $T_{1,n}=T_{2,n}$ for every $n\in \mathcal O$. 

The fitting part is invariant under interchanging the two factors because $\mu_1^o=\mu_2^o$. 

The metric regularizer is exchange-invariant because it depends on the two factors only through
\[
    Z_\pm(\mu_1,\mu_2)
    =
    W_2^2(\mu_1,\nu_\star^\pm)+W_2^2(\mu_2,\nu_\star^\pm).
\]

\par\noindent\textbf{Step 2: transform the characteristics and apply uniqueness.}
In particular, exchange symmetry gives
\[
 F(\mu_1,\mu_2)=F(\mu_2,\mu_1),\qquad
 g_1(\mu_1,\mu_2;w)=g_2(\mu_2,\mu_1;w),
\]
where \(g_i=\nabla_\Sphere(\delta F/\delta\mu_i)\), as in
\cref{eq:mf-branch-radial-notation}. For the swapped pair
\(\widetilde\mu_{1,t}=\mu_{2,t}\), \(\widetilde\mu_{2,t}=\mu_{1,t}\),
its first continuity equation is
\[
 \partial_t\widetilde\mu_{1,t}
 -\nabla_\Sphere\cdot
  (g_1(\widetilde\mu_{1,t},\widetilde\mu_{2,t};\cdot)\widetilde\mu_{1,t})
 =\partial_t\mu_{2,t}
 -\nabla_\Sphere\cdot(g_2(\mu_{1,t},\mu_{2,t};\cdot)\mu_{2,t})=0.
\]
The other equation follows by exchanging indices. Since
\(\mu_{1,0}=\mu_{2,0}\), the swapped pair has the same initial state.
Uniqueness on the stated interval gives
\(\widetilde\mu_{i,t}=\mu_{i,t}\), hence \(\mu_{1,t}=\mu_{2,t}\).
\end{proof}

\subsubsection{Positive curvature at and near the same-density teacher}

Let
\begin{equation}
 \nu^o:=\mu_1^o=\mu_2^o,
 \qquad
 \mu_+^o:=\nu^o\otimes\nu^o\in\mathcal M_+.
\end{equation}
At this state the student output equals the teacher output, so the fitting
residual vanishes. The next proposition combines the resulting nonnegative
fitting curvature with the strict metric curvature and shows that the margin
persists locally.

\begin{proposition}[Ambient positive curvature near a same-density teacher]
\label{prop:Mplus-teacher-local-positive-curvature}
\textbf{Assumptions used.}
\cref{assumption:on-sphere,assumption:odd-symmetric-teacher}.
Suppose
\[
 \nu^o\in\mathcal P_\pm(r_{\mathrm{cap}}),
 \qquad
 Z_o:=Z_\pm(\mu_+^o)
 =2W_{2,\Sphere}^2(\nu^o,\nu_\star^\pm)>0.
\]
Then there exist \(\rho_o>0\) and \(c_o>0\), depending only on the fixed
model parameters and the teacher, such that the following holds. For every
\(\mu=\mu_1\otimes\mu_2\in\mathcal X_\pm(r_{\mathrm{cap}})\) satisfying
\[
 W_{2,\Sphere}^2(\mu_1,\nu^o)
 +W_{2,\Sphere}^2(\mu_2,\nu^o)<\rho_o^2,
\]
and every smooth product tangent field \(v=(v_1,v_2)\) whose perturbation
paths preserve the fixed-assignment caps for small parameter values,
\begin{equation}
\label{eq:mf-teacher-local-positive-curvature}
 D_{W_2}^2F(\mu)[v,v]
 \ge c_o\|v\|_{L^2(\mu;\mathbb R^{d_w})}^2.
\end{equation}
No density assumption on \(\nu^o\) is required.
\end{proposition}

\begin{proof}
\noindent\textbf{Step 1: positive curvature at the exact-fit teacher.}
For smooth tangent fields \(v_i\) on the two factors, let
\[
 R_{i,h}(w):=\frac{w+hv_i(w)}{\|w+hv_i(w)\|_{\mathbb R^{d_w}}},
 \qquad
 \nu_{i,h}:=(R_{i,h})_\#\nu^o,
 \qquad
 \mu_h:=\nu_{1,h}\otimes\nu_{2,h}.
\]
The normalized retraction has the same second jet as the spherical EXP path
by \cref{eq:mf-retraction-second-jet}, so its second derivative computes
\(D_{W_2}^2F(\mu_+^o)[v,v]\). Set
\[
 Q(h,z):=h_{\nu_{1,h}}(z)h_{\nu_{2,h}}(z)^\top,
 \qquad
 B(z):=h_{\nu^o}(z)h_{\nu^o}(z)^\top.
\]
Exact fitting means \(Q(0,z)=B(z)\). Therefore, differentiating the squared
residual twice gives
\begin{align}
 D_{W_2}^2F_{\mathrm{fit}}(\mu_+^o)[v,v]
 &=\left.\frac{\dd^2}{\dd h^2}\right|_{h=0}
   \frac12\|Q(h)-B\|_{L^2_z(\mathrm F)}^2\notag\\
 &=\|\dot Q(0)\|_{L^2_z(\mathrm F)}^2\notag\\
 &=\sum_{p\ge0}\left\|
   \left.\frac{\dd}{\dd h}\right|_{h=0}
   \widetilde{\mathcal M}_p(\Theta(\mu_h))
   \right\|_{\mathrm{HS}}^2\ge0.
 \label{eq:mf-teacher-fitting-square}
\end{align}
The last equality is the finite active Hermite expansion from
\cref{lemma:EMN-expansion-F,eq:ambient-active-fit-functional}.

By \cref{lem:W2-power-regularizer-Hessian-on-fixed-caps}, the metric term at
the same state satisfies
\[
 D_{W_2}^2F_{\mathrm{reg}}(\mu_+^o)[v,v]
 \ge\lambdaReg\kappa_{\mathrm{cap}}Z_o^{(p_W-2)/2}
      \|v\|_{L^2(\mu_+^o;\mathbb R^{d_w})}^2.
\]
The coefficient is positive because \(Z_o>0\). Adding this inequality to
\cref{eq:mf-teacher-fitting-square} proves strict ambient curvature at the
teacher itself.

\par\noindent\textbf{Step 2: a lower bound at a nearby cap-supported state.}
Fix \(\mu=\mu_1\otimes\mu_2\in\mathcal X_\pm(r_{\mathrm{cap}})\), and use the
same normalized retractions based at each \(\mu_i\). Write
\[
 h_i(h,z):=\int\sigma(R_{i,h}(w)^\top z)\,\dd\mu_i(w),
 \qquad
 V_i:=\|v_i\|_{L^2(\mu_i;\mathbb R^{d_w})}.
\]
Dots below denote derivatives at \(h=0\). From
\cref{eq:mf-retraction-second-jet},
\(\dot R_{i,0}=v_i\) and
\(\ddot R_{i,0}=-\|v_i\|_{\mathbb R^{d_w}}^2w\), so differentiation under
the integral gives
\begin{equation}
\label{eq:mf-teacher-output-derivatives}
\begin{aligned}
 \dot h_i(z)
 &=\int\sigma'(w^\top z)(v_i(w)^\top z)\,\dd\mu_i(w),\\
 \ddot h_i(z)
 &=\int\left[
   \sigma''(w^\top z)(v_i(w)^\top z)^2
   -\|v_i(w)\|_{\mathbb R^{d_w}}^2
      \sigma'(w^\top z)(w^\top z)
   \right]\dd\mu_i(w).
\end{aligned}
\end{equation}
When \(v_i(w)\ne0\), the vectors
\(w\) and \(v_i(w)/\|v_i(w)\|_{\mathbb R^{d_w}}\) are orthogonal unit
vectors because \(v_i(w)\in T_w\Sphere^{d_w-1}\). Hence their projections
of \(z\sim N(0,I_{d_w})\) are independent standard Gaussians. Substitution
into the definitions of \(M_4,K_1,K_2\) in
\cref{eq:mf-gaussian-derivative-constants}, followed by Minkowski and
Cauchy--Schwarz, yields
\begin{equation}
\label{eq:mf-teacher-output-derivative-bounds}
 \|h_i(0)\|_{L^4_z}\le M_4,
 \qquad
 \|\dot h_i\|_{L^4_z}\le K_1V_i,
 \qquad
 \|\ddot h_i\|_{L^4_z}\le K_2V_i^2.
\end{equation}
At \(v_i(w)=0\), the corresponding integrands in
\cref{eq:mf-teacher-output-derivatives} vanish, so the same bounds hold
without division by zero. The finite polynomial activation and compact
sphere justify the differentiations and Gaussian integration.

For
\(Q(h,z):=h_1(h,z)h_2(h,z)^\top\), the product rule gives
\[
 \dot Q=\dot h_1h_2^\top+h_1\dot h_2^\top,
 \qquad
 \ddot Q=\ddot h_1h_2^\top+2\dot h_1\dot h_2^\top+h_1\ddot h_2^\top.
\]
Since \(\|xy^\top\|_{\mathrm F}=\|x\|_2\|y\|_2\), H\"older and
\cref{eq:mf-teacher-output-derivative-bounds} give, term by term,
\[
\begin{aligned}
 \|\ddot h_1h_2^\top\|_{L^2_z(\mathrm F)}&\le M_4K_2V_1^2,\\
 \|2\dot h_1\dot h_2^\top\|_{L^2_z(\mathrm F)}&\le2K_1^2V_1V_2,\\
 \|h_1\ddot h_2^\top\|_{L^2_z(\mathrm F)}&\le M_4K_2V_2^2.
\end{aligned}
\]
Using \(2V_1V_2\le V_1^2+V_2^2\), we obtain
\begin{equation}
\label{eq:mf-teacher-second-output-bound}
 \|\ddot Q\|_{L^2_z(\mathrm F)}
 \le (M_4K_2+K_1^2)
       \|v\|_{L^2(\mu;\mathbb R^{d_w})}^2.
\end{equation}

Recall \(B(z)=h_{\nu^o}(z)h_{\nu^o}(z)^\top\), and set
\[
 E(\mu):=\|h_{\mu_1}h_{\mu_2}^\top-B\|_{L^2_z(\mathrm F)},
 \qquad C:=M_4K_2+K_1^2.
\]
The constant \(C\) is proof-local and depends only on the fixed activation.
Differentiating
\(F_{\mathrm{fit}}(\mu_h)=\frac12\|Q(h)-B\|_{L^2_z(\mathrm F)}^2\),
dropping the nonnegative \(\|\dot Q\|_{L^2_z(\mathrm F)}^2\) term, and using
\cref{eq:mf-teacher-second-output-bound} gives
\begin{align*}
 D_{W_2}^2F_{\mathrm{fit}}(\mu)[v,v]
 &=\|\dot Q\|_{L^2_z(\mathrm F)}^2
   +\langle Q(0)-B,\ddot Q\rangle_{L^2_z(\mathrm F)}\\
 &\ge-CE(\mu)\|v\|_{L^2(\mu;\mathbb R^{d_w})}^2.
\end{align*}
Adding the regularizer estimate in
\cref{eq:W2-power-Hessian-lower-bound-fixed-caps} yields the key nearby
bound
\begin{equation}
\label{eq:mf-teacher-nearby-curvature-bound}
 D_{W_2}^2F(\mu)[v,v]
 \ge\left[
   \lambdaReg\kappa_{\mathrm{cap}}Z_\pm(\mu)^{(p_W-2)/2}
   -CE(\mu)
 \right]\|v\|_{L^2(\mu;\mathbb R^{d_w})}^2.
\end{equation}

\par\noindent\textbf{Step 3: keep the coefficient positive in a teacher neighborhood.}
Let
\[
 d:=\left[
 W_{2,\Sphere}^2(\mu_1,\nu^o)+
 W_{2,\Sphere}^2(\mu_2,\nu^o)
 \right]^{1/2}.
\]
For sphere points \(w,y\), let \(c:[0,1]\to\Sphere^{d_w-1}\) be a shortest
constant-speed arc from \(y\) to \(w\). Then
\(\|c'(s)\|_{\mathbb R^{d_w}}=d_\Sphere(w,y)\) and
\(c(s)^\top c'(s)=0\). Thus, when the speed is nonzero,
\(c(s)\) and \(c'(s)/\|c'(s)\|_{\mathbb R^{d_w}}\) are orthogonal unit
vectors. The same Gaussian calculation used in
\cref{eq:mf-teacher-output-derivative-bounds} gives
\[
 \|\sigma'(c(s)^\top z)(c'(s)^\top z)\|_{L^4_z}
 =K_1\|c'(s)\|_{\mathbb R^{d_w}}.
\]
The identity is immediate when the speed is zero. Integrating along the arc
and applying Minkowski gives
\[
 \|\sigma(w^\top z)-\sigma(y^\top z)\|_{L^4_z}
 \le K_1d_\Sphere(w,y).
\]
For an optimal coupling \(\pi_i\) of \(\mu_i\) and \(\nu^o\), another use
of Minkowski and Cauchy--Schwarz therefore yields
\[
\begin{aligned}
 \|h_{\mu_i}-h_{\nu^o}\|_{L^4_z}
 &\le K_1\int d_\Sphere(w,y)\,\dd\pi_i(w,y)\\
 &\le K_1W_{2,\Sphere}(\mu_i,\nu^o).
\end{aligned}
\]
Subtracting the two product outputs,
\[
 h_{\mu_1}h_{\mu_2}^\top-B
 =(h_{\mu_1}-h_{\nu^o})h_{\mu_2}^\top
 +h_{\nu^o}(h_{\mu_2}-h_{\nu^o})^\top.
\]
Since \(\|h_{\mu_i}\|_{L^4_z},\|h_{\nu^o}\|_{L^4_z}\le M_4\) by
\cref{eq:mf-output-fourth-moment-bound}, H\"older and the preceding feature
bound give \(E(\mu)\le\sqrt2M_4K_1d\). Moreover,
\[
 d=W_{2,\mathcal M}(\mu,\mu_+^o),\qquad
 \sqrt{Z_\pm(\mu)}=W_{2,\mathcal M}(\mu,\mu^\dagger),\qquad
 \sqrt{Z_o}=W_{2,\mathcal M}(\mu_+^o,\mu^\dagger),
\]
by the product metric in \cref{eq:product-Wasserstein-metric} and the
definition of $Z_\pm$ in \cref{eq:fixed-assignment-W2-square-formula},
together with the definition of $Z_o$ in the proposition statement.
Therefore the reverse triangle inequality gives
\(\left|\sqrt{Z_\pm(\mu)}-\sqrt{Z_o}\right|\le d\). Combining the two
estimates yields
\begin{equation}
\label{eq:mf-teacher-neighborhood-continuity}
 E(\mu)\le\sqrt2M_4K_1d,
 \qquad
 \left|\sqrt{Z_\pm(\mu)}-\sqrt{Z_o}\right|\le d.
\end{equation}

As \(d\to0\), \cref{eq:mf-teacher-neighborhood-continuity} shows that
\(E(\mu)\to0\) and \(Z_\pm(\mu)\to Z_o\). Hence the bracket in
\cref{eq:mf-teacher-nearby-curvature-bound} converges to the positive number
\(\lambdaReg\kappa_{\mathrm{cap}}Z_o^{(p_W-2)/2}\). Choose
\(\rho_o>0\) so that, whenever \(d<\rho_o\), the regularizer coefficient is
at least three quarters of this limit and \(CE(\mu)\) is at most one quarter
of it. Then \cref{eq:mf-teacher-nearby-curvature-bound} gives
\cref{eq:mf-teacher-local-positive-curvature} with
\[
 c_o:=\frac12\lambdaReg\kappa_{\mathrm{cap}}Z_o^{(p_W-2)/2}>0.
\]
The neighborhood is relative to \(\mathcal X_\pm(r_{\mathrm{cap}})\): small
Wasserstein distance alone does not force the entire support to remain in the
fixed-assignment caps. None of the estimates uses a teacher density.
\end{proof}

If \(Z_o=0\), the proposition does not apply. In that case the metric
regularizer is Hessian-flat at the antipodal Dirac target by
\cref{eq:mf-center-EXP-regularizer}.

%% file: appendix_numerical/20260926-023032JST_sanity_check.tex
\clearpage
\section{Numerical sanity checks}
\label{sec:numerical-sanity}

We examine the non-spikiness condition in
\Cref{assumption-main-coupled-particle-regularity}(iii) and the behavior of
adaptive PWGF with each system making its own perturbation decisions.
Following \citet[Algorithm~1]{yamamoto2025hessianguided}, the trigger uses the
\(L^2\) norm of the gradient under that system's current distribution.
We use a smooth matrix-factorization model, separate from the spherical model
of \Cref{sec:matrix-factorization}.

\subsection{Model and population--particle comparison}
\label{subsec:numerical-model}

\noindent\textbf{Objective and initial law.}
A particle is a matrix \(W\in\mathbb R^{32\times2}\), identified with
\(\mathbb R^{64}\) by row-major vectorization; matrix norms below are
Frobenius norms. With entrywise \(\tanh\), set
\begin{equation}
 F(\nu)=f(A_\nu),\qquad
 f(A)=\frac14\|AA^\top-M_*\|_F^2,\qquad
 A_\nu=\int\tanh(W)\,\nu(dW).
 \label{eq:numerical-objective}
\end{equation}
Here \(M_*=u_1u_1^\top+\tfrac12u_2u_2^\top\), where
\((u_1)_a=1/\sqrt{32}\) and
\((u_2)_a=(-1)^{a-1}/\sqrt{32}\) for \(1\le a\le32\).
Thus the objective factorizes the mean transformed particle.
Generate \(2^{14}\) matrices \(Z_a\) with independent standard normal entries
once (seed 20260925), and fix the realized population law
\[
 P_a(0-)=Z_a,\quad P_{a+2^{14}}(0-)=-Z_a,\qquad
 \mu_{0-}=2^{-15}\sum_{a=1}^{2^{14}}(\delta_{Z_a}+\delta_{-Z_a}).
\]
The notation \(0-\) denotes initialization before a possible jump at \(t=0\).
For each of three trials, sample labels \(I_i\) independently and uniformly
from \(\{1,\ldots,2^{15}\}\), independently of the population innovations, and
set \(X_{0-}^i=P_{I_i}(0-)\) before that jump. We use
\(N\in\{2048,4096,8192,16384,32768\}\), with nested label prefixes within each
trial. Sampling is with replacement, including at \(N=32768\).

\noindent\textbf{Continuous dynamics.}
The specified population has 32768 atoms, all of which are evolved; its
integrals are full atom averages. The empirical system uses only its sampled
particles:
\[
 \mu_t=2^{-15}\sum_{a=1}^{2^{15}}\delta_{P_a(t)},\qquad
 \widehat\mu_t^N=\frac1N\sum_{i=1}^N\delta_{X_t^i},\qquad
 Y_t^i=P_{I_i}(t).
\]
The ODEs and their shared velocity formula are
\begin{equation}
\begin{aligned}
 &\dot P_a(t)=-\nabla_WF(\mu_t,P_a(t)),\qquad
 \dot X_t^i=-\nabla_WF(\widehat\mu_t^N,X_t^i),\\
 &-\nabla_WF(\nu,W)=-(1-\tanh^2W)\odot
             [(A_\nu A_\nu^\top-M_*)A_\nu],
\end{aligned}
\label{eq:numerical-flow}
\end{equation}
where \(\odot\) is entrywise multiplication. In particular, for
\(F_N(\mathbf X)=F(\widehat\mu^N)\), the particle velocity is
\(-N\nabla_{X^i}F_N\). We integrate both systems by RK4; the population is
a fixed discrete measure, not a Monte Carlo approximation to a continuous law.

\noindent\textbf{Gaussian jumps.}
For a matrix \(W\), write
\(J_W=\operatorname{diag}(\operatorname{vec}(1-\tanh^2W))\) and
\(S_\nu=\int J_U^2\,\nu(dU)\).
Since \(\nabla f(A)=(AA^\top-M_*)A\), differentiation in a matrix direction
\(B\in\mathbb R^{32\times2}\) gives
\(\mathrm D(\nabla f)(A)[B]=(BA^\top+AB^\top)A+(AA^\top-M_*)B\).
Thus the symmetric matrix \(L_A\in\mathbb R^{64\times64}\) representing the
Hessian of \(f\) satisfies
\[
 L_A\operatorname{vec}(B)=\operatorname{vec}\bigl(
 (AA^\top-M_*)B+B(A^\top A)+AB^\top A\bigr).
\]
Thus the integral Hessian and its Gaussian covariance from
\Cref{subsec:population-pwgf} are
\begin{equation}
 \nabla_W^2F(\nu;W,V)=J_WL_{A_\nu}J_V,\qquad
 K_\nu(W,V)=J_WL_{A_\nu}S_\nu L_{A_\nu}^\top J_V.
 \label{eq:numerical-covariance}
\end{equation}
Since \(\operatorname{vec}(\mathrm D\tanh(V)[B])=J_V\operatorname{vec}(B)\),
the chain rule in \cref{eq:numerical-flow} yields the first identity.
The kernel composition in \Cref{subsec:population-pwgf} gives
\(K_\nu(W,V)=\int (J_WL_{A_\nu}J_U)(J_UL_{A_\nu}J_V)\,\nu(dU)
=J_WL_{A_\nu}S_\nu L_{A_\nu}J_V\), where
\(S_\nu=\int J_U^2\,\nu(dU)\) and \(L_{A_\nu}=L_{A_\nu}^\top\).
A draw of the field is
\(\operatorname{vec}(\xi_\nu(W))=J_WL_{A_\nu}S_\nu^{1/2}z\),
using one \(z\sim\mathcal N(0,I_{64})\) for all \(W\); its covariance is
\(K_\nu(W,V)\) because
\(S_\nu^{1/2}\mathbb E[zz^\top]S_\nu^{1/2}=S_\nu\).
The Gaussian rank is at most 64 even without finite support.
At their respective trigger times, the population and particle jumps have
marginal laws
\begin{equation}
\begin{aligned}
 P_a^+&=P_a^-+\eta\xi_{\mu^-}(P_a^-),\\
 \operatorname{vec}(X^{i,+}-X^{i,-})
 &=\frac\eta{\sqrt N}\sum_{j=1}^N
 \nabla_W^2F(\widehat\mu^{N,-};X^{i,-},X^{j,-})g_j,
 \qquad g_j\overset{\mathrm{iid}}\sim\mathcal N(0,I_{64}).
\end{aligned}
\label{eq:numerical-jumps}
\end{equation}
The implementation uses equivalent 64-dimensional factors.
The population path is generated first. Its \(j\)-th innovation is coupled,
by covariance-factor alignment with independent Gaussian completion, to the
empirical system's \(j\)-th attempt. An unmatched empirical attempt uses a
fresh independent field with its empirical covariance. This coupling uses
attempt index, not common physical time; the reference particles do not enter
the controller.

\subsection{Adaptive controller and recorded diagnostics}
\label{subsec:numerical-diagnostics}

Each system follows the gradient-trigger, wait, and decrease-test pattern of
\citet[Algorithm~1]{yamamoto2025hessianguided}. The controller acts separately
for the population and each particle system:
\begin{enumerate}
\item \textbf{Inspect and perturb.}
At \(t=0,0.1,\ldots,99.9\), if the system has not stopped and no waiting
window is active, test its pre-jump state using
\[
\begin{aligned}
\text{population:}\quad &
 \left(2^{-15}\sum_{a=1}^{2^{15}}
 \|\nabla_WF(\mu_{t-},P_a(t-))\|_F^2\right)^{1/2}\le0.01,\\
\text{particles:}\quad &
 \left(\frac1N\sum_{i=1}^N
 \|\nabla_WF(\widehat\mu_{t-}^N,X_{t-}^i)\|_F^2\right)^{1/2}\le0.01.
\end{aligned}
\]
These are the respective \(L^2\)-gradient norms, evaluated from
\cref{eq:numerical-flow}; each system uses its own measure in both the gradient
and the average. When its inequality holds, save the pre-jump measure and
apply the corresponding jump in \cref{eq:numerical-jumps} with \(\eta=0.2\).
Otherwise, continue the flow to the next inspection time.
\item \textbf{Wait and test the decrease.}
After a jump at \(s\), evolve for 10 time units without another jump, then
compare the objective with the saved pre-jump value:
\[
\begin{cases}
 \text{resume inspection at the next grid tick},
   &F(\nu_{s-})-F(\nu_{s+10})>0.001,\\
 \text{stop and return }\nu_{s-},
   &F(\nu_{s-})-F(\nu_{s+10})\le0.001.
\end{cases}
\]
Here \(\nu\) is the measure of the system being tested. A passed decrease test
does not trigger another jump at \(s+10\); inspection resumes at \(s+10.1\)
if time remains before the cap of 100. The returned candidate \(\nu_{s-}\)
is distinct from the executed state \(\nu_{s+10}\).
\end{enumerate}

\begin{center}
\begin{tabular}{@{}lclc@{}}
\toprule
Parameter & Value & Parameter & Value\\
\midrule
Jump amplitude \(\eta\) & 0.2 & Gradient threshold & 0.01\\
Waiting time & 10 & Decrease threshold & 0.001\\
RK4 step & 0.05 & Inspection interval & 0.1\\
Time cap & 100 & Trials & 3\\
\bottomrule
\end{tabular}
\end{center}

\begin{figure}[!ht]
\centering
\includegraphics[width=\linewidth]{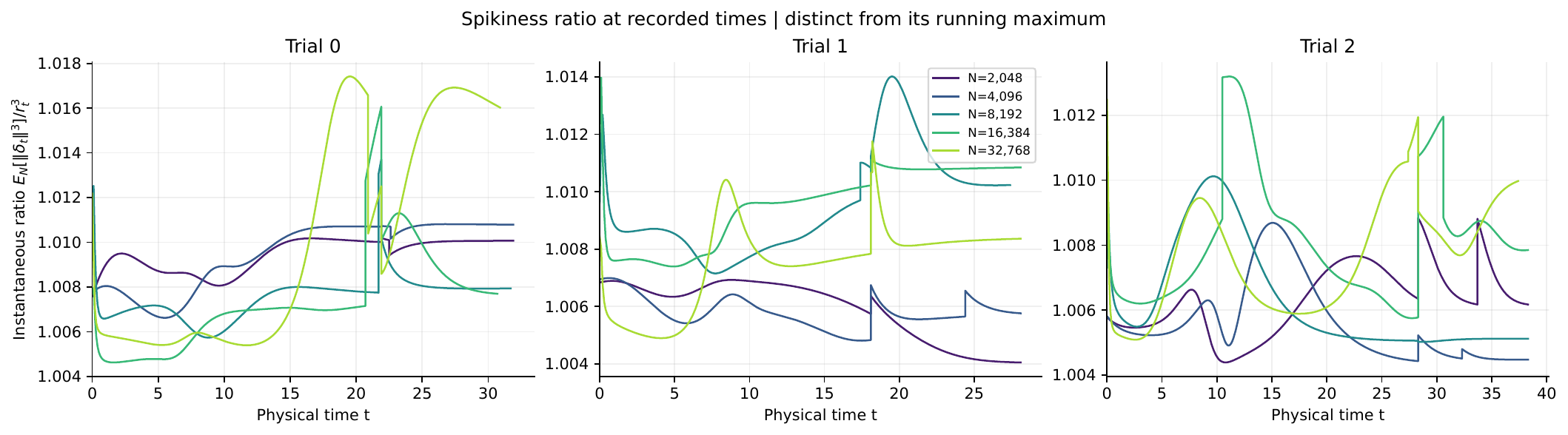}
\caption{The instantaneous cubic ratio \(s_t\) in
\cref{eq:numerical-spikiness}, with all three trials and five particle counts.
Curves end when either system stops; near-zero denominators are omitted.
The ratio measures the distribution of paired errors, not particle positions.}
\label{fig:numerical-spikiness}
\end{figure}

\noindent\textbf{Non-spikiness.}
At common physical times while both systems remain active, we compute the
paired error from \Cref{subsec:tracking-metric} and its cubic ratio:
\begin{equation}
 r_t=\left(\frac1N\sum_i\|X_t^i-Y_t^i\|_F^2\right)^{1/2},\qquad
 s_t=\frac{N^{-1}\sum_i\|X_t^i-Y_t^i\|_F^3}{r_t^3}.
 \label{eq:numerical-spikiness}
\end{equation}
We record inspection times and both sides of each jump, and omit \(s_t\)
when \(r_t\le10^{-12}\max\{1,(N^{-1}\sum_i\|Y_t^i\|_F^2)^{1/2}\}\).
Across all 15 conditions, the largest recorded ratio is 1.017416.
\Cref{fig:numerical-spikiness} therefore supports non-spiky paired errors in
these runs: the empirical analogue of
\Cref{assumption-main-coupled-particle-regularity}(iii) is close to one at
recorded times. This is not a bound over all continuous times or random paths.

\begin{figure}[!ht]
\centering
\includegraphics[width=\linewidth]{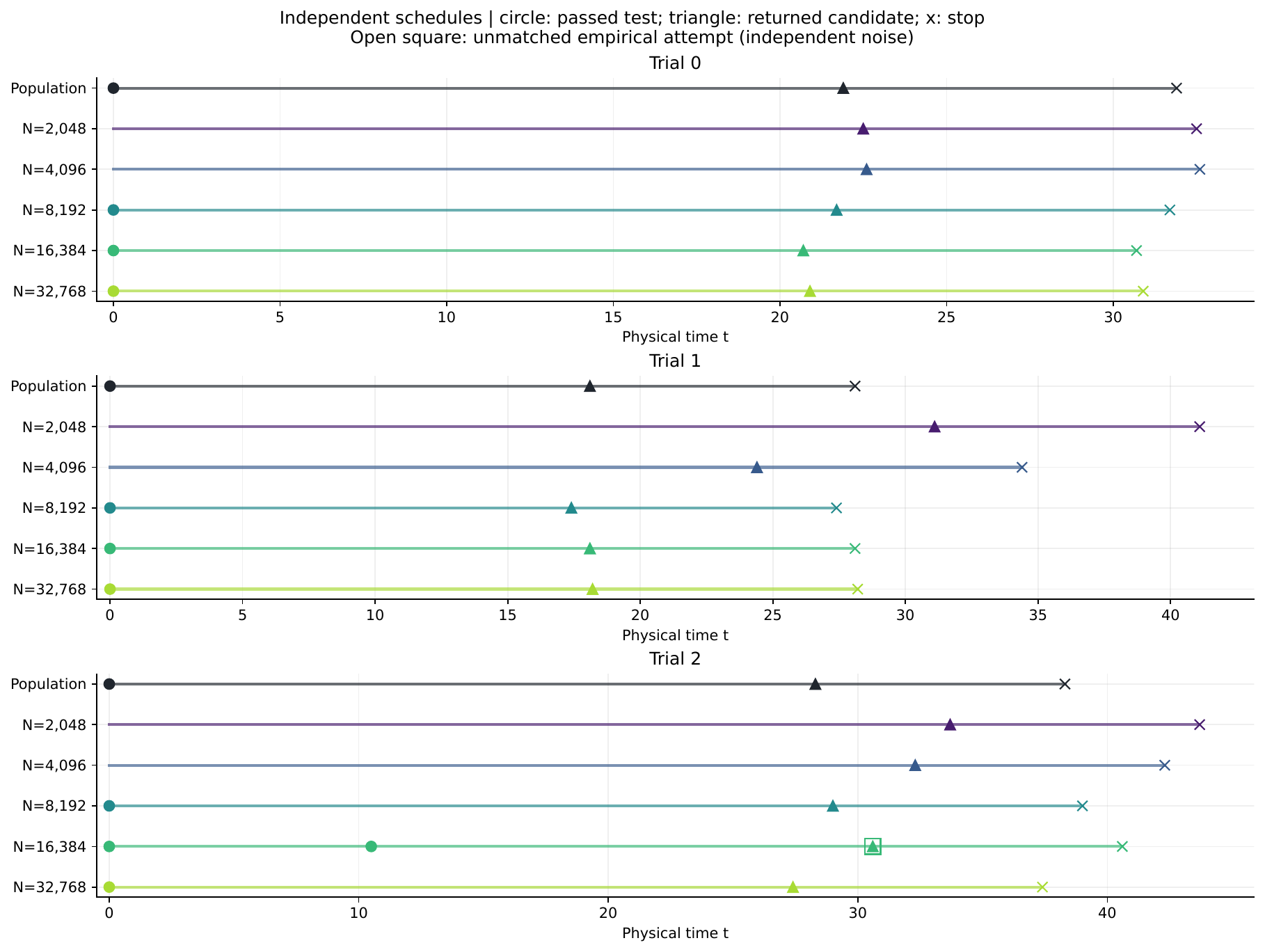}
\caption{Independently chosen adaptive schedules, showing every system.
Circles mark jumps whose decrease test passes; triangles mark the saved
pre-jump candidates returned after a failed test; crosses mark stopping times.
The open square marks an unmatched empirical attempt. Coupling by attempt
index does not force jump times to coincide.}
\label{fig:numerical-schedules}
\end{figure}

\noindent\textbf{Adaptive schedules.}
All 15 particle systems and three population paths stopped before the cap;
the empirical returned objectives ranged from \(1.12\times10^{-4}\) to
\(1.41\times10^{-4}\).
\Cref{fig:numerical-schedules} illustrates autonomous decisions based on each
system's own \(L^2\)-gradient test: all \(N=2048,4096\) systems skip the
population's initial jump, and trial 2 at \(N=16384\) makes a third attempt
without a population counterpart. The resulting differences in physical jump
times can produce temporary same-time discrepancies: at \(N=32768\), recorded
peaks of \(r_t\) range from 0.584 to 0.601. Discrepancies can be much larger
at smaller \(N\) and are not monotone in \(N\), despite the near-unit spikiness
ratios. These diagnostics complement \Cref{theorem-main-r_t}; its guarantee
remains for a prescribed common schedule.

\noindent\textbf{Numerical checks and reproduction.}
Halving the RK4 step preserved all event times and counts; halving the
inspection interval preserved counts but shifted some event times by 0.05.
The latter changed individual returned objectives by up to
\(4.89\times10^{-6}\), so small objective gaps are not used to infer a rate.
Code, configurations, exact seeds, and archived scalar records are provided
in the accompanying reproducibility bundle.

\clearpage
\subsection{Paired trajectories and objective values}
\label{subsec:numerical-tracking}

\begin{figure}[!ht]
\centering
\includegraphics[width=\linewidth]{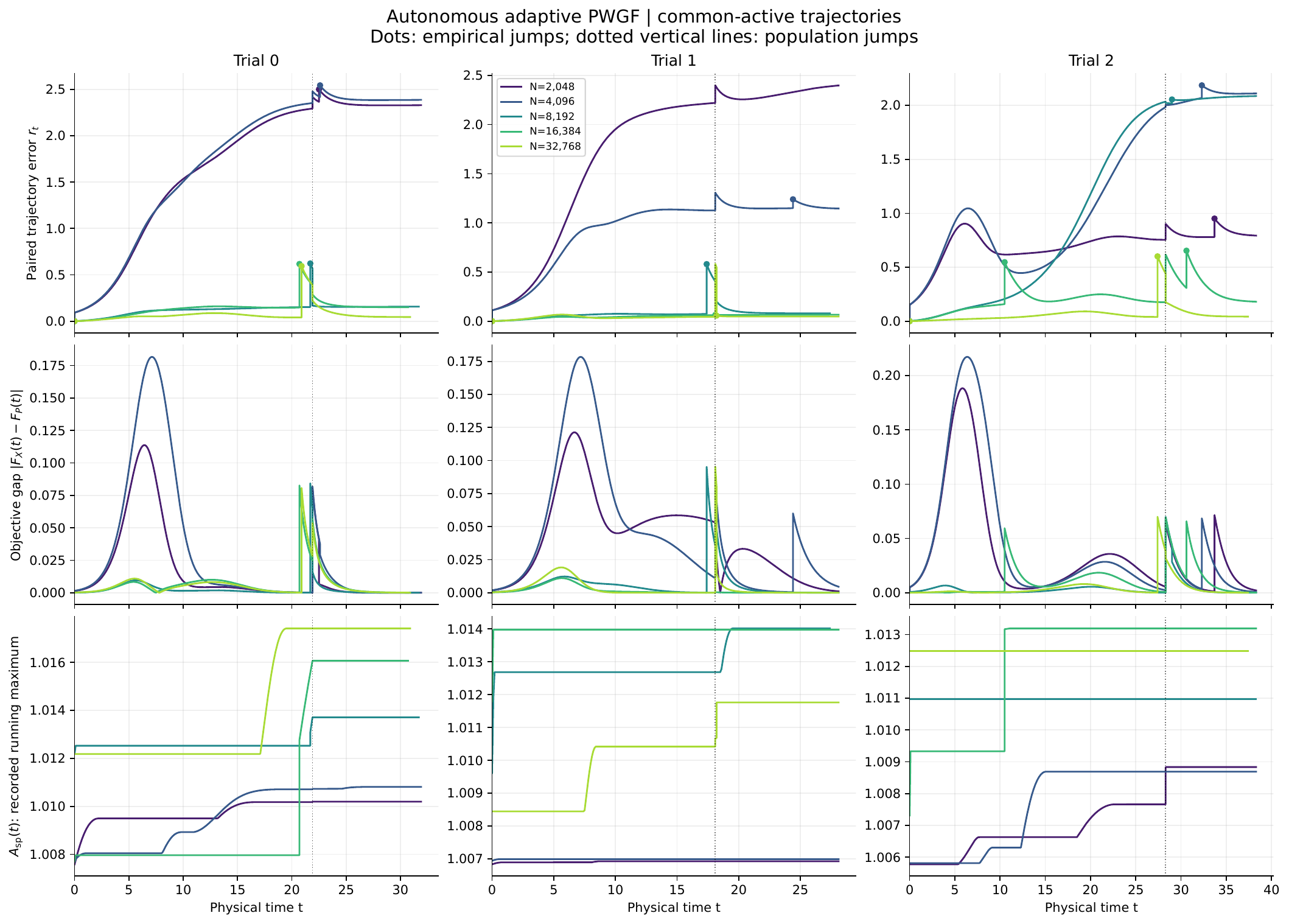}
\caption{Common-time diagnostics for all 15 conditions, with one column per
trial. Top: paired RMS error \(r_t\).
Middle: \(\lvert F_X(t)-F_P(t)\rvert\), where
\(F_X(t)=F(\widehat\mu_t^N)\) and \(F_P(t)=F(\mu_t)\).
Bottom: the running maximum of the recorded ratio \(s_t\), labeled
\(A_{\rm sp}(t)\), rather than its instantaneous value in
\cref{fig:numerical-spikiness}. Dots in the top row mark empirical jumps;
dotted vertical lines mark population jumps. Curves end when either system
stops; vertical segments connect recorded pre/post-jump values.}
\label{fig:numerical-tracking}
\end{figure}

\begin{figure}[!ht]
\centering
\includegraphics[width=\linewidth]{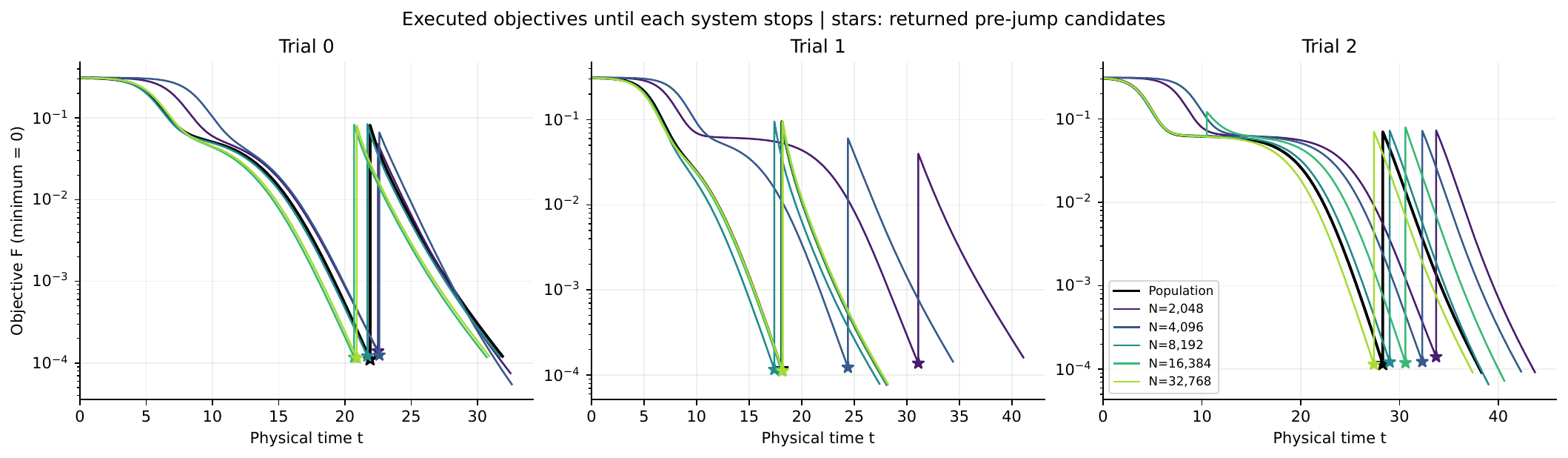}
\caption{Objective values \(F(\mu_t)\) (black) and
\(F(\widehat\mu_t^N)\) (colored) for every trial and particle count.
Each curve continues to that system's own stopping time. Stars mark the
saved pre-jump candidates at their candidate times, rather than the final
executed states. The minimum of \cref{eq:numerical-objective} is zero.}
\label{fig:numerical-objectives}
\end{figure}

\Cref{fig:numerical-tracking} shows the paired error \(r_t\) from
\cref{eq:numerical-spikiness} and the objective gap
\(\lvert F(\widehat\mu_t^N)-F(\mu_t)\rvert\) at the same physical time.
\Cref{fig:numerical-objectives} shows the objective values themselves, using
\cref{eq:numerical-objective}, until each system stops. Together they show
particle discrepancies and optimization progress under separately triggered
jumps.

